\documentclass[]{huawei_cloud}

\usepackage{minitoc}
\usepackage{xurl}
\usepackage{hyperref}

\usepackage{arydshln}
\usepackage{amsmath,amssymb}
\usepackage{algorithm}
\usepackage{algorithmicx}
\usepackage{algpseudocode}

\usepackage{booktabs}
\usepackage{multirow}
\usepackage{array}
\usepackage{tabularx}
\usepackage{makecell}
\usepackage{adjustbox}

\usepackage[dvipsnames,svgnames,table]{xcolor}
\usepackage{colortbl}
\usepackage{arydshln}

\usepackage{graphicx}
\usepackage{wrapfig}

\usepackage[most]{tcolorbox}
\tcbuselibrary{listings,breakable,skins}
\usepackage{listings}
\usepackage{tikz}

\usepackage{pifont}
\usepackage{fontawesome5}

\usepackage[toc,page,header]{appendix}

\usepackage{CJKutf8}

\definecolor{HeaderGray}{RGB}{242,244,247}
\definecolor{GroupGray}{RGB}{232,235,240}
\definecolor{FoundationBlue}{RGB}{233,242,255}
\definecolor{ScaleGreen}{RGB}{235,248,240}
\definecolor{SourceLavender}{RGB}{242,238,250}
\definecolor{AgentOrange}{RGB}{255,243,230}
\definecolor{RuleGray}{RGB}{180,185,192}
\colorlet{ProprietaryBand}{FoundationBlue}
\colorlet{OpenWeightBand}{ScaleGreen}
\colorlet{AgenticBand}{SourceLavender}
\colorlet{SourceOnlyBand}{GroupGray}

\definecolor{RoleUser}{RGB}{54,88,153}
\definecolor{RoleAssistant}{RGB}{46,125,93}
\definecolor{RoleTool}{RGB}{161,98,7}
\definecolor{ActionBlue}{RGB}{33,101,181}

\definecolor{HeaderBlue}{HTML}{EAF1FB}
\definecolor{OverallGray}{HTML}{F2F4F7}

\definecolor{DemoUserBg}{RGB}{245,248,255}
\definecolor{DemoUserFrame}{RGB}{86,116,185}

\definecolor{DemoTraceBg}{RGB}{246,250,247}
\definecolor{DemoTraceFrame}{RGB}{78,145,108}

\definecolor{DemoAgentBg}{RGB}{244,251,247}
\definecolor{DemoToolBg}{RGB}{255,248,238}
\definecolor{DemoMsgFrame}{RGB}{205,211,220}
\definecolor{DemoRubricBg}{RGB}{255,247,248}
\definecolor{DemoRubricFrame}{RGB}{190,92,108}
\definecolor{RoleRubric}{RGB}{153,54,70}
\definecolor{RubricRed}{RGB}{252,232,235}
\definecolor{MutedText}{RGB}{95,102,115}
\definecolor{GainGreen}{RGB}{38,126,83}

\definecolor{CodeBackground}{HTML}{F7F9FC}
\definecolor{CodeFrame}{HTML}{D8DEE9}
\definecolor{CodeString}{HTML}{245A8D}
\definecolor{CodeKeyword}{HTML}{8A3B72}
\definecolor{CodeComment}{HTML}{6B7280}

\definecolor{SystemPrimary}{HTML}{6E56CF}
\definecolor{SystemDark}{HTML}{4936A3}
\definecolor{SystemSoft}{HTML}{F5F2FF}
\definecolor{SystemRule}{HTML}{D9D0F5}

\definecolor{UserPrimary}{HTML}{3478D4}
\definecolor{UserDark}{HTML}{20599F}
\definecolor{UserSoft}{HTML}{F2F7FD}
\definecolor{UserRule}{HTML}{C9DBF2}

\definecolor{PromptText}{HTML}{263238}
\definecolor{PromptMuted}{HTML}{65717D}
\definecolor{PromptComment}{HTML}{83909C}
\definecolor{PromptString}{HTML}{17865D}
\definecolor{PromptKey}{HTML}{365DB5}
\definecolor{PromptNumber}{HTML}{C26420}
\definecolor{PromptPunctuation}{HTML}{6D7780}

\definecolor{SystemPurple}{HTML}{6C5CE7}
\definecolor{SystemPurpleDark}{HTML}{4635A8}
\definecolor{SystemPurpleSoft}{HTML}{F7F5FF}
\definecolor{SystemPurpleLine}{HTML}{DCD5F7}

\definecolor{UserBlue}{HTML}{3478D4}
\definecolor{UserBlueDark}{HTML}{20599F}
\definecolor{UserBlueSoft}{HTML}{F4F8FD}
\definecolor{UserBlueLine}{HTML}{CEDFF3}

\definecolor{CodeText}{HTML}{263238}
\definecolor{CodeString}{HTML}{18805D}
\definecolor{CodeComment}{HTML}{7B8794}

\lstdefinestyle{ModernPromptBase}{
  basicstyle=\ttfamily\fontsize{7.5pt}{10.2pt}\selectfont
             \color{PromptText},
  columns=fullflexible,
  keepspaces=true,
  showstringspaces=false,
  showtabs=false,
  tabsize=2,
  breaklines=true,
  breakatwhitespace=false,
  postbreak=\mbox{\textcolor{PromptMuted}{\(\hookrightarrow\)}\space},
  upquote=true,
  aboveskip=0pt,
  belowskip=0pt,
  xleftmargin=0pt,
  xrightmargin=0pt,
  frame=none,
  backgroundcolor=\color{white},
  literate=
    {**}{{\bfseries}}2
    {`}{{\textcolor{SystemPrimary}{\textasciigrave}}}1
}

\newtcblisting{SystemPromptCard}{
  enhanced,
  breakable,
  listing only,
  listing options={
    style=PromptJSON
  },
  colback=SystemPurpleSoft,
  colframe=SystemPurpleLine,
  boxrule=0.7pt,
  arc=3pt,
  outer arc=3pt,
  left=10pt,
  right=10pt,
  top=10pt,
  bottom=10pt,
  before skip=10pt,
  after skip=12pt,
  title={
    \sffamily\bfseries\color{SystemPurpleDark}
    SYSTEM PROMPT
    \hfill
    \scriptsize\color{SystemPurple}
    DIAGNOSTIC PROTOCOL
  },
  colbacktitle=white,
  coltitle=SystemPurpleDark,
  fonttitle=\sffamily\bfseries\small,
  boxed title style={
    colback=white,
    colframe=white,
    boxrule=0pt,
    sharp corners
  },
  attach boxed title to top left={
    xshift=8pt,
    yshift*=-\tcboxedtitleheight
  },
  top=13pt,
  borderline west={
    3pt
  }{
    0pt
  }{
    SystemPurple
  },
  overlay unbroken and first={
    \fill[SystemPurple]
      ([yshift=-1pt]frame.north west)
      rectangle
      ([xshift=4pt,yshift=1pt]frame.south west);
  }
}

\newtcblisting{UserPromptCard}{
  enhanced,
  breakable,
  listing only,
  listing options={
    style=PromptText
  },
  colback=UserBlueSoft,
  colframe=UserBlueLine,
  boxrule=0.7pt,
  arc=3pt,
  outer arc=3pt,
  left=10pt,
  right=10pt,
  top=10pt,
  bottom=10pt,
  before skip=10pt,
  after skip=12pt,
  title={
    \sffamily\bfseries\color{UserBlueDark}
    USER PROMPT
    \hfill
    \scriptsize\color{UserBlue}
    FAILED ROLLOUT EVIDENCE
  },
  colbacktitle=white,
  coltitle=UserBlueDark,
  fonttitle=\sffamily\bfseries\small,
  boxed title style={
    colback=white,
    colframe=white,
    boxrule=0pt,
    sharp corners
  },
  attach boxed title to top left={
    xshift=8pt,
    yshift*=-\tcboxedtitleheight
  },
  top=13pt,
  borderline west={
    3pt
  }{
    0pt
  }{
    UserBlue
  },
  overlay unbroken and first={
    \fill[UserBlue]
      ([yshift=-1pt]frame.north west)
      rectangle
      ([xshift=4pt,yshift=1pt]frame.south west);
  }
}

\lstdefinestyle{PromptJSON}{
  style=ModernPromptBase,
  keywordstyle=\color{PromptKey}\bfseries,
  stringstyle=\color{PromptString},
  commentstyle=\color{PromptComment}\itshape
}

\lstdefinestyle{PromptText}{
  style=ModernPromptBase
}

\lstdefinelanguage{AgentJSON}{
  sensitive=true,
  morestring=[b]",
  morecomment=[l]{//},
  morekeywords={true,false,null}
}

\lstdefinestyle{PromptBase}{
  basicstyle=\ttfamily\fontsize{8.5pt}{10.5pt}\selectfont,
  backgroundcolor=\color{CodeBackground},
  frame=single,
  rulecolor=\color{CodeFrame},
  framerule=0.4pt,
  framesep=6pt,
  xleftmargin=0.5em,
  xrightmargin=0.5em,
  showstringspaces=false,
  breaklines=true,
  breakatwhitespace=false,
  breakindent=1.5em,
  columns=fullflexible,
  keepspaces=true,
  tabsize=2,
  aboveskip=0.65\baselineskip,
  belowskip=0.65\baselineskip
}

\lstdefinestyle{PromptJSON}{
  style=PromptBase,
  language=AgentJSON,
  stringstyle=\color{CodeString},
  keywordstyle=\bfseries\color{CodeKeyword},
  commentstyle=\itshape\color{CodeComment}
}

\lstdefinestyle{PromptText}{
  style=PromptBase
}

\title{AgentOmnia: Scaling Agentic Models for Full-Scenario Applications}

\author{Huawei Cloud Post-Training Team}

\abstract{Large language model agents have advanced rapidly, yet progress remains fragmented across domains, capabilities, task difficulty, and interaction settings. We study this problem as \emph{full-scenario agentic scaling} and present \textbf{AgentOmnia}, a framework that coordinates task-space definition, data synthesis, post-training, evaluation, and iterative improvement for To-Consumer (ToC), To-Business (ToB), and To-Employee (ToE) applications. An extensible \emph{Domain $\times$ Capability $\times$ Atomic Difficulty} taxonomy aligns these stages and supports fine-grained diagnosis through the publicly released OmniaBench. AgentOmnia combines bidirectional environment--task synthesis with tool-dependency, program-structured, and solver-based task pipelines, constructing 5,018 code-driven, stateful environments with 255,375 tools and 52,361 tasks. Programs, solvers, and verifiers provide correctness signals for difficult tasks, while supervised fine-tuning, online agentic reinforcement learning, and a rollback curriculum support post-training. Evaluation failures can further be translated into Product Requirement Documents (PRDs) to guide targeted self-evolution. Starting from Qwen3-30B-A3B-Thinking-2507, AgentOmnia raises the task pass rate on the OmniaBench challenging subset from 9.16\% to 37.11\% and the macro-average over OmniaBench, $\tau^2$-Bench, DeepPlanning, and VitaBench from 22.86\% to 41.69\%. Under a unified evaluation protocol, it leads the evaluated agentic post-trained baselines on OmniaBench and retains the highest four-benchmark macro-average, despite trailing recent Qwen3.5-based Agents-A1 and Nex-N2-Mini on DeepPlanning. It also surpasses Qwen3-235B-A22B-Thinking-2507 on all four benchmarks and exceeds Qwen3.5-35B-A3B on the macro-average. Gains span 76 of 90 level-1 domains across ToC/ToB/ToE, all ten capability dimensions, and all eight atomic-difficulty factors, indicating broad rather than category-specific improvement. Finally, a preliminary one-round study provides initial evidence for PRD-guided self-evolution, motivating further validation at larger scales and in industrial settings.}
\date{\today}

\begin{document}

\maketitle

\begin{figure}[H]
    \centering
    \setlength{\abovecaptionskip}{10pt}
    \setlength{\belowcaptionskip}{0pt}
    \includegraphics[
        width=0.96\textwidth
    ]{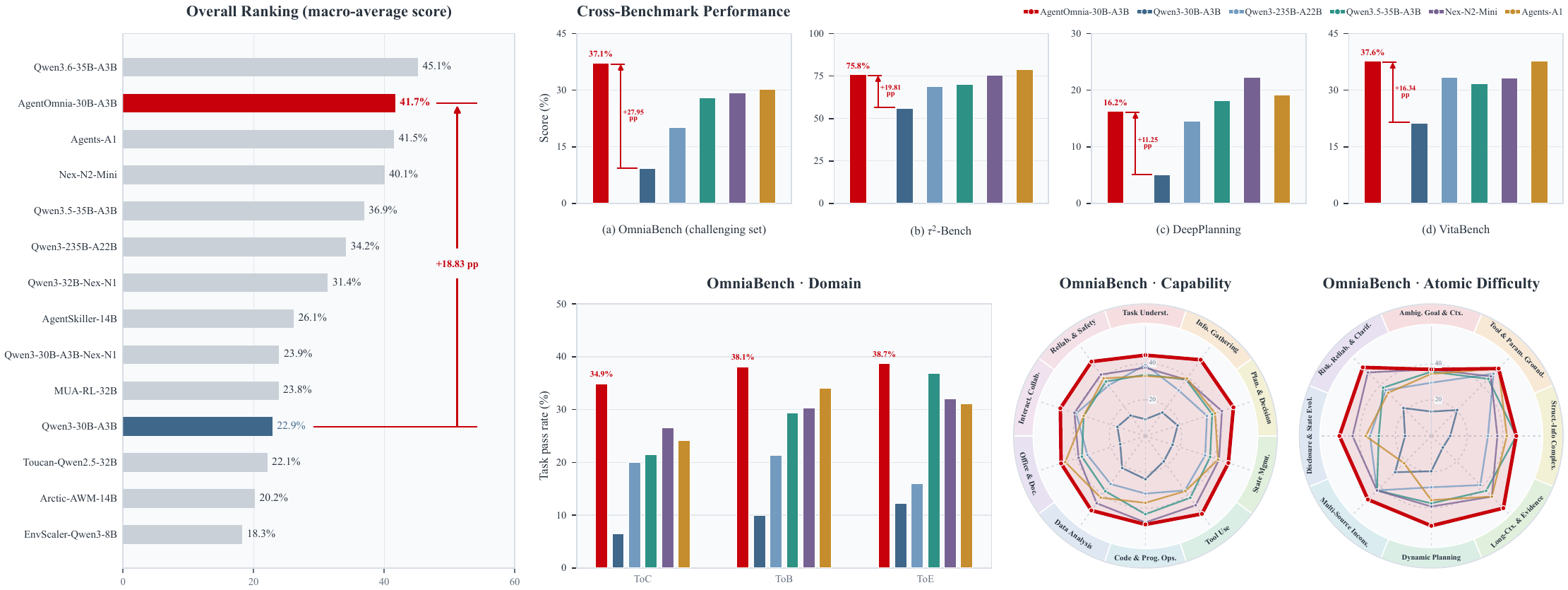}
    \caption{Performance overview of AgentOmnia. All displayed results are reproduced under a unified evaluation setting.}
    \label{fig:performance_overview}
\end{figure}

\noindent
{\scriptsize
We warmly welcome discussion, collaboration, and contributions to AgentOmnia.
Contact:
chenchong55@huawei.com, jianghao66@huawei.com
}

\newpage
\tableofcontents
\newpage

\section{Introduction}
\label{sec:intro}

Large language model (LLM) agents have progressed from reasoning-and-acting loops~\cite{yao2022react} to sustained interaction with tools, users, and external environments~\cite{schick2023toolformer,qin2023toolllm,2024patilgorilla,2023farntooltalk,li2023apibank}. Recent benchmarks increasingly test interactive software environments, service workflows, long-horizon planning, diverse tool ecosystems, and dynamic application settings~\cite{liu2023agentbench,zhou2023webarena,rawles2024androidworld,xie2024osworld,barres2025tau,zhang2026deepplanning,he2025vitabench,li2025toolathlon,froger2026gaia2}. Training efforts likewise draw on large-scale tool-use data, executable environments, broad domain coverage, and long-horizon interaction trajectories~\cite{zeng2023agenttuning,xu2025toucan,sun2026agentskiller,dong2026agentworld,wang2026agentworldmodel,bai2026agentsa1}. Yet benchmarks are commonly organized around a limited set of domains, platforms, or interaction protocols. Models with similar aggregate scores can therefore exhibit different strengths across scenarios and capabilities~\cite{shen2026omniabenchbenchmarkinggeneralai}. For example, a model that performs well in one tool ecosystem may still struggle in another with state tracking, constraint maintenance, document and data operations, user clarification, or error recovery. Progress on individual benchmarks alone thus does not establish reliable operation across heterogeneous real-world applications.

We frame this problem as \emph{full-scenario agentic scaling}: systematic and extensible progress across application domains, execution capabilities, task difficulty, and interaction modes. This setting spans three broad application contexts: To-Consumer (ToC), To-Business (ToB), and To-Employee (ToE). Representative settings within this scope include consumer services and app-based workflows~\cite{trivedi2024appworld}; enterprise systems and operational workflows~\cite{drouin2024workarena,huang2024crmarena}; and professional work involving office applications, documents, and spreadsheets~\cite{patwardhan2025gdpval,wang2024officebench,ma2024spreadsheetbench}. Agents in these settings interact with mutable state, domain rules, files, structured data, and users over extended trajectories. Scaling in this regime therefore requires more than collecting additional tool-call traces.

There are three main obstacles. \textbf{(1) Coverage and diagnosis.} Existing datasets and evaluations lack a shared coordinate system for application context, required capabilities, and sources of task difficulty. Data construction, training, evaluation, and subsequent improvement are therefore difficult to align, while aggregate scores provide limited guidance about what should be improved next. \textbf{(2) Scaling environments and tasks.} Constructing executable environments and tasks requires balancing coverage, difficulty, and correctness. Real APIs provide grounded behavior but are costly and restrictive to scale. Recent work has advanced task generation, programmatic environment synthesis, agent world models, graph-based construction, and verified tool-use data~\cite{2025xieagentsynth,song2026envscaler,guo2025genenv,wang2026agentworldmodel,2026chengraph2eval,li2025inftool}. Agent-World, for example, shows that realistic executable environments can be synthesized at scale to support general-agent evolution~\cite{dong2026agentworld}. A remaining challenge is to translate broader environment coverage into diverse and difficult tasks, while also allowing task requirements to drive environment construction or adaptation when the required capabilities are not yet supported. \textbf{(3) Learning from hard failures.} For difficult tasks, imitation may inherit teacher limitations and errors~\cite{gudibande2023false}, while hidden state transitions or globally coupled constraints may require planning beyond unaided language-model rollouts~\cite{hao2023reasoning,zhang2026deepplanning}. An on-policy learner also receives little useful signal when all attempts fail. Once observed, such failures must still be translated into controlled, verifiable objectives for subsequent data construction.

To overcome the problems mentioned above, we present \textbf{AgentOmnia}, a framework for full-scenario agentic scaling. It defines a shared task space that aligns data synthesis, post-training, evaluation, and PRD-guided self-evolution within a unified development loop. We train AgentOmnia-30B-A3B using Qwen3-30B-A3B-Thinking-2507~\cite{yang2025qwen3,qwenteam2025qwen330b2507} as the foundation model. Our evaluation spans OmniaBench, our companion benchmark for full-scenario evaluation, and three public agent benchmarks: $\tau^2$-Bench, DeepPlanning, and VitaBench~\cite{barres2025tau,zhang2026deepplanning,he2025vitabench}. Across all four benchmark families, AgentOmnia improves over its foundation model. Among the evaluated agentic post-training baselines, it obtains the strongest OmniaBench result and the highest four-benchmark macro-average, although recent Qwen3.5-based Agents-A1 and Nex-N2-Mini remain stronger on DeepPlanning~\cite{bai2026agentsa1,nexagi2026nexn2mini}. The OmniaBench results show gains across ToC, ToB, and ToE, with improvements distributed across capability dimensions and atomic-difficulty factors rather than concentrated in a few categories. A preliminary single-round study further indicates that PRD-guided synthesis can better align generated data with diagnosed weaknesses and yield modest aggregate gains. We view this as an initial validation of controllability, while the stability and returns of longer-horizon evolution remain to be studied.

In summary, our main contributions are as follows:
\begin{itemize}
    \item We formulate the problem of full-scenario agentic scaling and introduce an extensible three-axis taxonomy that aligns data construction, training, and diagnosis. We publicly release the companion OmniaBench for community use, providing broad and fine-grained evaluation over this space.
    \item We develop bidirectional environment--task synthesis with stateful executable environments and three complementary task pipelines. Solver-guided and solver-anchored synthesis extend task construction beyond local execution flows to planning and optimization under global constraints.
    \item We present a weak-to-strong synthesis and post-training recipe in which models generate candidates while programs, solvers, state-transition checks, rubrics, and verifiers provide correctness signals. Leakage-controlled privileged guidance supports difficult trajectory generation, while rollback-based curriculum learning recovers useful signals from otherwise all-fail tasks.
    \item We introduce PRD-guided self-evolution, adapting a widely used industrial specification format into a structured protocol that connects evaluation-derived diagnoses to targeted data synthesis while allowing stakeholder requirements to be incorporated through the same protocol.
    \item We train AgentOmnia-30B-A3B and observe broad improvements across the companion diagnostic benchmark and three external agent benchmarks, spanning application scenarios and capability dimensions rather than a single benchmark specialization.
\end{itemize}

\section{Framework Overview}
\label{sec:overview}

\begin{figure}[h]
    \centering
    \includegraphics[width=\linewidth]{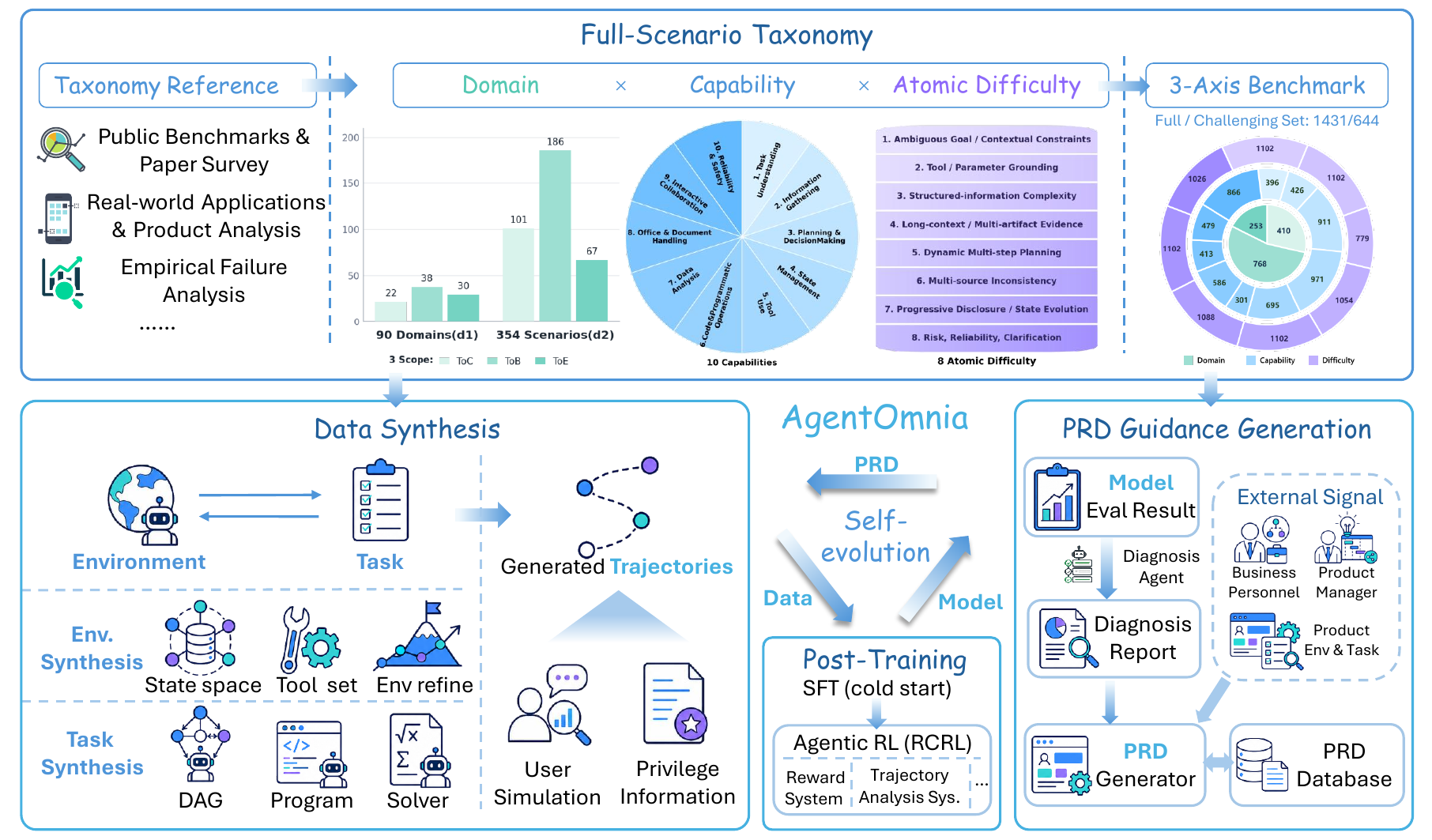}
    \caption{\textbf{Overview of AgentOmnia.} The upper panel presents a full-scenario taxonomy over domains, capabilities, and atomic difficulties, drawing on public research, analyses of real-world applications and products, and observed agent failures. The taxonomy defines the target task space, while OmniaBench measures model performance over that space. The lower-left module constructs executable environments and tasks, then synthesizes verified trajectories. The lower-center module curates verified trajectories into SFT examples; executable tasks and their associated environments support online agentic RL. The right module converts evaluation feedback into PRD-based guidance for the next development cycle; the dashed box denotes a product-facing extension through which external requirements may also be incorporated. Together, the four modules form a closed loop from task-space definition through data synthesis and model improvement to renewed diagnosis.}
    \label{fig:overview}
\end{figure}

Figure~\ref{fig:overview} presents the overall framework of AgentOmnia and the closed development loop formed by its four modules. This section summarizes their roles and interfaces before subsequent sections describe each component in detail.

\paragraph{Full-Scenario Taxonomy.}
At the foundation of AgentOmnia is a Domain $\times$ Capability $\times$ Atomic Difficulty taxonomy. The three axes distinguish where and for whom a task is performed, what the agent must do, and how the task is made difficult. The domain axis organizes ToC, ToB, and ToE scenarios into 90 level-1 and 354 level-2 domains, while the other two axes describe ten capability dimensions and eight atomic difficulty factors that can be composed within a task. Rather than defining a closed list of tasks, these coordinates provide a common indexing layer for data construction, evaluation, and diagnosis. The taxonomy remains extensible: its entries and mappings can be refined as products, interaction environments, and model capabilities evolve. Our companion work, \textbf{OmniaBench}~\cite{shen2026omniabenchbenchmarkinggeneralai}, instantiates this design as a general-agent benchmark with 1,431 tasks and a challenging subset of 644 tasks for cost-efficient evaluation. Its tasks are deduplicated against the AgentOmnia training corpus and manually curated for solvability and evaluation validity. During synthesis, the same coordinates are used to track coverage and specify task difficulty. By reporting model performance along these coordinates, OmniaBench also supports mapping observed failures to capability targets. Its analyses reveal substantial rank variation across scenarios and capabilities, motivating taxonomy-level analysis alongside aggregate benchmark scores. Section~\ref{sec:taxonomy} formalizes the taxonomy and its associated metadata.

\paragraph{Scalable Data Synthesis.}
AgentOmnia instantiates this task space with executable environments and tasks. Building on recent programmatic environment and agentic data-scaling efforts~\cite{2025xieagentsynth,song2026envscaler,xu2025toucan,sun2026agentskiller,dong2026agentworld,wang2026agentworldmodel}, it adopts a bidirectional synthesis paradigm that connects capability supply from environments with capability demand from tasks. The environment-oriented route first constructs an environment and reuses it to generate grounded tasks, amortizing the cost of environment construction. The task-oriented route starts from a task specification and constructs or adapts a supporting environment, broadening the diversity of goals and workflows. We construct code-driven, stateful environments from heterogeneous seeds and validate their initialization, tool behavior, and global state transitions. Task synthesis uses three complementary pipelines: \textbf{DAG-based synthesis} captures tool dependencies and long-horizon workflows; \textbf{program-based synthesis} represents branches, loops, and data-dependent execution; and \textbf{solver-based synthesis} addresses planning and optimization under global constraints through solver-guided and solver-anchored strategies. Across the three pipelines, tasks are retained only when their execution traces, state changes, and evaluation criteria are mutually consistent. The resulting environments and tasks support trajectory generation through direct rollout, user simulation, or privileged guidance. Section~\ref{sec:data-synthesis} describes the synthesis framework in detail.

\paragraph{Weak-to-Strong Synthesis and Post-Training.}
Previous work on \emph{weak-to-strong generalization} asks whether weak supervision can elicit capabilities beyond the supervisor~\cite{burns2023weak}. We use this idea more narrowly to construct reliable training signals when a teacher model cannot consistently solve a task through direct rollout. In this process, language models generate candidate environments, tasks, and trajectories, while programs, solvers, state-transition checks, and structured verifiers provide correctness signals. Privileged planning structures, solver outputs, and rubric constraints can further guide trajectory generation. Only trajectories that pass correctness and groundedness checks and are verified to be leakage-free are retained; privileged content is excluded from both user-facing tasks and retained trajectories. In post-training, verified trajectories are curated into examples for supervised fine-tuning, while agentic reinforcement learning~\cite{bai2024digirl,xi2024agentgym,zhao2025muarl} improves the policy through online rollouts on executable tasks in their associated environments, with rewards computed from task-specific rules and rubrics. For all-fail rollout groups, rollback-based curriculum reinforcement learning resumes exploration from an adaptive prefix of a golden trajectory, using longer prefixes when a task remains too difficult and shorter ones as the policy improves. See Sections~\ref{sec:trajectory-synthesis} and~\ref{sec:posttraining} for details.

\paragraph{PRD-Guided Self-Evolution.}
Previous work has explored model-generated data and feedback~\cite{wang2022selfinstruct,yuan2024selfreward}, instance-level reflection~\cite{shinn2023reflexion,madaan2023selfrefine,lu2023self}, and the evolution of agents or their learning environments~\cite{xi2024agentgym,guo2025genenv,zhai2025agentevolver,hu2026seal,yan2026openskill}. AgentOmnia takes a complementary, product-facing view by repurposing the Product Requirements Document (PRD), a widely used specification artifact in product development, as a structured protocol for model evolution. On the internal path, evaluation failures are mapped to taxonomy coordinates and aggregated into capability-level diagnosis reports. These reports are then translated into PRDs that specify target scenarios, capability gaps, environment semantics, synthesis constraints, and measurable success conditions. On the external path, business stakeholders, product managers, and domain experts can directly provide requirements or supporting product artifacts, such as environment and task specifications. Inputs from both paths are normalized into PRDs, which guide the next round of environment, task, and trajectory construction. This provides an interpretable and traceable interface connecting evaluation evidence, stakeholder requirements, and model development. Section~\ref{sec:prd} details the guidance-generation and self-evolution process.

\section{A Full-Scenario Taxonomy for General Agents}
\label{sec:taxonomy}

\begin{figure}
    \centering
    \includegraphics[width=1\linewidth]{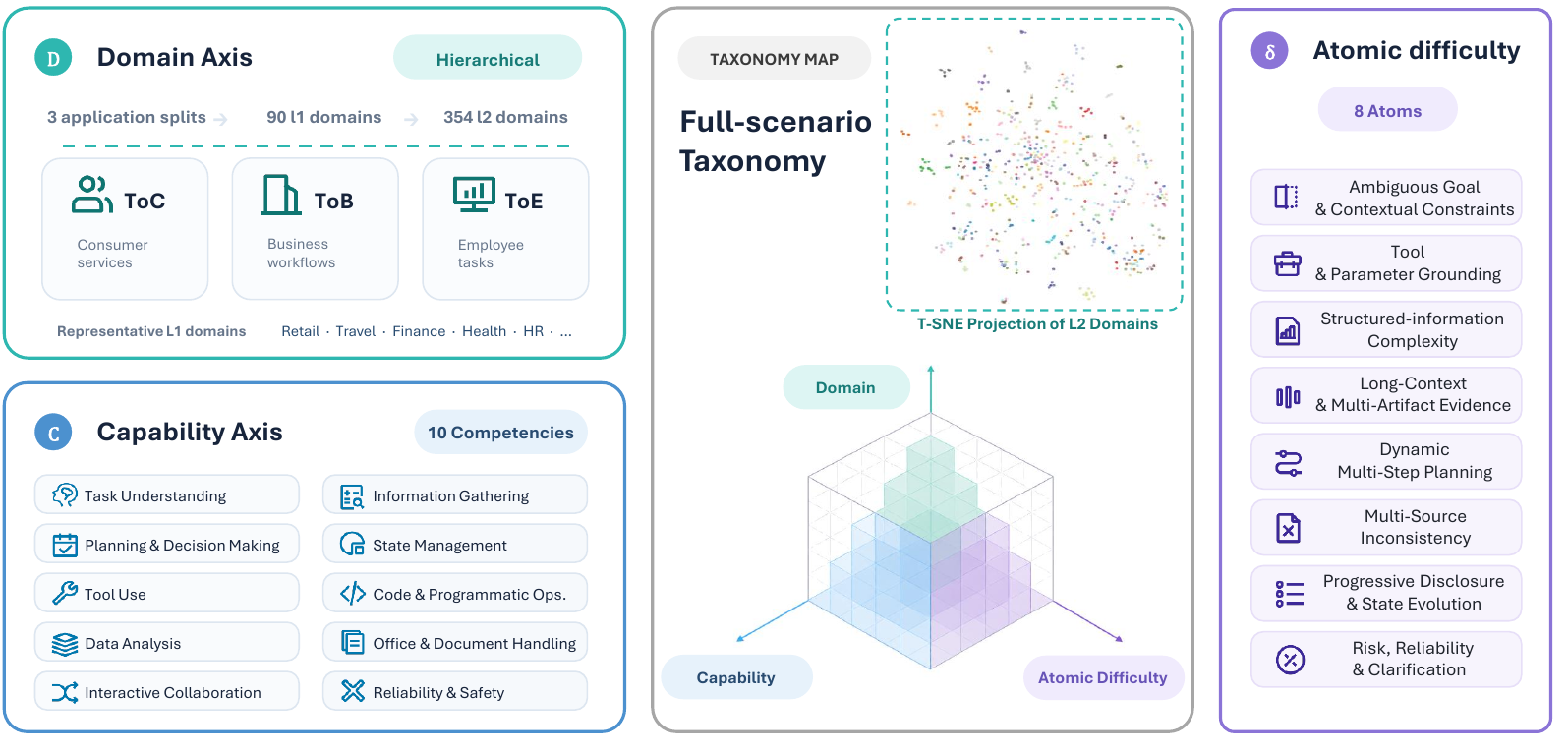}
    \caption{\textbf{Full-scenario taxonomy of AgentOmnia.} The taxonomy jointly indexes tasks by hierarchical domain, capability profile, and compositional atomic difficulty. The inset shows a t-SNE visualization of embeddings of the level-2 domain descriptions.}
    \label{fig:taxonomy_overview}
\end{figure}

The central design choice of AgentOmnia is to define the task space before constructing concrete environments and tasks. A benchmark or training corpus intended to cover the full scenario space should not be organized as a flat collection of tool-call traces, since such an organization provides limited control over coverage and data distribution. Instead, each task should be characterized by the real-world domain it models, the agent capabilities it requires, and the difficulty factors deliberately introduced into its design. Accordingly, we organize the taxonomy as \textbf{a three-axis coordinate system: Domain × Capability × Atomic Difficulty}. Each task, together with its associated environment, trajectory, verifier, and rubric, is indexed by
\begin{equation}
Z = \left(\mathcal{D}, C, \boldsymbol{\delta}\right),
\qquad
\mathcal{D} = \left(s,d_1,d_2\right),
\end{equation}
where \(\mathcal{D}\) denotes the hierarchical domain coordinate, \(C\) denotes the capability profile, and \(\boldsymbol{\delta}\) denotes the atomic difficulty profile.
The domain coordinate \(\mathcal{D}\) consists of an application split \(s\), a level-1 domain \(d_1\), and a level-2 domain \(d_2\). The capability profile \(C\) may include multiple capabilities, while the atomic difficulty profile \(\boldsymbol{\delta}\) may activate multiple atomic difficulties. Capability analyses retain the full multi-label profile, whereas each task designates one primary atomic difficulty for mutually exclusive difficulty statistics.

\subsection{Taxonomy Construction}

We construct the taxonomy along three complementary axes: domain, capability, and atomic difficulty. For the \textbf{domain axis}, we organize application settings into ToC, ToB, and ToE. For ToC, category systems and functional descriptions collected from major app stores are decomposed into executable user-facing domains. ToB is grounded in representative industries and occupational tasks from GDPval~\cite{patwardhan2025gdpval}, supplemented by standard industrial classification schemes. For ToE, recurring employee activities drawn from representative industries and workplace templates are abstracted into industry-general domains, such as reporting, project coordination, travel arrangements, data analysis, and administrative operations. Model-assisted organization and human review are jointly used to split overly broad categories, merge redundant entries, normalize naming, align hierarchical granularity, and validate split assignments.

The \textbf{capability axis} is manually designed with reference to representative agent benchmarks~\cite{barres2025tau,li2025toolathlon,he2025vitabench} and existing formulations of agent abilities. It contains ten dimensions covering task understanding, information gathering, planning and decision making, state management, tool use, code and programmatic operations, data analysis, office and document handling, interactive collaboration, and reliability and safety. The \textbf{atomic difficulty axis} is derived from an analysis of internal single-turn and multi-turn datasets, focusing on their interaction patterns, execution trajectories, tool dependencies, and failure conditions. This analysis yields eight reusable difficulty factors that can be compositionally assigned to tasks.

The resulting taxonomy contains 90 level-1 and 354 level-2 domains: 22/101 for ToC, 38/186 for ToB, and 30/67 for ToE, alongside ten capability dimensions and eight compositional difficulty factors. Across all three axes, semantic analysis and expert review are used to refine and validate the taxonomy. Together, the domain hierarchy, capability dimensions, and difficulty factors provide a unified structure for data construction, coverage analysis, and fine-grained diagnosis. For qualitative inspection of the domain hierarchy, we additionally visualize the t-SNE embeddings of level-2 domain descriptions to identify semantic clusters, local overlaps, and potential outliers (Figure~\ref{fig:taxonomy_overview}).

\begin{table}[t]
\centering
\small
\setlength{\tabcolsep}{4pt}
\renewcommand{\arraystretch}{1.10}
\caption{Overview of the full-scenario taxonomy structure.}
\label{tab:split-taxonomy}
\begin{tabularx}{\linewidth}{@{}l l X c@{}}
\toprule
\textbf{View} & \textbf{Type} & \textbf{Source / Basis} & \textbf{Scale} \\
\midrule
\multirow{4}{*}{Domain}
& ToC & Multiple app stores and consumer-facing app categories & 22 L1 / 101 L2 \\
& ToB & GDPval-style~\cite{patwardhan2025gdpval} tasks and industry classifications & 38 L1 / 186 L2 \\
& ToE & Industry-general employee tasks from GDPval and industry templates & 30 L1 / 67 L2 \\
\cmidrule(lr){2-4}
& Total & Normalized real-world domain taxonomy & 90 L1 / 354 L2 \\
\midrule
Capability
& Dims. & General-agent execution requirements & 10 dims. \\
\midrule
Atomic Difficulty
& Factors & Eight challenge factors across user, environment, tool use, and interaction & 8 factors \\
\bottomrule
\end{tabularx}
\end{table}

\subsection{Domain Axis}

The domain axis \(\mathcal{D}=(s,d_1,d_2)\) specifies the application context and target actor of an agent task. The split \(s\) distinguishes three complementary settings. \textbf{ToC} covers consumer-facing applications and life-service scenarios, including shopping, travel, booking, payments, personal scheduling, and after-sales services. These tasks typically involve user preferences, service policies, resource search, temporal and budget constraints, multi-turn clarification, and exception handling. \textbf{ToB} covers industry-specific business systems and operational workflows, such as finance, procurement, manufacturing, logistics, inventory, customer management, and IT operations. These tasks emphasize structured business entities, domain rules, cross-system state, multi-step dependencies, data verification, and workflow completion. \textbf{ToE} covers industry-general employee activities, including email, calendar, documents, reporting, project coordination, approval processes, reimbursement, knowledge management, and data analysis. These tasks focus on common workplace tools, organizational processes, multi-artifact handling, collaboration, and deliverable quality. Within each split, \(d_1\) identifies a broad domain, while \(d_2\) refines it into a concrete domain that can support environment and task construction.

A level-2 domain is retained only when it can be grounded in an executable and stateful setting. Specifically, each domain should admit identifiable entities and attributes, mutable or queryable states, operational constraints, and a meaningful set of agent actions. We therefore associate each domain with concise metadata describing its domain path, representative workflows, state objects, business rules, typical operations, and real-world references. This criterion prevents the domain axis from degenerating into a collection of topical labels and ensures that every taxonomy entry can support realistic interactions, state transitions, and verifiable task execution.

\begin{table}[t]
\centering
\small
\setlength{\tabcolsep}{4pt}
\renewcommand{\arraystretch}{1.10}
\caption{\textbf{Capability dimensions.} Each task may involve multiple dimensions, which are retained jointly for multi-label diagnosis.}
\label{tab:capability-axis}
\begin{tabularx}{\linewidth}{@{}p{0.31\linewidth}X@{}}
\toprule
\textbf{Capability} & \textbf{What it evaluates} \\
\midrule

Task Understanding
& Identifying user goals, implicit requirements, priorities, domain constraints, and expected outcomes. \\

Information Gathering
& Locating, retrieving, and integrating relevant evidence from environment states, tools, files, databases, and external information sources. \\

Planning \& Decision Making
& Decomposing goals, selecting execution strategies, respecting dependencies and constraints, and revising plans as new observations become available. \\

State Management
& Tracking intermediate progress and maintaining consistency across entities, environment states, and long or multi-turn trajectories. \\

Tool Use
& Selecting appropriate tools, constructing valid arguments, interpreting outputs, and coordinating multiple tool calls. \\

Code \& Programmatic Operations
& Writing and executing code for computation, data transformation, automation, file manipulation, and programmatic task completion. \\

Data Analysis
& Filtering, aggregating, comparing, reconciling, and reasoning over structured or semi-structured data. \\

Office \& Document Handling
& Reading, extracting, editing, merging, validating, and producing documents, spreadsheets, presentations, and other file-based artifacts. \\

Interactive Collaboration
& Requesting missing information, clarifying ambiguous goals, confirming actions, incorporating user feedback, and coordinating across interaction turns. \\

Reliability \& Safety
& Detecting and recovering from failures, maintaining constraint compliance, avoiding unsafe or invalid actions, and completing tasks robustly under uncertainty. \\

\bottomrule
\end{tabularx}
\end{table}

\subsection{Capability Axis}
\label{sec:capability-axis}

The capability axis describes the core abilities required for an agent to complete a task, independently of the domain in which the task is instantiated. Let
\begin{equation}
\mathcal{C}=\{c_1,\ldots,c_{10}\}
\end{equation}
denote the set of ten capability dimensions defined in
Table~\ref{tab:capability-axis}. Each task is assigned a non-empty subset of capability dimensions \(C\subseteq\mathcal{C}\). The full profile is retained for multi-label coverage and diagnosis. This formulation reflects the compositional nature of general-agent tasks: for example, completing a procurement request may jointly require task understanding, information gathering, planning, tool use, and state management.

The capability taxonomy is defined at a level that remains applicable across different domains and interaction modes. It separates understanding and information acquisition from downstream planning and execution, distinguishes state tracking from general tool use, and treats document processing, data analysis, and coding as separate operational abilities. Interactive collaboration is modeled explicitly because realistic agents must handle tool failures, incomplete information, user feedback, and changing requirements rather than merely follow a fixed, successful trajectory.

The same capability can be instantiated across different domains and under different atomic difficulty profiles, allowing the taxonomy to distinguish capability deficiencies from domain-specific or difficulty-specific failures.

\subsection{Atomic Difficulty Axis}
\label{sec:difficulty-axis}

\begin{table}[t]
\centering
\small
\setlength{\tabcolsep}{4pt}
\renewcommand{\arraystretch}{1.20}
\caption{\textbf{Atomic difficulty axis.} Multiple atoms may be composed within a single task.}
\label{tab:difficulty-atoms}
\begin{tabularx}{\linewidth}{@{}p{0.35\linewidth}X@{}}
\toprule
\textbf{Atomic Difficulty} & \textbf{Instantiation} \\
\midrule

Ambiguous Goal and Contextual Constraints
& The request is underspecified, indirect, or conditioned on implicit preferences, policies, priorities, or professional constraints. \\

Tool and Parameter Grounding
& The agent must distinguish similar or redundant tools, infer arguments from context, or request missing parameters before execution. \\

Structured-information Complexity
& The environment contains numerous structured entities, attributes, relations, or records that must be filtered, joined, compared, or reconciled. \\

Long-context and Multi-artifact Evidence
& Relevant evidence is distributed across long tool outputs, documents, files, attachments, logs, or multiple heterogeneous artifacts. \\

Dynamic Multi-step Planning
& Completion requires long chains of dependencies, conditional branches, intermediate decisions, or replanning after new observations. \\

Multi-source Inconsistency
& Information from users, tools, files, or environment states is incomplete, duplicated, outdated, or mutually conflicting. \\

Progressive Disclosure and State Evolution
& Critical information or constraints are revealed gradually through user turns, tool results, approval stages, or state transitions. \\

Risk, Reliability, and Clarification
& The task involves irreversible actions, insufficient evidence, conflicting instructions, or operations that require explicit confirmation, recovery procedures, or refusal. \\

\bottomrule
\end{tabularx}
\end{table}

The atomic difficulty axis characterizes how a task becomes challenging, independently of its domain and required capabilities. Rather than estimating difficulty solely from trajectory length, tool-call count, or model performance, we represent each case with an explicit atomic difficulty profile
\begin{equation}
\boldsymbol{\delta}
=
(\delta_1,\ldots,\delta_8)
\in \{0,1\}^{8},
\end{equation}
where \(\delta_i=1\) indicates that the \(i\)-th atomic difficulty is present in the task by design. A task may activate multiple atoms simultaneously. We additionally
designate one active atom as its primary difficulty for mutually
exclusive coverage statistics, while retaining the complete profile
\(\boldsymbol{\delta}\) for compositional difficulty analysis.

Atomic difficulties describe properties of the request, environment, tool space, information structure, and interaction protocol. They are therefore distinct from the capability axis: for example, document processing is an agent capability, whereas long-context and multi-artifact evidence specifies the conditions under which that capability is tested. Similarly, data analysis denotes an ability, while structured-information complexity controls the amount, organization, and relational complexity of the information that must be analyzed.

Atomic difficulties can be combined into reusable profiles for task construction. For instance, one task may combine ambiguous goals and contextual constraints, tool and parameter grounding, dynamic multi-step planning, and multi-source inconsistency, while another task with the same domain and capability coordinates may instead introduce long-context and multi-artifact evidence. This design allows tasks with the same domain and capability requirements to vary systematically along the difficulty axis, supporting fine-grained diagnosis of agent failures.

Beyond the three primary axes, each task retains lightweight auxiliary metadata, including its interaction mode and specialized execution setting. These fields support data filtering and analysis but do not constitute additional taxonomy dimensions. Each task and its associated trajectory, rubric, and verifier share a task-level coordinate, while environments are linked to the coordinates they support. This common indexing layer enables consistent coverage measurement and failure diagnosis throughout the data life cycle, including synthesis, training, and evaluation. The taxonomy itself remains extensible, allowing its domains and mappings to be refined as product requirements, interaction environments, and model capabilities evolve.

\section{Data Synthesis Framework}
\label{sec:data-synthesis}

AgentOmnia organizes agentic data synthesis around three complementary components: environments, tasks, and trajectories. For environment construction, we introduce code-driven environments that offer greater scalability and execution stability than approaches based on real API invocation or LLM-based simulation. For task construction, we design three synthesis strategies—DAG-based synthesis, program-based synthesis, and solver-based synthesis—to broaden the coverage of tasks with diverse reasoning structures. For trajectory construction, we leverage task-specific privileged guidance to generate reliable trajectories for post-training.

\subsection{Data Synthesis Overview}
 
\begin{figure}[t]
    \centering
    \includegraphics[width=\linewidth]{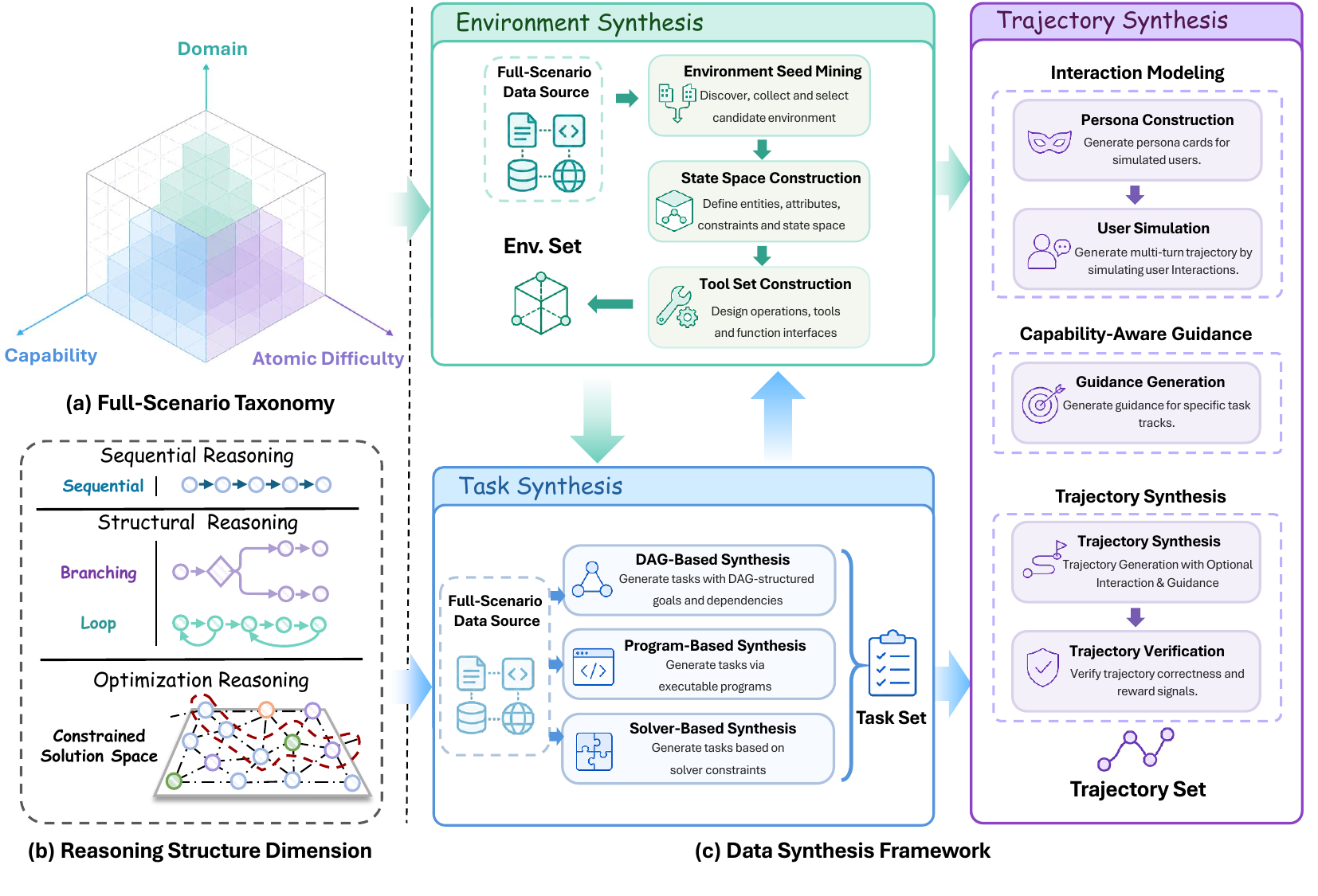}
    \caption{
        \textbf{Overview of the AgentOmnia synthesis pipeline.}
        Environment synthesis constructs diverse executable interaction spaces,
        task synthesis creates executable tasks with different reasoning structures,
        and trajectory synthesis generates verifiable trajectories for agent post-training.
    }
    \label{fig:synthesis-concept-main}
\end{figure}

Figure~\ref{fig:synthesis-concept-main} illustrates the AgentOmnia synthesis pipeline, which supports bidirectional synthesis between environments and tasks. \textbf{Environment synthesis} (Section~\ref{sec:environment-synthesis}) involves constructing stateful interaction spaces and tool repositories, while \textbf{Task synthesis} (Section~\ref{sec:task-synthesis}) generates tasks across a reasoning spectrum encompassing: (1) \textit{sequential reasoning}, requiring the management of multi-step tool dependencies and intermediate cognitive operations; (2) \textit{structural reasoning}, involving complex control flows such as conditional branching and loops; and (3) \textit{optimization reasoning}, where agents must navigate intricate constraints to achieve global objectives. These reasoning structures are operationalized via three synthesis paradigms: DAG-based, program-based, and solver-based synthesis. \textbf{Trajectory synthesis} (Section~\ref{sec:trajectory-synthesis}) generates interaction data through environment rollouts for post-training.

Before describing these components, we first introduce the basic notation used throughout this section. An interactive environment is defined as $E = (\mathcal{S}, T)$, where $\mathcal{S}$ denotes the state space and $T$ denotes the set of executable tools. Each individual tool is denoted by $t \in T$. Given the current state $s \in \mathcal{S}$ and the tool arguments $x$, executing tool $t$ produces an updated state $s' \in \mathcal{S}$ and an observation $o$, formally written as $t(s, x) = (s', o)$.

An executable task is denoted by $\tau = (E, s_0, D, R)$, where $E$ is the associated environment, $s_0 \in \mathcal{S}$ is the initial state, $D$ denotes the task description, and $R$ denotes the success criterion used to evaluate task completion. Executing a task $\tau$ in environment $E$ produces an interaction trajectory $\xi = \left(s_0, t_1, o_1, s_1, \ldots, t_l, o_l, s_l\right)$, where $t_i \in T$ is the tool executed at the $i$-th interaction step, $o_i$ is the corresponding observation, and $s_i$ is the resulting environment state. The trajectory $\xi$ records the complete interaction process from the initial state $s_0$ to the final state $s_l$.

\subsection{Interactive Environment Synthesis}
\label{sec:environment-synthesis}

\begin{figure}[t]
\centering
  \includegraphics[width=0.97\linewidth]{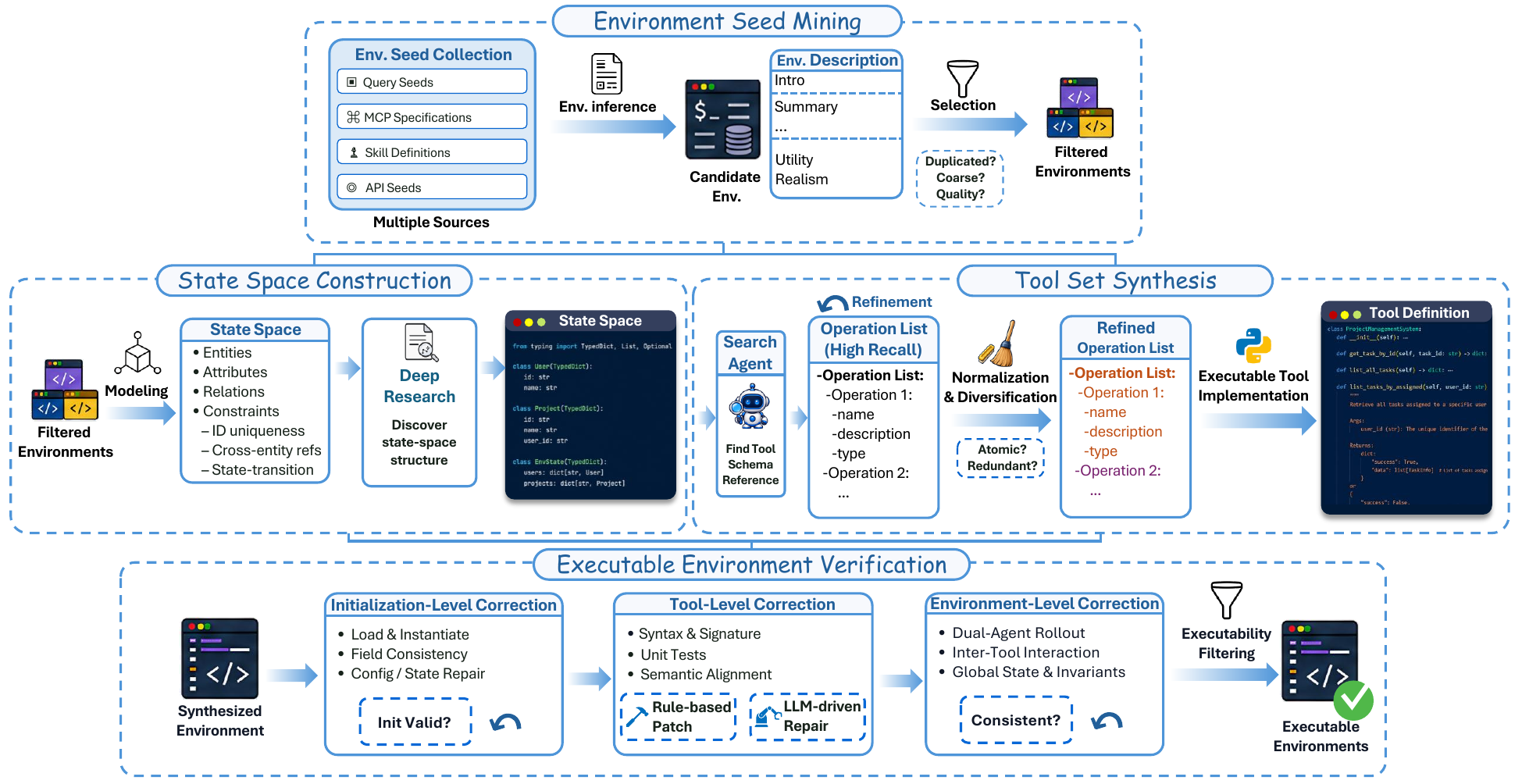}

\caption{\textbf{Overview of the proposed interactive environment synthesis pipeline.} The pipeline consists of the four sequential stages: environment seed mining, state space construction, tool set synthesis, and environment executability verification.}
\label{fig: env_synthesis_pipeline}
\end{figure}

An interactive environment $E$ is a programmatic interaction space that encapsulates a persistent state space $\mathcal{S}$ and a set of executable tools $T$, enabling agents to perform actions and receive verifiable feedback. Building upon prior research~\cite{song2026envscaler,cai2025autoforge,fang2026towards,wang2026agentworldmodel}, the proposed interactive environment synthesis pipeline, as illustrated in Figure~\ref{fig: env_synthesis_pipeline}, systematically transforms heterogeneous seeds into executable environments $E$ through four sequential stages. First, \textbf{Environment Seed Mining} identifies potential domains and selects representative environments according to their practical utility and realism. Second, \textbf{State Space Construction} formalizes the entities, attributes, and structural constraints that define the environment's persistent state space $\mathcal{S}$. Third, \textbf{Tool Set Synthesis} generates and refines a set of executable tools $T$ where each tool $t \in T$ is semantically grounded in the established state space $\mathcal{S}$. Finally, \textbf{Environment Executability Verification} validates the reliability and behavioral consistency of the synthesized environments $E$ via rigorous checks.

\subsubsection{Environment Seed Mining}

Environment seeds serve as foundational blueprints, providing the essential context and domain knowledge required for the synthesis process. To establish a robust and comprehensive basis for the pipeline, we curate an extensive and diverse pool of candidate seeds, thereby maximizing both domain coverage and scenario variety.

\paragraph{Candidate Environment Discovery.}

The discovery of candidate environments is conducted through a systematic pipeline consisting of two sequential phases: environment seed collection and environment inference.

\begin{itemize}

\item \textbf{Environment seed collection.}
Environment seed data is harvested from a variety of sources and formats, as detailed in Table~\ref{tab:envgen_data_sources}. Specifically, we aggregate \textbf{Query Seeds}, \textbf{Skill Definitions}, \textbf{MCP Specifications}, and \textbf{API Seeds} from established web repositories and ecosystems. These complementary corpora provide the requisite domain breadth and realism for subsequent synthesis stages.

\item \textbf{Environment inference.}
By analyzing the compiled seed data, we extract the essential features and technical requirements of the target environments. For every candidate seed, our inference module produces a brief summary and a detailed overview, accompanied by quantitative scores for utility and realism. The summary categorizes the domain, while the overview specifies the persistent state components, typical operations, and the broader functional objectives associated with the environment.

\end{itemize}

\begin{table}[t]
    \centering
    \caption{Summary of environment seed sources.}
    \label{tab:envgen_data_sources}
    \begin{tabular}{p{0.168\linewidth} p{0.5562\linewidth} p{0.12\linewidth}}
        \toprule
        \textbf{Seed Category} & \textbf{Source / Description} & \textbf{Quantity} \\
        \midrule
        Query Seeds 
        & User queries and intents adapted from public web sources, covering diverse user needs and interaction scenarios.
        & 197K \\

        Skill
        & Skill specifications describing high-level tool capabilities and their intended usage contexts.
        & 20K \\

        MCP
        & MCP-style specifications collected from public MCP registries, including tool definitions and interface descriptions.
        & 2.3K \\

        API Seeds 
        & Real-world tool seeds derived from existing APIs, reflecting practical tool functionalities and invocation patterns.
        & 1.3K \\
        \bottomrule
    \end{tabular}
\end{table}

\paragraph{Candidate Environment Selection.}

The discovered environment seeds exhibit substantial heterogeneity in both practical utility and structural complexity. Accordingly, we perform a quality-aware selection procedure to curate candidates prior to the construction of executable environments.

\begin{itemize}

\item \textbf{Granularity control.}
A primary design consideration is the granularity of the environment. Overly expansive environments (e.g., generic enterprise management systems) present challenges in modeling coherent state spaces and tool boundaries, whereas overly specialized environments often support only isolated task instances. We therefore prioritize environments that are sufficiently broad to encompass a family of related tasks while remaining sufficiently constrained to admit well-defined entities, operations, and executable logic.

\item \textbf{Quality scoring.}
Each candidate environment is evaluated across two primary dimensions: \textit{utility} and \textit{realism}. \textit{Utility} assesses the functional richness and task-solving potential, ensuring the environment provides sufficient affordances to accommodate diverse and complex user objectives. In contrast, \textit{realism} examines the structural integrity and domain-specific fidelity, verifying that the internal logic, entity relationships, and operational constraints remain strictly consistent with established real-world practices.

\item \textbf{Deduplication and selection.}
A multi-stage procedure is used to remove semantic overlap. We perform exact deduplication on summaries, retaining candidates with superior realism scores, and filter out environments below minimum quality thresholds. Finally, remaining candidates are clustered using latent text embeddings, with the medoid environment selected as the representative.

\end{itemize}

\subsubsection{State Space Construction}
\label{sec:state-space}

State-space construction formalizes the persistent memory of each synthesized environment $E$ by formulating a structured state specification $\mathcal{S}$. This specification defines the entities maintained within the environment, their associated attributes, inter-entity relations, and the structural constraints governing valid states and legal transitions. As the semantic foundation of the executable environment $E$, this representation enables a unified lifecycle: initialization generates valid states $s \in \mathcal{S}$, tool synthesis operates over these states, and dynamic verification evaluates whether tool executions $t(s, x)$ induce correct state transitions to $s'$.

\paragraph{Knowledge-Augmented State Generation.}
For each candidate environment, we prompt an LLM with the environment information to generate a high-recall state-space specification. Rather than generating task-specific minimal schemas, the model leverages two-round deep research to mine potential state spaces and infer reusable domain-level entities representative of the target real-world system. To enhance realism, the generated state space is refined with external domain knowledge, eliminating implausible fields while incorporating commonly used entities and attributes.

\paragraph{Executable State Compilation.}
The resulting state specification is compiled into executable Python state containers through an interleaved process of generation, loading, and validation. Entities and attributes are mapped to persistent dictionary-like attributes and structured field definitions. By concurrently loading and checking the generated code, we identify and correct syntactically invalid implementations, ensuring that downstream tool synthesis originates from a verified executable environment representation.

\subsubsection{Tool Set Synthesis}
\label{sec:tool-set}

Following the construction of the executable state space, we synthesize the tool interface through which agents interact with the environment. This phase involves generating executable operations grounded in the state representation, refining the action space into a compact and composable toolset, and finally compiling the resulting operations into callable Python implementations with standardized interfaces.

\paragraph{Tool Generation.}

We construct candidate toolsets by deriving operations from the environment information and state-space specification. Each operation is defined by its name and description, categorized as either a state-querying or state-modifying action. To ensure breadth, the model is prompted to generate a comprehensive range of reusable interactions across major entities, relations, and state transitions.

\paragraph{Tool Refinement.}

Before implementation, the generated operation space is refined with the aim of enhancing its overall coverage, composability, and behavioral consistency.

\begin{itemize}

\item \textbf{Validator-guided refinement.}
An LLM-based validator evaluates whether the generated tools adequately cover the environment $E$, remain grounded in the synthesized state space $\mathcal{S}$, eliminate redundancy, and facilitate complex multi-step interactions. Operations that fail to satisfy one or more of these evaluation criteria are iteratively regenerated based on the feedback provided by the validator.

\item \textbf{Operation normalization and diversification.}
Validated operations undergo normalization to ensure a consistent representation: redundant operations are merged, overly broad functions are decomposed, and unsafe updates or superficial shortcut tools are removed. To ensure traceability from the raw operation space to the final executable toolset $T$, each candidate operation is explicitly tracked as \texttt{kept}, \texttt{merged}, \texttt{split}, \texttt{rewritten}, or \texttt{removed}. 
Furthermore, to increase the evaluative challenge, we move beyond atomic tools with single-argument inputs and scalar outputs, and systematically diversify the granularity complexity of the synthesized tools. Specifically, we incorporate multi-branch functions conditioned on mode-selector arguments, multi-argument inputs with inter-parameter constraints, and structured multi-field outputs requiring downstream field extraction. We also introduce operations with overlapping naming or parameter structures but divergent functional logic, such that correct tool selection demands reasoning over functional semantics rather than lexical matching.

\end{itemize}

\paragraph{Executable Tool Implementation.}

The normalized operations are compiled into executable Python methods within the environment class. All tools adhere to a unified return protocol that standardizes responses and error handling, streamlining behavior verification. Tools are iteratively regenerated until they pass unit tests and satisfy the state space specifications. Finally, executable tool schemas are extracted to provide standardized interfaces for agent interaction.

\subsubsection{Executable Environment Verification}
\label{sec:env-verification}

\begin{wraptable}[11]{r}{0.45\textwidth}
\vspace{-20pt}
\centering
\caption{Statistics of the synthesized environments.}
\label{tab:dataset_overview}

\small
\setlength{\tabcolsep}{4pt}
\renewcommand{\arraystretch}{1.18}

\begin{tabularx}{\linewidth}{
    @{}
    >{\raggedright\arraybackslash}X
    >{\centering\arraybackslash}p{0.27\linewidth}
    @{}
}
\toprule
\textbf{Metric} & \textbf{Value} \\
\midrule
Total Environments          & 5,018         \\
Domain Categories (L1)      & 90            \\
Domain Subcategories (L2)   & 354           \\
Tools (total)                & 255,375       \\
Tools (mean $\pm$ std)       & $50.9 \pm 13.4$ \\
Entities (mean $\pm$ std)    & $13.3 \pm 3.9$  \\
Attributes (mean $\pm$ std)  & $77.3 \pm 22.5$ \\
Attrs / Entity               & 5.9           \\
\bottomrule
\end{tabularx}

\end{wraptable}

To ensure reliability, synthesized environments undergo a multi-level self-correction mechanism designed to address failures identified at three hierarchical layers, followed by rigorous executability filtering.

\paragraph{Initialization-Level Correction.} 
At the initialization level, we focus on resolving failures that occur during the loading and instantiation of the environment $E$. If instantiation fails or triggers field-level consistency errors, the system iteratively refines the configuration or the environment's internal state logic, seeking to obtain a valid starting state $s_0 \in \mathcal{S}$ for all subsequent interactions.

\paragraph{Tool-Level Correction.} 
At the tool level, the system addresses both syntactic and functional failures within individual tool $t \in T$. This involves applying rule-based patches for deterministic errors, such as missing imports or API signature mismatches. Furthermore, we implement LLM-driven functional alignment: when automated rollouts detect that a tool's execution trace $t(s, x) = (s', o)$ deviates from its semantic specification, the error trace and tool code are fed back into a repair model to realign the implementation with its intended behavior.

\paragraph{Environment-Level Correction.} 
At the environment level, the system fixes global inconsistencies that arise from complex inter-tool interactions. Even if individual tools $t$ pass unit tests, their combined execution may lead to invalid global states $s \notin \mathcal{S}$ or broken relational invariants. 

\begin{figure}[t]
    \centering
    \includegraphics[width=\textwidth]{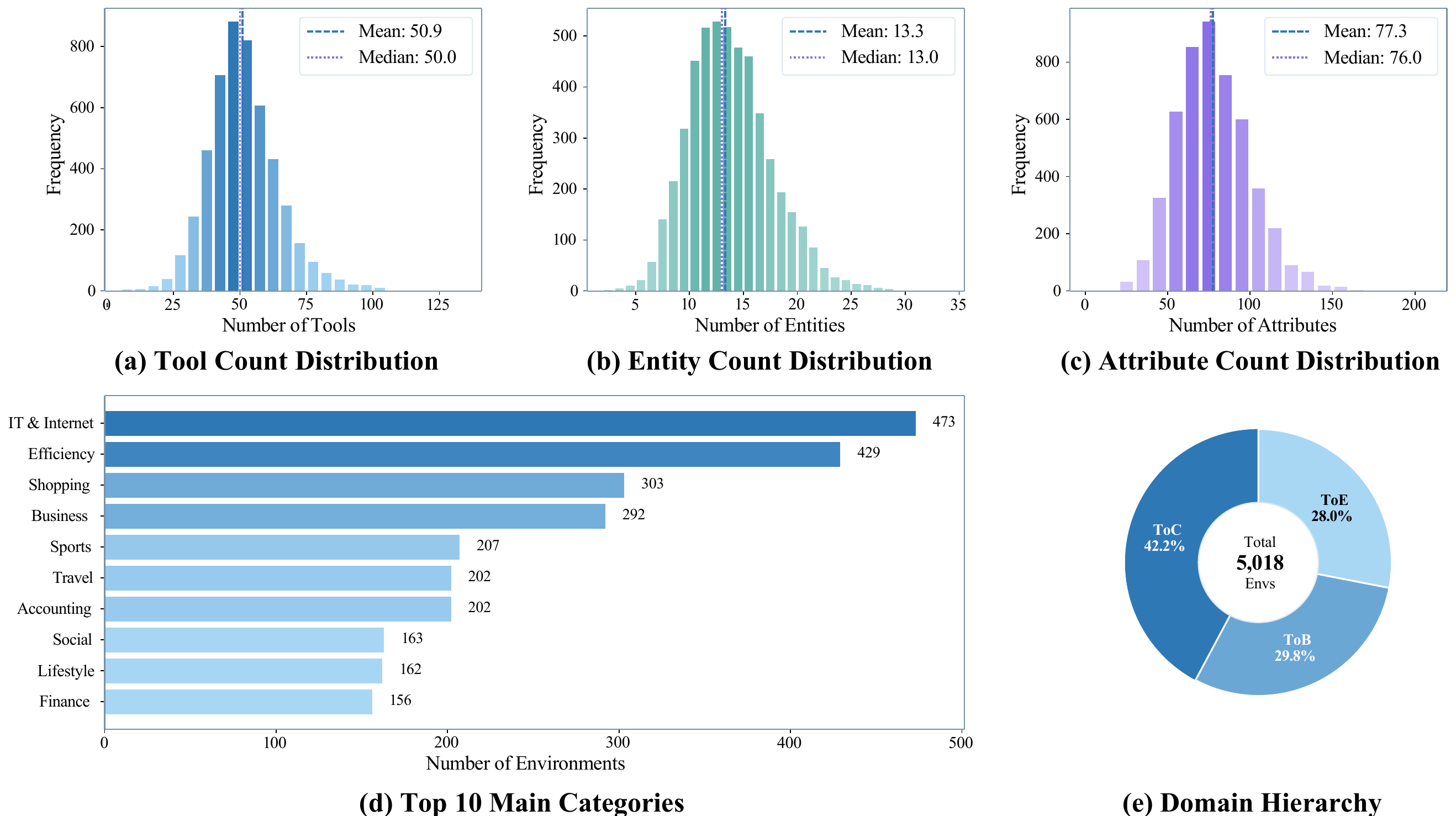}
    \caption{Statistics of the synthesized environments following the verification and filtering process.}
    \label{fig:env_datacard}
\end{figure}

\paragraph{Executability Filtering.}
Finally, we apply a rigorous executability filtering process to maximize the yield of high-quality data. Components that remain non-executable or logically inconsistent after several repair attempts are discarded. An environment $E$ is finalized and retained only if its constituent tools $t$ and global transitions $t(s, x) = (s', o)$ successfully pass the aforementioned verification. The statistical profile of the environments synthesized through this pipeline is presented in Table~\ref{tab:dataset_overview} and Figure~\ref{fig:env_datacard}.

\subsection{Environment Sandbox}
\label{sec:sandbox}

AgentOmnia incorporates a sandbox infrastructure for large-scale agentic post-training. The system orchestrates isolated, stateful environment instances within user space, enabling deterministic resets and high-concurrency execution to support complex, multi-step agentic workflows.
The architecture comprises three hierarchical layers: cluster-level scheduling, execution interfaces, and runtime management. A centralized scheduling gateway dispatches requests to worker nodes, which instantiate environments from predefined configurations and manage their end-to-end lifecycles.

\begin{itemize}
    \item \textbf{Cluster-level scheduling.}
    The gateway dynamically distributes instances based on real-time resource occupancy. It employs admission control mechanisms to regulate throughput and coordinate resource allocation across concurrent requests during peak execution loads.
    
    \item \textbf{Runtime-level isolation.}
    This layer comprises an environment loader and a runtime manager. The loader implements on-demand loading to minimize resource overhead, while the manager maintains in-memory instances to facilitate rapid state resets and efficient resource reclamation.
    
    \item \textbf{Environment interface.}
    A unified API provides a high-level abstraction for heterogeneous environments. The sandbox employs a standardized protocol to ensure seamless integration with downstream training pipelines.
\end{itemize}

\subsection{Task Synthesis}
\label{sec:task-synthesis}

Task construction synthesizes executable task instances through three paradigms, including DAG-based synthesis (Section~\ref{sec:DAG-based-synthesis}) for long-horizon tool-use tasks, program-based synthesis (Section~\ref{sec:program-based-synthesis}) for structured executable programs, and solver-based synthesis (Section~\ref{sec:solver-based-synthesis}) for optimization tasks with verifiable solutions. Across these three paradigms, we synthesize 52,361 tasks: 45,855 DAG-based, 2,204 program-based, and 4,302 solver-based.

\begin{figure*}[t]
\centering
\vspace{0pt}
\includegraphics[width=\linewidth]{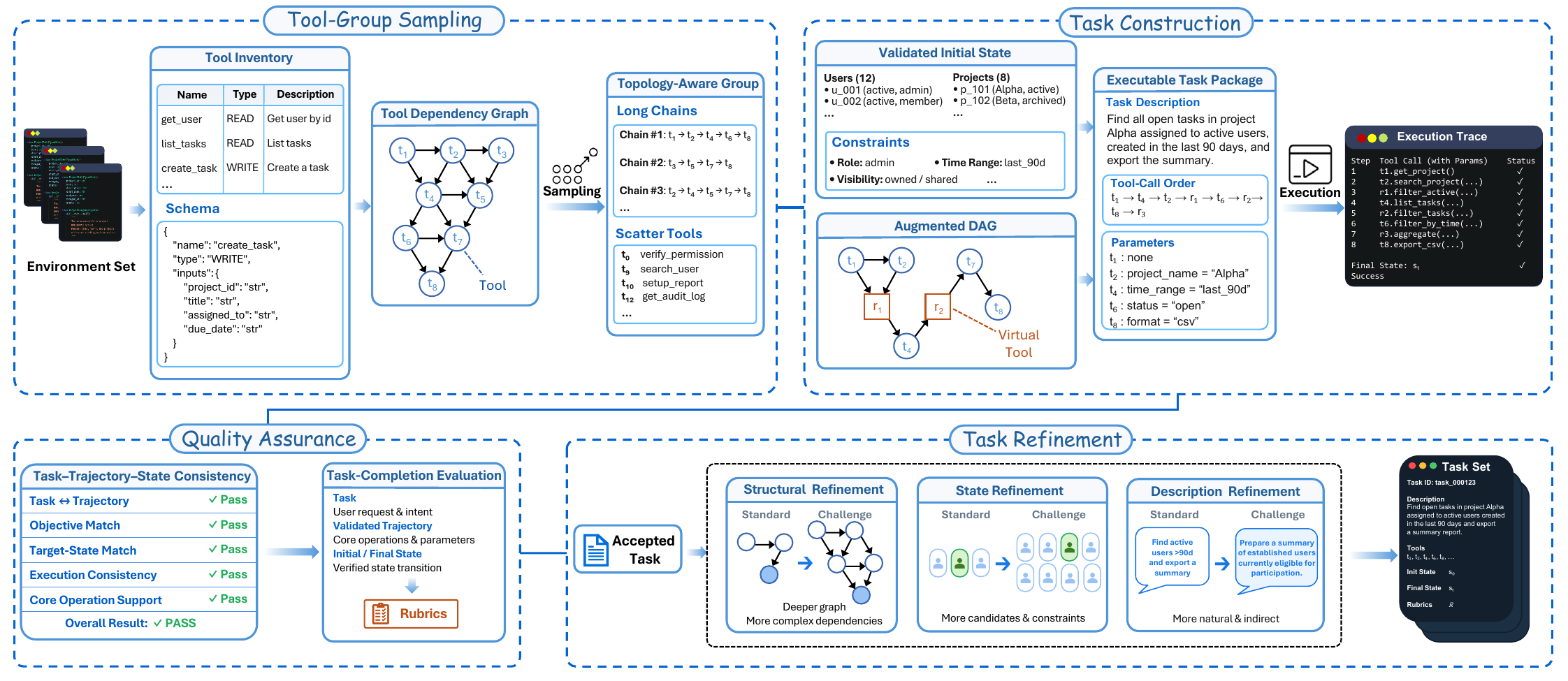}
\caption{Overview of the DAG-based task synthesis pipeline.}
\label{fig:dag-synthesis}
\end{figure*}

\subsubsection{DAG-Based Task Synthesis}
\label{sec:DAG-based-synthesis}

As shown in Figure~\ref{fig:dag-synthesis}, the pipeline consists of tool-group sampling, task construction, quality assurance, and task refinement. Given a set of tools, the pipeline first models pairwise dependencies among them and samples topology-aware groups that combine multiple long tool chains with scattered tools. It then inserts virtual tools to convert each sampled group into an augmented DAG. The DAG representation allows tool calls to branch, merge, and share prerequisites without imposing a single linear execution order. Based on these DAGs and their initial states, the framework generates candidate tasks. To support scalable scenario expansion across a broad range of difficulty levels, we apply different refinement strategies to construct two task categories. DAG-Standard emphasizes broad coverage and cost-effective generation, whereas DAG-Challenge strengthens structural and state complexity to increase execution difficulty. This separation allows the same synthesis backbone to support large-scale coverage expansion and the targeted construction of harder tasks without changing the underlying environments or tool interfaces. The resulting dataset contains 43,851 DAG-Standard tasks and 2,004 DAG-Challenge tasks.

\paragraph{Tool-Group Sampling.}

Tool-group sampling identifies relations among tools and uses them to construct compatible long chains. It then combines these chains with scattered tools that can support auxiliary operations, producing diverse groups with different dependency structures.

\begin{itemize}

\item \textbf{Tool dependency graph.}
For each environment, we construct a directed graph over the available tools to represent candidate dependencies between tool pairs. The graph includes three types of dependencies: \textbf{parameter dependency}, \textbf{entity-anchor dependency}, and \textbf{state-transition dependency}. A parameter dependency indicates that the output of an upstream tool can supply a value required by a downstream tool. An entity-anchor dependency represents the identification or disambiguation of an entity. A state-transition dependency connects an operation that changes the environment state to a subsequent tool that reads or depends on the resulting state. Candidate dependencies are proposed from tool descriptions, operation types, and input schemas, and are then deterministically filtered to remove self-loops, duplicates, unsupported relation types, and low-confidence candidates. The resulting graph captures plausible tool dependencies rather than simple tool co-occurrence.

\item \textbf{Topology-aware group sampling.}
The sampler uses the dependency graph to construct a pool of long tool chains through depth-first search. The search begins from tools with no incoming dependencies or with many outgoing dependencies. Random walks provide additional variation in the starting tools and chain lengths. Each sampled group contains several long tool chains and a set of scattered tools. The long chains provide dependency-compatible multistep structures, while the scattered tools can support auxiliary queries, verification, state inspection, or plausible distractions. To limit redundancy, we compute the Jaccard overlap between each candidate group and previously retained groups and discard groups with excessive overlap.

\end{itemize}

\paragraph{Task Construction.}

For each sampled tool group, task construction inserts virtual tools to form an augmented DAG with an initial state, and generates candidate task descriptions that are subsequently validated through execution.

\begin{itemize}

\item \textbf{Augmented DAG.}
For each sampled tool group, we insert virtual tools to connect relevant scattered tools with the long tool chains, forming a unified dependency structure. Each virtual tool belongs to one of seven predefined types: COMPUTE, LOGIC, EXTRACT, TRANSFORM, AGGREGATE, VALIDATE, or FILTER. Table~\ref{tab:node-types} summarizes their functions. Virtual tools represent intermediate operations needed to integrate scattered tools into the dependency structure, but they cannot be called during execution. The resulting DAG contains tools from the long chains, scattered tools, virtual tools, and the directed dependencies among them. We preserve the original order of tools within each long chain and insert virtual tools only along directions consistent with the existing dependency structure. This construction introduces no backward edges with respect to the original topological order, ensuring that the augmented graph remains acyclic and admits a dependency-consistent topological ordering.

\begin{table}[t]
\centering
\small
\setlength{\tabcolsep}{1pt}
\renewcommand{\arraystretch}{1.15}
\caption{Node types and their functional descriptions.}
\label{tab:node-types}

\begin{tabularx}{0.72\columnwidth}{
    @{}
    >{\raggedright\arraybackslash\ttfamily}p{0.15\columnwidth}
    >{\raggedright\arraybackslash}X
    @{}
}
\toprule
\textnormal{\textbf{Node Type}}
& \textbf{Description} \\
\midrule

COMPUTE
& Perform numerical computations or derive required intermediate values. \\

LOGIC
& Evaluate rules, logical conditions, or decision criteria. \\

EXTRACT
& Extract specific fields or values from an output. \\

TRANSFORM
& Convert data between formats or types. \\

AGGREGATE
& Combine multiple outputs into a unified result. \\

VALIDATE
& Check prerequisites, constraints, or validity conditions before execution. \\

FILTER
& Select a subset of results based on specified criteria. \\

\bottomrule
\end{tabularx}
\end{table}

\item \textbf{Initial-state construction.}
The state generator uses the environment class, state containers, and domain description to propose an initial state $s_0$. It checks the proposed state for invalid fields, type mismatches, illegal enum values, missing containers, and inconsistent cross-object references, and filters out any invalid state. Valid states can subsequently be augmented with additional records and constraints, but every augmented state must pass the same initialization checks before it is used for task generation.

\item \textbf{Execution-grounded task generation.}
Given the augmented DAGs, scattered tools, and a validated initial state $s_0$, the generator uses topological orderings of the DAGs as structural guidance to produce a task description $D$, an ordered sequence of tool calls, and the arguments for each call. The system executes these calls sequentially and records the resulting interaction trajectory $\xi$. A candidate task description is retained only when every call succeeds and the trajectory reaches a valid final state $s_l$. The trajectory serves as execution evidence for task validation and rubric construction, and is not used for post-training.

\end{itemize}

\paragraph{Quality Assurance.}

We retain only candidates whose task descriptions, execution trajectories, and resulting state changes are mutually consistent, and derive an outcome-focused rubric from the validated execution evidence.

\begin{itemize}

\item \textbf{Task-trajectory-state consistency.}
Successful execution alone does not guarantee that the resulting trajectory fulfills the task description. We therefore check whether the task description is consistent with both the interaction trajectory $\xi$ and the state transition from $s_0$ to $s_l$. Auxiliary calls, such as search, read, validation, and state inspection, need not be explicitly mentioned in the description. However, the core operations performed along the trajectory and the resulting state changes must be explicitly requested or logically implied. A candidate is rejected if the trajectory targets a different objective, performs an unsupported operation, or modifies an unrelated object.

\item \textbf{Task-completion evaluation.}
We construct the rubric $R$ from the task description, initial state, validated trajectory, and final state. These elements determine the task-specific outcomes, state changes, and constraints that the rubric must evaluate. The rubric assesses whether a downstream agent fulfills the task description and its associated constraints without requiring it to reproduce the reference order of tool calls. The environment, initial state, task description, and validated rubric jointly define the executable task $\tau$.

\end{itemize}

\paragraph{Task Refinement.}

As shown in Figure~\ref{fig:dag-synthesis}, task refinement adjusts the DAG structure, initial state, and task description to produce standard and challenge variants. The refinement process is integrated into the synthesis pipeline rather than applied as a separate post-processing stage.

\begin{itemize}
\item \textbf{Structural refinement.} During group sampling, we vary the number and depth of long tool chains as well as the number of scattered tools. These changes increase the length of the required tool-call sequence and the complexity of the dependencies that the agent must resolve.

\item \textbf{State refinement.} We enrich an existing valid initial state with additional candidate entities, eligibility or permission constraints, threshold conditions, historical evidence, and conflict-resolution cases. These additions introduce plausible distractors and make it more difficult to identify the entities and conditions within the enriched state that are directly relevant to the task description.

\item \textbf{Description refinement.} We rewrite the task description in a more natural and indirect form while removing explicit execution prompts. This makes the description less procedural and requires the agent to infer the necessary operations and constraints from the request.

\end{itemize}

\begin{figure}[h]
    \centering
    \includegraphics[width=1\linewidth]{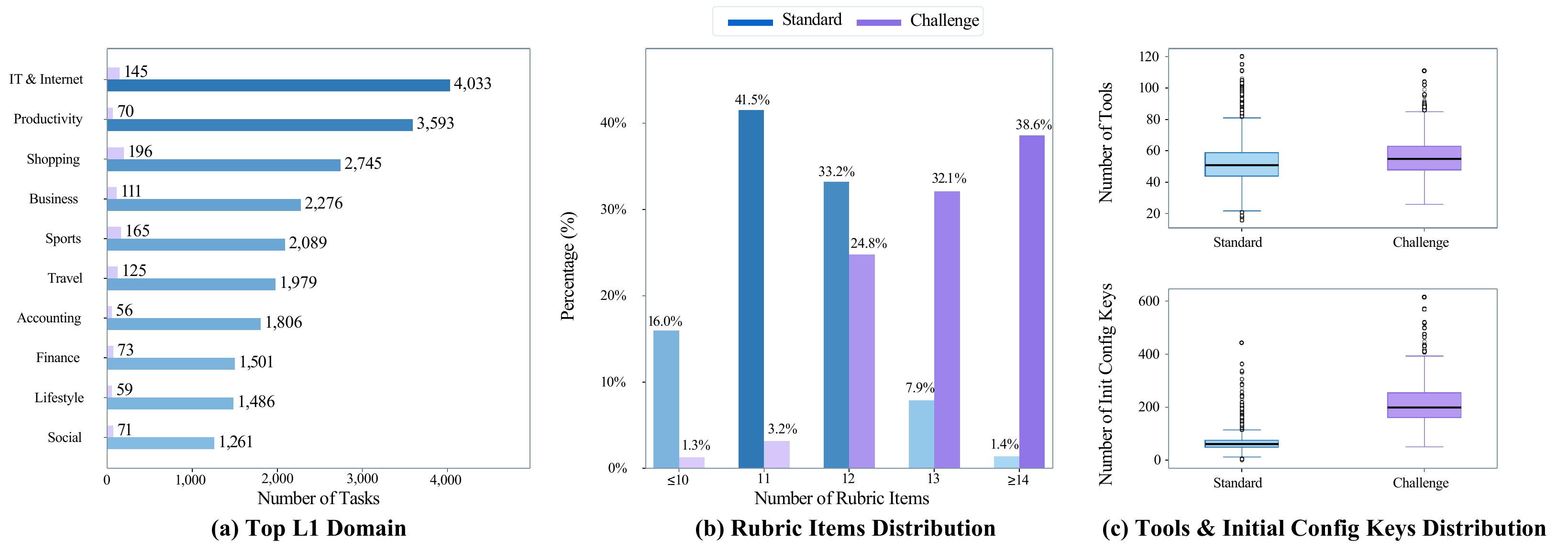}
    \caption{Statistics of the synthesized DAG tasks.}
    \label{fig:DAG_datacard}
\end{figure}

Together, these controls increase structural complexity, state ambiguity, and linguistic indirectness without weakening initialization, execution, consistency, or evaluation checks. We use different refinement profiles to construct two task variants, DAG-Standard and DAG-Challenge. Their statistics are reported in Figure~\ref{fig:DAG_datacard}.

\begin{figure*}[t]
\centering
\vspace{0pt}
\centering
\includegraphics[width=\linewidth]{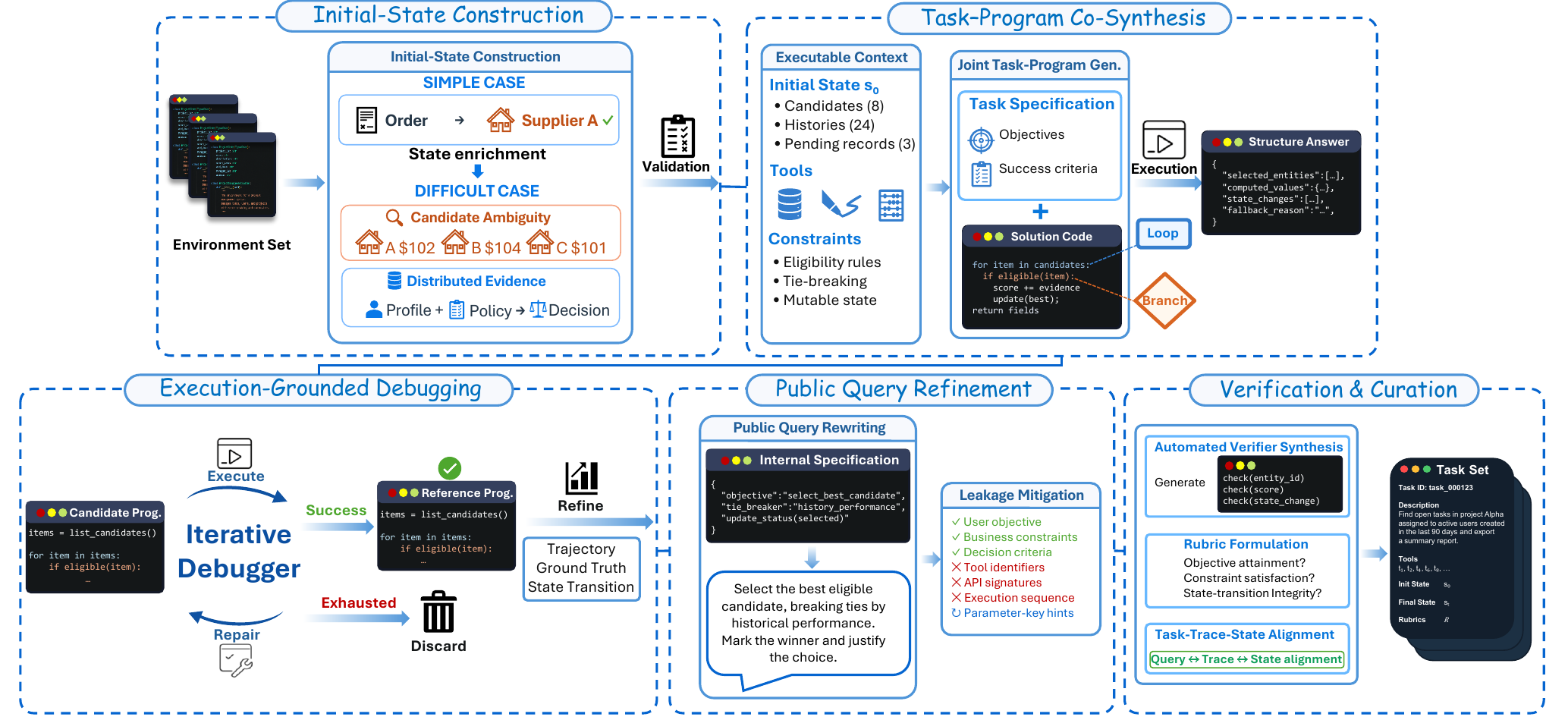}
\caption{Overview of the program-based task synthesis pipeline.}
\label{fig:program-synthesis}
\end{figure*}

\subsubsection{Program-Based Task Synthesis}
\label{sec:program-based-synthesis}

Building upon the methodology established in~\cite{dong2026agentworld}, our program-based synthesis framework generalizes DAG-based synthesis by incorporating iterative and conditional logic. The framework utilizes a pipeline that integrates state-aware initialization with joint task-program co-synthesis, ensuring that synthesized tasks $\tau$ are both structurally intricate and semantically consistent with the environment $E$. Using this framework, we construct 2,204 program-based tasks that require iterative and conditional execution. These tasks extend beyond static DAG structures by introducing loops, branching decisions, and state-dependent control flow, thereby supporting more complex and realistic agent interactions.

\paragraph{Initial-State Construction.}
Prior to synthesizing program-based tasks, we construct an executable initial state for each environment that strictly adheres to its specification while maintaining sufficient structural complexity to facilitate non-trivial execution. As illustrated in Figure~\ref{fig:program-synthesis}, this process commences with difficulty-oriented state synthesis, where initialization difficulty is structured across two dimensions: \textbf{candidate ambiguity}, which precludes direct target identification through distractors and secondary decision rules; \textbf{distributed evidence}, which necessitates the aggregation of task-relevant information across multiple entities and records. The concrete operators employed to instantiate these dimensions are detailed in Appendix~\ref{app:program_init_state}. Subsequently, each candidate state undergoes executable initialization validation via a conformance test against the target environment interface. States failing this test are either repaired or discarded, ensuring that all downstream tasks are grounded in states that are both structurally challenging and fully executable within the target environment implementation.

\paragraph{Task-Program Co-Synthesis.}

Building upon the validated initial state and tool set, we jointly synthesize an internal task specification and a corresponding solution program. While the specification formalizes the objective and success criteria, the program encodes the procedural logic required for completion. 

\begin{itemize}
    \item \textbf{Joint task-program generation.} 
    To support high structural complexity, the program incorporates control structures such as loops for iterative processing and conditional branches for state-dependent decision making. Once the program is generated, the internal task specification is derived from its logic, ensuring that the observable objectives and constraints are intrinsically linked to an executable solution.
    
    \item \textbf{Structured answer specification.} 
    To facilitate automated verification, task-relevant execution results are aggregated into structured output fields. These fields capture selected entities, computed values, and state-transition outcomes, along with justifications for any fallback decisions. Taken together, they define what information is expected in the response and serve as the ground truth used to construct both programmatic verifiers and fine-grained evaluation rubrics.
\end{itemize}

\paragraph{Execution-Grounded Program Debugging.}

Synthesized programs are treated as candidate solutions, with task specifications remaining provisional until validation within the target environment. These candidates may exhibit failure modes such as syntax errors, schema-inconsistent tool invocations, invalid arguments, or erroneous control logic. To address these, an iterative debugging framework refines programs based on environmental feedback. 

\begin{itemize}
    \item \textbf{Iterative repair.}
    Candidate programs are executed within the target environment. In each iteration, runtime errors, tool outputs, and execution traces are recorded. These observations, combined with the environment description, initial state, and tool set, are used to diagnose failures. A rectified program is then generated for the subsequent iteration. This cycle continues until the program executes successfully or the repair budget is exhausted. Unresolved programs are excluded from the synthesis pipeline.

    \item \textbf{Execution-grounded task refinement.} Following verification, the program is designated as a reference and the task specification is finalized. Re-execution from the initial state generates a canonical trajectory and structured ground-truth response. These artifacts are used to refine the task description, ensuring consistency with the verified execution dynamics and environment state.
\end{itemize}

\paragraph{Public Query Refinement.}

Internal task specifications generated during synthesis often contain details specific to the implementation that are unsuitable for downstream training and evaluation. Following reference program verification and the finalization of supervision grounded in execution, the task is distilled into a query intended for public use. This process preserves core objectives and constraints while removing metadata related to implementation details.

\begin{itemize}
    \item \textbf{Public query rewriting.} The internal specification is reformulated into a natural language request. This transformation preserves semantic integrity by eliding references exclusive to the synthesis phase, ensuring the query remains consistent with the verified execution trajectory.

    \item \textbf{Leakage mitigation.} We remove artifacts specific to the implementation that could leak the reference solution, such as tool identifiers, API signatures, and explicit execution sequences. Additionally, hints within parameter keys are replaced with descriptive natural language to prevent models from relying on superficial pattern matching or heuristics based on tool names.
\end{itemize}

\begin{figure}[t]
    \centering
    \includegraphics[width=\textwidth]{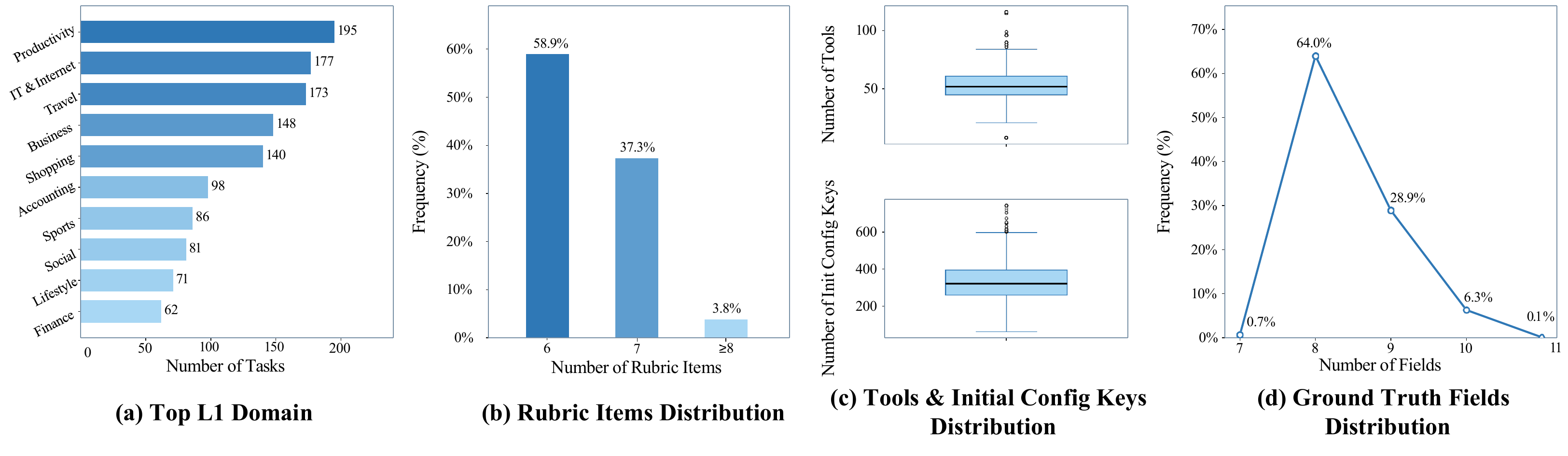}
    \caption{Statistics of the synthesized program-based tasks.}
    \label{fig:program_datacard}
\end{figure}

\paragraph{Verification and Curation.}
The verification and curation process consists of multiple stages to produce execution-grounded tasks. This pipeline integrates deterministic validation of structured outputs, rubric-based assessment, and semantic consistency checks across queries, reference executions, and state transitions. Specifically, \textbf{automated verifier synthesis} generates code for task-critical fields for deterministic assessment. To address criteria beyond field-level validation, \textbf{rubric formulation} derives task-specific rubrics from the query, initial state, and execution trajectory. Furthermore, \textbf{task-trace-state alignment} checks semantic consistency between the query and reference execution, removing tasks that diverge from environment behavior. Statistics for the resulting program-based tasks retained after these verification and curation stages are summarized in Figure~\ref{fig:program_datacard}.

\subsubsection{Solver-Based Task Synthesis}
\label{sec:solver-based-synthesis}

\begin{figure}
    \centering
    \includegraphics[width=1\linewidth]{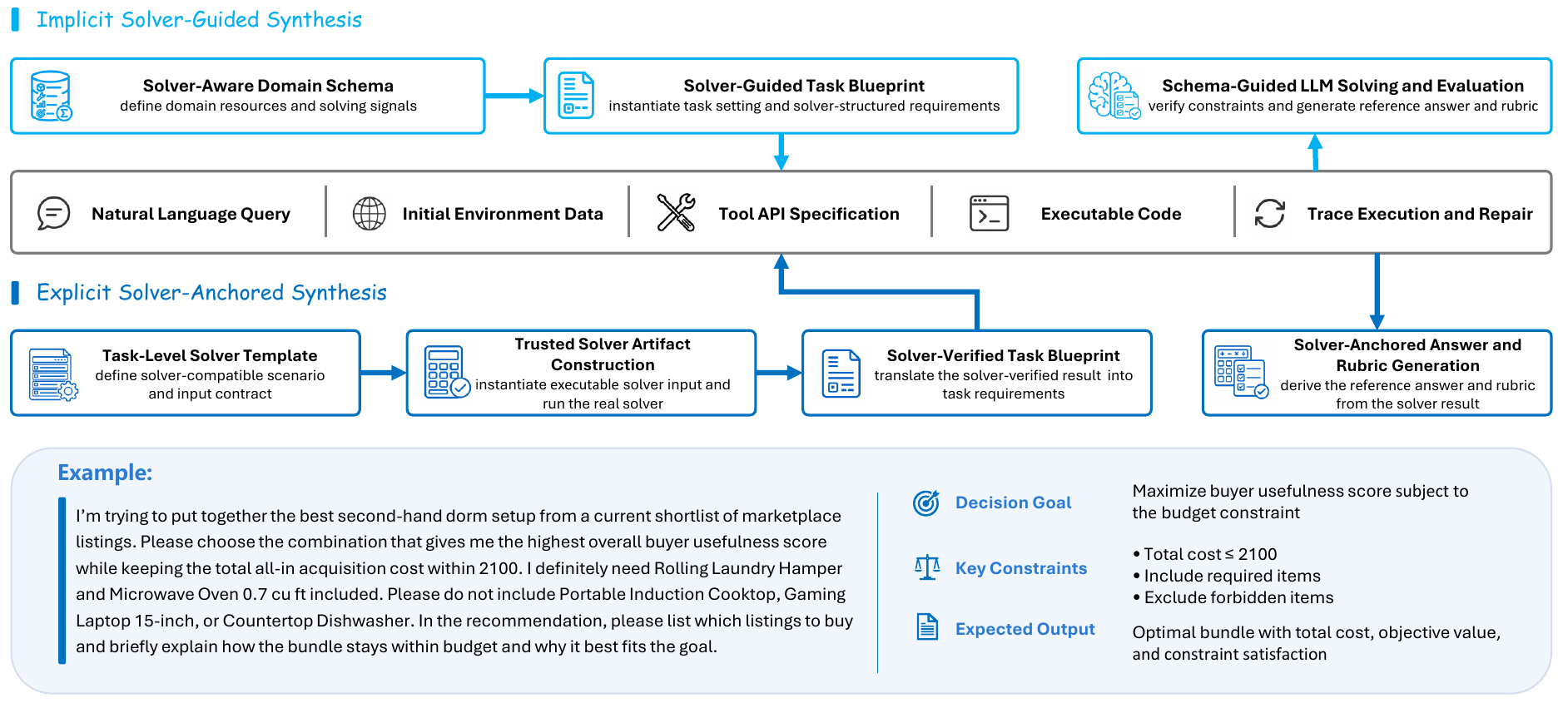}
    \caption{Overview of the solver-based task synthesis pipeline.}
    \label{fig:solver_overview}
\end{figure}

To systematically construct tasks that elicit the complex reasoning capabilities of agentic models, we adopt a solver-based task synthesis paradigm. This paradigm leverages the decision variables, constraints, and optimization objectives inherent in solvers to design agentic problems that require multi-step information gathering, constraint checking, candidate comparison, and objective optimization. Inspired by the task-first environment synthesis principle in Agent World Model~\cite{wang2026agentworldmodel}, we adopt a task-oriented synthesis process: we first generate a structured task specification $D$, and then construct the hidden environment, tool interfaces, executable environment, ground truth, and evaluation rubric $R$ around that task. This paradigm is mainly applicable to domains involving selection, allocation, scheduling, and planning, especially when the underlying tasks contain resource constraints or explicit optimization objectives. Based on this paradigm, we implement two strategies, as illustrated in Figure~\ref{fig:solver_overview}, that differ in how solving is performed. Implicit Solver-Guided Synthesis encodes solver structure into a \textbf{solver-aware domain schema and task blueprint}, but relies on \textbf{the LLM to perform solving and verification}. In contrast, Explicit Solver-Anchored Synthesis \textbf{executes a real solver during synthesis} and uses the resulting solver-verified artifact to anchor subsequent task construction and evaluation. After the task specification is obtained, the tasks synthesized by both strategies are further instantiated through common agentic task construction steps, including query generation, initial environment data construction, tool API specification, executable code generation, and trace execution and repair. Using these two strategies, we construct 1,848 tasks through Implicit Solver-Guided Synthesis and 2,454 tasks through Explicit Solver-Anchored Synthesis.

\paragraph{Implicit Solver-Guided Synthesis.}

In this strategy, we represent solver structure implicitly through a solver-aware domain schema and task blueprint, without executing a domain-specific solver. The ground truth and rubric are therefore produced through schema-guided LLM solving rather than real solver execution. This design makes the strategy easier to extend across domains, but provides weaker optimality guarantees because the reference solution still depends on LLM reasoning.

\begin{itemize}
\item \textbf{Solver-aware domain schema.}
We first perform solver-aware structured modeling of the target domain and construct a reusable domain-level schema. This schema provides a unified description of the domain semantics, business entities involved in decision-making, available environment resources and tool capabilities, as well as the solving signals used for constraint verification and objective computation. Through this schema, subsequent task blueprints can be instantiated around decision variables, constraints, and optimization objectives.

\item \textbf{Schema-guided LLM solving and evaluation.}
As the final stage of the implicit strategy, the LLM generates the ground truth and evaluation rubric according to a predefined solving and verification schema. It performs constraint checking, candidate comparison, objective-value calculation, and feasibility verification based on the task blueprint and environment data returned by tools. Since optimality judgments still rely on the LLM, the rubric does not treat the reference solution as absolutely optimal. Therefore, the rubric allows a candidate answer to pass if it satisfies the explicit constraints, is consistent with the environment data returned by tools, and produces a solution that is equivalent to the reference solution or achieves a better objective value.

\end{itemize}

\begin{figure}[t]
    \centering
    \includegraphics[width=\textwidth]{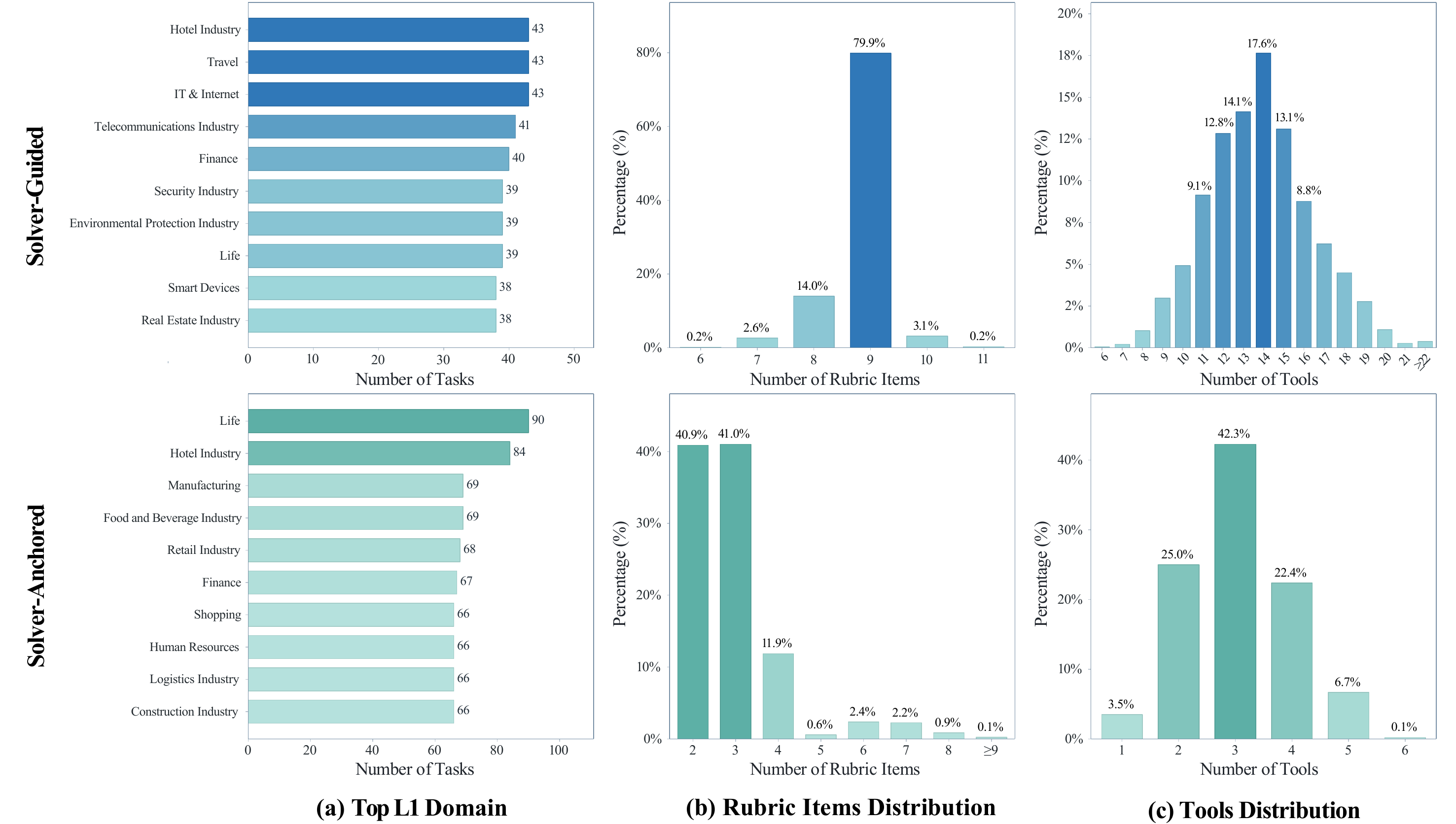}
    \caption{Statistics of the synthesized solver-based tasks.}
    \label{fig:solver_datacard}
\end{figure}

\paragraph{Explicit Solver-Anchored Synthesis.}

In this strategy, we prepare a real executable solver in advance and invoke it during synthesis to construct a trusted solver artifact that anchors the task blueprint, environment, ground truth, and rubric. Because the reference answer is derived directly from solver execution, the strategy offers greater accuracy and reproducibility, along with stronger optimality guarantees. The trade-off is the upfront implementation and adaptation of the solver required for each target task type, which increases the engineering effort needed to extend the approach to new domains. Once implemented, the solver can be reused across tasks that share the same input contract. This reuse also helps maintain consistent execution and verification across independently generated instances.

\begin{itemize}
\item \textbf{Task-level solver template.}
We construct a task-level template for the current task that is aligned with the input structure of the real solver. We select a solver that matches the business scenario and define the task scenario, user role, and solver input contract, specifying the fields required by the solver and their business meanings. The template also specifies the scale, value ranges, and feasibility conditions for subsequent input generation. It does not generate concrete data, the query, environment, tools, or answer; instead, it provides a structured foundation for executable solver inputs.

\item \textbf{Trusted solver artifact construction.}
We instantiate the solver input contract into a solver input instance executable by the real solver, and invoke the corresponding solver to solve and verify the instance. The instance must conform to the solver’s input format and contain a complete set of constraints and a meaningful comparison space. The real solver then produces the solving status, objective value, and optimal solution, forming a trusted solver artifact that anchors the subsequent task blueprint, environment, ground truth, and rubric.

\item \textbf{Constraint- and objective-based task blueprint.}
We convert the solver-verified result into a task blueprint expressed in business semantics, specifying explicit constraints, implicit constraints, and optimization objectives. Each constraint and objective is traced back to the input fields of the real solver and filled with concrete values or thresholds to provide a clear solving basis. This blueprint provides the task specification for subsequent generation of the query, environment, tools, ground truth, and rubric.

\item \textbf{Solver-anchored answer and rubric generation.}
We generate the reference answer and evaluation rubric from the solver-verified trusted solving record. The reference answer is consistent with the real solver result and explains the final decision, objective value, and satisfaction of explicit constraints in business language. The rubric converts explicit constraints and optimization objectives into self-contained judge rules. It evaluates the candidate answer by checking whether its final decision, key metrics, and objective value match or are equivalent to the solver result, rather than assessing the solving process or tool-call path.

\end{itemize}

Together, these solver-based synthesis strategies introduce decision variables, constraints, and explicit optimization objectives into agentic tasks while maintaining verifiable ground truth and evaluation criteria. Dataset statistics for the resulting
solver-based tasks are summarized in Figure~\ref{fig:solver_datacard}.

\subsection{Trajectory Synthesis}
\label{sec:trajectory-synthesis}

\begin{figure}
    \centering
    \includegraphics[width=\linewidth]{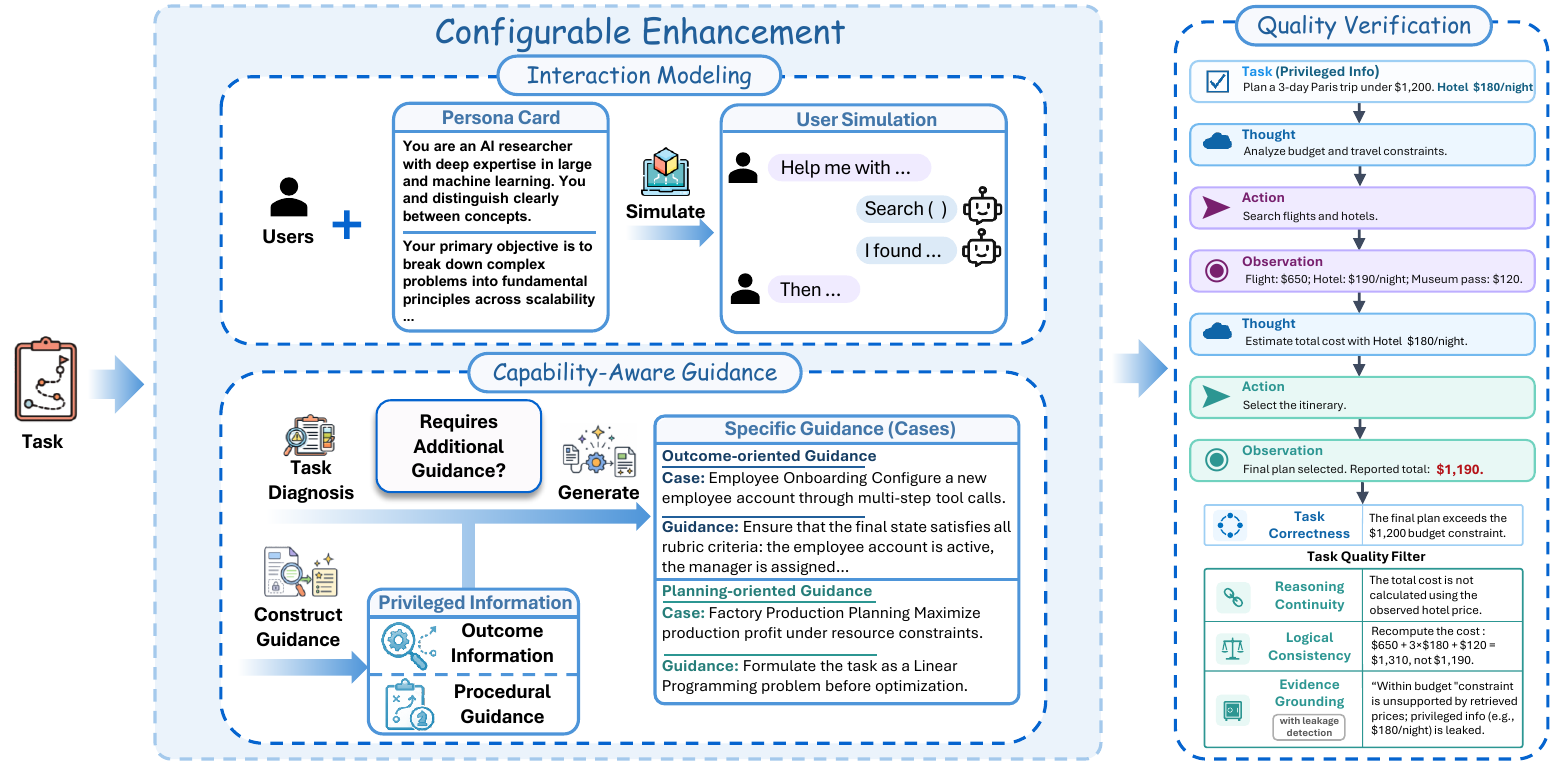}
    \caption{Overview of the trajectory synthesis pipeline.}
    \label{fig:trajectory_overview}
\end{figure}

After the task and environment synthesis stages, we obtain a diverse set of executable tasks with corresponding environments and evaluation systems, covering a broad range of task complexities. Based on these executable tasks, we further synthesize interaction trajectories for agent training. In this section, we describe our trajectory synthesis framework, which generates trajectories that satisfy task objectives and execution constraints.

For tasks with simple execution patterns, direct rollout with frontier models is often sufficient to generate valid trajectories. However, synthesizing high-quality training trajectories for complex tasks requires addressing two complementary requirements: First, real-world interactions involve diverse user behaviors, preferences, communication patterns, and partially specified intentions. To improve the coverage and diversity of synthesized trajectories, existing approaches have explored interaction modeling strategies to simulate diverse user behaviors and interaction contexts. Second, even for a fixed user task, long-horizon execution and complex decision-making remain challenging for current models. Agents may struggle to maintain execution validity and consistency across multiple dependent steps with evolving constraints, and to discover effective strategies for tasks with implicit complexity beyond their surface descriptions.

To address these requirements, we incorporate Interaction Modeling to capture diverse interaction patterns, following prior approaches that model user-side diversity, and introduce Capability-Aware Guidance to improve execution reliability by providing additional task-specific guidance during trajectory synthesis. As illustrated in Figure~\ref{fig:trajectory_overview}, these two components serve as configurable enhancements to a unified trajectory generation and verification pipeline.

\subsubsection{User Interaction Modeling}

Existing approaches such as $\tau^2$-Bench~\cite{barres2025tau}, VitaBench~\cite{he2025vitabench}, and OmniaBench~\cite{shen2026omniabenchbenchmarkinggeneralai} have explored user simulation and interaction modeling, where user profiles or behavioral factors are used to characterize diverse interaction patterns, enabling simulated users to interact with agents through natural and imperfect communication. Inspired by these approaches, we adopt a persona-driven user simulation process and model task-level interaction factors that commonly affect real-world user-agent interactions. Specifically, we consider request granularity, information completeness, and inconsistent or misleading details, as these factors capture diverse user behaviors that can affect trajectory evolution, thereby enhancing the diversity and realism of synthesized interactions while capturing user-side uncertainty during agent trajectory generation.

\subsubsection{Capability-Aware Privileged Guidance}

While interaction modeling improves user-side diversity, generating high-quality trajectories for challenging tasks remains limited by the capability of the teacher model itself. For challenging tasks, even frontier models often achieve relatively low success rates through direct rollout due to insufficient task-specific knowledge and ineffective long-horizon exploration. Since supervised fine-tuning largely relies on the quality of synthesized supervision trajectories rather than the rollout policy itself, unreliable rollouts directly limit the quality of supervision data and consequently the effectiveness of downstream training.

Recent studies such as OPSD \cite{zhao2026selfdistilledreasoneronpolicyselfdistillation} and EDGE-OPD \cite{lazaridis2026edgeopdinternalizingprivilegedcontext} have explored leveraging privileged information during training to improve policy learning by providing auxiliary signals unavailable at inference time. However, existing approaches typically generate privileged trajectories using the same policy being optimized, making the resulting supervision fundamentally bounded by the reasoning and exploration capabilities of that policy. Other methods like \cite{wang2026hindsighthintdistillationscaffolded} employ hindsight-based refinement to repair erroneous trajectories after rollout. While effective for correcting local reasoning errors, such post-hoc refinement becomes considerably less effective for long-horizon agent tasks, where execution failures accumulate across multiple dependent steps and often cannot be recovered through local corrections.

We focus on improving the quality of supervision trajectories for SFT. Since teacher models may fail due to different capability bottlenecks across different types of complex tasks, we first diagnose representative failure patterns for each task type and design capability-aware \textit{privileged guidance} to assist trajectory synthesis. Such guidance is available only to the teacher model during generation and is not accessible to the target model. The synthesized trajectories are further filtered by a multi-stage verification pipeline, ensuring that only reliable and grounded trajectories are retained for supervision.

\paragraph{Privileged Guidance Decomposition.}

The effectiveness of privileged guidance depends on whether the provided information matches the capability bottlenecks responsible for unsuccessful rollouts. Therefore, instead of applying a unified guidance format, we analyze teacher model failures from different capability perspectives and design task-adaptive privileged guidance with different levels of abstraction. Following the ReAct~\cite{yao2022react} paradigm, we view agent trajectories as an iterative process involving planning, reasoning, execution, and outcome verification, and design guidance to address deficiencies at different stages of this process.
Specifically, we identify two complementary types of privileged guidance according to the underlying capability limitations:

\begin{itemize}

    \item \textbf{Planning-oriented guidance.}
    When failures mainly originate from insufficient planning and exploration capability, we provide high-level planning abstractions, decomposition strategies, and execution heuristics to improve the organization of the solution process. Such guidance focuses on improving how the model formulates and executes solution strategies, while leaving subsequent interactions with the environment unconstrained.

    \item \textbf{Outcome-oriented guidance.}
    When the teacher model is capable of planning but fails to satisfy task-specific requirements, we provide outcome-level information, including desired completion criteria, execution constraints, target states, and verification requirements. Such guidance helps the model align its execution results with task objectives without prescribing the intermediate reasoning process.

\end{itemize}

For both types of guidance, intermediate trajectory components, including reasoning, actions, and observations, are generated through natural interaction with the environment. 
Moreover, the teacher model is instructed to use privileged guidance implicitly and avoid explicitly mentioning or bypassing necessary tool interactions or observations in generated trajectories. This design improves synthesis reliability while reducing the risk of privileged information exposure in the resulting supervision data.

\paragraph{Task-Specific Guidance.}

Instead of applying a unified guidance format, we adopt a diagnosis-driven strategy that identifies dominant failure patterns for each task category and designs corresponding privileged guidance to improve trajectory synthesis.

\begin{itemize}

\item \textbf{DAG-challenge / program-based tasks.}
Although these tasks involve different underlying structures, frontier models generally possess sufficient high-level planning capability to decompose the overall objectives. Their failures mainly arise from overlooking execution details, missing intermediate requirements, or violating task-specific constraints during long-horizon interactions. Based on this observation, we provide outcome-oriented guidance, including rubric-based criteria and structured target states (e.g., expected JSON specifications), to help the teacher model maintain execution consistency and satisfy critical requirements without constraining its planning process.

\item \textbf{Solver-based tasks.}
Solver tasks present a different challenge, where models may fail to identify feasible or optimal solutions even when provided with solution references. These failures are often caused by insufficient reasoning depth, incomplete constraint consideration, or ineffective exploration of the solution space. To address these limitations, we introduce planning-oriented guidance that encourages more systematic reasoning during solution generation. In addition, solution-level hints are provided as auxiliary references to help the teacher model explore candidate solutions and improve solution quality through deeper analysis.

\end{itemize}

\subsubsection{Trajectory Quality Verification}

High-quality supervision trajectories must both satisfy the target task requirements and exhibit reliable reasoning. After synthesis, we first assess task correctness using task-specific evaluation rubrics; only trajectories that satisfy the required criteria proceed to trajectory quality verification. We then assess trajectory quality against three complementary criteria: reasoning continuity, logical consistency, and evidence grounding. Together, these checks verify that task-correct trajectories are also internally coherent and grounded in observable evidence. Evidence grounding is particularly important because privileged guidance is available during synthesis but must not leak into the retained supervision. This criterion verifies that each reasoning step can be attributed to information observable to the assistant.

\paragraph{Reasoning Continuity.}

Reasoning continuity evaluates whether the reasoning process progresses through sufficiently supported intermediate steps. We examine whether each reasoning step can be naturally inferred from previously available observations, executed actions, or established intermediate conclusions. Trajectories containing abnormal reasoning jumps, omitted intermediate reasoning, or conclusions unsupported by preceding execution are regarded as violating reasoning continuity and are therefore discarded.

\paragraph{Logical Consistency.}

Logical consistency evaluates whether the reasoning process remains internally coherent throughout execution. We verify that every reasoning step is compatible with preceding observations, environment states, executed actions, and intermediate conclusions. Trajectories containing contradictory reasoning, inconsistent state transitions, or conclusions conflicting with earlier reasoning are removed.

\paragraph{Evidence Grounding.}

Evidence grounding evaluates whether every reasoning step is supported exclusively by evidence observable to the assistant model. Valid evidence includes the task description, environment specifications, environment observations, and tool execution results. Every reasoning step should be attributable to these observable sources. Trajectories introducing unsupported facts, hallucinated evidence, explicit references to privileged guidance, or reasoning relying on information unavailable to the assistant model are regarded as evidence violations and discarded.
Only trajectories passing all three quality checks are retained for supervised fine-tuning.

\begin{wrapfigure}[20]{r}{0.35\textwidth}
    \vspace{-0.5em}
    \centering
    \includegraphics[width=\linewidth]{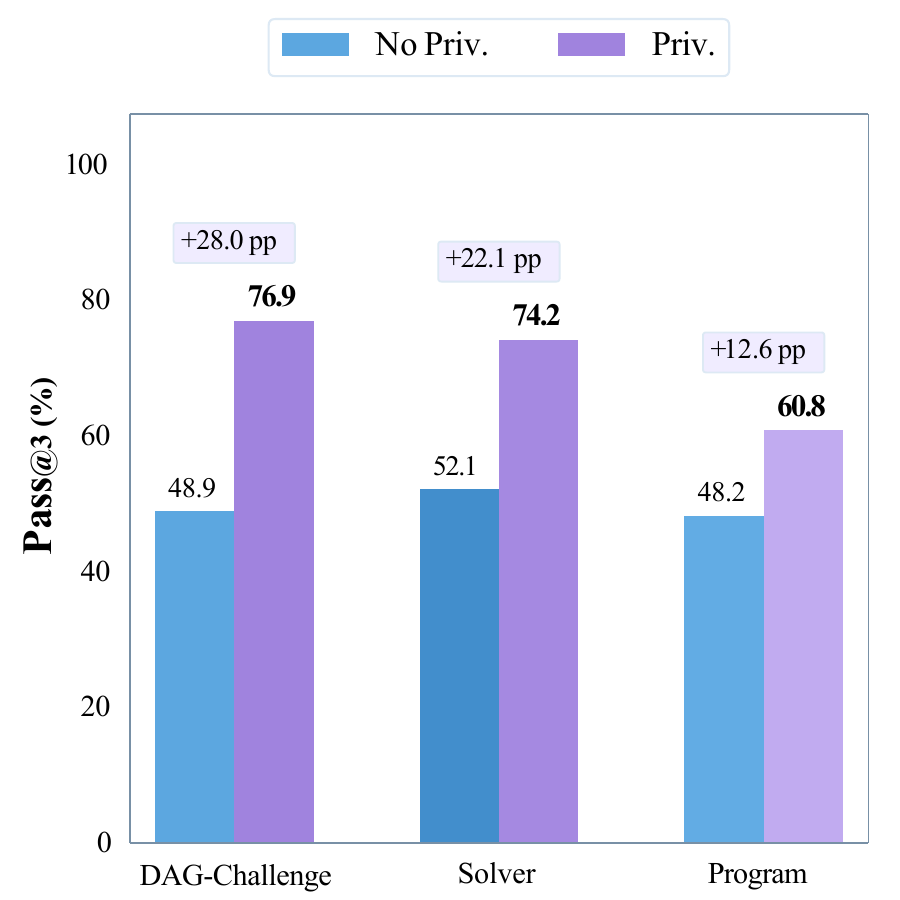}
    \caption{\textbf{Effect of privileged guidance on verified trajectory synthesis.} Verified Pass@3 comparison between synthesis with and without privileged guidance.}
    \label{fig:data_card_env}
    \vspace{-1em}
\end{wrapfigure}

\paragraph{Effect of Privileged Guidance.}

Figure~\ref{fig:data_card_env} reports the verified Pass@3 trajectory synthesis results under the complete verification pipeline. Compared with synthesis without privileged guidance, privileged guidance consistently improves trajectory synthesis performance across all evaluated datasets, producing more trajectories that satisfy both task requirements and quality criteria.

We further evaluate whether privileged guidance introduces information that is inaccessible during actual task execution into the final supervision. As shown in Table~\ref{tab:evidence-grounding-verification}, trajectories generated with privileged guidance maintain high evidence grounding performance across all paradigms. Although the Solver paradigm exhibits a more noticeable decrease compared with synthesis without guidance, its grounding pass rate remains high after verification. This indicates that the proposed pipeline allows privileged guidance to improve synthesis quality without compromising the reliability of the resulting supervision data.

The remaining reasoning quality criteria remain stable across different task paradigms and synthesis settings. Since these criteria are not directly related to privileged information leakage, we omit further discussion and focus on evidence grounding, which is directly related to privileged guidance leakage.

\begin{table}[t]
\centering
\scriptsize
\setlength{\tabcolsep}{8pt}
\renewcommand{\arraystretch}{1.15}
\caption{\textbf{Evidence grounding verification under different trajectory synthesis settings.} Pass rates measure the proportion of synthesized trajectories whose reasoning is fully supported by information accessible to the assistant model, excluding unsupported evidence, hallucinated facts, and privileged guidance leakage.}
\label{tab:evidence-grounding-verification}

\begin{tabular}{lccc}
\toprule[0.9pt]

\textbf{Synthesis Setting}
& \textbf{DAG-challenge}
& \textbf{Solver}
& \textbf{Program} \\

\midrule

Without privileged guidance
& 95.84\%
& 98.00\%
& 98.45\% \\

With privileged guidance
& 94.28\%
& 92.01\%
& 97.00\% \\

\bottomrule[0.9pt]

\end{tabular}

\end{table}
\section{Agentic Post-Training}
\label{sec:posttraining}

\begin{figure}[h]
    \centering
    \includegraphics[width=\linewidth]{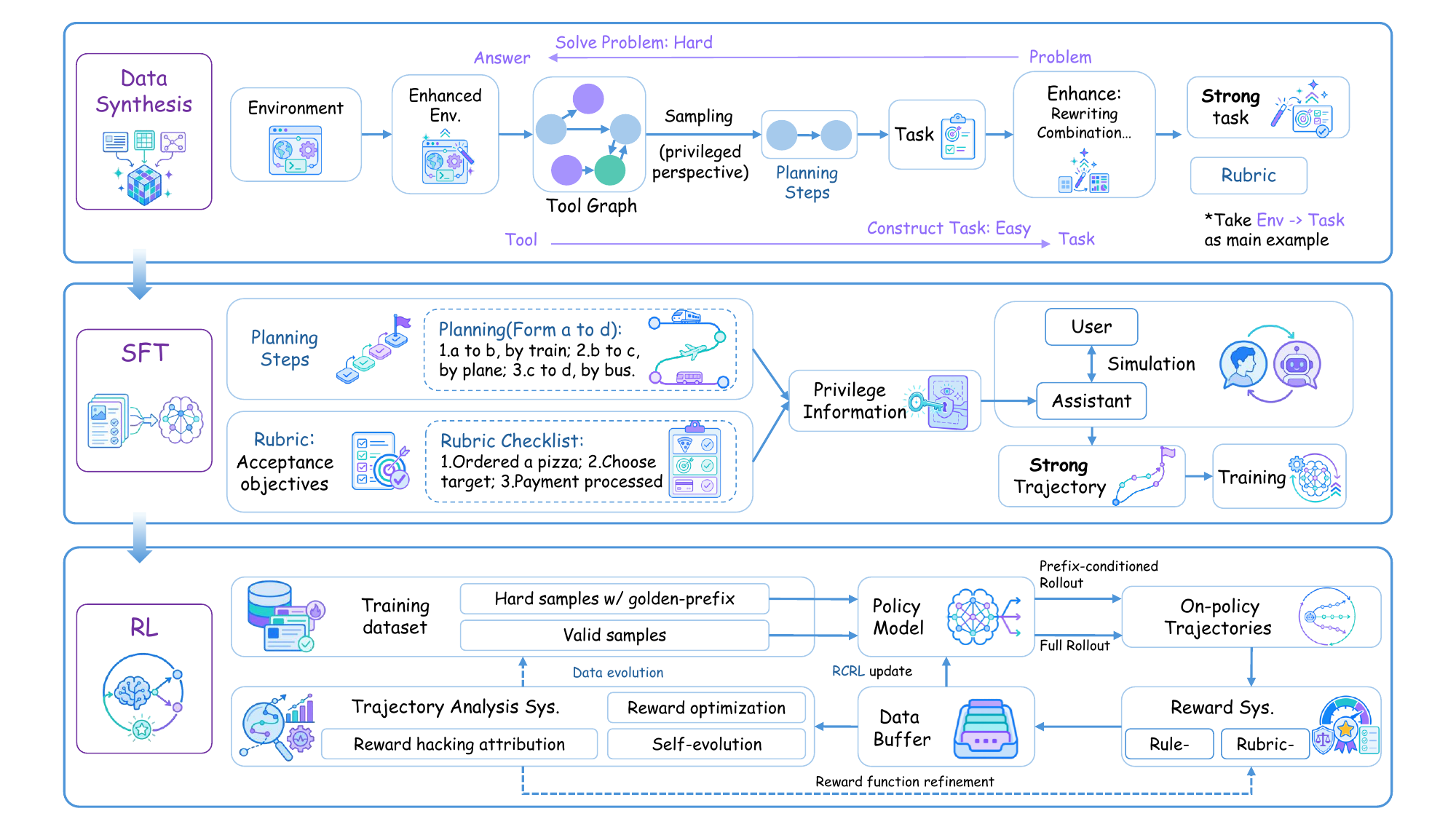}
    \caption{\textbf{Overview of weak-to-strong data synthesis and post-training paradigm.} Starting from the enhanced executable environments, AgentOmnia synthesizes privileged tool-call chains into validated, user-facing tasks that are easy to verify but hard to solve, ensuring retained samples are both challenging and solvable. \textbf{In SFT}, the capability-aware privileged guidance framework combines user interaction modeling with planning- and outcome-oriented guidance to synthesize accurate, efficient, and diverse supervision trajectories for challenging tasks. \textbf{In RL}, a hybrid rule- and rubric-based reward system scores rollout trajectories, and RCRL performs progressive policy optimization based on the resulting rewards. Subsequently, a separate trajectory analysis system performs failure-mode analysis and reward-hacking audits, identifying capability gaps from rollout trajectories to refine the reward function and guide the synthesis of targeted training instances.}
    \label{fig:posttrain_overview}
\end{figure}

AgentOmnia adopts a weak-to-strong data synthesis and post-training paradigm. Privileged information first assists the teacher model in solving tasks that exceed its own capability boundary, and post-training then distills this verified, executable supervision into tangible policy improvement. As illustrated in Figure~\ref{fig:posttrain_overview}, AgentOmnia follows a two-stage alignment strategy in which supervised fine-tuning (SFT) first instills complex, interleaved reasoning patterns into the policy, followed by rollback-based curriculum reinforcement learning (RCRL), which enables progressive and efficient on-policy learning over challenging samples.

\subsection{SFT Capability Bootstrapping}

Building upon the synthesized trajectories described in Section~\ref{sec:trajectory-synthesis}, we construct the SFT training corpus through three processing steps. We first apply \textbf{Format validation} to remove invalid trajectories, and then perform \textbf{Convergence-aware sample budgeting} to balance the data composition, obtaining 53K SFT training instances. Finally, we apply \textbf{Context alignment} to handle multi-turn trajectory contexts before optimizing the agent policy with the standard causal language modeling objective for RL initialization. The following sections describe these procedures in detail.

\paragraph{Format Validation.}
We filter trajectories whose tool interactions violate the predefined interface specifications. Specifically, we validate the parsability of
actions, the correctness of function names, and the consistency of arguments with
corresponding tool schemas. This step removes malformed interactions and improves the reliability of the resulting training data.

\paragraph{Convergence-Aware Sample Budgeting.}
Different trajectory sources exhibit different data scales and learning dynamics during supervised fine-tuning. Directly mixing these sources according to their original proportions may cause over-represented sources to dominate optimization, while under-represented sources may receive insufficient supervision to fully acquire the corresponding capabilities. Such imbalance can lead to suboptimal capability acquisition across heterogeneous tasks. Therefore, instead of allocating training samples based on the original data distribution, we adjust the sampling budget according to the estimated convergence requirements of each trajectory source.

To estimate the convergence requirement, we independently fine-tune the model on each trajectory source and analyze the corresponding validation performance curves. We define the convergence requirement as the amount of supervision needed for the model performance to reach a stable state, providing an empirical estimate of the learning budget required by each trajectory source.

Based on the estimated convergence requirements, we allocate training samples proportionally across trajectory sources. In this way, each source receives a supervision budget that better matches its learning difficulty, allowing different capabilities to progress through more balanced optimization during training. This convergence-based allocation strategy directly derives the sampling budget from empirical learning dynamics, rather than relying on iterative searches for mixture ratios as in previous data mixture optimization methods \cite{xie2023doremioptimizingdatamixtures,li2025datamixingoptimizationsupervised}.

\paragraph{Context Alignment.} We perform trajectory context alignment in the inner training framework by converting the filtered
trajectories into SFT samples following the chat template of the base model. For
multi-turn trajectories, only the reasoning process associated with the final user
query is retained, while reasoning traces from earlier assistant turns are
removed. This design aligns the training context with the inference-time reasoning
context and avoids introducing irrelevant historical reasoning patterns during
training.

\subsection{Reinforcement Learning}
Building on SFT initialization, we develop an RL framework designed for robust and scalable agentic policy optimization. It comprises the following components:
\textbf{Data preparation} curates high-quality training instances with appropriate difficulty distribution and stable execution environments, ensuring sustained and stable RL training; \textbf{Reward system} provides rule- and rubric-based reward computation with routing safeguards, efficient asynchronous judge serving, and RL-oriented reward model calibration to deliver reliable and scalable optimization signals;
\textbf{Rollout trajectory analysis system} monitors and analyzes RL rollouts to diagnose reward behaviors, identify reward hacking patterns, and uncover capability gaps, providing actionable insights for reward refinement and the synthesis of targeted training instances for continual model improvement;
\textbf{RL training} builds upon GRPO with rollback-based curriculum learning for challenging tasks, and integrates targeted techniques including dynamic filtering and training-inference mismatch correction to improve optimization stability and policy effectiveness.

\subsubsection{Data Preparation}
Training data quality, difficulty, and reward reliability critically govern RL optimization stability and policy performance. We construct our training corpus via a three-stage pipeline comprising basic filtering, difficulty estimation with quality validation, and curriculum-aware data composition, progressively eliminating noise, resolving reward ambiguity, and aligning task distribution with policy capacity.

\paragraph{\textbf{Data Filtering.}} We filter data along two dimensions: at the prompt level, redundant examples are removed via N-gram and embedding similarity, and prompts below a minimum length threshold are discarded. At the rubric level, LLM-based validation eliminates ambiguous, non-atomic, conflicting, or incomplete criteria that fail to provide sufficient evaluation coverage, improving reward reliability and preventing reward hacking.

\paragraph{\textbf{Difficulty Grading and Quality Validation.}} We estimate the intrinsic difficulty of each instance via $\text{pass}@K$ evaluation using the SFT checkpoint. The resulting $K$ rollout logs are further leveraged for quality validation: samples with environment anomalies (e.g., API failures, abnormal termination) and excessive sequence truncation are discarded to ensure reward reliability and rollout quality.

\paragraph{\textbf{Data Construction.}} The RL training set is constructed by uniformly sampling across difficulty levels within $\text{Pass}@K\in[0\%, 80\%]$, yielding 5K training instances, each paired with a golden trajectory. To stabilize policy optimization, we maintain a 10\%--20\% overlap with the SFT dataset, which acts as an implicit regularizer against policy drift.

\subsubsection{Reward System}
\label{sec:reward}
Our task-adaptive reward system synergizes rule/rubric-based verification and efficient judge-serving, providing fine-grained, stable, and scalable signals for robust RL optimization under large-scale trajectory sampling.

\paragraph{\textbf{Reward Design.}} We organize reward computation into two complementary types conditioned on metadata. Rule-based rewards apply extraction and comparison functions over candidate and reference answers to check schema validity and exact or fuzzy matching, providing high-precision anchors for verifiable constraints. Rubric-based rewards evaluate decision criteria that resist direct matching via general rubrics for cross-task discipline and task-specific rubrics for benchmark-dependent operational logic, supplying multidimensional evaluation for long-horizon agentic behaviors. In our experiments, both rule- and rubric-based rewards are converted into binary signals in $\{0,1\}$.

\paragraph{\textbf{Reward Routing and Safeguards.}} The reward router dynamically assigns each rollout trajectory to rule-based verification, rubric-based scoring, or a combination thereof, conditioned on the associated metadata. For rule-based verification, any execution failure triggers a safeguard mechanism that invokes an LLM judge for re-evaluation, ensuring reward coverage and scoring stability.

\paragraph{\textbf{System Efficiency.}} As rubric-based RL scales, generative reward computation has emerged as a primary throughput bottleneck~\cite{wang2025let}. Our reward system addresses this at two levels: at the \textit{sample level}, scoring requests are dispatched as individual responses complete rather than waiting for full-batch accumulation; at the \textit{step level}, process validations are activated mid-generation without requiring complete trajectories, eliminating end-of-trajectory synchronization latency. Requests are further routed across reward model (RM) instances by real-time load and prefix-sharing affinity to maximize KV-cache reuse and minimize NPU idle time.

\paragraph{\textbf{Reward Model Calibration.}}
Standard evaluation metrics, such as accuracy and mean absolute error, are insufficient for selecting a reliable reward model in agentic RL, as they may obscure optimization-critical failure modes~\cite{gu2026survey}. These include errors on key agentic behaviors, corrupted within-group rankings, advantage sign flips, distorted update magnitudes, and score inconsistency. 

Given queries $\{x_n\}_{n=1}^N$ each paired with a group of responses $\{y_i\}_{i=1}^G$, each query $x_n$ is further associated with $K_n$ rubrics, with $g_{n,i,k},\hat{g}_{n,i,k}\in\{0,1\}$ denoting the gold and predicted judgments of whether response $y_i$ satisfies rubric $k$. Let $\hat{R}_{n,i}\in[0, 1]$ denote the response-level normalized predicted reward. We evaluate the candidate reward model across five complementary dimensions that directly probe scoring fidelity under group-relative RL methods.
\begin{itemize}
    \item \textbf{Advanced-balanced rubric reliability score} measures rubric-level correctness. Standard RMs often perform well on basic rubrics (e.g., format compliance, value validation) but struggle with advanced rubrics (e.g., complex procedural constraints). When aggregated naively, the dominance of basic rubrics can obscure these failures. We therefore partition rubrics into a basic set $\mathcal{B}$ and an advanced set $\mathcal{A}$, compute per-category accuracy independently, and aggregate them with a weighted harmonic mean.
    
    \item \textbf{Kendall tau-b} measures whether the RM preserves within-group trajectory rankings under the same prompt. Since group-relative RL methods optimize from intra-group comparisons rather than absolute reward values, rank reversals risk reinforcing inferior trajectories while suppressing superior ones. For each prompt $x_i$, let $N_i^{Con}$ and $N_i^{Dis}$ denote the number of concordant and discordant trajectory pairs, $N_i^{Tie_{G}}$ and $N_i^{Tie_{P}}$ denote ties exclusive to golden and predicted rewards, respectively:
    \begin{equation}
        C_{\mathrm{Rank}} = \frac{1}{2}\left(1 + \frac{1}{N}\sum_{i=1}^{N} \tau_{i} \right),
        \quad
        \tau_{i} = \frac{N_i^{Con} - N_i^{Dis}}{
        \sqrt{ (N_i^{Con} + N_i^{Dis} + N_i^{Tie_{G}})(N_i^{Con} + N_i^{Dis} + N_i^{Tie_{P}}) }
        },
    \end{equation}
    where $C_{\mathrm{Rank}}\in[0,1]$ with larger values indicating better ranking consistency.
    
    \item \textbf{Advantage direction reliability} measures whether the RM preserves the sign of normalized advantages. During RL training, advantage sign governs the direction of policy updates, making sign flips a source of fundamentally incorrect gradient updates even when relative rankings are partially preserved. We quantify two sign-error rates, AdvFNR (gold-positive trajectories receiving negative predicted advantage) and AdvFPR (gold-negative trajectories receiving positive predicted advantage), and aggregate them with a weighted harmonic mean.
    \begin{equation}
        C_{\mathrm{Dir}} = \left(\frac{\lambda}{\max(1-\mathrm{AdvFPR}, \epsilon)} + 
        \frac{1-\lambda}{\max(1-\mathrm{AdvFNR}, \epsilon)}\right)^{-1},
    \end{equation}
    where $\lambda$ balances the relative cost of false-positive and false-negative sign errors, and $\epsilon$ prevents division by zero. $C_{\mathrm{Dir}}\in(0,1]$, with lower values driven by whichever sign-error type is worse.
    
    \item \textbf{Advantage magnitude consistency} measures whether the RM preserves training signal strength after group normalization. Correct advantage direction alone is insufficient. Over-amplifying weak positive trajectories or under-penalizing strongly negative ones distorts the policy gradient weighting, leading to unstable or biased updates. We therefore measure the mean squared error between normalized advantages $\hat{A}_{n,i}^{\mathrm{pre}}$ and $\hat{A}_{n,i}^{\mathrm{gold}}$:
    \begin{equation}
        C_{\mathrm{Mag}} = \max\left( 0, 1-\frac{\operatorname{Mean}_{n,i}\left[\left( \hat{A}_{n,i}^{\mathrm{pre}} - \hat{A}_{n,i}^{\mathrm{gold}} \right)^2\right]}{\max\left(\operatorname{Var}_{n,i}\left[\hat{A}_{n,i}^{\text{gold}}\right], \epsilon\right)}\right),
    \end{equation}
    where $C_{\text{Mag}}\in[0,1]$ with larger values indicating better magnitude consistency.
    
    \item \textbf{Scoring consistency} measures the reproducibility of predicted rewards under repeated evaluation of the same trajectory. Score variability across identical inputs introduces stochastic noise into advantage estimation, destabilizing policy gradient updates. For $T$ repeated evaluations of trajectory $(x_n,y_i)$, the average repeated-scoring variance is defined as:
    \begin{equation}
        C_\text{Con}
        =
        1-\frac{4}{N\cdot G\cdot T}
        \sum_{n,i}
        \sum_{t=1}^{T}
        \left(
        \hat{R}_{n,i}^{(t)}
        -
        \operatorname{Mean}_t[\hat{R}_{n,i}^{(t)}]
        \right)^2,
    \end{equation}
    where $C_{\text{Con}}\in[0,1]$ with larger values reflecting greater scoring reproducibility.
\end{itemize}

Together, these five dimensions provide a fine-grained diagnostic profile of a reward model's reliability under group-relative RL. We aggregate them into an overall score $C_{\text{RM}}=\sum_{j=1}^{5} w_j \cdot C_j$ via a weighted sum, and additionally require each individual dimension to exceed a minimum threshold, preventing a single high-scoring dimension from masking a critical failure on another.

\subsubsection{Rollout Trajectory Analysis System}
Scalar rewards in conventional RL pipelines are inherently opaque, masking behavioral anomalies and hindering precise credit assignment over long-horizon rollouts~\cite{zhang2026reasoning}. We introduce a trajectory analysis system that bridges this gap by diagnosing complete execution trajectories at the behavioral level: identifying policy failures, attributing reward hacking patterns, and uncovering capability gaps to guide reward refinement and self-evolving data synthesis. This closed-loop design allows the reward scheme to co-evolve with growing agent capability~\cite{wang2026verification, wang2026reproducing}.

\paragraph{\textbf{Reward Optimization.}} Successful and unsuccessful agent rollouts often diverge along measurable behavioral indicators. We compare positive trajectories against zero- and low-reward executions to identify recurring behavioral gaps, converting them into auxiliary rewards that provide more informative optimization signals. For example, tool-call anomalies (e.g., nonexistent tool names, missing or ill-typed arguments, invalid serialization formats) account for up to 10\% of all tool calls in certain tasks, with over 70\% occurring in failed rollouts, motivating explicit format-validity rewards. We also observe that failed rollouts frequently repeat tool invocations until context exhaustion instead of requesting missing information, suggesting rewards that discourage redundant tool usage and promote clarification-seeking behaviors.

\paragraph{\textbf{Reward Hacking Attribution.}} We audit suspicious high-reward rollouts via LLM-based classification and pattern induction, distinguishing genuine reward hacking from judge failures, environment errors, and overly permissive rubrics. For example, in a procurement-planning task, the model fabricated prices for unlisted grocery items and falsely claimed the total had been verified; in a supplier-selection task, the model happened to produce a feasible plan matching the optimal reference, yet without exhaustive enumeration or a valid pruning proof to substantiate the claimed optimality.

\paragraph{\textbf{Self-Evolution.}} Trajectory analysis identifies capability gaps and transforms them into targeted training instances, enabling continual self-improvement through iterative post-training, as detailed in Section~\ref{sec:prd}.

\subsubsection{Reinforcement Learning with Rollback Curriculum}
\label{sec:RCRL}
\paragraph{\textbf{RL Algorithm Backbone.}}
Our reinforcement learning algorithm is built upon GRPO~\cite{shao2024deepseekmath}, with the KL divergence regularization term removed to fully exploit the policy improvement capacity of RL training. To address the distributional discrepancy between the training and inference engines during RL optimization~\cite{liu-li-2025-rl-collapse, yao2025offpolicy}, we replace the conventional proximal policy in the PPO ratio with the raw behavior policy from the inference engine, which preserves training stability while improving computational efficiency. Moreover, we introduce Routing Replay~\cite{ma2025stabilizing} to ensure consistency between routed experts used in training and those used during rollout.
Concretely, for each query $q$, given a group of responses $\{y_1,...,y_G\}$ sampled from the old policy $\pi_{\theta_\text{old}}$, the policy model $\pi_{\theta}$ is optimized by maximizing the following objective:
\begin{equation}
    \mathcal{J}_{\text{RL}}(\theta) = \mathbb{E}_{x\sim\mathcal{D},{\{y_i\}}^G_{i=1}\sim\pi_{\theta_\text{old}}^{\text{infer}}(\cdot|x)}\left[\frac{1}{\sum_{i=1}^{G}|y_i|}\sum_{i=1}^{G}\sum_{t=1}^{|y_i|}\min\left(r_{i,t}\hat{A}_{i,t}, \text{clip}(r_{i,t}, 1-\epsilon_{\text{low}},  1+\epsilon_{\text{high}})\hat{A}_{i,t}\right)\right],
\end{equation}
where
\begin{equation}
    r_{i,t} = \frac{\pi_\theta^{\text{train}}(y_{i,t}|x_i,y_{i,<t})}{\pi_{\theta_\text{old}}^{\text{infer}}(y_{i,t}|x_i,y_{i,<t})},\qquad \hat{A}_{i,t}=\frac{R_i-\operatorname{mean}(R_1,\ldots,R_G)}{\operatorname{std}(R_1,\ldots,R_G)}.
\end{equation}
$\pi_\theta^{\text{train}}$ denotes the policy hosted by the training engine (e.g., FSDP, Megatron) for gradient updates, while $\pi_\theta^{\text{infer}}$ denotes the policy hosted by the inference engine (e.g., vLLM, SGLang) for generating rollouts. $\epsilon_{\text{low}}=0.2$ and $\epsilon_{\text{high}}=0.28$ are asymmetric hyperparameters that control the clipping range~\cite{yu2026dapo}.

\paragraph{Dynamic Rollback-Based Curriculum.}
\label{sec:rollback}
For group-relative RL methods, sufficiently challenging tasks can leave the policy without a useful learning signal. A natural remedy, long studied in classical RL, is curriculum learning via state rollback. In maze navigation, for instance, learning a policy directly from the start state is difficult; a common strategy is to initialize the agent near the goal and progressively expand the initialization region until it covers the original start state~\cite{duan2023hand}. We adopt an analogous strategy for LLM agents: rather than rolling out from the raw prompt, the policy is rolled out from the prompt concatenated with a golden-response prefix, which effectively reduces task difficulty by shortening the exploration horizon~\cite{wang2025let, li2026saveloadlearn}.

\begin{algorithm}[t]
\caption{Rollback-based Curriculum Reinforcement Learning}
\label{alg:rcrl}
\begin{algorithmic}[1]
\Require Policy model $\pi_\theta$; reward model $R$; task prompt set $\mathcal{D}$;
gold-prefix increment ratios $\alpha$; max retry count $T_{\max}$; group size $G$
\State Initialize gold-prefix turn count $P_q \gets 0$, group average reward $\Bar{R}_{q} \gets 0$ and coarse increment $M_q^{(0)} \gets \alpha |\tau_q^*|$ for all prompts $q \in \mathcal{D}$
\For{epoch $= 1, \dots, N$}
    \For{step $= 1, \dots, K$}
        \State Sample a mini-batch $\mathcal{D}_b \subset \mathcal{D}$
        \State Update behavior policy: $\pi_{\theta_{\text{old}}} \gets \pi_\theta$
        \For{each prompt $q \in \mathcal{D}_b$}
            \State Initialize retry counter $t \gets 0$
            \Repeat
                \State Sample $G$ rollouts $\{o_i\}_{i=1}^{G} \sim \pi_{\theta_{\text{old}}}(\cdot \mid q, P_q)$
                \State Compute rewards $\{R_q(o_i)\}_{i=1}^{G}$
                \State Compute group average reward: $\Bar{R}_{q} \gets \frac{1}{G}\sum_i{R_q(o_i)}$
                \If{$\Bar{R}_{q} = 0$}
                    \State $P_q \gets P_q + M_q^{(\text{epoch})}$ \Comment{extend golden prefix by $M_q$ turns to avoid all-fail groups}
                    \State $t \gets t + 1$
                \EndIf
            \Until{$\Bar{R}_{q} \neq 0$ \textbf{or} $t \geq T_{\max}$}
            \State Update golden-prefix turn count $P_q \gets P_q + \delta_q$ (Eq.~\eqref{func:rollback}) \Comment{rollback based on latest $\bar{R}_q$}
        \EndFor
        \For{iteration $= 1, \dots, J$}
            \State Update the policy model $\pi_\theta$ (Eq.~\eqref{func:rcrl})
        \EndFor
    \EndFor
\EndFor
\State \Return $\pi_\theta$
\end{algorithmic}
\end{algorithm}

In our experiments, each RL sample is paired with a golden trajectory. We first verify that all environments satisfy a consistent resettable property, i.e., replaying the same successful trajectory always leads to the same state. The proposed RCRL (Algorithm~\ref{alg:rcrl}) combines two complementary rollback strategies during rollout. The first is a \textbf{retry-triggered} mechanism: whenever a group yields zero accuracy, we forcibly extend the golden prefix by $M$ turns and re-rollout, repeating this process until the group accuracy exceeds zero. In the first epoch, $M^{(0)}$ is set to a coarse granularity (e.g., $M^{(0)}$=20\% of the total golden-trajectory length $|\tau^*|$) to quickly localize the bottleneck region of each trajectory. The second is a \textbf{round-wise adaptive} mechanism applied after every rollout round, where the prefix increment $\delta_q$ for prompt $q$ is dynamically determined by comparing its group average reward $\bar{R}_q$ against the threshold $R_{\mathrm{th}}$:
\begin{equation}
\label{func:rollback}
\delta_q =
\begin{cases}
-\delta_{max}, & \text{if } \Bar{R}_{q} \gg R_{\text{th}},\\
-1, & \text{if } \Bar{R}_{q} \approx R_{\text{th}},\\
1, & \text{if } \Bar{R}_{q} \ll R_{\text{th}}.
\end{cases}
\end{equation}
Intuitively, an average reward $\bar{R}_q$ greatly exceeding the threshold indicates the task has become too easy, so the prefix is aggressively shortened by the maximum step size $\delta_{max}$ to expose more of the trajectory to on-policy exploration; a rate near the threshold is shortened by a small step for fine-grained adjustment, gradually increasing difficulty as the policy improves; and a group whose $\bar{R}_q$ falls well below the threshold indicates the task is still too hard, so the prefix is extended to provide additional guidance and keep the learning signal non-degenerate. From the second epoch onward, since every sample has already received an informative gradient signal, we replace the coarse-grained constant with this adaptive increment, i.e., $M_q^{(>0)}\leftarrow \delta_q$.

Furthermore, to accelerate convergence and ensure the policy can still learn to generate the full trajectory even when trained on prefix-conditioned rollouts, we adopt the hybrid training objective
\begin{equation}
\label{func:rcrl}
\mathcal{L}_{\text{RCRL}}(\theta) = 
\underbrace{\lambda_{\text{CE}} \cdot \mathcal{L}_{\text{SFT}}(\tau^*_{\le P_q})}_{\text{golden-prefix supervision}}
\;+\;
\underbrace{\lambda_{\text{RL}} \cdot \mathcal{L}_{\text{GRPO}}(\tau_{>P_q})}_{\text{on-policy exploration}},
\end{equation}
where $\tau^*$ denotes the golden-prefix trajectory, $\tau$ the on-policy trajectory, $P_q$ the current golden-prefix turn count for prompt $q$, and $\lambda_{\text{CE}}$, $\lambda_{\text{RL}}$ are weighting coefficients. The objective imposes a cross-entropy loss on the golden-prefix segment to ensure the policy remains capable of reproducing it under its own parameterization.

\paragraph{\textbf{Training Stabilization and Systems Alignment.}}
To enhance both performance and training stability in reinforcement learning, we introduce and integrate the following key components.
\begin{itemize}
    \item \textbf{Dynamic filtering.} Multi-turn agent interactions may introduce \textbf{environmental noise} (e.g., transient API failures, environment initialization errors, unexpected shutdowns, reward evaluation timeouts). Trajectories corrupted by these policy-extrinsic factors are strictly discarded. For \textbf{length-truncated rollouts}, the final outcome is often unidentifiable. Prior work typically masks or filters such samples, which may inadvertently bias the policy toward longer responses. We instead use a reward-based filtering scheme that discards truncated trajectories passing low-level validity checks (e.g., redundant or repeated tool calls) and assigns negative rewards to the rest.
    \item \textbf{Removing KL and entropy regularization.} Unlike reasoning-only tasks, 
    agentic tasks concatenate heterogeneous OOD inputs (e.g., tool outputs and user responses), 
    yielding inherently higher policy entropy. We empirically find that KL penalties impede 
    policy improvement, while entropy bonuses promote verbosity without increasing effective 
    trajectory diversity; we therefore omit both regularization terms.
    \item \textbf{Training-inference mismatch correction.} Discrepancies between 
    training and inference engines introduce off-policy bias. We address this on three fronts: 
    (1)~replacing the proximal policy with the raw behavior policy to correct 
    distribution mismatch; (2)~applying routing replay to enforce consistent expert routing; and (3)~setting $\text{top-p}=1$ during rollout to align action spaces 
    between training and inference.
    \item \textbf{Token-in/token-out alignment.} Each rollout step's input is formed by directly 
    concatenating the previous step's output tokens, with no intermediate decode-encode 
    reprocessing, ensuring strict token-level alignment between training and inference. For 
    Qwen-family models, a newline token is appended after each inference-engine EOS to maintain TI/TO~\cite{skyrl_gym_generator}.
\end{itemize}

\section{PRD-Guided Self-Evolution}
\label{sec:prd}

\begin{figure*}[t]
    \centering
    \includegraphics[width=\linewidth]{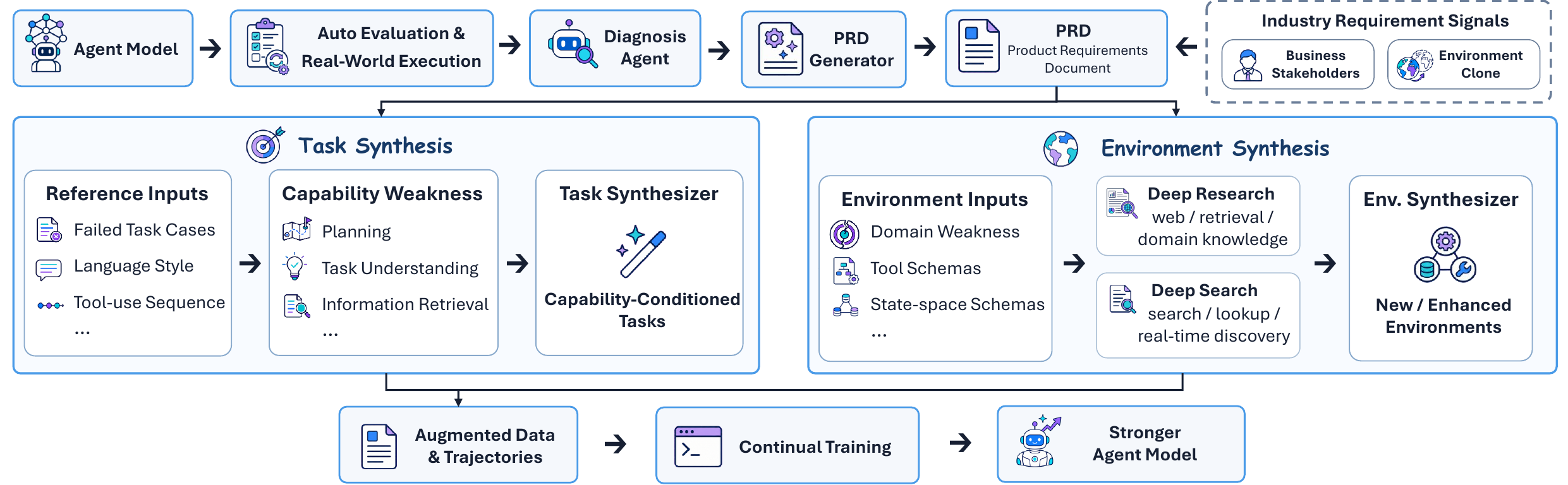}
    \caption{\textbf{Overview of PRD-guided self-evolution.} Evaluation results are transformed into diagnosis reports and then converted into structured Product Requirement Documents (PRDs). PRDs serve as capability specifications that guide environment construction, task synthesis, trajectory generation, and continual learning.}
    \label{fig:PRD_overview}
\end{figure*}

AgentOmnia uses PRDs as a structured specification interface for targeted data synthesis and post-training. Taxonomy coordinates and diagnosis reports identify where and why the model fails, but they do not by themselves specify the environments, tasks, constraints, and evaluation requirements needed to address those failures. A PRD packages these elements into a human-readable artifact that downstream generators can consume consistently and product stakeholders can author or review. The protocol accepts two input paths: internal signals from evaluation and execution failures, and external requirements from industrial applications. Both are normalized into PRDs that define the domains and capabilities to target in a subsequent training round, as illustrated in Figure~\ref{fig:PRD_overview}. We evaluate the internal, diagnosis-driven path in Section~\ref{sec:experiments}; the external path is introduced as a product-facing extension and remains under validation.

\subsection{PRD-Protocol Guidance Generation}

\subsubsection{Diagnosis Report Generation}

Our diagnostic framework proceeds in two stages, moving from failures in individual tasks to broader capability gaps. For each stage, we define a structured protocol that specifies the report format and evidence requirements. Reports that do not pass protocol validation are revised by the LLM until all requirements are met. The complete protocols and prompt templates are provided in Appendix~\ref{app:prd-cases}.

\paragraph{Task-Level Analysis.} 

Task-level analysis examines each task independently. Given the task description, execution trajectory, and evaluation results, the model reconstructs how the task was executed and determines why it failed. The analysis focuses on errors specific to the task, including incorrect tool selection, misunderstanding of tool functionality, invalid argument construction, and improper invocation order.

\paragraph{Capability-Level Analysis.} 

Capability-level analysis aggregates task-level diagnoses to identify recurring failure patterns across tasks. The model groups related failures, summarizes their shared causes, and maps them to broader capability gaps. The resulting analysis directly informs targeted data synthesis.

\subsubsection{PRD Generation}

PRDs convert diagnostic findings and real-world requirements into specifications for targeted data synthesis. They may address domain-specific weaknesses in areas such as finance, law, and software engineering, or general agent capabilities such as multi-step reasoning, tool coordination, and error recovery.

To construct a PRD, the system analyzes recurring failures using the associated task descriptions, execution trajectories, and evaluation results. It compares the observed trajectory with the expected execution process to identify where and why the failure occurred. The PRD then records the target scenario and functional requirements, summarizes the diagnosed cause, and defines corresponding synthesis guidance. For multi-tool composition, for example, the requirements may include correct information transfer between tools and consistent state updates. The resulting guidance may modify the environment by introducing distractors or unavailable entities, or increase task difficulty through deeper dependencies and more complex data transformations.

PRDs are prioritized by failure frequency and severity so that synthesis and training focus on the most important capability gaps. Table~\ref{tab:example_prd} shows a PRD example for multi-tool composition. Operationally, each PRD separates mandatory core specifications, which define the target scenario and required behavior, from optional diagnosis-derived guidance, which records failure evidence and specifies how environment and task synthesis should cover it.

% ------------------------------------------------
\definecolor{PRDCoreBlue}{HTML}{E8F1F8}
\definecolor{PRDGuideGreen}{HTML}{EAF3EC}
\definecolor{PRDGroupGray}{HTML}{F2F3F5}

\newcolumntype{L}[1]{%
    >{\raggedright\arraybackslash}p{#1}%
}

\newcolumntype{B}[1]{%
    >{\raggedright\arraybackslash\bfseries}p{#1}%
}

\renewcommand{\tabularxcolumn}[1]{p{#1}}

\newcolumntype{Y}{%
    >{\raggedright\arraybackslash}X%
}

\begin{table*}[t]
\centering
\scriptsize
\setlength{\tabcolsep}{4pt}
\renewcommand{\arraystretch}{1.18}

\caption{Example of a structured PRD targeting multi-tool composition gaps.}
\label{tab:example_prd}

\begin{tabularx}{\linewidth}{
    B{0.17\linewidth}
    L{0.19\linewidth}
    Y
}
\specialrule{0.9pt}{0pt}{0pt}
\rowcolor{PRDCoreBlue}
\multicolumn{3}{c}{
    \rule{0pt}{2.5ex}
    \textbf{Part I: Core Specifications}
    \hspace{0.5em}
    \textit{(Mandatory)}
}
\\

ID
& \texttt{PRD-TU-042}
& \textbf{Priority:} High
  \qquad
  \textbf{Status:} Active
\\

Scenario
& Multi-tool Composition
& Output of preceding tools serves as input for subsequent invocations.
\\

Requirements
& Cross-tool Integrity
&
  \textbullet\ Map $T_n$ outputs to the corresponding $T_{n+1}$ parameters.
  \par
  \textbullet\ Perform necessary schema transformation and normalization.
  \par
  \textbullet\ Maintain state consistency across the execution trace.
\\

Case Study
& Trip Planning
&
  \textbf{Query:}
  ``Book a flight to NYC and a hotel near the arrival airport.''

  \smallskip

  \textbf{Logic:}
  \texttt{SearchFlight}
  $\rightarrow$
  extract \textit{arrival\_airport}
  $\rightarrow$
  \texttt{SearchHotel}.
\\

\specialrule{0.6pt}{2pt}{0pt}
\rowcolor{PRDGuideGreen}
\multicolumn{3}{c}{
    \rule{0pt}{2.5ex}
    \textbf{Part II: Synthesis Guidance}
    \hspace{0.5em}
    \textit{(Optional)}
}
\\
\specialrule{0.35pt}{0pt}{0pt}
\rowcolor{PRDGroupGray}
\multicolumn{3}{c}{\rule{0pt}{2.3ex}\textbf{Deep Analysis}}
\\

Failure Analysis
& Parameter Binding
& Failure to propagate the \textit{arrival\_airport} entity
  to the hotel search module.
\\

Failure Trajectory
& \textbf{Incorrect}
&
  \texttt{SearchFlight}
  $\rightarrow$
  \texttt{arr\_airport=JFK}
  $\rightarrow$
  \texttt{SearchHotel(loc=DepCity)}
\\

Correct Trajectory
& \textbf{Correct}
&
  \texttt{SearchFlight}
  $\rightarrow$
  \texttt{arr\_airport=JFK}
  $\rightarrow$
  \texttt{SearchHotel(loc=JFK)}
\\

\specialrule{0.45pt}{2pt}{0pt}
\rowcolor{PRDGroupGray}
\multicolumn{3}{c}{\rule{0pt}{2.3ex}\textbf{Synthesis Guidance}}
\\

Environment Synthesis
& Distractor Injection
& Inject multiple candidate identifiers to validate extraction precision.
\\

&
Dynamic Availability
& Simulate entity unavailability to necessitate fallback reasoning.
\\

Task Synthesis
& Dependency Depth
&
  Enforce at least three sequential dependencies, e.g.,
  \texttt{Flight}
  $\rightarrow$
  \texttt{Hotel}
  $\rightarrow$
  \texttt{Ride}.
\\

&
Implicit References
& Replace explicit literals with referential aliases,
  such as ``the Big Apple.''
\\

Extra Information
& Contextual Metadata
& Provide auxiliary domain knowledge or API documentation
  to support task synthesis.
\\

\specialrule{0.9pt}{2pt}{0pt}
\end{tabularx}

\end{table*}

\subsection{PRD-Guided Data Synthesis}

As illustrated in Figure~\ref{fig:PRD_overview}, PRD-guided data synthesis conditions the pipeline introduced in Section~\ref{sec:data-synthesis} on the requirements specified by each PRD. The target scenario and functional requirements define the capability to be improved, while the diagnosis and synthesis guidance specify the failure conditions and data characteristics that should be covered.

\paragraph{Task Synthesis.}

Task synthesis converts the synthesis guidance in each PRD into concrete task instances. Task generation is conditioned on the target scenario, functional requirements, and the contrast between failed and expected trajectories identified through diagnostic analysis. Together, these inputs determine the capability being exercised and the required task difficulty, including tool dependencies, parameter propagation, data transformations, implicit references, and state changes. The synthesized tasks vary in their descriptions and execution contexts while retaining the targeted failure conditions, yielding training examples that directly address the diagnosed capability gap.

\paragraph{Environment Synthesis.}

Environment synthesis constructs executable settings that match the target domain and the conditions associated with the identified capability gap. Based on the PRD, it defines the required tools, data schemas, initial states, and operational constraints, and introduces controlled variations such as distracting entities, unavailable resources, and conflicting states. These environments provide suitable execution contexts for the synthesized tasks and help reduce the gap between the existing post-training distribution and the target scenarios.

\subsection{Iterative Post-Training Loop}

AgentOmnia uses a two-stage post-training process that combines SFT cold-start with RL fine-tuning. Within an iteration, SFT is performed on expert trajectories selected according to prioritized PRDs. This stage corrects recurring execution errors and teaches appropriate patterns of tool use, state transition, and environment interaction. The resulting model provides a stable initialization for RL and reduces the need for costly exploration.

RL fine-tuning combines challenging PRD-generated tasks with samples from the base distribution. The former target diagnosed weaknesses, while the latter help preserve capabilities beyond the targeted scenarios. Together, they improve multi-step execution, tool coordination, state consistency, and failure recovery without over-specializing the model to the latest synthesis round. After post-training, newly observed failures update the PRDs for another round, closing the loop from evaluation and data specification to synthesis and model improvement.

\subsection{Product-Facing Industrial Extension}

Industrial deployments require agents to follow scenario-specific workflows, tool interfaces, data structures, permissions, and business rules underrepresented in general-purpose post-training data. In the product-facing extension, product managers or domain experts can capture these requirements in PRDs that specify the application context, expected behavior, available tools, operational constraints, and evaluation criteria. For example, a PRD for email management may cover message retrieval, information extraction, reply drafting, and email organization, while a database administration PRD may cover record queries, entity resolution, result validation, and state changes subject to permission constraints.

These PRDs provide the requirements for constructing environments and tasks that reflect the target scenario. After deployment, interaction logs, user feedback, and newly observed failures can be incorporated into the PRDs, so that later rounds of data synthesis and training remain aligned with evolving business needs. The framework therefore supports both continued improvement in existing scenarios and adaptation to new applications.
\section{Experiments}
\label{sec:experiments}

We evaluate AgentOmnia at three levels: aggregate performance across four benchmark families, fine-grained OmniaBench diagnostics, and a one-round study of PRD-guided self-evolution. We first describe the baselines, benchmarks, and reporting protocol (Section~\ref{sec:exp-settings}), followed by the results and analyses (Section~\ref{sec:exp-main-results}).

\subsection{Experimental Settings}
\label{sec:exp-settings}

\paragraph{Baseline Models.}
We compare AgentOmnia with the four groups shown in Table~\ref{tab:benchmark-comparison}:
\begin{itemize}
    \item \textbf{Proprietary General-Purpose Models:} GPT-5.5 (xhigh), Claude Opus 4.7, Gemini 3.5 Flash, and Qwen3.7-Max~\cite{openai2026gpt55,anthropic2026claudeopus47,googledeepmind2026gemini35flash,qwenteam2026qwen37max}. These models serve as frontier reference points rather than like-for-like comparisons.
    \item \textbf{Open-Weight General-Purpose Models:} DeepSeek-V4-Pro-Max~\cite{deepseekai2026deepseekv4}, Qwen3-30B-A3B-Thinking-2507~\cite{qwenteam2025qwen330b2507}, Qwen3-235B-A22B-Thinking-2507~\cite{qwenteam2025qwen3235b2507}, Qwen3.5-35B-A3B~\cite{qwen35blog}, and Qwen3.6-35B-A3B~\cite{qwenteam2026qwen36}. Qwen3-30B-A3B-Thinking-2507 is the foundation checkpoint from which AgentOmnia is trained.
    \item \textbf{Open-Weight Agentic Post-Trained Models:} MUA-RL~\cite{zhao2025muarl}, Toucan~\cite{xu2025toucan}, the Qwen3-based Nex-N1 variants~\cite{cai2025nexn1}, AgentSkiller~\cite{sun2026agentskiller}, Arctic-AWM~\cite{wang2026agentworldmodel}, EnvScaler-Qwen3-8B~\cite{song2026envscaler}, Nex-N2-Mini~\cite{nexagi2026nexn2mini}, and Agents-A1~\cite{bai2026agentsa1}. This group provides the most direct comparison with other agentic post-training recipes, although the foundation models and parameter scales still differ.
    \item \textbf{Source-Reported Agentic Post-Trained Models:} AgentScaler~\cite{fang2026towards}, AutoForge~\cite{cai2025autoforge}, Qwen3-SE from ScaleEnv~\cite{tu2026scaleenv}, and Agent-World~\cite{dong2026agentworld}. Because no matched checkpoint or evaluation setup is available, their published values are retained only as source-reported references.
\end{itemize}

\paragraph{Evaluation Benchmarks.}
We evaluate AgentOmnia on the companion OmniaBench and three external benchmarks:
\begin{itemize}
    \item \textbf{Companion diagnostic benchmark.} OmniaBench~\cite{shen2026omniabenchbenchmarkinggeneralai} comprises 1,431 tasks and a fixed 644-task challenging subset. Its tasks are deduplicated against the AgentOmnia post-training corpus and manually curated for solvability and evaluation validity. We use the challenging subset for aggregate comparison and fine-grained diagnosis, as it retains broad scenario coverage at a lower evaluation cost. Its taxonomy-aligned annotations support analysis across application splits, capability dimensions, and atomic difficulty factors, providing diagnostic signals for PRD-guided self-evolution.
    \item \textbf{External benchmarks.} $\tau^2$-Bench evaluates tool-agent-user interaction across Airline, Retail, and Telecom~\cite{barres2025tau}. DeepPlanning measures long-horizon planning in Shopping and Travel settings~\cite{zhang2026deepplanning}, while VitaBench covers Cross-domain, Delivery, In-store, and OTA life-service tasks~\cite{he2025vitabench}. Together, they provide external comparisons across distinct interaction protocols and application settings.
\end{itemize}

\paragraph{Implementation Details.}
We initialize AgentOmnia from Qwen3-30B-A3B-Thinking-2507~\cite{qwenteam2025qwen330b2507} and conduct a two-stage post-training procedure. During cold-start SFT, we optimize the model with the standard cross-entropy loss, masking tool-response tokens since these are supplied by the environment rather than generated by the model. Training uses AdamW with a global batch size of 128 and a maximum sequence length of 64K tokens. The learning rate follows cosine decay with a 2\% warmup ratio, decreasing from $2\times10^{-6}$ to $1\times10^{-6}$. During RL, we adopt GRPO~\cite{shao2024deepseekmath} with a constant learning rate of $1\times10^{-6}$. Each batch contains 64 tasks, with 8 rollouts per task sampled at temperature 1.0 and top-$p$ 1.0. The maximum sequence length remains 64K tokens. We further apply asymmetric clipping with $\epsilon_{\mathrm{low}}=0.2$ and $\epsilon_{\mathrm{high}}=0.28$, following the clip-higher strategy of~\cite{yu2026dapo}.

\paragraph{Reporting Protocol.}
We use the environments, tasks, and trajectories described in Section~\ref{sec:data-synthesis}, together with 53K SFT samples and 5K RL tasks described in Section~\ref{sec:posttraining}. SFT learns from curated verified trajectories. Agentic RL instead performs online rollouts on executable tasks in their associated environments, using task-specific rule- and rubric-based rewards; RCRL supplies a verified-trajectory prefix only when a challenging group requires curriculum support. For self-evolution, OmniaBench diagnostics define the PRD-based targets, while the three external benchmarks are excluded from target construction and used only to assess transfer.

Unless otherwise noted, external-benchmark evaluation configurations follow the June 2026 leaderboard snapshot, which uses DeepSeek-V4-Flash with thinking disabled as the user simulator. DeepPlanning uses the pinned Qwen-Agent adapter with a high LLM-call budget and reports Shopping / Match and Travel / Comp / CS / PS. For open-weight general-purpose and agentic post-trained baselines, Table~9 prioritizes our local reruns under a unified evaluation setup, while retaining official paper or model-card results in dark gray for reference. Proprietary models use the corresponding leaderboard snapshot, and source-only agentic baselines report official results only. This design reduces confounding from deployment configurations, benchmark and framework revisions, and user-simulation models.

\subsection{Main Results}
\label{sec:exp-main-results}

\providecommand{\reportedscore}[2]{\makecell[c]{#1\\[-0.15em]{\tiny\textcolor{MutedText}{(#2)}}}}
\providecommand{\derivedreportedscore}[2]{\makecell[c]{#1\\[-0.15em]{\tiny\textcolor{MutedText}{(#2$^\dagger$)}}}}
\providecommand{\sourcereportedscore}[1]{\textcolor{MutedText}{#1}}
\providecommand{\derivedsourcereportedscore}[1]{\textcolor{MutedText}{#1$^\dagger$}}
\providecommand{\agenttablegroupsep}{\specialrule{0.5pt}{1.1pt}{0pt}}
\providecommand{\agenttablebottomrule}{\specialrule{0.9pt}{0.35pt}{0pt}}
\providecommand{\agenttablegroupstrut}{\rule[-0.55ex]{0pt}{2.7ex}}
\providecommand{\agenttabletwolinestrut}{\vphantom{\makecell[c]{00.00\\[-0.15em]{\tiny\itshape (00.00$^\dagger$)}}}}
\providecommand{\gain}[1]{{\tiny\textcolor{GainGreen}{\hspace{0.15em}(+#1)}}}

\begin{table*}[!t]
\centering
\scriptsize
\setlength{\tabcolsep}{3.1pt}
\renewcommand{\arraystretch}{1.16}
\caption{\textbf{Comprehensive benchmark comparison.} Black upright values are unified-evaluation or leaderboard-snapshot results; dark-gray values are source-reported and shown in parentheses when paired. Among agentic post-trained models and AgentOmnia, bold and underlined upright values mark the highest and second-highest eligible scores, with ties sharing the same style. Source-reported values are excluded from ranking; the highest is gray-bolded only when it exceeds all eligible upright scores. Avg. is the macro-average across the four benchmarks and is shown only when complete. OmniaBench uses Pass@1 on the fixed 644-task challenging subset. $\ddagger$: foundation checkpoint; S/T: Shopping/Travel; $\dagger$: unweighted track mean; ``--'': unavailable results.}
\label{tab:benchmark-comparison}
\begin{adjustbox}{max width=\linewidth}
\begin{tabular}{l c c cccc ccccccc ccccc}
\arrayrulecolor{black}
\toprule[0.9pt]
\multirow{2}{*}{\textbf{Model}} &
\multirow{2}{*}{\textbf{Avg.}} &
\multirow{2}{*}{\makecell{\textbf{OmniaBench}\\{\tiny\normalfont (challenging set)}}} &
\multicolumn{4}{c}{\textbf{$\tau^2$-Bench}} &
\multicolumn{7}{c}{\textbf{DeepPlanning}} &
\multicolumn{5}{c}{\textbf{VitaBench}} \\
\cmidrule(lr){4-7}\cmidrule(lr){8-14}\cmidrule(lr){15-19}
& & & \textbf{Avg.} & \textbf{Airline} & \textbf{Retail} & \textbf{Telecom} &
\textbf{Avg.} & \textbf{S~Avg.} & \textbf{S~Match} &
\textbf{T~Avg.} & \textbf{T~Comp.} & \textbf{T~CS} & \textbf{T~PS} &
\textbf{Avg.} & \textbf{Cross} & \textbf{Delivery} & \textbf{In-store} & \textbf{OTA} \\
\agenttablegroupsep
\rowcolor{ProprietaryBand}
\multicolumn{19}{c}{\agenttabletwolinestrut\textbf{Proprietary General-Purpose Models}} \\
\agenttabletwolinestrut GPT-5.5 (xhigh) & 68.56 & 57.61 & 86.94 & 81.00 & 82.24 & \reportedscore{97.59}{98.0} & 72.50 & 77.50 & 92.24 & 67.50 & 85.67 & 98.54 & 72.79 & 57.19 & 39.88 & 64.75 & 67.75 & 56.38 \\
\agenttabletwolinestrut Claude Opus 4.7 (Thinking) & 56.80 & 54.19 & 82.36 & 81.50 & 83.99 & 81.58 & 36.31 & 58.33 & 86.22 & 14.29 & 89.57 & 94.11 & 85.01 & 54.34 & 40.00 & 63.13 & 61.00 & 53.25 \\
\agenttabletwolinestrut Gemini 3.5 Flash & 53.99 & 45.65 & 84.60 & 81.00 & 80.04 & 92.76 & 29.99 & 42.50 & 77.39 & 17.48 & 73.25 & 79.99 & 66.50 & 55.72 & 42.00 & 63.50 & 65.13 & 52.25 \\
\agenttabletwolinestrut Qwen3.7-Max & 59.79 & 49.69 & 86.05 & 77.00 & 81.80 & 99.34 & 50.42 & 51.67 & 84.10 & 49.17 & 89.20 & 93.26 & 85.10 & 53.00 & 36.50 & 64.13 & 64.13 & 47.25 \\
\agenttablegroupsep
\rowcolor{OpenWeightBand}
\multicolumn{19}{c}{\agenttabletwolinestrut\textbf{Open-Weight General-Purpose Models}} \\
\agenttabletwolinestrut DeepSeek-V4-Pro-Max & 61.09 & 54.50 & 83.47 & 82.00 & 83.55 & 84.87 & 46.67 & 58.33 & 85.71 & 35.00 & 82.56 & 87.48 & 77.62 & 59.72 & 46.13 & 67.75 & 71.13 & 53.87 \\
\agenttabletwolinestrut Qwen3-30B-A3B-Thinking-2507$^\ddagger$ & 22.86 & 9.16 & \derivedreportedscore{55.98}{47.70} & \reportedscore{67.50}{58.0} & \reportedscore{67.11}{58.8} & \reportedscore{33.33}{26.3} & 5.00 & 10.00 & 47.15 & 0.00 & 19.87 & 33.05 & 6.69 & 21.28 & 6.63 & 35.25 & 26.00 & 17.25 \\
% Qwen3-235B aggregate/DeepPlanning/Vita values follow the Qwen3.5 model card; per-domain TAU2 values follow the Qwen3-235B model card.
\agenttabletwolinestrut Qwen3-235B-A22B-Thinking-2507 & 34.17 & 20.03 & \reportedscore{68.67}{58.5} & \reportedscore{65.00}{58.0} & \reportedscore{76.75}{71.9} & \reportedscore{64.25}{45.6} & \reportedscore{14.58}{17.1} & 29.17 & 72.26 & 0.00 & 28.41 & 40.44 & 16.39 & \reportedscore{33.38}{31.6} & 20.75 & 47.50 & 40.75 & 24.50 \\
\agenttabletwolinestrut Qwen3.5-35B-A3B & 36.92 & 27.95 & \reportedscore{69.89}{81.2} & 48.50 & 66.45 & 94.74 & \reportedscore{18.16}{22.8} & 34.17 & 73.53 & 2.14 & 65.51 & 61.05 & 69.96 & \reportedscore{31.69}{31.9} & 18.63 & 40.75 & 37.75 & 29.63 \\
\agenttabletwolinestrut Qwen3.6-35B-A3B & 45.15 & 37.27 & 87.27 & 82.00 & 81.14 & 98.68 & \reportedscore{24.17}{25.9} & 44.17 & 79.62 & 4.17 & 60.46 & 59.41 & 61.51 & \reportedscore{31.88}{35.6} & 16.63 & 43.63 & 36.13 & 31.13 \\
\agenttablegroupsep
\rowcolor{AgenticBand}
\multicolumn{19}{c}{\agenttabletwolinestrut\textbf{Open-Weight Agentic Post-Trained Models}} \\
\agenttabletwolinestrut MUA-RL-32B & 23.82 & 14.13 & \derivedreportedscore{53.93}{47.00} & \reportedscore{46.00}{45.4} & \reportedscore{66.89}{67.3} & \reportedscore{48.90}{28.3} & 6.67 & 13.33 & 49.27 & 0.00 & 18.84 & 32.85 & 4.83 & 20.56 & 7.00 & 31.62 & 25.75 & 17.88 \\
\agenttabletwolinestrut Toucan-Qwen2.5-32B & 22.09 & 17.39 & \reportedscore{43.28}{31.60} & \reportedscore{34.00}{22.00} & \reportedscore{60.31}{52.60} & \reportedscore{35.53}{20.20} & 4.17 & 8.33 & 53.10 & 0.00 & 21.71 & 31.43 & 12.00 & 23.50 & 11.13 & 31.88 & 34.00 & 17.00 \\
\agenttabletwolinestrut Qwen3-30B-A3B-Nex-N1 & 23.92 & 10.40 & \reportedscore{70.54}{65.3} & 57.00 & 60.09 & \textbf{94.52} & 1.67 & 3.33 & 39.36 & 0.00 & 13.78 & 23.61 & 3.97 & 13.06 & 3.00 & 28.00 & 15.13 & 6.13 \\
\agenttabletwolinestrut Qwen3-32B-Nex-N1 & 31.37 & 23.60 & \reportedscore{72.90}{72.1} & 54.00 & \underline{71.49} & \underline{93.20} & 7.50 & 15.00 & 54.69 & 0.00 & 19.61 & 26.89 & 12.35 & 21.47 & 10.63 & 33.50 & 27.13 & 14.63 \\
\agenttabletwolinestrut AgentSkiller-14B & 26.13 & 17.70 & \reportedscore{58.51}{79.1} & \reportedscore{49.00}{56.0} & \reportedscore{70.18}{\textbf{77.2}} & \reportedscore{56.36}{91.2} & 5.20 & 10.00 & 30.31 & \underline{0.40} & 26.05 & 34.35 & 17.76 & 23.09 & 8.00 & 35.75 & 31.62 & 17.00 \\
\agenttabletwolinestrut Arctic-AWM-14B & 20.19 & 12.27 & \reportedscore{42.60}{39.03} & \reportedscore{37.00}{31.50} & \reportedscore{45.18}{63.60} & \reportedscore{45.61}{17.76} & 7.92 & 15.83 & 57.39 & 0.00 & 17.76 & 28.02 & 7.50 & 17.97 & 6.75 & 26.88 & 24.00 & 14.25 \\
\agenttabletwolinestrut EnvScaler-Qwen3-8B & 18.26 & 12.11 & 37.36 & 39.50 & 49.12 & 23.46 & 7.92 & 15.83 & 57.82 & 0.00 & 18.28 & 28.06 & 8.50 & 15.66 & 3.13 & 25.75 & 23.00 & 10.75 \\
\agenttabletwolinestrut Nex-N2-Mini & 40.06 & 29.35 & 75.42 & \underline{67.50} & 66.01 & 92.76 & \textbf{22.29} & \textbf{38.33} & \textbf{77.98} & \textbf{6.25} & \textbf{69.38} & \textbf{70.92} & \textbf{67.83} & 33.19 & 15.13 & \underline{48.25} & 41.38 & 28.00 \\
\agenttabletwolinestrut Agents-A1 & \underline{41.52} & \underline{30.28} & \reportedscore{\textbf{78.96}}{\textbf{79.81}} & \textbf{74.50} & \textbf{74.67} & 87.72 & \underline{19.17} & \textbf{38.33} & \underline{75.49} & 0.00 & \underline{46.69} & 39.57 & \underline{53.81} & \reportedscore{\textbf{37.66}}{\textbf{38.75}} & \textbf{24.19} & 46.38 & \underline{45.25} & \textbf{34.81} \\
\agenttablegroupsep
\rowcolor{SourceOnlyBand}
\multicolumn{19}{c}{\agenttabletwolinestrut\textbf{Source-Reported Agentic Post-Trained Models}} \\
\agenttabletwolinestrut AgentScaler-30B-A3B & -- & -- & \sourcereportedscore{62.5} & \sourcereportedscore{60.0} & \sourcereportedscore{70.2} & \sourcereportedscore{55.3} & -- & -- & -- & -- & -- & -- & -- & -- & -- & -- & -- & -- \\
\agenttabletwolinestrut AutoForge-30B-A3B & -- & -- & \derivedsourcereportedscore{71.03} & \sourcereportedscore{62.0} & \sourcereportedscore{74.8} & \sourcereportedscore{76.3} & -- & -- & -- & -- & -- & -- & -- & \derivedsourcereportedscore{35.50} & \sourcereportedscore{17.5} & \sourcereportedscore{46.0} & \textbf{\sourcereportedscore{54.5}} & \sourcereportedscore{24.0} \\
\agenttabletwolinestrut Qwen3-SE-32B & -- & -- & \derivedsourcereportedscore{47.50} & \sourcereportedscore{48.0} & \sourcereportedscore{63.6} & \sourcereportedscore{30.9} & -- & -- & -- & -- & -- & -- & -- & \derivedsourcereportedscore{22.28} & \sourcereportedscore{10.8} & \sourcereportedscore{31.3} & \sourcereportedscore{34.5} & \sourcereportedscore{12.5} \\
\agenttabletwolinestrut Agent-World-14B & -- & -- & \sourcereportedscore{65.4} & \sourcereportedscore{52.0} & \sourcereportedscore{74.5} & \sourcereportedscore{56.1} & -- & -- & -- & -- & -- & -- & -- & -- & -- & -- & -- & -- \\
\agenttablegroupsep
\rowcolor{AgentOrange!80}
\agenttabletwolinestrut\textbf{AgentOmnia-30B-A3B (Ours)} & \textbf{41.69} & \textbf{37.11} & \underline{75.79} & \underline{67.50} & 70.39 & 89.47 & 16.25 & \underline{32.50} & 67.62 & 0.00 & 35.25 & \underline{41.86} & 28.61 & \underline{37.62} & \underline{20.37} & \textbf{49.63} & \textbf{48.63} & \underline{31.88} \\
\agenttablebottomrule
\end{tabular}
\end{adjustbox}
\vspace{-0.5em}
\end{table*}

\paragraph{Overall Benchmark Performance.}
AgentOmnia scores 37.11\% on the OmniaBench challenging subset and achieves a macro-average of 41.69\% across the four benchmarks (Table~\ref{tab:benchmark-comparison}), compared with 22.86\% for its Qwen3-30B-A3B-Thinking-2507 foundation checkpoint. The gains span OmniaBench (+27.95 points), $\tau^2$-Bench (+19.81), DeepPlanning (+11.25), and VitaBench (+16.34), rather than concentrating on one evaluation format. The larger OmniaBench gain should be interpreted in light of its taxonomy alignment and partial reuse of the synthesis methodology, although its task instances are deduplicated from the training corpus. Improvements on all three independently developed external benchmarks provide complementary evidence that the effect extends beyond the companion suite. Among comparable 30--32B agentic models built on Qwen3 or earlier foundations, AgentOmnia achieves the strongest OmniaBench score and four-benchmark average. It also exceeds Qwen3-235B-A22B-Thinking-2507 on all four benchmarks. 

The comparison with more recent Qwen3.5-based agentic models is mixed but competitive. AgentOmnia retains the strongest OmniaBench result and the highest four-benchmark average, narrowly ahead of Agents-A1 (41.69 vs.\ 41.52) and more clearly ahead of Nex-N2-Mini (40.06). Agents-A1 leads on $\tau^2$-Bench and DeepPlanning and is effectively tied on VitaBench (37.66 vs.\ 37.62), while Nex-N2-Mini also leads on DeepPlanning. This ordering is consistent with the 13.16-point DeepPlanning gap between Qwen3.5-35B-A3B and our Qwen3 foundation checkpoint, suggesting sensitivity to foundation-model reasoning and planning strength. Nevertheless, AgentOmnia improves its own foundation checkpoint by 11.25 points on DeepPlanning. At the same time, Qwen3.6-35B-A3B remains ahead by 3.46 points on the four-benchmark average, and wider gaps remain to DeepSeek-V4-Pro-Max and proprietary frontier systems. Taken together, these results support the value of full-scenario post-training at the present model scale, while also highlighting substantial headroom from stronger foundation models and greater inference-time capacity.

\begin{figure*}[!t]
\centering
\includegraphics[width=\linewidth]{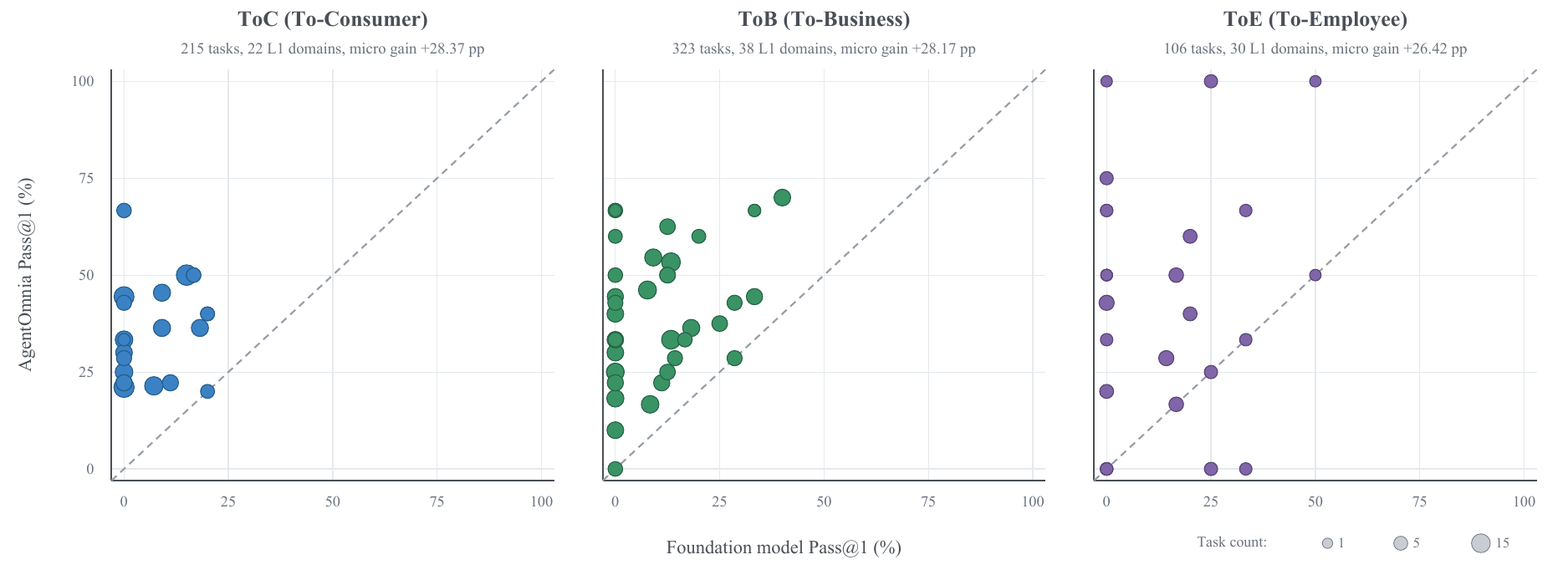}
\caption{\textbf{Level-1 domain performance on OmniaBench.} Each point represents one of the 90 level-1 domains in the challenging subset (some overlap due to similar scores), and its area reflects the number of tasks. The axes show Pass@1 for Qwen3-30B-A3B-Thinking-2507 and AgentOmnia-30B-A3B. A task passes only when its route-level score equals 1.0. Points above the dashed diagonal favor AgentOmnia, while the panel annotations give the micro-averaged gain for each application split.}
\label{fig:omnia-domain-l1-scatter}
\end{figure*}

\paragraph{Full-Scenario Improvements on OmniaBench.}
We use the three taxonomy views of OmniaBench to isolate the effect of AgentOmnia post-training relative to its foundation checkpoint. Table~\ref{tab:OmniaBench-domain} summarizes the three application splits, while Figure~\ref{fig:omnia-domain-l1-scatter} resolves the comparison across all 90 level-1 domains. Tables~\ref{tab:OmniaBench-capability} and~\ref{tab:OmniaBench-difficulty} report the capability and atomic-difficulty views. All results use Pass@1 (\%) to measure task success.

At the split level, AgentOmnia improves by 28.37, 28.17, and 26.42 points on ToC, ToB, and ToE, respectively, indicating that the gains are not confined to a particular application scenario. This pattern holds at finer granularity as well: among all 90 level-1 domains, 76 (84\%) improve, 12 remain unchanged, and only 2 decline. The improvements are similarly broad-based across the ten capability dimensions, ranging from 25.00 points for Code \& Programmatic Operations to 36.77 points for Reliability \& Safety, and across all eight atomic-difficulty factors, which improve with gains ranging from 21.43 points for Multi-source Inconsistency to 47.73 points for Long-context and Multi-artifact Evidence. Taken together, these results confirm that the overall OmniaBench gain is distributed across domains, capabilities, and difficulty factors rather than driven by improvement in a single category. The relatively small gain on Multi-source Inconsistency and the low absolute score on Ambiguous Goal and Contextual Constraints identify concrete targets for further synthesis.

\begin{table*}[!t]
\centering
\scriptsize
\setlength{\tabcolsep}{5pt}
\renewcommand{\arraystretch}{1.12}
\caption{OmniaBench results across ToC, ToB, and ToE.}
\label{tab:OmniaBench-domain}
\begin{tabularx}{0.78\linewidth}{@{}>{\raggedright\arraybackslash}X*{3}{>{\centering\arraybackslash}p{0.145\linewidth}}@{}}
\arrayrulecolor{black}
\toprule[0.9pt]
\textbf{Model} & \mbox{\textbf{ToC} {\normalfont (To-Consumer)}} & \mbox{\textbf{ToB} {\normalfont (To-Business)}} & \mbox{\textbf{ToE} {\normalfont (To-Employee)}} \\
\agenttablegroupsep
Qwen3-30B-A3B-Thinking-2507 & 6.51 & 9.91 & 12.26 \\
\agenttablegroupstrut AgentOmnia-30B-A3B & 34.88\gain{28.37} & 38.08\gain{28.17} & 38.68\gain{26.42} \\
\agenttablebottomrule
\end{tabularx}
\vspace{0.25em}
\end{table*}

\begin{table*}[!t]
\centering
\scriptsize
\setlength{\tabcolsep}{1.6pt}
\renewcommand{\arraystretch}{1.18}
\caption{OmniaBench results by capability dimension.}
\label{tab:OmniaBench-capability}
\begin{adjustbox}{max width=\linewidth}
\begin{tabular}{>{\raggedright\arraybackslash}p{0.25\linewidth}*{10}{c}}
\arrayrulecolor{black}
\toprule[0.9pt]
\textbf{Model}
& \textbf{\makecell{Task\\Understanding}}
& \textbf{\makecell{Information\\Gathering}}
& \textbf{\makecell{Planning \& Decision\\Making}}
& \textbf{\makecell{State\\Management}}
& \textbf{\makecell{Tool\\Use}}
& \textbf{\makecell{Code \& Programmatic\\Operations}}
& \textbf{\makecell{Data\\Analysis}}
& \textbf{\makecell{Office \& Document\\Handling}}
& \textbf{\makecell{Interactive\\Collaboration}}
& \textbf{\makecell{Reliability \&\\Safety}} \\
\agenttablegroupsep
Qwen3-30B-A3B-Thinking-2507
& 9.23 & 16.00 & 18.73 & 15.76 & 17.30 & 23.86 & 21.86 & 14.65 & 16.30 & 13.97 \\
\agenttablegroupstrut AgentOmnia-30B-A3B
& 44.62\gain{35.39} & 52.00\gain{36.00} & 51.17\gain{32.44} & 48.23\gain{32.47} & 53.16\gain{35.86} & 48.86\gain{25.00} & 50.82\gain{28.96} & 49.04\gain{34.39} & 49.46\gain{33.16} & 50.74\gain{36.77} \\
\agenttablebottomrule
\end{tabular}
\end{adjustbox}
\vspace{0.25em}
\end{table*}

\begin{table*}[!t]
\centering
\fontsize{6.6pt}{7.5pt}\selectfont
\setlength{\tabcolsep}{1.8pt}
\renewcommand{\arraystretch}{1.12}
\caption{OmniaBench results by atomic-difficulty factor.}
\label{tab:OmniaBench-difficulty}
\begin{adjustbox}{max width=\linewidth}
\begin{tabular}{>{\raggedright\arraybackslash}p{0.23\linewidth}*{8}{c}}
\arrayrulecolor{black}
\toprule[0.8pt]
\textbf{Model}
& \textbf{\makecell{Ambig.\\Goal \& Ctx.}}
& \textbf{\makecell{Tool \& Param.\\Ground.}}
& \textbf{\makecell{Struct.-Info\\Complex.}}
& \textbf{\makecell{Long-Ctx.\\\& Evidence}}
& \textbf{\makecell{Dynamic\\Planning}}
& \textbf{\makecell{Multi-Source\\Incons.}}
& \textbf{\makecell{Disclosure \&\\State Evol.}}
& \textbf{\makecell{Risk, Reliab.\\\& Clarif.}} \\
\specialrule{0.45pt}{1.1pt}{0pt}
Qwen3-30B-A3B-Thinking-2507 & 13.58 & 20.41 & 10.53 & 9.09 & 19.64 & 28.57 & 14.63 & 22.00 \\
\agenttablegroupstrut AgentOmnia-30B-A3B & 37.04\gain{23.46} & 53.06\gain{32.65} & 47.37\gain{36.84} & 56.82\gain{47.73} & 50.00\gain{30.36} & 50.00\gain{21.43} & 51.22\gain{36.59} & 54.00\gain{32.00} \\
\specialrule{0.8pt}{0.35pt}{0pt}
\end{tabular}
\end{adjustbox}
\vspace{0.25em}
\end{table*}

\subsection{PRD-Guided Self-Evolution}
\label{sec:exp-prd-evolution}

\paragraph{Targeting of PRD-Guided Synthesis.}
We evaluate PRD-guided self-evolution by examining whether PRD guidance shifts synthesized data toward diagnosed weaknesses and whether training on these data improves the model beyond the diagnostic benchmark. The current experiment covers one evolution round, in which OmniaBench failures are converted into PRD-based targets that guide the construction of 121 environments and 804 tasks.

\begin{figure*}[!t]
    \centering
    \includegraphics[width=.95\linewidth]{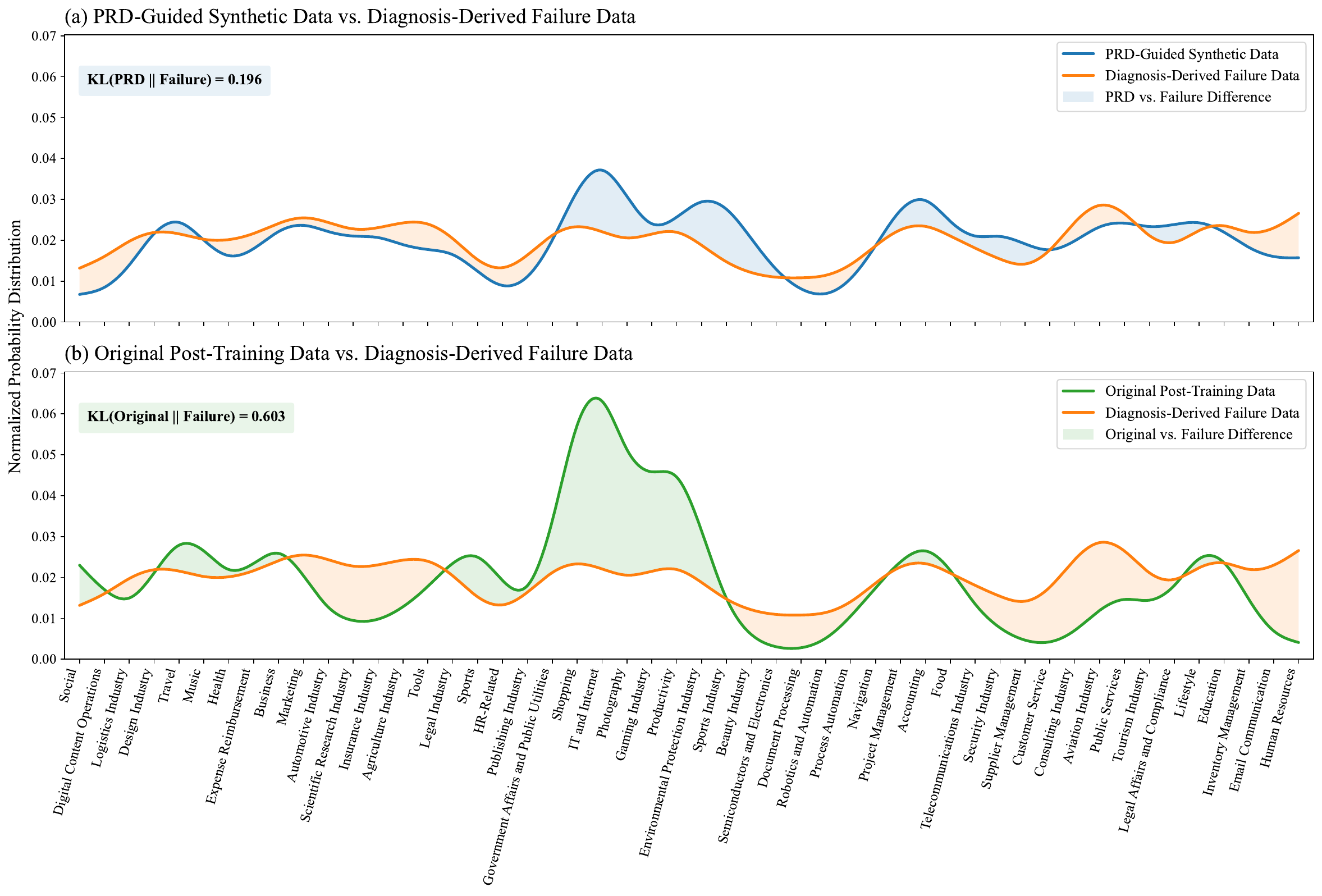}
    \caption{\textbf{Domain alignment of PRD-guided synthesis.} Normalized distributions over the displayed top 50 level-1 domains for PRD-guided synthetic data, original post-training data, and failed evaluation tasks used for diagnosis. The KL divergence to the failure distribution is 0.196 for PRD-guided data and 0.603 for the original post-training data.}
    \label{fig:KL}
\end{figure*}

Figure~\ref{fig:KL} compares the level-1 domain distributions of the PRD-guided synthetic data, the original post-training data, and the failed evaluation tasks used for diagnosis. The PRD-guided distribution is more closely aligned with the failure distribution, with $D_{\mathrm{KL}}(P_{\mathrm{PRD}}\|P_{\mathrm{failure}})=0.196$, compared with $D_{\mathrm{KL}}(P_{\mathrm{original}}\|P_{\mathrm{failure}})=0.603$. This comparison indicates that PRDs steer synthesis toward the domains identified during diagnosis. It evaluates target alignment rather than downstream model improvement, which is examined next.

\begin{table*}[!t]
\centering
\scriptsize
\setlength{\tabcolsep}{3.1pt}
\renewcommand{\arraystretch}{1.16}
\caption{\textbf{Preliminary evaluation of PRD-guided self-evolution.} Results compare AgentOmnia before and after one evolution round. OmniaBench uses the challenging subset that provides the diagnostic signals for PRD target construction, while $\tau^2$-Bench, DeepPlanning, and VitaBench are excluded from target construction and used to assess transfer. Ext. Avg. is the macro-average of these three external benchmarks.}
\label{tab:self-evolution-1plus3}
\begin{adjustbox}{max width=\linewidth}
\begin{tabular}{l c c cccc ccccccc ccccc}
\arrayrulecolor{black}
\toprule[0.9pt]
\multirow{2}{*}{\textbf{Model}} &
\multirow{2}{*}{\makecell{\textbf{OmniaBench}\\{\tiny\normalfont (challenging set)}}} &
\multirow{2}{*}{\textbf{Ext. Avg.}} &
\multicolumn{4}{c}{\textbf{$\tau^2$-Bench}} &
\multicolumn{7}{c}{\textbf{DeepPlanning}} &
\multicolumn{5}{c}{\textbf{VitaBench}} \\
\cmidrule(lr){4-7}\cmidrule(lr){8-14}\cmidrule(lr){15-19}
& & & \textbf{Avg.} & \textbf{Airline} & \textbf{Retail} & \textbf{Telecom} &
\textbf{Avg.} & \textbf{S~Avg.} & \textbf{S~Match} &
\textbf{T~Avg.} & \textbf{T~Comp.} & \textbf{T~CS} & \textbf{T~PS} &
\textbf{Avg.} & \textbf{Cross} & \textbf{Delivery} & \textbf{In-store} & \textbf{OTA} \\
\agenttablegroupsep
\agenttablegroupstrut AgentOmnia-30B-A3B & 37.11 & 43.22 & 75.79 & 67.50 & 70.39 & \textbf{89.47} & 16.25 & 32.50 & \textbf{67.62} & \textbf{0.00} & \textbf{35.25} & \textbf{41.86} & 28.61 & 37.62 & 20.37 & 49.63 & \textbf{48.63} & 31.88 \\
\agenttablegroupsep
\agenttablegroupstrut AgentOmnia-30B-A3B-evo & \textbf{38.49} & \textbf{44.14} & \textbf{77.48} & \textbf{69.50} & \textbf{73.46} & \textbf{89.47} & \textbf{17.08} & \textbf{34.17} & 67.47 & \textbf{0.00} & 35.13 & 41.25 & \textbf{29.00} & \textbf{37.87} & \textbf{21.11} & \textbf{49.88} & 47.75 & \textbf{32.75} \\
\agenttablebottomrule
\end{tabular}
\end{adjustbox}
\end{table*}

\paragraph{Model Improvement and Transfer.}
Table~\ref{tab:self-evolution-1plus3} reports the downstream effect of training on the PRD-guided data. On the OmniaBench challenging subset, AgentOmnia-30B-A3B-evo improves from 37.11\% to 38.49\%, a gain of 1.38 percentage points. The external-benchmark average increases from 43.22\% to 44.14\%, with gains of 1.69 points on $\tau^2$-Bench, 0.83 on DeepPlanning, and 0.25 on VitaBench. Given the limited scale of this one-round study, these modest gains provide preliminary evidence that diagnosis-guided synthesis can improve aggregate performance and transfer beyond the benchmark used for diagnosis. Several fine-grained metrics remain unchanged or decrease, and substantially more data and repeated evolution rounds are needed to characterize the attainable gains. We therefore plan to scale diagnosis and PRD-guided synthesis and to examine settings farther from the original training distribution, including industrial applications in which PRDs are authored by product or business teams or derived from product documentation and representative user queries.

\section{Related Work}
\label{sec:related}

\paragraph{LLM-Based Autonomous Agents.}
LLMs have demonstrated substantial reasoning ability under chain-of-thought prompting, zero-shot reasoning, verifier-guided reasoning, and systematic reasoning benchmarks~\cite{wei2022chain,kojima2022large,2021cobbegsm8k,2023fureasoninghub}. Building on these capabilities, agents interleave reasoning, action, and observation during task execution. ReAct~\cite{yao2022react} combines reasoning traces with actions in environments. Toolformer~\cite{schick2023toolformer}, ToolLLM~\cite{qin2023toolllm}, Gorilla~\cite{2024patilgorilla}, ToolTalk~\cite{2023farntooltalk}, and API-Bank~\cite{li2023apibank} investigate API selection, function calling, and conversational tool use. Recent surveys~\cite{wang2023survey,xi2023rise} further organize agent systems around planning, memory, tool use, feedback, and interaction. Together, these works establish core patterns for agent reasoning and tool interaction. AgentOmnia focuses on scaling agents across full-scenario applications, where domain coverage, capability diagnosis, stateful execution, and training signals must be organized jointly.

\paragraph{Agent Benchmarks and Interactive Environments.}
Interactive agent benchmarks cover web, GUI, mobile, and desktop environments. WebShop~\cite{yao2022webshop}, AgentBench~\cite{liu2023agentbench}, Mind2Web~\cite{deng2023mind2web}, WebArena~\cite{zhou2023webarena}, VisualWebArena~\cite{koh2024visualwebarena}, AndroidWorld~\cite{rawles2024androidworld}, and OSWorld~\cite{xie2024osworld} evaluate capabilities such as web interaction, visual grounding, mobile control, and operating-system manipulation. Other benchmarks target professional or domain-specific workflows. SWE-bench~\cite{jimenez2023swebench} evaluates software issue resolution, while WorkArena~\cite{drouin2024workarena}, OfficeBench~\cite{wang2024officebench}, CRMArena~\cite{huang2024crmarena}, SpreadsheetBench~\cite{ma2024spreadsheetbench}, and AppWorld~\cite{trivedi2024appworld} cover enterprise software, office automation, customer relationship management, spreadsheets, and API ecosystems. Benchmarks such as $\tau$-bench~\cite{yao2024tau}, DeepPlanning~\cite{zhang2026deepplanning}, VitaBench~\cite{he2025vitabench}, BFCL-v4~\cite{bfcl2024}, and Toolathlon~\cite{li2025toolathlon} further assess service-domain interaction, long-horizon planning, life-service tasks, function calling, and diverse tool execution. These benchmarks make agent evaluation increasingly realistic, but their task organizations generally remain local to individual suites. Economically grounded evaluations such as GDPval~\cite{patwardhan2025gdpval} provide a complementary view of application domains. AgentOmnia is evaluated on a suite comprising OmniaBench~\cite{shen2026omniabenchbenchmarkinggeneralai}, $\tau^2$-Bench~\cite{barres2025tau}, DeepPlanning, and VitaBench. OmniaBench also instantiates the domain--capability--difficulty taxonomy, enabling fine-grained diagnosis in the same coordinates used to organize data synthesis and PRD-guided self-evolution.

\paragraph{Agentic Data Synthesis.} Early research, exemplified by ToolAlpaca~\cite{tang2023toolalpaca}, APIGen~\cite{liu2024apigen}, and ToolACE~\cite{liu2025toolace}, primarily focused on enhancing tool-use capabilities through the synthesis of API specifications, user instructions, and function call annotations. While these methods introduced scalable frameworks for function calling data, they frequently conceptualized tools as isolated interfaces, providing limited support for persistent states or long-horizon interactions. Consequently, recent efforts have transitioned from the synthesis of discrete tool calls toward the instantiation of fully executable environments. AgentScaler~\cite{fang2026towards} structures extensive API collections via tool graphs and materializes tools for specific domains as read and write operations atop structured databases. EnvScaler~\cite{song2026envscaler} programmatically constructs environment skeletons, initial states, and task scenarios, facilitating both supervised fine-tuning and reinforcement learning within stateful sandboxes. Similarly, Agent World Model~\cite{wang2026agentworldmodel} synthesizes environments implemented in code and supported by databases with consistent state transitions, while AutoForge derives interaction structures from tool dependency graphs by constructing environment states and tool implementations directly from documentation~\cite{cai2025autoforge}. EnvFactory~\cite{xu2026envfactory} further integrates the discovery and verification of executable environments with trajectory synthesis informed by topology; meanwhile, Agent-World~\cite{dong2026agentworld} leverages themes drawn from real environments, databases, and tool ecosystems to foster continuous coevolution between task generation and agent training. At the task level, synthesis has matured from isolated prompts into compositional and verifiable workflows. Methodologies based on graphs and programs derive tasks from valid tool dependencies or executable solution paths, enabling precise control over task complexity through tool composition, state constraints, and interaction topology~\cite{cai2025autoforge,song2026envscaler,dong2026agentworld}. AgentSkiller~\cite{sun2026agentskiller} further establishes semantically coherent domains through ontologies, entity graphs, and service blueprints, generating natural user requests only after validating their underlying solution paths. At the trajectory level, research emphasis has shifted toward grounding supervision in empirical execution. Toucan~\cite{xu2025toucan} synthesizes large-scale trajectories over real MCP servers and applies rigorous filtering. Departing from the conventional paradigm that begins with a query, DIVE~\cite{chen2026dive} prioritizes the execution of diverse tools in real settings to collect evidence and subsequently derives tasks supported by the resulting traces to ensure inherent executability and verifiability. Collectively, these advancements represent a paradigm shift from fragmented function call synthesis toward the holistic construction of environments, tasks, and trajectories. AgentOmnia builds on this direction with taxonomy-guided, bidirectional environment--task synthesis and execution-grounded validation of environments, tasks, and trajectories.

\paragraph{Agentic Reinforcement Learning.}
With the emergence of reasoning models, reinforcement learning has become a standard component of large-model post-training pipelines~\cite{shao2024deepseekmath, guo2025deepseek}, particularly for agentic tasks~\cite{feng2025retoolreinforcementlearningstrategic, qian2025toolrlrewardtoollearning, research2026composer2technicalreport}. Recent agentic RL has rapidly evolved from optimizing individual tools~\cite{jin2025searchr1trainingllmsreason, zhang2026aster} to training general-purpose agents capable of long-horizon decision making across diverse environments~\cite{li2026compactionrlreinforcementlearningcontext, zhang2026landscapeagenticreinforcementlearning, hou2026singlerolloutasynchronousoptimizationagentic, wang2026openclawrltrainagentsimply}. On the algorithmic side, group-relative and REINFORCE-style methods, such as GRPO~\cite{shao2024deepseekmath, zeng2026glm, liu2025deepseek}, CISPO~\cite{minimax2025minimaxm1scalingtesttimecompute} and IPA~\cite{wang2025let}, have become widely used optimization approaches. Subsequent studies have further improved training stability and efficiency through sequence-level importance sampling~\cite{zheng2025group}, alternatives to hard clipping~\cite{minimax2025minimaxm1scalingtesttimecompute, gao2025softadaptivepolicyoptimization}, dynamic sampling~\cite{yu2026dapo, hong2025glm}, asymmetric policy optimization~\cite{roux2025tapered, wang2025let}, and environment dynamics modeling~\cite{shrivastava2026echo}. Meanwhile, increasing attention has been devoted to training–inference consistency, including rollout correction for training-inference mismatch~\cite{liu-li-2025-rl-collapse, yao2025offpolicy, kpop}, expert routing replay~\cite{ma2025stabilizing}, and activated-vocabulary space alignment~\cite{liu2025deepseek}. Despite this progress, most existing approaches remain confined to policy optimization over a static training distribution. Within this line of work, AgentOmnia combines rule- and rubric-based rewards, rollout trajectory analysis, training--inference alignment, and rollback-based curriculum learning for otherwise all-fail tasks.

\paragraph{Agent Systems with Self-Evolution.}
Expanding beyond static training pipelines, agent systems with self-evolution aim for autonomous refinement through feedback. Foundational frameworks, such as Reflexion~\cite{shinn2023reflexion} and ExpeL~\cite{zhao2024expel}, incorporate linguistic critiques or abstract reusable insights into memory, allowing agents to adapt across successive trials without explicit parameter updates. EigenData~\cite{gao2026self} employs a hierarchical multi-agent system to synthesize tool-grounded multi-turn dialogues and executable instance-level verifiers. The resulting data further supports policy optimization through reinforcement learning with verifiable rewards. Recent systems strive to integrate task generation and policy optimization into a unified loop. For instance, AgentEvolver~\cite{zhai2025agentevolver} improves exploration efficiency by allowing agents to formulate their own questions and assigning rewards with greater granularity, while Agent0~\cite{xia2025agent0} enables a curriculum agent and an executor agent to evolve jointly. Agent-World~\cite{dong2026agentworld} further identifies capability gaps through dynamic task synthesis and uses them to drive targeted learning, fostering the joint evolution of policies and training environments. In AgentOmnia, evaluation-derived diagnoses are converted into structured PRDs that guide targeted data synthesis and iterative policy refinement; the same interface can also accept external product requirements.

\section{Conclusion}
\label{sec:conclusion}

We presented AgentOmnia, a framework for full-scenario agentic scaling across ToC, ToB, and ToE applications. It connects a Domain $\times$ Capability $\times$ Atomic Difficulty taxonomy with bidirectional environment--task synthesis, verified trajectory construction, SFT, online agentic RL, and PRD-guided iterative improvement. Starting from Qwen3-30B-A3B-Thinking-2507, AgentOmnia improves the OmniaBench challenging-set score from 9.16\% to 37.11\% and raises the macro-average over OmniaBench, $\tau^2$-Bench, DeepPlanning, and VitaBench from 22.86\% to 41.69\%, with gains distributed across application splits, capability dimensions, and atomic-difficulty factors. A preliminary one-round study further supports the potential of PRD-guided self-evolution. At the same time, stronger foundation and proprietary models remain ahead on several comparisons, and the current self-evolution evidence is limited to one round at modest scale. These limitations motivate applying the framework to stronger foundation models, with the aim of achieving stronger overall agent performance and extending the benefits of full-scenario post-training to newer model generations. Future work will also scale synthesis and self-evolution across repeated rounds, strengthen environment and verifier construction, and investigate broader challenges in distribution transfer and product-driven industrial deployment.

\section{Authors}
\label{sec:authors}

\textbf{Core Contributors:} Hao Jiang, Gangtao Xin, Yingdi Huang, Guojie Zhu, Jiangshan Zhang, Xinyuan Lin, Yunkun Xu, Chengyu Shen, Wenlong Fei, Jiawei Li, Yujie Fu, Sichen Kang, Tingyu Xie, Yedi Hu, Jingren Zhang, Hongcheng Gao, Jianshu Zeng, Chong Chen$^{\dagger}$

\textbf{Contributors} (ordered alphabetically): Chang Guo, Chao Feng, Feng Wang, Fulin Lin, Jinchao Ma, Lang Mei, Li Huang, Liyan Liu, Qing He, Shuting Tao, Siyu Mo, Xiangnan Chen, Xiaohan Yu, Xiaoyang Li, Yanheng Hou, Yanyu Wu, Zhihan Yang

\textbf{Academic Contributors} (ordered alphabetically): Wentao Zhang (Peking University), Yang Gao (Beijing Institute of Technology), Zhao Cao (Renmin University of China).

\medskip

{\small \textsuperscript{$\dagger$}Team Lead.}

\clearpage

\bibliographystyle{unsrtnat}
\bibliography{main}

@misc{shen2026omniabenchbenchmarkinggeneralai,
      title={OmniaBench: Benchmarking General AI Agents Across Diverse Scenarios}, 
      author={Chengyu Shen and Yujie Fu and Gangtao Xin and Yanheng Hou and Wenlong Fei and Guojie Zhu and Jiawei Li and Hongcheng Gao and Runming He and Zhen Hao Wong and Meiyi Qiang and Hao Liang and Zhao Cao and Hao Jiang and Chong Chen and Wentao Zhang},
      year={2026},
      eprint={2607.14989},
      archivePrefix={arXiv},
      primaryClass={cs.CL},
      url={https://arxiv.org/abs/2607.14989}, 
}

@article{froger2026gaia2,
  title   = {{Gaia2}: Benchmarking {LLM} Agents on Dynamic and Asynchronous Environments},
  author  = {Froger, Romain and Andrews, Pierre and Bettini, Matteo and Budhiraja, Amar and Cabral, Ricardo Silveira and Do, Virginie and Garreau, Emilien and Gaya, Jean-Baptiste and Lauren{\c{c}}on, Hugo and Lecanu, Maxime and Malkan, Kunal and Mekala, Dheeraj and M{\'e}nard, Pierre and Moreno-Torres Bertran, Gerard and Piterbarg, Ulyana and Plekhanov, Mikhail and Rita, Mathieu and Rusakov, Andrey and Vorotilov, Vladislav and Wang, Mengjue and Yu, Ian and Benhalloum, Amine and Mialon, Gr{\'e}goire and Scialom, Thomas},
  journal = {arXiv preprint arXiv:2602.11964},
  year    = {2026},
  url     = {https://arxiv.org/abs/2602.11964}
}

@article{bai2026agentsa1,
  title   = {Scaling the Horizon, Not the Parameters: Reaching Trillion-Parameter Performance with a 35B Agent},
  author  = {Bai, Lei and Cao, Zongsheng and Chen, Yang and Cui, Zhiyao and Du, Shangheng and Fan, Yue and Feng, Shiyang and Guo, Zijie and He, Haonan and He, Liang and others},
  journal = {arXiv preprint arXiv:2606.30616},
  year    = {2026},
  url     = {https://arxiv.org/abs/2606.30616}
}

@article{hu2026seal,
  title   = {{SEAL}: Synergistic Co-Evolution of Agents and Learning Environments},
  author  = {Hu, Yihao and Wen, Zhihao and Liu, Xiujin and Wang, Pan and Zhang, Xin and Wu, Wei},
  journal = {arXiv preprint arXiv:2605.24426},
  year    = {2026},
  url     = {https://arxiv.org/abs/2605.24426}
}

@article{li2025inftool,
  title   = {Close the Loop: Synthesizing Infinite Tool-Use Data via Multi-Agent Role-Playing},
  author  = {Li, Yuwen and Zhang, Wei and Huang, Zelong and Yang, Mason and Wu, Jiajun and Guo, Shawn and Hu, Huahao and Sun, Lingyi and Yang, Jian and Tang, Mingjie and Dai, Byran},
  journal = {arXiv preprint arXiv:2512.23611},
  year    = {2025},
  url     = {https://arxiv.org/abs/2512.23611}
}

@article{zhai2025agentevolver,
  title   = {{AgentEvolver}: Towards Efficient Self-Evolving Agent System},
  author  = {Zhai, Yunpeng and Tao, Shuchang and Chen, Cheng and Zou, Anni and Chen, Ziqian and Fu, Qingxu and Mai, Shinji and Yu, Li and Deng, Jiaji and Cao, Zouying and Liu, Zhaoyang and Ding, Bolin and Zhou, Jingren},
  journal = {arXiv preprint arXiv:2511.10395},
  year    = {2025},
  url     = {https://arxiv.org/abs/2511.10395}
}

@article{yan2026openskill,
  title   = {{OpenSkill}: Open-World Self-Evolution for {LLM} Agents},
  author  = {Yan, Zhiling and Song, Dingjie and Zhang, Hanrong and Liang, Wei and Zhang, Yuxuan and Dai, Yutong and He, Lifang and Yu, Philip S. and Xu, Ran and Li, Xiang and Sun, Lichao},
  journal = {arXiv preprint arXiv:2606.06741},
  year    = {2026},
  url     = {https://arxiv.org/abs/2606.06741}
}

@misc{qwen35blog,
  title  = {{Qwen3.5}: Towards Native Multimodal Agents},
  author = {{Qwen Team}},
  month  = {February},
  year   = {2026},
  url    = {https://qwen.ai/blog?id=qwen3.5}
}

@misc{qwenteam2025qwen330b2507,
  title        = {{Qwen3-30B-A3B-Thinking-2507}},
  author       = {{Qwen Team}},
  year         = {2025},
  howpublished = {Hugging Face model card},
  url          = {https://huggingface.co/Qwen/Qwen3-30B-A3B-Thinking-2507}
}

@misc{qwenteam2025qwen3235b2507,
  title        = {{Qwen3-235B-A22B-Thinking-2507}},
  author       = {{Qwen Team}},
  year         = {2025},
  howpublished = {Hugging Face model card},
  url          = {https://huggingface.co/Qwen/Qwen3-235B-A22B-Thinking-2507}
}

@misc{openai2026gpt55,
  title        = {{GPT-5.5} System Card},
  author       = {{OpenAI}},
  year         = {2026},
  month        = {April},
  howpublished = {System card},
  url          = {https://openai.com/index/gpt-5-5-system-card/}
}

@misc{anthropic2026claudeopus47,
  title        = {{Claude Opus 4.7} Model Report},
  author       = {{Anthropic}},
  year         = {2026},
  month        = {April},
  howpublished = {Anthropic Transparency Hub},
  url          = {https://www.anthropic.com/transparency/model-report}
}

@misc{googledeepmind2026gemini35flash,
  title        = {{Gemini 3.5 Flash} Model Card},
  author       = {{Google DeepMind}},
  year         = {2026},
  month        = {May},
  howpublished = {Model card},
  url          = {https://deepmind.google/models/model-cards/gemini-3-5-flash/}
}

@misc{qwenteam2026qwen37max,
  title        = {{Qwen3.7}: The Agent Frontier},
  author       = {{Qwen Team}},
  year         = {2026},
  month        = {May},
  howpublished = {Qwen blog},
  url          = {https://qwen.ai/blog?id=qwen3.7}
}

@misc{deepseekai2026deepseekv4,
  title        = {{DeepSeek-V4}: Towards Highly Efficient Million-Token Context Intelligence},
  author       = {{DeepSeek-AI}},
  year         = {2026},
  eprint       = {2606.19348},
  archivePrefix = {arXiv},
  primaryClass = {cs.CL},
  url          = {https://huggingface.co/deepseek-ai/DeepSeek-V4-Pro}
}

@inproceedings{wei2022chain,
  title     = {Chain-of-Thought Prompting Elicits Reasoning in Large Language Models},
  author    = {Wei, Jason and Wang, Xuezhi and Schuurmans, Dale and Bosma, Maarten and Ichter, Brian and Xia, Fei and Chi, Ed and Le, Quoc V and Zhou, Denny},
  booktitle = {Advances in Neural Information Processing Systems},
  volume    = {35},
  pages     = {24824--24837},
  year      = {2022}
}

@inproceedings{kojima2022large,
  title     = {Large Language Models are Zero-Shot Reasoners},
  author    = {Kojima, Takeshi and Gu, Shixiang Shane and Reid, Machel and Matsuo, Yutaka and Iwasawa, Yusuke},
  booktitle = {Advances in Neural Information Processing Systems},
  volume    = {35},
  pages     = {22199--22213},
  year      = {2022}
}

@inproceedings{schick2023toolformer,
  title     = {Toolformer: Language Models Can Teach Themselves to Use Tools},
  author    = {Schick, Timo and Dwivedi-Yu, Jane and Dess{\`\i}, Roberto and Raileanu, Roberta and Lomeli, Maria and Hambro, Eric and Zettlemoyer, Luke and Cancedda, Nicola and Scialom, Thomas},
  booktitle = {Advances in Neural Information Processing Systems},
  volume    = {36},
  year      = {2023}
}

@inproceedings{qin2023toolllm,
  title     = {{ToolLLM}: Facilitating Large Language Models to Master 16000+ Real-World {APIs}},
  author    = {Qin, Yujia and Liang, Shihao and Ye, Yining and Zhu, Kunlun and Yan, Lan and Lu, Yaxi and Lin, Yankai and Cong, Xin and Tang, Xiangru and Qian, Bill and Zhao, Sihan and Hong, Lauren and Tian, Runchu and Xie, Ruobing and Zhou, Jie and Gerstein, Mark and Li, Dahai and Liu, Zhiyuan and Sun, Maosong},
  booktitle = {The Twelfth International Conference on Learning Representations},
  year      = {2024}
}

@article{wang2023survey,
  title   = {A Survey on Large Language Model Based Autonomous Agents},
  author  = {Wang, Lei and Ma, Chen and Feng, Xueyang and Zhang, Zeyu and Yang, Hao and Zhang, Jingsen and Chen, Zhiyuan and Tang, Jiakai and Chen, Xu and Lin, Yankai and Zhao, Wayne Xin and Wei, Zhewei and Wen, Ji-Rong},
  journal = {Frontiers of Computer Science},
  volume  = {18},
  number  = {6},
  pages   = {186345},
  year    = {2024}
}

@article{xi2023rise,
  title   = {The Rise and Potential of Large Language Model Based Agents: A Survey},
  author  = {Xi, Zhiheng and Chen, Wenxiang and Guo, Xin and He, Wei and Ding, Yiwen and Hong, Boyang and Zhang, Ming and Wang, Junzhe and Jin, Senjie and Zhou, Enyu and others},
  journal = {arXiv preprint arXiv:2309.07864},
  year    = {2023}
}

@inproceedings{yao2022react,
  title     = {{ReAct}: Synergizing Reasoning and Acting in Language Models},
  author    = {Yao, Shunyu and Zhao, Jeffrey and Yu, Dian and Du, Nan and Shafran, Izhak and Narasimhan, Karthik and Cao, Yuan},
  booktitle = {The Eleventh International Conference on Learning Representations},
  year      = {2023}
}

@inproceedings{liu2023agentbench,
  title     = {{AgentBench}: Evaluating {LLMs} as Agents},
  author    = {Liu, Xiao and Yu, Hao and Zhang, Hanchen and Xu, Yifan and Lei, Xuanyu and Lai, Hanyu and Gu, Yu and Ding, Hangliang and Men, Kaiwen and Yang, Kejuan and Zhang, Shudan and Deng, Xiang and Zeng, Aohan and Du, Zhengxiao and Zhang, Chenhui and Shen, Sheng and Zhang, Tianjun and Su, Yu and Sun, Huan and Huang, Minlie and Dong, Yuxiao and Tang, Jie},
  booktitle = {The Twelfth International Conference on Learning Representations},
  year      = {2024}
}

@inproceedings{deng2023mind2web,
  title     = {{Mind2Web}: Towards a Generalist Agent for the Web},
  author    = {Deng, Xiang and Gu, Yu and Zheng, Boyuan and Chen, Shijie and Stevens, Samuel and Wang, Boshi and Sun, Huan and Su, Yu},
  booktitle = {Advances in Neural Information Processing Systems},
  volume    = {36},
  year      = {2023}
}

@inproceedings{zhou2023webarena,
  title     = {{WebArena}: A Realistic Web Environment for Building Autonomous Agents},
  author    = {Zhou, Shuyan and Xu, Frank F and Zhu, Hao and Zhou, Xuhui and Lo, Robert and Sridhar, Abishek and Cheng, Xianyi and Ou, Tianyue and Bisk, Yonatan and Fried, Daniel and Alon, Uri and Neubig, Graham},
  booktitle = {The Twelfth International Conference on Learning Representations},
  year      = {2024}
}

@inproceedings{jimenez2023swebench,
  title     = {{SWE-bench}: Can Language Models Resolve Real-World {GitHub} Issues?},
  author    = {Jimenez, Carlos E and Yang, John and Wettig, Alexander and Yao, Shunyu and Pei, Kexin and Press, Ofir and Narasimhan, Karthik},
  booktitle = {The Twelfth International Conference on Learning Representations},
  year      = {2024}
}

@article{yao2024tau,
  title   = {$\tau$-bench: A Benchmark for Tool-Agent-User Interaction in Real-World Domains},
  author  = {Yao, Shunyu and Shinn, Noah and Razavi, Pedram and Narasimhan, Karthik},
  journal = {arXiv preprint arXiv:2406.12045},
  year    = {2024}
}

@inproceedings{yao2022webshop,
  title     = {{WebShop}: Towards Scalable Real-World Web Interaction with Grounded Language Agents},
  author    = {Yao, Shunyu and Chen, Howard and Yang, John and Narasimhan, Karthik},
  booktitle = {Advances in Neural Information Processing Systems},
  volume    = {35},
  year      = {2022}
}

@article{koh2024visualwebarena,
  title   = {{VisualWebArena}: Evaluating Multimodal Agents on Realistic Visual Web Tasks},
  author  = {Koh, Jing Yu and Lo, Robert and Jang, Lawrence and Duvvur, Vikram and Lim, Ming and Huang, Po-Yao and Neubig, Graham and Zhou, Shuyan and Salakhutdinov, Ruslan and Fried, Daniel},
  journal = {arXiv preprint arXiv:2401.13649},
  year    = {2024}
}

@article{barres2025tau,
  title={{$\tau^2$-Bench: Evaluating Conversational Agents in a Dual-Control Environment}},
  author={Barres, Victor and Dong, Honghua and Ray, Soham and Si, Xujie and Narasimhan, Karthik},
  journal={arXiv preprint arXiv:2506.07982},
  year={2025}
}

@article{rawles2024androidworld,
  title   = {{AndroidWorld}: A Dynamic Benchmarking Environment for Autonomous Agents},
  author  = {Rawles, Christopher and Clinckemaillie, Sarah and Chang, Yifan and Waltz, Jonathan and Lau, Gabrielle and Fair, Marybeth and Li, Alice and Bishop, William and Li, Wei and Campbell-Ajala, Folawiyo and Toyama, Daniel and Berry, Robert and Tyamagundlu, Divya and Lillicrap, Timothy and Riva, Oriana},
  journal = {arXiv preprint arXiv:2405.14573},
  year    = {2024},
  url     = {https://arxiv.org/abs/2405.14573}
}

@article{xie2024osworld,
  title   = {{OSWorld}: Benchmarking Multimodal Agents for Open-Ended Tasks in Real Computer Environments},
  author  = {Xie, Tianbao and Zhang, Danyang and Chen, Jixuan and Li, Xiaochuan and Zhao, Siheng and Cao, Ruisheng and Hua, Toh Jing and Cheng, Zhoujun and Shin, Dongchan and Lei, Fangyu and Liu, Yitao and Xu, Yiheng and Zhou, Shuyan and Savarese, Silvio and Xiong, Caiming and Zhong, Victor and Yu, Tao},
  journal = {arXiv preprint arXiv:2404.07972},
  year    = {2024},
  url     = {https://arxiv.org/abs/2404.07972}
}

@article{drouin2024workarena,
  title   = {{WorkArena}: How Capable Are Web Agents at Solving Common Knowledge Work Tasks?},
  author  = {Drouin, Alexandre and Gasse, Maxime and Caccia, Massimo and Laradji, Issam H. and Del Verme, Manuel and Marty, Tom and Boisvert, L{\'e}o and Thakkar, Megh and Cappart, Quentin and Vazquez, David and Chapados, Nicolas and Lacoste, Alexandre},
  journal = {arXiv preprint arXiv:2403.07718},
  year    = {2024},
  url     = {https://arxiv.org/abs/2403.07718}
}

@article{wang2024officebench,
  title   = {{OfficeBench}: Benchmarking Language Agents across Multiple Applications for Office Automation},
  author  = {Wang, Zilong and Cui, Yuedong and Zhong, Li and Zhang, Zimin and Yin, Da and Lin, Bill Yuchen and Shang, Jingbo},
  journal = {arXiv preprint arXiv:2407.19056},
  year    = {2024},
  url     = {https://arxiv.org/abs/2407.19056}
}

@article{huang2024crmarena,
  title   = {{CRMArena}: Understanding the Capacity of {LLM} Agents to Perform Professional {CRM} Tasks in Realistic Environments},
  author  = {Huang, Kung-Hsiang and Prabhakar, Akshara and Dhawan, Sidharth and Mao, Yixin and Wang, Huan and Savarese, Silvio and Xiong, Caiming and Laban, Philippe and Wu, Chien-Sheng},
  journal = {arXiv preprint arXiv:2411.02305},
  year    = {2024},
  url     = {https://arxiv.org/abs/2411.02305}
}

@article{ma2024spreadsheetbench,
  title   = {{SpreadsheetBench}: Towards Challenging Real World Spreadsheet Manipulation},
  author  = {Ma, Zeyao and Zhang, Bohan and Zhang, Jing and Yu, Jifan and Zhang, Xiaokang and Zhang, Xiaohan and Luo, Sijia and Wang, Xi and Tang, Jie},
  journal = {arXiv preprint arXiv:2406.14991},
  year    = {2024},
  url     = {https://arxiv.org/abs/2406.14991}
}

@article{trivedi2024appworld,
  title   = {{AppWorld}: A Controllable World of Apps and People for Benchmarking Interactive Coding Agents},
  author  = {Trivedi, Harsh and Khot, Tushar and Hartmann, Mareike and Manku, Ruskin and Dong, Vinty and Li, Edward and Gupta, Shashank and Sabharwal, Ashish and Balasubramanian, Niranjan},
  journal = {arXiv preprint arXiv:2407.18901},
  year    = {2024},
  url     = {https://arxiv.org/abs/2407.18901}
}

@article{li2023apibank,
  title   = {{API-Bank}: A Comprehensive Benchmark for Tool-Augmented {LLMs}},
  author  = {Li, Minghao and Zhao, Yingxiu and Yu, Bowen and Song, Feifan and Li, Hangyu and Yu, Haiyang and Li, Zhoujun and Huang, Fei and Li, Yongbin},
  journal = {arXiv preprint arXiv:2304.08244},
  year    = {2023},
  url     = {https://arxiv.org/abs/2304.08244}
}

@inproceedings{gudibande2023false,
  title     = {The False Promise of Imitating Proprietary Language Models},
  author    = {Gudibande, Arnav and Wallace, Eric and Snell, Charlie and Geng, Xinyang and Liu, Hao and Abbeel, Pieter and Levine, Sergey and Song, Dawn},
  booktitle = {The Twelfth International Conference on Learning Representations},
  year      = {2024}
}

@inproceedings{zeng2023agenttuning,
  title     = {{AgentTuning}: Enabling Generalized Agent Abilities for {LLMs}},
  author    = {Zeng, Aohan and Liu, Mingdao and Lu, Rui and Wang, Bowen and Liu, Xiao and Dong, Yuxiao and Tang, Jie},
  booktitle = {Findings of the Association for Computational Linguistics: ACL 2024},
  pages     = {3053--3077},
  year      = {2024}
}

@inproceedings{wang2022selfinstruct,
  title     = {Self-Instruct: Aligning Language Models with Self-Generated Instructions},
  author    = {Wang, Yizhong and Kordi, Yeganeh and Mishra, Swaroop and Liu, Alisa and Smith, Noah A and Khashabi, Daniel and Hajishirzi, Hannaneh},
  booktitle = {Proceedings of the 61st Annual Meeting of the Association for Computational Linguistics},
  pages     = {13484--13508},
  year      = {2023}
}

@inproceedings{yuan2024selfreward,
  title     = {Self-Rewarding Language Models},
  author    = {Yuan, Weizhe and Pang, Richard Yuanzhe and Cho, Kyunghyun and Li, Xian and Sukhbaatar, Sainbayar and Xu, Jing and Weston, Jason},
  booktitle = {Proceedings of the 41st International Conference on Machine Learning},
  year      = {2024}
}

@article{shinn2023reflexion,
  title   = {Reflexion: Language Agents with Verbal Reinforcement Learning},
  author  = {Shinn, Noah and Cassano, Federico and Berman, Edward and Gopinath, Ashwin and Narasimhan, Karthik and Yao, Shunyu},
  journal = {arXiv preprint arXiv:2303.11366},
  year    = {2023}
}

@article{madaan2023selfrefine,
  title   = {Self-Refine: Iterative Refinement with Self-Feedback},
  author  = {Madaan, Aman and Tandon, Niket and Gupta, Prakhar and Hallinan, Skyler and Gao, Luyu and Wiegreffe, Sarah and Alon, Uri and Dziri, Nouha and Prabhumoye, Shrimai and Yang, Yiming and Gupta, Shashank and Majumder, Bodhisattwa Prasad and Hermann, Katherine and Welleck, Sean and Yazdanbakhsh, Amir and Clark, Peter},
  journal = {arXiv preprint arXiv:2303.17651},
  year    = {2023}
}

@article{lu2023self,
  title   = {{SELF}: Self-Evolution with Language Feedback},
  author  = {Lu, Jianqiao and Zhong, Wanjun and Huang, Wenyong and Wang, Yufei and Zhu, Qi and Mi, Fei and Wang, Baojun and Wang, Weichao and Zeng, Xingshan and Shang, Lifeng and Jiang, Xin and Liu, Qun},
  journal = {arXiv preprint arXiv:2310.00533},
  year    = {2023}
}

@inproceedings{burns2023weak,
  title     = {Weak-to-Strong Generalization: Eliciting Strong Capabilities With Weak Supervision},
  author    = {Burns, Collin and Izmailov, Pavel and Kirchner, Jan Hendrik and Baker, Bowen and Gao, Leo and Aschenbrenner, Leopold and Chen, Yining and Ecoffet, Adrien and Joglekar, Manas and Leike, Jan and Sutskever, Ilya and Wu, Jeffrey},
  booktitle = {Proceedings of the 41st International Conference on Machine Learning},
  year      = {2024}
}

@inproceedings{hao2023reasoning,
  title     = {Reasoning with Language Model is Planning with World Model},
  author    = {Hao, Shibo and Gu, Yi and Ma, Haodi and Hong, Joshua Jiahua and Wang, Zhen and Wang, Daisy Zhe and Hu, Zhiting},
  booktitle = {Proceedings of the 2023 Conference on Empirical Methods in Natural Language Processing},
  pages     = {8154--8173},
  year      = {2023}
}

@inproceedings{bai2024digirl,
  title     = {{DigiRL}: Training In-The-Wild Device-Control Agents with Autonomous Reinforcement Learning},
  author    = {Bai, Hao and Zhou, Yifei and Cemri, Mert and Pan, Jiayi and Suhr, Alane and Levine, Sergey and Kumar, Aviral},
  booktitle = {Advances in Neural Information Processing Systems},
  volume    = {37},
  year      = {2024}
}

@article{xi2024agentgym,
  title   = {{AgentGym}: Evolving Large Language Model-based Agents across Diverse Environments},
  author  = {Xi, Zhiheng and Ding, Yiwen and Chen, Wenxiang and Hong, Boyang and Guo, Honglin and Wang, Junzhe and Yang, Dingwen and Liao, Chenyang and Guo, Xin and He, Wei and others},
  journal = {arXiv preprint arXiv:2406.04151},
  year    = {2024}
}

@article{shrivastava2026echo,
  title={Echo: Terminal agents learn world models for free},
  author={Shrivastava, Vaishnavi and Kauffmann, Piero and Awadallah, Ahmed and Papailiopoulos, Dimitris},
  journal={arXiv preprint arXiv:2605.24517},
  year={2026}
}

@article{dong2026agentworld,
  title   = {{Agent-World}: Scaling Real-World Environment Synthesis for Evolving General Agent Intelligence},
  author  = {Dong, Guanting and Lu, Junting and Huang, Junjie and Zhong, Wanjun and Liu, Longxiang and Huang, Shijue and Li, Zhenyu and Zhao, Yang and Song, Xiaoshuai and Li, Xiaoxi and Jin, Jiajie and Zhu, Yutao and Wang, Hanbin and Lei, Fangyu and Luo, Qinyu and Chen, Mingyang and Chen, Zehui and Feng, Jiazhan and Wen, Ji-Rong and Dou, Zhicheng},
  journal = {arXiv preprint arXiv:2604.18292},
  year    = {2026}
}

@article{wang2026agentworldmodel,
  title   = {Agent World Model: Infinity Synthetic Environments for Agentic Reinforcement Learning},
  author  = {Wang, Zhaoyang and Xu, Canwen and Liu, Boyi and Wang, Yite and Han, Siwei and Yao, Zhewei and Yao, Huaxiu and He, Yuxiong},
  journal = {arXiv preprint arXiv:2602.10090},
  year    = {2026}
}

@article{zhang2026deepplanning,
  title   = {{DeepPlanning}: Benchmarking Long-Horizon Agentic Planning with Verifiable Constraints},
  author  = {Zhang, Yinger and Jiang, Shutong and Li, Renhao and Tu, Jianhong and Su, Yang and Deng, Lianghao and Guo, Xudong and Lv, Chenxu and Lin, Junyang},
  journal = {arXiv preprint arXiv:2601.18137},
  year    = {2026}
}

@article{he2025vitabench,
  title   = {{VitaBench}: Benchmarking {LLM} Agents with Versatile Interactive Tasks in Real-world Applications},
  author  = {He, Wei and Sun, Yueqing and Hao, Hongyan and Hao, Xueyuan and Xia, Zhikang and Gu, Qi and Han, Chengcheng and Zhao, Dengchang and Su, Hui and Zhang, Kefeng and Gao, Man and Su, Xi and Cai, Xiaodong and Cai, Xunliang and Yu, Yang and Zhao, Yunke},
  journal = {arXiv preprint arXiv:2509.26490},
  year    = {2025}
}

@article{li2025toolathlon,
  title   = {The Tool Decathlon: Benchmarking Language Agents for Diverse, Realistic, and Long-Horizon Task Execution},
  author  = {Li, Junlong and Zhao, Wenshuo and Zhao, Jian and Zeng, Weihao and Wu, Haoze and Wang, Xiaochen and Ge, Rui and Cao, Yuxuan and Huang, Yuzhen and Liu, Wei and Liu, Junteng and Su, Zhaochen and Guo, Yiyang and Zhou, Fan and Zhang, Lueyang and Michelini, Juan and Wang, Xingyao and Yue, Xiang and Zhou, Shuyan and Neubig, Graham and He, Junxian},
  journal = {arXiv preprint arXiv:2510.25726},
  year    = {2025}
}

@misc{bfcl2024,
  title        = {Berkeley Function-Calling Leaderboard},
  author       = {{Berkeley Gorilla Team}},
  howpublished = {\url{https://gorilla.cs.berkeley.edu/leaderboard.html}},
  year         = {2024},
  note         = {Accessed 2026-06-11}
}

@article{2026chengraph2eval,
  author       = {Yurun Chen and
                  Xavier Hu and
                  Yuhan Liu and
                  Ziqi Wang and
                  Zeyi Liao and
                  Lin Chen and
                  Feng Wei and
                  Yuxi Qian and
                  Bo Zheng and
                  Keting Yin and
                  Shengyu Zhang},
  title        = {Graph2Eval: Automatic Multimodal Task Generation for Agents via Knowledge
                  Graphs},
  journal      = {CoRR},
  volume       = {abs/2510.00507},
  year         = {2025},
  url          = {https://doi.org/10.48550/arXiv.2510.00507},
  doi          = {10.48550/ARXIV.2510.00507},
  eprinttype   = {arXiv},
  eprint       = {2510.00507},
  bibsource    = {dblp computer science bibliography, https://dblp.org}
}

@article{2021cobbegsm8k,
  author       = {Karl Cobbe and
                  Vineet Kosaraju and
                  Mohammad Bavarian and
                  Mark Chen and
                  Heewoo Jun and
                  Lukasz Kaiser and
                  Matthias Plappert and
                  Jerry Tworek and
                  Jacob Hilton and
                  Reiichiro Nakano and
                  Christopher Hesse and
                  John Schulman},
  title        = {Training Verifiers to Solve Math Word Problems},
  journal      = {CoRR},
  volume       = {abs/2110.14168},
  year         = {2021},
  url          = {https://arxiv.org/abs/2110.14168},
  eprinttype   = {arXiv},
  eprint       = {2110.14168},
  bibsource    = {dblp computer science bibliography, https://dblp.org}
}

@article{2023fureasoninghub,
  author       = {Yao Fu and
                  Litu Ou and
                  Mingyu Chen and
                  Yuhao Wan and
                  Hao Peng and
                  Tushar Khot},
  title        = {Chain-of-Thought Hub: {A} Continuous Effort to Measure Large Language
                  Models' Reasoning Performance},
  journal      = {CoRR},
  volume       = {abs/2305.17306},
  year         = {2023},
  url          = {https://doi.org/10.48550/arXiv.2305.17306},
  doi          = {10.48550/ARXIV.2305.17306},
  eprinttype   = {arXiv},
  eprint       = {2305.17306},
  bibsource    = {dblp computer science bibliography, https://dblp.org}
}

@inproceedings{2024patilgorilla,
  author       = {Shishir G. Patil and
                  Tianjun Zhang and
                  Xin Wang and
                  Joseph E. Gonzalez},
  editor       = {Amir Globersons and
                  Lester Mackey and
                  Danielle Belgrave and
                  Angela Fan and
                  Ulrich Paquet and
                  Jakub M. Tomczak and
                  Cheng Zhang},
  title        = {Gorilla: Large Language Model Connected with Massive APIs},
  booktitle    = {Advances in Neural Information Processing Systems 37: Annual Conference
                  on Neural Information Processing Systems 2024, NeurIPS 2024, Vancouver,
                  BC, Canada, December 10 - 15, 2024},
  year         = {2024},
  url          = {http://papers.nips.cc/paper\_files/paper/2024/hash/e4c61f578ff07830f5c37378dd3ecb0d-Abstract-Conference.html},
  bibsource    = {dblp computer science bibliography, https://dblp.org}
}

@article{2023farntooltalk,
  author       = {Nicholas Farn and
                  Richard Shin},
  title        = {ToolTalk: Evaluating Tool-Usage in a Conversational Setting},
  journal      = {CoRR},
  volume       = {abs/2311.10775},
  year         = {2023},
  url          = {https://doi.org/10.48550/arXiv.2311.10775},
  doi          = {10.48550/ARXIV.2311.10775},
  eprinttype   = {arXiv},
  eprint       = {2311.10775},
  bibsource    = {dblp computer science bibliography, https://dblp.org}
}

@article{2025xieagentsynth,
  title={AgentSynth: Scalable Task Generation for Generalist Computer-Use Agents},
  author={Xie, Jingxu and Xu, Dylan and Zhao, Xuandong and Song, Dawn},
  year={2025},
  journal={arXiv (Cornell University)},
  volume={abs/2506.14205},
  doi={10.48550/arxiv.2506.14205},
}

@misc{song2026envscaler,
      title={EnvScaler: Scaling Tool-Interactive Environments for LLM Agent via Programmatic Synthesis}, 
      author={Xiaoshuai Song and Haofei Chang and Guanting Dong and Yutao Zhu and Ji-Rong Wen and Zhicheng Dou},
      year={2026},
      eprint={2601.05808},
      archivePrefix={arXiv},
      primaryClass={cs.CL},
      url={https://arxiv.org/abs/2601.05808}, 
}

@article{tu2026scaleenv,
  title   = {{ScaleEnv}: Scaling Environment Synthesis from Scratch for Generalist Interactive Tool-Use Agent Training},
  author  = {Tu, Dunwei and Hao, Hongyan and Yang, Hansi and Chen, Yihao and Zhang, Yi-Kai and Xia, Zhikang and Yang, Yu and Sun, Yueqing and Liu, Xingchen and Shen, Furao and Gu, Qi and Su, Hui and Cai, Xunliang},
  journal = {arXiv preprint arXiv:2602.06820},
  year    = {2026},
  url     = {https://arxiv.org/abs/2602.06820}
}

@article{yang2025qwen3,
  title   = {{Qwen3} Technical Report},
  author  = {Yang, An and Li, Anfeng and Yang, Baosong and Zhang, Beichen and Hui, Binyuan and Zheng, Bo and Yu, Bowen and Gao, Chang and Huang, Chengen and Lv, Chenxu and others},
  journal = {CoRR},
  volume  = {abs/2505.09388},
  year    = {2025},
  url     = {https://doi.org/10.48550/arXiv.2505.09388},
  doi     = {10.48550/ARXIV.2505.09388}
}

@article{xu2025toucan,
  title   = {{Toucan}: Synthesizing 1.5M Tool-Agentic Data from Real-World {MCP} Environments},
  author  = {Xu, Zhangchen and Meza Soria, Adriana and Tan, Shawn and Roy, Anurag and Agrawal, Ashish Sunil and Poovendran, Radha and Panda, Rameswar},
  journal = {arXiv preprint arXiv:2510.01179},
  year    = {2025},
  url     = {https://arxiv.org/abs/2510.01179}
}

@article{sun2026agentskiller,
  title   = {{AgentSkiller}: Scaling Generalist Agent Intelligence through Semantically Integrated Cross-Domain Data Synthesis},
  author  = {Sun, Zexu and Ji, Bokai and Cai, Hengyi and Wang, Shuaiqiang and Wang, Lei and Li, Guangxia and Chen, Xu},
  journal = {arXiv preprint arXiv:2602.09372},
  year    = {2026},
  url     = {https://arxiv.org/abs/2602.09372}
}

@article{zhao2025muarl,
  title   = {{MUA-RL}: Multi-Turn User-Interacting Agent Reinforcement Learning for Agentic Tool Use},
  author  = {Zhao, Weikang and Wang, Xili and Ma, Chengdi and Kong, Lingbin and Yang, Zhaohua and Tuo, Mingxiang and Shi, Xiaowei and Zhai, Yitao and Cai, Xunliang},
  journal = {arXiv preprint arXiv:2508.18669},
  year    = {2025},
  url     = {https://arxiv.org/abs/2508.18669}
}

@article{cai2025nexn1,
  title   = {{Nex-N1}: Agentic Models Trained via a Unified Ecosystem for Large-Scale Environment Construction},
  author  = {Cai, Yuxuan and Chen, Lu and Chen, Qiaoling and Ding, Yuyang and Fan, Liwen and Fu, Wenjie and Gao, Yufei and Guo, Honglin and others},
  journal = {arXiv preprint arXiv:2512.04987},
  year    = {2025},
  url     = {https://arxiv.org/abs/2512.04987}
}

@misc{nexagi2026nexn2mini,
  title        = {{Nex-N2-mini}},
  author       = {{Nex AGI}},
  year         = {2026},
  howpublished = {Hugging Face model card},
  url          = {https://huggingface.co/nex-agi/Nex-N2-mini}
}

@article{guo2025genenv,
  author       = {Jiacheng Guo and
                  Ling Yang and
                  Peter Chen and
                  Qixin Xiao and
                  Yinjie Wang and
                  Xinzhe Juan and
                  Jiahao Qiu and
                  Ke Shen and
                  Mengdi Wang},
  title        = {GenEnv: Difficulty-Aligned Co-Evolution Between {LLM} Agents and Environment
                  Simulators},
  journal      = {CoRR},
  volume       = {abs/2512.19682},
  year         = {2025},
  url          = {https://doi.org/10.48550/arXiv.2512.19682},
  doi          = {10.48550/ARXIV.2512.19682},
  eprinttype   = {arXiv},
  eprint       = {2512.19682},
  bibsource    = {dblp computer science bibliography, https://dblp.org}
}

@article{wang2025let,
  title={Let it flow: Agentic crafting on rock and roll, building the rome model within an open agentic learning ecosystem},
  author={Wang, Weixun and Xu, XiaoXiao and An, Wanhe and Dai, Fangwen and Gao, Wei and He, Yancheng and Huang, Ju and Ji, Qiang and Jin, Hanqi and Li, Xiaoyang and others},
  journal={arXiv preprint arXiv:2512.24873},
  year={2025}
}

@article{zeng2026glm,
  title={Glm-5: from vibe coding to agentic engineering},
  author={Zeng, Aohan and Lv, Xin and Hou, Zhenyu and Du, Zhengxiao and Zheng, Qinkai and Chen, Bin and Yin, Da and Ge, Chendi and Huang, Chenghua and Xie, Chengxing and others},
  journal={arXiv preprint arXiv:2602.15763},
  year={2026}
}

@article{liu2025deepseek,
  title={Deepseek-v3.2: Pushing the frontier of open large language models},
  author={Liu, Aixin and Mei, Aoxue and Lin, Bangcai and Xue, Bing and Wang, Bingxuan and Xu, Bingzheng and Wu, Bochao and Zhang, Bowei and Lin, Chaofan and Dong, Chen and others},
  journal={arXiv preprint arXiv:2512.02556},
  year={2025}
}

@article{hong2025glm,
  title={Glm-4.5 v and glm-4.1 v-thinking: Towards versatile multimodal reasoning with scalable reinforcement learning},
  author={Hong, Wenyi and Yu, Wenmeng and Gu, Xiaotao and Wang, Guo and Gan, Guobing and Tang, Haomiao and Cheng, Jiale and Qi, Ji and Ji, Junhui and Pan, Lihang and others},
  journal={arXiv preprint arXiv:2507.01006},
  year={2025}
}

@article{roux2025tapered,
  title={Tapered off-policy reinforce: Stable and efficient reinforcement learning for llms},
  author={Roux, Nicolas Le and Bellemare, Marc G and Lebensold, Jonathan and Bergeron, Arnaud and Greaves, Joshua and Fr{\'e}chette, Alex and Pelletier, Carolyne and Thibodeau-Laufer, Eric and Toth, S{\'a}ndor and Work, Sam},
  journal={arXiv preprint arXiv:2503.14286},
  year={2025}
}

@misc{gao2025softadaptivepolicyoptimization,
      title={Soft Adaptive Policy Optimization}, 
      author={Chang Gao and Chujie Zheng and Xiong-Hui Chen and Kai Dang and Shixuan Liu and Bowen Yu and An Yang and Shuai Bai and Jingren Zhou and Junyang Lin},
      year={2025},
      eprint={2511.20347},
      archivePrefix={arXiv},
      primaryClass={cs.LG},
      url={https://arxiv.org/abs/2511.20347}, 
}

@article{zheng2025group,
  title={Group sequence policy optimization},
  author={Zheng, Chujie and Liu, Shixuan and Li, Mingze and Chen, Xiong-Hui and Yu, Bowen and Gao, Chang and Dang, Kai and Liu, Yuqiong and Men, Rui and Yang, An and others},
  journal={arXiv preprint arXiv:2507.18071},
  year={2025}
}

@article{gu2026survey,
  title={A survey on llm-as-a-judge},
  author={Gu, Jiawei and Jiang, Xuhui and Shi, Zhichao and Tan, Hexiang and Zhai, Xuehao and Xu, Chengjin and Li, Wei and Shen, Yinghan and Ma, Shengjie and Liu, Honghao and others},
  journal={The Innovation},
  volume={7},
  number={6},
  year={2026},
  publisher={Elsevier}
}

@article{zhang2026reasoning,
  title={From reasoning to agentic: Credit assignment in reinforcement learning for large language models},
  author={Zhang, Chenchen},
  journal={arXiv preprint arXiv:2604.09459},
  year={2026}
}

@article{shao2024deepseekmath,
  title={Deepseekmath: Pushing the limits of mathematical reasoning in open language models},
  author={Shao, Zhihong and Wang, Peiyi and Zhu, Qihao and Xu, Runxin and Song, Junxiao and Bi, Xiao and Zhang, Haowei and Zhang, Mingchuan and Li, YK and Wu, Yang and others},
  journal={arXiv preprint arXiv:2402.03300},
  year={2024}
}

@article{guo2025deepseek,
  title={Deepseek-r1: Incentivizing reasoning capability in llms via reinforcement learning},
  author={Guo, Daya and Yang, Dejian and Zhang, Haowei and Song, Junxiao and Wang, Peiyi and Zhu, Qihao and Xu, Runxin and Zhang, Ruoyu and Ma, Shirong and Bi, Xiao and others},
  journal={arXiv preprint arXiv:2501.12948},
  year={2025}
}

@misc{research2026composer2technicalreport,
      title={Composer 2 Technical Report}, 
      author={Cursor Research Team},
      year={2026},
      eprint={2603.24477},
      archivePrefix={arXiv},
      primaryClass={cs.SE},
      url={https://arxiv.org/abs/2603.24477}, 
}

@misc{minimax2025minimaxm1scalingtesttimecompute,
      title={MiniMax-M1: Scaling Test-Time Compute Efficiently with Lightning Attention}, 
      author={MiniMax Team},
      year={2025},
      eprint={2506.13585},
      archivePrefix={arXiv},
      primaryClass={cs.CL},
      url={https://arxiv.org/abs/2506.13585}, 
}

@article{zhang2026aster,
  title={ASTER: Agentic Scaling with Tool-integrated Extended Reasoning},
  author={Zhang, Xuqin and He, Quan and Zheng, Zhenrui and Zhang, Zongzhang and He, Xu and Li, Dong},
  journal={arXiv preprint arXiv:2602.01204},
  year={2026}
}

@misc{wang2026openclawrltrainagentsimply,
      title={OpenClaw-RL: Train Any Agent Simply by Talking}, 
      author={Yinjie Wang and Xuyang Chen and Xiaolong Jin and Mengdi Wang and Ling Yang},
      year={2026},
      eprint={2603.10165},
      archivePrefix={arXiv},
      primaryClass={cs.CL},
      url={https://arxiv.org/abs/2603.10165}, 
}

@misc{zhang2026landscapeagenticreinforcementlearning,
      title={The Landscape of Agentic Reinforcement Learning for LLMs: A Survey}, 
      author={Guibin Zhang and Hejia Geng and Xiaohang Yu and Zhenfei Yin and Zaibin Zhang and Zelin Tan and Heng Zhou and Zhongzhi Li and Xiangyuan Xue and Yijiang Li and Yifan Zhou and Yang Chen and Chen Zhang and Yutao Fan and Zihu Wang and Songtao Huang and Francisco Piedrahita-Velez and Yue Liao and Hongru Wang and Mengyue Yang and Heng Ji and Jun Wang and Shuicheng Yan and Philip Torr and Lei Bai},
      year={2026},
      eprint={2509.02547},
      archivePrefix={arXiv},
      primaryClass={cs.AI},
      url={https://arxiv.org/abs/2509.02547}, 
}

@misc{hou2026singlerolloutasynchronousoptimizationagentic,
      title={Single-Rollout Asynchronous Optimization for Agentic Reinforcement Learning}, 
      author={Zhenyu Hou and Yujiang Li and Jie Tang and Yuxiao Dong},
      year={2026},
      eprint={2607.07508},
      archivePrefix={arXiv},
      primaryClass={cs.LG},
      url={https://arxiv.org/abs/2607.07508}, 
}

@misc{li2026compactionrlreinforcementlearningcontext,
      title={CompactionRL: Reinforcement Learning with Context Compaction for Long-Horizon Agents}, 
      author={Yujiang Li and Zhenyu Hou and Yi Jing and Jie Tang and Yuxiao Dong},
      year={2026},
      eprint={2607.05378},
      archivePrefix={arXiv},
      primaryClass={cs.LG},
      url={https://arxiv.org/abs/2607.05378}, 
}

@misc{jin2025searchr1trainingllmsreason,
      title={Search-R1: Training LLMs to Reason and Leverage Search Engines with Reinforcement Learning}, 
      author={Bowen Jin and Hansi Zeng and Zhenrui Yue and Jinsung Yoon and Sercan Arik and Dong Wang and Hamed Zamani and Jiawei Han},
      year={2025},
      eprint={2503.09516},
      archivePrefix={arXiv},
      primaryClass={cs.CL},
      url={https://arxiv.org/abs/2503.09516}, 
}

@misc{qian2025toolrlrewardtoollearning,
      title={ToolRL: Reward is All Tool Learning Needs}, 
      author={Cheng Qian and Emre Can Acikgoz and Qi He and Hongru Wang and Xiusi Chen and Dilek Hakkani-Tür and Gokhan Tur and Heng Ji},
      year={2025},
      eprint={2504.13958},
      archivePrefix={arXiv},
      primaryClass={cs.LG},
      url={https://arxiv.org/abs/2504.13958}, 
}

@misc{feng2025retoolreinforcementlearningstrategic,
      title={ReTool: Reinforcement Learning for Strategic Tool Use in LLMs}, 
      author={Jiazhan Feng and Shijue Huang and Xingwei Qu and Ge Zhang and Yujia Qin and Baoquan Zhong and Chengquan Jiang and Jinxin Chi and Wanjun Zhong},
      year={2025},
      eprint={2504.11536},
      archivePrefix={arXiv},
      primaryClass={cs.CL},
      url={https://arxiv.org/abs/2504.11536}, 
}

@article{ma2025stabilizing,
  title={Stabilizing moe reinforcement learning by aligning training and inference routers},
  author={Ma, Wenhan and Zhang, Hailin and Zhao, Liang and Song, Yifan and Wang, Yudong and Sui, Zhifang and Luo, Fuli},
  journal={arXiv preprint arXiv:2510.11370},
  year={2025}
}

@misc{liu-li-2025-rl-collapse,
  title = {When Speed Kills Stability: Demystifying {RL} Collapse from the Training-Inference Mismatch},
  author = {Liu, Jiacai and Li, Yingru and Fu, Yuqian and Wang, Jiawei and Liu, Qian and Shen, Yu},
  year = {2025},
  url = {https://richardli.xyz/rl-collapse}
}

@misc{skyrl_gym_generator,
  author = {{SkyRL Team}},
  title = {SkyRL Gym Generator Tutorial},
  howpublished = {\url{https://docs.skyrl.ai/docs/tutorials/skyrl_gym_generator}},
  year = {2025}
}

@misc{yao2025offpolicy,
  title = {Your Efficient RL Framework Secretly Brings You Off-Policy RL Training},
  url = {https://fengyao.notion.site/off-policy-rl},
  author = {Yao, Feng and Liu, Liyuan and Zhang, Dinghuai and Dong, Chengyu and Shang, Jingbo and Gao, Jianfeng},
  year = {2025},
}

@misc{kpop,
  title = {K-pop: Taming Training–Inference Mismatch in Reinforcement Learning with Adaptive Masking Regions},
  url = {https://ringtech.notion.site/kpop},
  author = {Jia Guo and Yan Sun and Zhenyu Huang and Zihao Wang and Zujie Wen and Zhiqiang Zhang and Jun Zhou and Stanley Kok},
  year = {2026},
}

@article{yu2026dapo,
  title={Dapo: An open-source llm reinforcement learning system at scale},
  author={Yu, Qiying and Zhang, Zheng and Zhu, Ruofei and Yuan, Yufeng and Zuo, Xiaochen and Yue, Yu and Dai, Weinan and Fan, Tiantian and Liu, Gaohong and Liu, Lingjun and others},
  journal={Advances in Neural Information Processing Systems},
  volume={38},
  pages={113222--113244},
  year={2026}
}

@article{wang2026verification,
  title={The Verification Horizon: No Silver Bullet for Coding Agent Rewards},
  author={Wang, Binghai and Zhang, Chenlong and Liu, Dayiheng and Zhang, Jiajun and Chen, Jiawei and Chen, Mouxiang and Fang, Rongyao and Zhang, Siyuan and Wang, Xuwu and Jing, Yuheng and others},
  journal={arXiv preprint arXiv:2606.26300},
  year={2026}
}

@article{wang2026reproducing,
  title={Reproducing, Analyzing, and Detecting Reward Hacking in Rubric-Based Reinforcement Learning},
  author={Wang, Xuekang and Hao, Zhuoyuan and Hou, Shuo and Peng, Hao and Li, Juanzi and Wang, Xiaozhi},
  journal={arXiv preprint arXiv:2606.04923},
  year={2026}
}

@article{patwardhan2025gdpval,
  title={Gdpval: Evaluating ai model performance on real-world economically valuable tasks},
  author={Patwardhan, Tejal and Dias, Rachel and Proehl, Elizabeth and Kim, Grace and Wang, Michele and Watkins, Olivia and Fishman, Sim{\'o}n Posada and Aljubeh, Marwan and Thacker, Phoebe and Fauconnet, Laurance and others},
  journal={arXiv preprint arXiv:2510.04374},
  year={2025}
}

@article{duan2023hand,
  title={Hand-in-Hand guidance: an explore-exploit based reinforcement learning method for performance driven assembly-adjustment},
  author={Duan, Guifang and Xu, Yunkun and Liu, Zhenyu and Tan, Jianrong},
  journal={IEEE Transactions on Industrial Informatics},
  volume={19},
  number={10},
  pages={10045--10055},
  year={2023},
  publisher={IEEE}
}

@misc{li2026saveloadlearn,
  title        = {Save, Load and Learn: Boosting Agentic LLMs via Rollback-based Curriculum Learning},
  author       = {Xiaoyang Li and Weixun Wang and Yancheng He},
  year         = {2026},
  howpublished = {\url{https://warm-pajama-44a.notion.site/Save-Load-and-Learn-Boosting-Agentic-LLMs-via-Rollback-based-Curriculum-Learning-687a76d7970e831a91c501bafd9c7b2b}},
}

@misc{lazaridis2026edgeopdinternalizingprivilegedcontext,
      title={EDGE-OPD: Internalizing Privileged Context with Evidence Guided On-Policy Distillation}, 
      author={Aristotelis Lazaridis and Dylan Bates and Aman Sharma and Brian King and Vincent Lu and Jack FitzGerald},
      year={2026},
      eprint={2605.23493},
      archivePrefix={arXiv},
      primaryClass={cs.AI},
      url={https://arxiv.org/abs/2605.23493}, 
}

@misc{zhao2026selfdistilledreasoneronpolicyselfdistillation,
      title={Self-Distilled Reasoner: On-Policy Self-Distillation for Large Language Models}, 
      author={Siyan Zhao and Zhihui Xie and Mengchen Liu and Jing Huang and Guan Pang and Feiyu Chen and Aditya Grover},
      year={2026},
      eprint={2601.18734},
      archivePrefix={arXiv},
      primaryClass={cs.LG},
      url={https://arxiv.org/abs/2601.18734}, 
}

@misc{wang2026hindsighthintdistillationscaffolded,
      title={Hindsight Hint Distillation: Scaffolded Reasoning for SWE Agents from CoT-free Answers}, 
      author={Shengjie Wang and Guanghe Li and Zonghan Yang and Yang Gao},
      year={2026},
      eprint={2605.11556},
      archivePrefix={arXiv},
      primaryClass={cs.AI},
      url={https://arxiv.org/abs/2605.11556}, 
}

@article{liu2024apigen,
  title={Apigen: Automated pipeline for generating verifiable and diverse function-calling datasets},
  author={Liu, Zuxin and Hoang, Thai and Zhang, Jianguo and Zhu, Ming and Lan, Tian and Kokane, Shirley and Tan, Juntao and Yao, Weiran and Liu, Zhiwei and Feng, Yihao and others},
  journal={Advances in Neural Information Processing Systems},
  volume={37},
  pages={54463--54482},
  year={2024}
}

@inproceedings{liu2025toolace,
  title={Toolace: Winning the points of llm function calling},
  author={Liu, Weiwen and Huang, Xu and Zeng, Xingshan and Yu, Shuai and Li, Dexun and Wang, Shuai and Gan, Weinan and Liu, Zhengying and Yu, Yuanqing and WANG, Zezhong and others},
  booktitle={International Conference on Learning Representations},
  volume={2025},
  pages={41359--41381},
  year={2025}
}

@article{tang2023toolalpaca,
  title={Toolalpaca: Generalized tool learning for language models with 3000 simulated cases},
  author={Tang, Qiaoyu and Deng, Ziliang and Lin, Hongyu and Han, Xianpei and Liang, Qiao and Cao, Boxi and Sun, Le},
  journal={arXiv preprint arXiv:2306.05301},
  year={2023}
}

@inproceedings{fang2026towards,
  title={Towards general agentic intelligence via environment scaling},
  author={Fang, Runnan and Cai, Shihao and Li, Baixuan and Wu, Jialong and Li, Guangyu and Yin, Wenbiao and Wang, Xinyu and Wang, Xiaobin and Su, Liangcai and Zhang, Zhen and others},
  booktitle={Findings of the Association for Computational Linguistics: ACL 2026},
  pages={17610--17621},
  year={2026}
}

@article{cai2025autoforge,
  title={AutoForge: Automated Environment Synthesis for Agentic Reinforcement Learning},
  author={Cai, Shihao and Fang, Runnan and Wu, Jialong and Li, Baixuan and Wang, Xinyu and Jiang, Yong and Su, Liangcai and Zhang, Liwen and Yin, Wenbiao and Zhang, Zhen and others},
  journal={arXiv preprint arXiv:2512.22857},
  year={2025}
}

@article{xu2026envfactory,
  title={EnvFactory: Scaling Tool-Use Agents via Executable Environments Synthesis and Robust RL},
  author={Xu, Minrui and Wang, Zilin and Deng, Mengyi and Li, Zhiwei and Yang, Zhicheng and Zhu, Xiao and Liu, Yinhong and Zhu, Boyu and Huang, Baiyu and Chen, Chao and others},
  journal={arXiv preprint arXiv:2605.18703},
  year={2026}
}

@article{chen2026dive,
  title={Dive: Scaling diversity in agentic task synthesis for generalizable tool use},
  author={Chen, Aili and Zhang, Chi and Liu, Junteng and Chen, Jiangjie and Du, Chengyu and Li, Yunji and Zhong, Ming and Wang, Qin and Zhu, Zhengmao and Song, Jiayuan and others},
  journal={arXiv preprint arXiv:2603.11076},
  year={2026}
}

@inproceedings{zhao2024expel,
  title={Expel: Llm agents are experiential learners},
  author={Zhao, Andrew and Huang, Daniel and Xu, Quentin and Lin, Matthieu and Liu, Yong-Jin and Huang, Gao},
  booktitle={Proceedings of the AAAI Conference on Artificial Intelligence},
  pages={19632--19642},
  year={2024}
}

@article{xia2025agent0,
  title={Agent0: Unleashing self-evolving agents from zero data via tool-integrated reasoning},
  author={Xia, Peng and Zeng, Kaide and Liu, Jiaqi and Qin, Can and Wu, Fang and Zhou, Yiyang and Xiong, Caiming and Yao, Huaxiu},
  journal={arXiv preprint arXiv:2511.16043},
  year={2025}
}

@misc{qwenteam2026qwen36,
    title = {{Qwen3.6-35B-A3B}: Agentic Coding Power, Now Open to All},
    url = {https://qwen.ai/blog?id=qwen3.6-35b-a3b},
    author = {{Qwen Team}},
    month = {April},
    year = {2026}
}

@misc{xie2023doremioptimizingdatamixtures,
      title={DoReMi: Optimizing Data Mixtures Speeds Up Language Model Pretraining}, 
      author={Sang Michael Xie and Hieu Pham and Xuanyi Dong and Nan Du and Hanxiao Liu and Yifeng Lu and Percy Liang and Quoc V. Le and Tengyu Ma and Adams Wei Yu},
      year={2023},
      eprint={2305.10429},
      archivePrefix={arXiv},
      primaryClass={cs.CL},
      url={https://arxiv.org/abs/2305.10429}, 
}

@misc{li2025datamixingoptimizationsupervised,
      title={Data Mixing Optimization for Supervised Fine-Tuning of Large Language Models}, 
      author={Yuan Li and Zhengzhong Liu and Eric Xing},
      year={2025},
      eprint={2508.11953},
      archivePrefix={arXiv},
      primaryClass={cs.AI},
      url={https://arxiv.org/abs/2508.11953}, 
}

@article{gao2026self,
  title={From Self-Evolving Synthetic Data to Verifiable-Reward RL: Post-Training Multi-turn Interactive Tool-Using Agents},
  author={Gao, Jiaxuan and Chen, Jiaao and He, Chuyi and Wang, Wei-Chen and Xu, Shusheng and Wang, Hanrui and Jin, Di and Wu, Yi},
  journal={arXiv preprint arXiv:2601.22607},
  year={2026}
}

\clearpage

\beginappendix

\definecolor{SoftMainColor}{HTML}{E8F1FC}
\definecolor{MidMainColor}{HTML}{709BCC}
\definecolor{LightMainColor}{HTML}{C6DAF2}
\definecolor{SoftSecondColor}{HTML}{EBFCFB}
\definecolor{MainColor}{HTML}{2165B5}
\definecolor{SecondColor}{HTML}{39C5BB}
\definecolor{MidSecondColor}{HTML}{7DCCC6}
\definecolor{LightSecondColor}{HTML}{CCF2F0}
\definecolor{DarkSecondColor}{HTML}{0F3330}
\definecolor{ThirdColor}{HTML}{B7A7FF}
\definecolor{LightThirdColor}{HTML}{E3DFF2}
\definecolor{AppendixBackGray}{HTML}{FCFCFC}
\providecommand{\toolentry}[2]{}

\renewcommand{\toolentry}[2]{%
  \par\noindent
  {\textbf\ttfamily\bfseries\detokenize{#1}}%
  \nobreak\enspace\textemdash\enspace
  #2%
  \par\vspace{0.46em}%
}
\newcolumntype{L}{>{\RaggedRight\arraybackslash}X}
\lstdefinestyle{PromptJSON}{
  basicstyle=\scriptsize,
  showstringspaces=false,
  breaklines=true,
  breakatwhitespace=false,
  columns=fullflexible,
  keepspaces=true,
  tabsize=2,
  frame=none,
  numbers=none,
  aboveskip=0pt,
  belowskip=0pt,
  xleftmargin=0pt
}
\newtcbox{\highlightword}{
  on line,
  colback=SoftMainColor,
  colframe=SoftMainColor,
  boxrule=0pt,
  arc=2pt,
  left=3pt,
  right=3pt,
  top=1pt,
  bottom=1pt,
  boxsep=0pt
}
\newtcbox{\highlightwordDarker}{
  on line,
  colback=LightMainColor,
  colframe=LightMainColor,
  boxrule=0pt,
  arc=2pt,
  left=3pt,
  right=3pt,
  top=1pt,
  bottom=1pt,
  boxsep=0pt
}
\newtcbox{\highlightwordsecond}{
  on line,
  colback=SoftSecondColor,
  colframe=SoftSecondColor,
  boxrule=0pt,
  arc=2pt,
  left=3pt,
  right=3pt,
  top=1pt,
  bottom=1pt,
  boxsep=0pt
}
\newtcbox{\highlightwordsecondDarker}{
  on line,
  colback=LightSecondColor,
  colframe=LightSecondColor,
  boxrule=0pt,
  arc=2pt,
  left=3pt,
  right=3pt,
  top=1pt,
  bottom=1pt,
  boxsep=0pt
}
\newtcbox{\highlightwordthirdDarker}{
  on line,
  colback=LightThirdColor,
  colframe=LightThirdColor,
  boxrule=0pt,
  arc=2pt,
  left=3pt,
  right=3pt,
  top=1pt,
  bottom=1pt,
  boxsep=0pt
}
\NewDocumentCommand{\StateEntityPair}{m m m +m}{%
  \noindent
  \begin{tcbitemize}[
    raster columns=2,
    raster equal height=rows,
    raster width=\linewidth,
    raster column skip=0.01\linewidth,
    raster row skip=0pt,
    raster before skip=2pt,
    raster after skip=5pt
  ]
  \tcbitem[
    colback=SoftSecondColor,
    colframe=MidSecondColor,
    boxrule=0.4pt,
    arc=1pt,
    title={\textbf{#1}},
    colbacktitle=LightSecondColor,
    coltitle=DarkSecondColor,
    fonttitle=\scriptsize,
    left=3pt,
    right=5pt,
    top=3pt,
    bottom=3pt
  ]
  \textit{<Description>}\\[1pt]
  #2\\[2pt]

  \textit{<Attributes>}\\[1pt]
  #3
  \tcbitem[
    colback=SoftSecondColor,
    colframe=MidSecondColor,
    boxrule=0.4pt,
    arc=1pt,
    title={\textbf{Class Definition}},
    colbacktitle=LightSecondColor,
    coltitle=DarkSecondColor,
    fonttitle=\scriptsize,
    left=3pt,
    right=5pt,
    top=3pt,
    bottom=3pt
  ]
  #4

  \end{tcbitemize}
}
\newcommand{\fielditem}[2]{%
  \par\noindent
  \hangindent=1em
  \hangafter=1
  \makebox[0.8em][l]{\raisebox{0.2ex}{\tiny$\cdot$}}%
  \textbf{#1}: #2%
  \par\vspace{0.15em}%
}

\lstdefinestyle{TraceJSON}{
  basicstyle=\ttfamily\scriptsize,
  showstringspaces=false,
  breaklines=true,
  breakatwhitespace=false,
  columns=fullflexible,
  keepspaces=true,
  tabsize=2,
  frame=none,
  numbers=none,
  aboveskip=0pt,
  belowskip=0pt,
  xleftmargin=0pt
}
\lstdefinestyle{AppendixPython}{
  language=Python,
  basicstyle=\ttfamily\scriptsize,
  keywordstyle=\bfseries,
  commentstyle=\itshape\color{MutedText},
  stringstyle=\color{ScaleGreen!70!black},
  showstringspaces=false,
  breaklines=true,
  breakatwhitespace=false,
  columns=fullflexible,
  keepspaces=true,
  tabsize=4,
  numbers=left,
  numberstyle=\tiny\color{MutedText},
  numbersep=8pt,
  xleftmargin=8pt,
  framexleftmargin=0pt,
  frame=none,
  aboveskip=0pt,
  belowskip=0pt
}
\newtcblisting{AppendixCode}[2][]{%
  enhanced,
  breakable,
  listing only,
  listing engine=listings,
  listing options={
    style=AppendixPython
  },
  colback=SoftSecondColor,
  colframe=MidSecondColor,
  boxrule=0.4pt,
  arc=1pt,
  left=8pt,
  right=6pt,
  top=5pt,
  bottom=5pt,
  title={\textbf{#2}},
  colbacktitle=LightSecondColor,
  coltitle=DarkSecondColor,
  fonttitle=\scriptsize,
  attach boxed title to top left={
    xshift=0pt,
    yshift=0pt
  },
  boxed title style={
    boxrule=0pt,
    arc=1pt
  },
  #1
}
\tcbset{
    threecolumnbox/.style={
        enhanced,
        coltitle=black,
        colbacktitle=LightMainColor,
        colback=white,
        colframe=MidMainColor,
        boxrule=0.6pt,
        arc=2pt,
        outer arc=2pt,
        left=1pt,
        right=1pt,
        top=2pt,
        bottom=2pt,
         boxed title style={
            boxrule=0pt,
            arc=1pt
          },
        attach boxed title to top left={
            xshift=0pt,
            yshift=0pt
        }
    }
}
\tcbset{
  ToolTraceStyle/.style={
    enhanced,
    colback=SoftSecondColor,
    colframe=MidSecondColor,
    boxrule=0.4pt,
    arc=1pt,
    colbacktitle=LightSecondColor,
    coltitle=DarkSecondColor,
    fonttitle=\scriptsize,
    left=5pt,
    right=5pt,
    top=4pt,
    bottom=4pt,
    before skip=0pt,
    after skip=0pt
  }
}

\section{Supplementary Environment and Task Synthesis}
\subsection{Hard Initial-State Construction}
\label{app:program_init_state}

We instantiate difficulty-oriented initial states using seven complementary construction operators:

\begin{itemize}
\item \textbf{Multi-candidate construction:} creates multiple plausible candidates for the target role or decision.
\item \textbf{Distractor injection:} introduces superficially relevant records that are excluded by status, capability, evidence, or relational constraints.
\item \textbf{Boundary and tie construction:} places candidates near decision thresholds or ties them on primary criteria, requiring secondary rules for resolution.
\item \textbf{Cross-entity evidence distribution:} distributes required evidence across primary entities, associated records, configuration objects, logs, and identifier references.
\item \textbf{Aggregatable history construction:} introduces historical records that require counting, summation, averaging, ranking, or threshold comparison.
\item \textbf{Mutable-object construction:} provides pending, assignable, updatable, or archivable objects that support valid state transitions.
\item \textbf{Closed-loop opportunity construction:} ensures that the state supports a complete workflow involving evidence collection, decision making, and subsequent environment modification.
\end{itemize}
\subsection{Executable Environment Example}
\providecommand{\caseusertag}{}
\renewcommand{\caseusertag}{%
  \fcolorbox{RoleUser}{FoundationBlue!12}{\scriptsize\textsf{\textcolor{RoleUser}{User}}}%
}
\providecommand{\caseassistanttag}{}
\renewcommand{\caseassistanttag}{%
  \fcolorbox{RoleAssistant}{ScaleGreen!12}{\scriptsize\textsf{\textcolor{RoleAssistant}{Agent}}}%
}
\providecommand{\casetooltag}{}
\renewcommand{\casetooltag}{%
  \fcolorbox{RoleTool}{AgentOrange!12}{\scriptsize\textsf{\textcolor{RoleTool}{Tool}}}%
}
{\small
\begin{tcolorbox}[
  enhanced,
  breakable,
  colback=white,
  colframe=MidMainColor,
  boxrule=0.6pt,
  arc=1.5pt,
  title={\textbf{Environment Case Example: Healthcare Provider Registry System}},
  colbacktitle=SoftMainColor,
  coltitle=black,
  fonttitle=\bfseries\small,
  left=4pt, right=4pt, top=3pt, bottom=3pt
]
\scriptsize

\textbf{\color{MainColor}[1] Environment Introduction} \\[4pt]
This environment maintains a centralized database of healthcare providers, their administrative details, and contractual partnerships with insurance carriers. The system stores clinic registration information including names, affiliation codes, physical addresses, and associated postal codes (CEPs), while also tracking which providers are credentialed with specific insurance networks. Typical operations involve querying provider-insurance relationships, validating address data against postal code databases, and updating provider status information. 

\tcbline
\textbf{\color{MainColor}[2] State Space}\\[1pt]
\StateEntityPair
  {Insurance Carrier}
  {%
    Health insurance provider organization that offers services and enters into partnerships.
  }
  {%
    carrier\_code, name, website, registration\_number, status
  }
  {%
    \texttt{class InsuranceCarrierDict(TypedDict):}\\[2pt]
    \texttt{\hspace*{1em}carrier\_code: str}\\[2pt]
    \texttt{\hspace*{1em}name: str}\\[2pt]
    \texttt{\hspace*{1em}website: str}\\[2pt]
    \texttt{\hspace*{1em}registration\_number: str}\\[2pt]
    \texttt{\hspace*{1em}status: str}
  }
\StateEntityPair
  {Partnership Review}
  {%
    A formal review of an active partnership to ensure ongoing compliance with contractual terms.
  }
  {%
    review\_id, partnership\_id, reviewed\_by\_user\_id, review\_timestamp, review\_result
  }
  {%
    \texttt{class PartnershipReviewDict(TypedDict):}\\[2pt]
    \texttt{\hspace*{1em}review\_id: str}\\[2pt]
    \texttt{\hspace*{1em}partnership\_id: str}\\[2pt]
    \texttt{\hspace*{1em}reviewed\_by\_user\_id: str}\\[2pt]
    \texttt{\hspace*{1em}review\_timestamp: str}\\[2pt]
    \texttt{\hspace*{1em}review\_result: str}
  }

\StateEntityPair {Clinic} 
    {% 
    Registered healthcare provider facility with administrative details and lifecycle status. } 
    {% 
    clinic\_id, name, cep, street\_address, unit\_number, status, phone\_number, email\_address 
    } 
    {% 
    \texttt{class ClinicDict(TypedDict):}\\[2pt] 
    \texttt{\hspace*{1em}clinic\_id: str}\\[2pt] 
    \texttt{\hspace*{1em}name: str}\\[2pt] 
    \texttt{\hspace*{1em}cep: str}\\[2pt] 
    \texttt{\hspace*{1em}street\_address: str}\\[2pt] 
    \texttt{\hspace*{1em}unit\_number: str}\\[2pt] 
    \texttt{\hspace*{1em}status: str}\\[2pt] 
    \texttt{\hspace*{1em}phone\_number: str}\\[2pt] 
    \texttt{\hspace*{1em}email\_address: str} }
{\color{MutedText}\textit{[\ldots additional states omitted \ldots]}}

\tcbline

\textbf{\color{MainColor}[3] Tools} \\[1pt]
\toolentry{get_clinic}
{Retrieve clinic record(s) by clinic ID or name.}
\toolentry{list_all_clinics}
{List all clinic records currently in the system.}

\toolentry{get_insurance_carrier}
{Retrieve insurance carrier record(s) by code or name.}

\toolentry{list_all_insurance_carriers}
{List all insurance carrier records in the system.}

\toolentry{get_partnership}
{Retrieve partnership record(s) by ID or by clinic and insurer combination.}

\toolentry{list_partnerships_for_clinic}
{List all partnerships associated with a given clinic ID.}

\toolentry{list_partnerships_for_insurer}
{List all partnerships associated with a given insurance carrier code.}

\toolentry{get_postal_code_by_cep}
{Retrieve a postal code record by its CEP.}

\toolentry{search_postal_codes_by_city}
{Search for postal codes in a given city.}

\toolentry{check_address_verification}
{Check whether a clinic's address is verified by verifying that its CEP exists in the PostalCode database.}

\toolentry{get_user}
{Retrieve a user record by user ID.}

\toolentry{get_user_role}
{Retrieve a user role record by role ID.}
{\color{MutedText}\textit{[\ldots additional tools omitted \ldots]}}
\clearpage
\textbf{\color{ActionBlue}[4] Tool Examples} \\[1pt]
\begin{AppendixCode}{\texttt{list\_all\_clinics}}
def list_all_clinics(self) -> Dict:
    clinics_list = list(self.clinics.values())
    return {
        "data": clinics_list
    }
\end{AppendixCode}
\begin{AppendixCode}{\texttt{list\_partnerships\_for\_insurer}}
def list_partnerships_for_insurer(self, insurer_code: str) -> Dict:
    if not insurer_code or not isinstance(insurer_code, str):
        return {
            "error": "Invalid insurer_code: must be a non-empty string"
        }
    if insurer_code not in self.insurance_carriers:
        return {
            "error": (
                f"Insurance carrier with code "
                f"'{insurer_code}' not found"
            )
        }
    matching_partnerships = [
        partnership
        for partnership in self.partnerships.values()
        if partnership["insurer_code"] == insurer_code
    ]
    return {
        "data": matching_partnerships
    }
\end{AppendixCode}
\color{MutedText}\textit{[\ldots additional tools omitted \ldots]}
\end{tcolorbox}
}

\subsection{DAG-Based Task Example}
{\small
\begin{tcolorbox}[
  enhanced,
  breakable,
  colback=white,
  colframe=MidMainColor,
  boxrule=0.6pt,
  arc=1.5pt,
  title={\textbf{DAG-Task Case Example: Healthcare Provider Registry System}},
  colbacktitle=SoftMainColor,
  coltitle=black,
  fonttitle=\bfseries\small,
  left=4pt, right=4pt, top=3pt, bottom=3pt
]
\scriptsize
\textbf{\color{MainColor}[1] Task Description} \\[4pt]
\begin{tabularx}{\linewidth}{
    @{}
    L
    @{\hspace{8pt}}
    L
    @{\hspace{8pt}}
    L
    @{}
}
    \highlightwordsecondDarker{Domain:}\newline
    ~~~~~~ - Healthcare \newline
    ~~~~~~ - Healthcare Services
    &
    \highlightwordDarker{Capability:} \newline
    ~~~~~~- Planning \& Decision Making \newline
    ~~~~~~- Reliability \& Safety
    &
    \highlightwordthirdDarker{Atomic Difficulty:}\newline
    ~~~~~~- Dynamic Multi-step Planning\newline
    ~~~~~~- Risk, Reliability, and Clarification
\end{tabularx}\\[4pt]
Using insurance administrator \texttt{user-002}, handle a narrowly scoped insurer/clinic compliance reconciliation centered on \textbf{Blue Cross Blue Shield} and \textbf{Oakwood Community Clinic}, but make sure the clinic does not move out of its current onboarding state until the discrepancy work and partnership corrections support that transition. Before any insurer or clinic relationship maintenance occurs, establish that this operator's authority is appropriate for insurer administration and leave auditable evidence on the Blue Cross Blue Shield insurer record that \texttt{user-002} initiated a targeted insurer and clinic partnership compliance review. In the insurer portion, the Blue Cross Blue Shield record should remain \textbf{active}, but its public link needs to be corrected to \texttt{https://providers.bcbs.com}. Separately, confirm that \textbf{CEP} \texttt{90210} aligns with \textbf{Los Angeles} and that the clinic's location has already been validated. Once that reconciliation is settled, add a new future arrangement with \textbf{HealthGuard}: \textbf{PPO}, \texttt{NET-D}, beginning \texttt{2026-06-01} and ending \texttt{2027-05-31}. Separately, update Oakwood's existing Blue Cross relationship by extending its expiration through \texttt{2027-06-01} and changing the benefit model to \textbf{PPO}. After the Blue Cross correction is in place, register a successful assessment for that relationship using \texttt{REV-010} at \texttt{2026-05-20T09:20:00}. If the transition is valid, move Oakwood Community Clinic to \textbf{active} status, then confirm the refreshed clinic profile, review its insurer relationships, verify that the corrected Blue Cross relationship is active, and retrieve audit evidence for both the clinic and the partnership.
\tcbline

\textbf{\color{MainColor}[2] Execution Trajectory} \\[1pt]
\begin{tcolorbox}[
  ToolTraceStyle,
  title={  
    \makebox[\linewidth][l]{%
    \parbox[t]{0.46\linewidth}{\textbf{Step 1}}%
    \hfill
    \parbox[t]{0.48\linewidth}{\texttt{get\_user}}%
    }
  }
]
\noindent
\begin{minipage}[t]{0.46\linewidth}
\vspace{0pt}
\textit{<Parameters>}\\[2pt]
\begin{lstlisting}[style=TraceJSON]
{
    "user_id": "user-002"
}
\end{lstlisting}
\end{minipage}
\hfill
{\color{MidSecondColor}\vrule width 0.4pt}
\hfill
\begin{minipage}[t]{0.48\linewidth}
\vspace{0pt}
\textit{<Observation>}\\[2pt]
\begin{lstlisting}[style=TraceJSON]
{
    "data": {
        "user_id": "user-002",
        "username": "john.smith",
        "role_id": "role-insurance-admin"
    }
}
\end{lstlisting}
\end{minipage}
\end{tcolorbox}

\begin{tcolorbox}[
  ToolTraceStyle,
  title={  
    \makebox[\linewidth][l]{%
    \parbox[t]{0.46\linewidth}{\textbf{Step 2}}%
    \hfill
    \parbox[t]{0.48\linewidth}{\texttt{check\_user\_permission}}%
    }
  }
]
\noindent
\begin{minipage}[t]{0.46\linewidth}
\vspace{0pt}
\textit{<Parameters>}\\[2pt]
\begin{lstlisting}[style=TraceJSON]
{
    "user_id": "user-002",
    "permission": "admin_insurance"
}
\end{lstlisting}
\end{minipage}
\hfill
{\color{MidSecondColor}\vrule width 0.4pt}
\hfill
\begin{minipage}[t]{0.48\linewidth}
\vspace{0pt}
\textit{<Observation>}\\[2pt]
\begin{lstlisting}[style=TraceJSON]
{
    "data": {
        "has_permission": true,
        "role_name": "Insurance Administrator",
        "permissions": [
            "admin_insurance",
            "manage_insurance_carriers",
            "view_clinics"
        ]
    }
}
\end{lstlisting}
\end{minipage}
\end{tcolorbox}

\begin{tcolorbox}[
  ToolTraceStyle,
  title={  
    \makebox[\linewidth][l]{%
    \parbox[t]{0.46\linewidth}{\textbf{Step 3}}%
    \hfill
    \parbox[t]{0.48\linewidth}{\texttt{log\_access\_attempt}}%
    }
  }
]
\noindent
\begin{minipage}[t]{0.46\linewidth}
\vspace{0pt}
\textit{<Parameters>}\\[2pt]
\begin{lstlisting}[style=TraceJSON]
{
    "verified_entity_id": "BCBS-01",
    "verified_entity_type": "InsuranceCarrier",
    "field_changed": "access_attempt",
    "old_value": "",
    "new_value": "compliance_review_and_network_update",
    "changed_by_user_id": "user-002",
    "change_timestamp": "2026-05-20T09:00:00",
    "verification_method": "api_call",
    "verification_notes": "Initiated targeted insurer and clinic partnership compliance review."
}
\end{lstlisting}
\end{minipage}
\hfill
{\color{MidSecondColor}\vrule width 0.4pt}
\hfill
\begin{minipage}[t]{0.48\linewidth}
\vspace{0pt}
\textit{<Observation>}\\[2pt]
\begin{lstlisting}[style=TraceJSON]
{
    "message": "Access attempt logged successfully with log ID LOG_000004."
}
\end{lstlisting}
\end{minipage}
\end{tcolorbox}

\begin{tcolorbox}[
  ToolTraceStyle,
  title={  
    \makebox[\linewidth][l]{%
    \parbox[t]{0.46\linewidth}{\textbf{Step 4}}%
    \hfill
    \parbox[t]{0.48\linewidth}{\texttt{list\_all\_insurance\_carriers}}%
    }
  }
]
\noindent
\begin{minipage}[t]{0.46\linewidth}
\vspace{0pt}
\textit{<Parameters>}\\[2pt]
\begin{lstlisting}[style=TraceJSON]
{}
\end{lstlisting}
\end{minipage}
\hfill
{\color{MidSecondColor}\vrule width 0.4pt}
\hfill
\begin{minipage}[t]{0.48\linewidth}
\vspace{0pt}
\textit{<Observation>}\\[2pt]
\begin{lstlisting}[style=TraceJSON]
{
    "data": [
        {
            "carrier_code": "BCBS-01",
            "name": "Blue Cross Blue Shield",
            "website": "https://www.bcbs.com",
            "registration_number": "REG-1001",
            "status": "Active"
        },
        ...
    ]
}
\end{lstlisting}
\end{minipage}
\end{tcolorbox}

\begin{tcolorbox}[
  ToolTraceStyle,
  title={  
    \makebox[\linewidth][l]{%
    \parbox[t]{0.46\linewidth}{\textbf{Step 5}}%
    \hfill
    \parbox[t]{0.48\linewidth}{\texttt{get\_insurance\_carrier}}%
    }
  }
]
\noindent
\begin{minipage}[t]{0.46\linewidth}
\vspace{0pt}
\textit{<Parameters>}\\[2pt]
\begin{lstlisting}[style=TraceJSON]
{
    "carrier_code": "BCBS-01",
    "name": null
}
\end{lstlisting}
\end{minipage}
\hfill
{\color{MidSecondColor}\vrule width 0.4pt}
\hfill
\begin{minipage}[t]{0.48\linewidth}
\vspace{0pt}
\textit{<Observation>}\\[2pt]
\begin{lstlisting}[style=TraceJSON]
{
    "data": {
        "carrier_code": "BCBS-01",
        "name": "Blue Cross Blue Shield",
        "website": "https://www.bcbs.com",
        "registration_number": "REG-1001",
        "status": "Active"
    }
}
\end{lstlisting}
\end{minipage}
\end{tcolorbox}
\vspace{3pt}
\color{MutedText}\textit{[\ldots additional steps omitted \ldots]}
\clearpage
\textbf{\color{MainColor}[3] Rubrics}\\[1pt]

\begin{tabularx}{\linewidth}{@{}p{0.485\linewidth}@{\hspace{0.01\linewidth}}p{0.485\linewidth}@{}}
\begin{minipage}[t]{\linewidth}
\begin{tcolorbox}[
  colback=SoftSecondColor,
  colframe=MidSecondColor,
  boxrule=0.4pt,
  arc=1pt,
  title={\textbf{Rubric G1}\hfill\textbf{3/3 pts}},
  colbacktitle=LightSecondColor,
  coltitle=DarkSecondColor,
  fonttitle=\scriptsize,
  left=3pt,right=5pt,top=3pt,bottom=3pt
]
\textit{<Criterion>}\\[1pt]
Whether the work was performed under user-002 after validating that this operator has insurance-administration authority relevant to the case, rather than making insurer and partnership changes under an unauthorized or different user. \\[2pt]
\end{tcolorbox}
\end{minipage}
\newline
\color{MutedText}\textit{[\ldots additional general rubrics omitted \ldots]}
&
\begin{minipage}[t]{\linewidth}
\begin{tcolorbox}[
  colback=SoftSecondColor,
  colframe=MidSecondColor,
  boxrule=0.4pt,
  arc=1pt,
  title={\textbf{Rubric T1}\hfill\textbf{3/3 pts}},
  colbacktitle=LightSecondColor,
  coltitle=DarkSecondColor,
  fonttitle=\scriptsize,
  left=3pt,right=5pt,top=3pt,bottom=3pt
]
\textit{<Criterion>}\\[1pt]
Whether an audit log entry was created against the Blue Cross Blue Shield insurer record for the access attempt by user-002, with notes indicating that the action was an initiated targeted insurer and clinic partnership compliance review. \\[2pt]
\end{tcolorbox}
\end{minipage}
\newline
\color{MutedText}\textit{[\ldots additional task-specific rubrics omitted \ldots]}
\end{tabularx} 
\end{tcolorbox}
}

\subsection{Program-Based Task Example}
{\small
\begin{tcolorbox}[
  enhanced,
  breakable,
  colback=white,
  colframe=MidMainColor,
  boxrule=0.6pt,
  arc=1.5pt,
  title={\textbf{Program-Task Case Example: FloraDB Wiki and Data File System}},
  colbacktitle=SoftMainColor,
  coltitle=black,
  fonttitle=\bfseries\small,
  left=4pt, right=4pt, top=3pt, bottom=3pt
]

\scriptsize
\textbf{\color{MainColor}[1] Task Description} \\[4pt]
\begin{tabularx}{\linewidth}{
    @{}
    L
    @{\hspace{8pt}}
    L
    @{\hspace{8pt}}
    L
    @{}
}
    \highlightwordsecondDarker{Domain:}\newline
    ~~~~~~ - Books \newline
    ~~~~~~ - E-books
    &
    \highlightwordDarker{Capability:} \newline
    ~~~~~~- Data Analysis \newline
    ~~~~~~- Planning \& Decision Making
    &
    \highlightwordthirdDarker{Atomic Difficulty:}\newline
    ~~~~~~- Structured-information Complexity\newline
    ~~~~~~- Dynamic Multi-step Planning
\end{tabularx}\\[4pt]
I want to clear \textbf{exactly one worthwhile piece} of the lingering species-reference cleanup on our draft wiki workbench, not do a broad sweep. Please look across the \textbf{unfinished sessions that belong to people who can actually edit} in this staff group: \texttt{helen.carter}, \texttt{jonah.lee}, \texttt{priya.shah}, \texttt{samuel.reed}, \texttt{nora.kim}, \texttt{alice.nguyen}, \texttt{marco.silva}. Pick \textbf{just one session and its page} where finishing the linking work now is genuinely the best use of effort. I’m not looking for the first unfinished item on the list. We’ve had a mix of drafts, published pages, side tasks, and sessions that sound close to done already, so weigh the choice a bit: \textbf{the owner should have edit rights}, \textbf{the page should still be a draft}, the task should really be about \textbf{species-linking or reference cleanup}, the page should contain a \textbf{nontrivial set of detectable species mentions}, \textbf{some of those mentions should already be formally connected while others are still missing}, and the session history should suggest we’re close enough that closing the gap now makes sense. If something is vague, tied to the wrong kind of page, already effectively resolved, or not clearly about this linking cleanup, leave it out of consideration. Once you decide which single draft page is the strongest candidate, finish the formal page-to-species linking on that page for \textbf{every detectable species mention that already has a matching species record but is still missing its formal connection}. \textbf{Keep any existing formal links in place and don’t duplicate them.} If a mention doesn’t have a matching species record, leave it alone. I do want the work to leave a \textbf{short audit trail on the chosen session} so someone reviewing later can see both that \textbf{the missing links were checked} and that \textbf{the completion work was done}. If, after your work, \textbf{nothing remains unlinked on that page among the detectable matched species}, close that session out as \textbf{completed}. If there really isn’t a clearly suitable session/page under these rules, \textbf{make no changes} and say that nothing was selected.

\vspace{2pt} Return the result as a JSON object with \textbf{exactly these fields and meanings}:
\vspace{2pt}
\fielditem{selected\_session\_id}
{string or null --- the chosen task session ID, or null if no eligible session was acted on.}

\fielditem{selected\_page\_id}
{string or null --- the selected draft wiki page ID, or null if none was acted on.}

\fielditem{selected\_page\_title}
{string or null --- the title of the selected page, or null if none was acted on.}

\fielditem{links\_created\_count}
{number --- count of new formal page-to-species links actually created.}

\fielditem{species\_detected\_count}
{number --- total detectable species mentions on the selected page; use 0 if no page was selected.}

\fielditem{missing\_links\_before\_count}
{number --- how many detectable matched species mentions lacked formal links before work began; use 0 if no page was selected.}

\fielditem{session\_completed}
{boolean --- true if the selected session was moved to completed because no missing formal links remained afterward; false otherwise.}

\fielditem{state\_modified}
{boolean --- true if any state-changing action was successfully performed; false if nothing was changed.}

\tcbline

\textbf{\color{MainColor}[2] Execution Trajectory} \\[1pt]
\begin{AppendixCode}{Ground Truth}
    selected_session_id                                TS010
    selected_page_id                                   WP011
    selected_page_title                                Low-Maintenance Dry Border Trials
    links_created_count                                3
    species_detected_count                             6
    missing_links_before_count                         3
    session_completed                                  true
    state_modified                                     true
\end{AppendixCode}
\vspace{3pt}
\begin{tcolorbox}[
  ToolTraceStyle,
  title={  
    \makebox[\linewidth][l]{%
    \parbox[t]{0.46\linewidth}{\textbf{Step 1}}%
    \hfill
    \parbox[t]{0.48\linewidth}{\texttt{get\_user\_by\_username}}%
    }
  }
]
\noindent
\begin{minipage}[t]{0.46\linewidth}
\vspace{0pt}
\textit{<Parameters>}\\[2pt]
\begin{lstlisting}[style=TraceJSON]
{
    "username": "helen.carter"
}
\end{lstlisting}
\end{minipage}
\hfill
{\color{MidSecondColor}\vrule width 0.4pt}
\hfill
\begin{minipage}[t]{0.48\linewidth}
\vspace{0pt}
\textit{<Observation>}\\[2pt]
\begin{lstlisting}[style=TraceJSON]
{
    "data": {
        "id": "USR003",
        "username": "helen.carter",
        "role": "editor"
    }
}
\end{lstlisting}
\end{minipage}
\end{tcolorbox}

\begin{tcolorbox}[
  ToolTraceStyle,
  title={  
    \makebox[\linewidth][l]{%
    \parbox[t]{0.46\linewidth}{\textbf{Step 2}}%
    \hfill
    \parbox[t]{0.48\linewidth}{\texttt{list\_all\_wiki\_pages}}%
    }
  }
]
\noindent
\begin{minipage}[t]{0.46\linewidth}
\vspace{0pt}
\textit{<Parameters>}\\[2pt]
\begin{lstlisting}[style=TraceJSON]
{}
\end{lstlisting}
\end{minipage}
\hfill
{\color{MidSecondColor}\vrule width 0.4pt}
\hfill
\begin{minipage}[t]{0.48\linewidth}
\vspace{0pt}
\textit{<Observation>}\\[2pt]
\begin{lstlisting}[style=TraceJSON]
{
    "data": [
        {
            "id": "WP011",
            "title": "Low-Maintenance Dry Border Trials"
        },
        ...
    ]
}
\end{lstlisting}
\end{minipage}
\end{tcolorbox}
\vspace{3pt}
\color{MutedText}\textit{[\ldots intermediate steps omitted \ldots]}\\
\begin{tcolorbox}[
  ToolTraceStyle,
  title={  
    \makebox[\linewidth][l]{%
    \parbox[t]{0.46\linewidth}{\textbf{Step 4}}%
    \hfill
    \parbox[t]{0.48\linewidth}{\texttt{get\_task\_sessions\_by\_user}}%
    }
  }
]
\noindent
\begin{minipage}[t]{0.46\linewidth}
\vspace{0pt}
\textit{<Parameters>}\\[2pt]
\begin{lstlisting}[style=TraceJSON]
{
    "user_id": "USR003"
}
\end{lstlisting}
\end{minipage}
\hfill
{\color{MidSecondColor}\vrule width 0.4pt}
\hfill
\begin{minipage}[t]{0.48\linewidth}
\vspace{0pt}
\textit{<Observation>}\\[2pt]

\begin{lstlisting}[style=TraceJSON,escapeinside={(*@}{@*)}]
{
    "data": [
        {
            (*@\highlightwordDarker{"id": "TS010"}@*),
            "user_id": "USR003",
            "task_description": "Finalize low-maintenance dry border draft by checking parsed species against existing content references.",
            "status": "in_progress",
            "current_step": 3,
            "step_results": [
                {
                    "action": "opened_page",
                    "output": "Loaded WP011 draft for cross-check"
                },
                {
                    "action": "parsed_species",
                    "output": "Detected Lavandula angustifolia, Salvia officinalis, Thymus vulgaris, Origanum vulgare, Achillea millefolium, and Rosmarinus officinalis"
                },
                {
                    "action": "reference_check",
                    "output": "Existing references present for Lavender, Thyme, and Yarrow only"
                }
            ]
        },
        ...
    ]
}
\end{lstlisting}
\end{minipage}
\end{tcolorbox}
\vspace{3pt}
\color{MutedText}\textit{[\ldots intermediate steps omitted \ldots]}\\
\begin{tcolorbox}[
  ToolTraceStyle,
  title={  
    \makebox[\linewidth][l]{%
    \parbox[t]{0.46\linewidth}{\textbf{Step 17}}%
    \hfill
    \parbox[t]{0.48\linewidth}{\texttt{get\_wiki\_page\_by\_id}}%
    }
  }
]
\noindent
\begin{minipage}[t]{0.46\linewidth}
\vspace{0pt}
\textit{<Parameters>}\\[2pt]
\begin{lstlisting}[style=TraceJSON]
{
    "page_id": "WP011"
}
\end{lstlisting}
\end{minipage}
\hfill
{\color{MidSecondColor}\vrule width 0.4pt}
\hfill
\begin{minipage}[t]{0.48\linewidth}
\vspace{0pt}
\textit{<Observation>}\\[2pt]
\begin{lstlisting}[style=TraceJSON,escapeinside={(*@}{@*)}]
{
    "data": {
        (*@\highlightwordDarker{"id": "WP011"}@*),
        (*@\highlightwordDarker{"title": "Low-Maintenance Dry Border Trials"}@*),
        "content": "Trial notes compare Lavandula angustifolia, Salvia officinalis, Thymus vulgaris, Origanum vulgare, and Achillea millefolium in exposed beds. Lavender, Thyme, and Yarrow performed well through a dry summer, while Rosemary remained acceptable in sheltered plots.",
        "created_at": "2024-04-05T08:30:00",
        "modified_at": "2024-05-07T17:45:00",
        "status": "draft",
        "author_id": "USR003"
    }
}
\end{lstlisting}
\end{minipage}
\end{tcolorbox}
~\\
\color{MutedText}\textit{[\ldots intermediate steps omitted \ldots]}\\

\begin{tcolorbox}[
  ToolTraceStyle,
  title={  
    \makebox[\linewidth][l]{%
    \parbox[t]{0.46\linewidth}{\textbf{Step 28}}%
    \hfill
    \parbox[t]{0.48\linewidth}{\texttt{get\_content\_references\_by\_wiki\_page}}%
    }
  }
]
\noindent
\begin{minipage}[t]{0.46\linewidth}
\vspace{0pt}
\textit{<Parameters>}\\[2pt]
\begin{lstlisting}[style=TraceJSON]
{
    "page_id": "WP011"
}
\end{lstlisting}
\end{minipage}
\hfill
{\color{MidSecondColor}\vrule width 0.4pt}
\hfill
\begin{minipage}[t]{0.48\linewidth}
\vspace{0pt}
\textit{<Observation>}\\[2pt]
\begin{lstlisting}[style=TraceJSON,escapeinside={(*@}{@*)}]
{
    "data": [
        {
            "id": "CR022",
            "wiki_page_id": "WP011",
            "species_record_id": "SR0001"
        },
        {
            "id": "CR023",
            "wiki_page_id": "WP011",
            "species_record_id": "SR0007"
        },
        {
            "id": "CR024",
            "wiki_page_id": "WP011",
            "species_record_id": "SR0015"
        }
    ]
}
\end{lstlisting}
\end{minipage}
\end{tcolorbox}
\begin{tcolorbox}[
  ToolTraceStyle,
  title={  
    \makebox[\linewidth][l]{%
    \parbox[t]{0.46\linewidth}{\textbf{Step 29}}%
    \hfill
    \parbox[t]{0.48\linewidth}{\texttt{create\_content\_reference}}%
    }
  }
]
\noindent
\begin{minipage}[t]{0.46\linewidth}
\vspace{0pt}
\textit{<Parameters>}\\[2pt]
\begin{lstlisting}[style=TraceJSON]
{
    "wiki_page_id": "WP011",
    "species_record_id": "SR0002",
    "user_id": "USR003"
}
\end{lstlisting}
\end{minipage}
\hfill
{\color{MidSecondColor}\vrule width 0.4pt}
\hfill
\begin{minipage}[t]{0.48\linewidth}
\vspace{0pt}
\textit{<Observation>}\\[2pt]
\begin{lstlisting}[style=TraceJSON,escapeinside={(*@}{@*)}]
{
    "message": "Content reference created successfully.",
    "content_reference_id": "cr_27"
}
\end{lstlisting}
\end{minipage}
\end{tcolorbox}
\vspace{3pt}
\color{MutedText}\textit{[\ldots step 30-31: create\_content\_reference omitted \ldots]}\\
\begin{tcolorbox}[
  ToolTraceStyle,
  title={  
    \makebox[\linewidth][l]{%
    \parbox[t]{0.46\linewidth}{\textbf{Step 32}}%
    \hfill
    \parbox[t]{0.48\linewidth}{\texttt{update\_task\_session\_step}}%
    }
  }
]
\noindent
\begin{minipage}[t]{0.46\linewidth}
\vspace{0pt}
\textit{<Parameters>}\\[2pt]
\begin{lstlisting}[style=TraceJSON,escapeinside={(*@}{@*)}]
{
    "session_id": "TS010",
    "user_id": "USR003",
    "step_result": {(*@~~\highlightwordDarker{"output": "Created 3 new formal species links}\\
            \highlightwordDarker{for Low-Maintenance Dry Border Trials."}
        @*)
        "action": "links_completed"
    }
}
\end{lstlisting}
\end{minipage}
\hfill
{\color{MidSecondColor}\vrule width 0.4pt}
\hfill
\begin{minipage}[t]{0.48\linewidth}
\vspace{0pt}
\textit{<Observation>}\\[2pt]
\begin{lstlisting}[style=TraceJSON,escapeinside={(*@}{@*)}]
{
    "message": "Task session step updated successfully."
}
\end{lstlisting}
\end{minipage}
\end{tcolorbox}
\begin{tcolorbox}[
  ToolTraceStyle,
  title={  
    \makebox[\linewidth][l]{%
    \parbox[t]{0.46\linewidth}{\textbf{Step 33}}%
    \hfill
    \parbox[t]{0.48\linewidth}{\texttt{parse\_species\_names\_from\_wiki\_content}}%
    }
  }
]
\noindent
\begin{minipage}[t]{0.46\linewidth}
\vspace{0pt}
\textit{<Parameters>}\\[2pt]
\begin{lstlisting}[style=TraceJSON,escapeinside={(*@}{@*)}]
{
    "wiki_page_id": "WP011"
}
\end{lstlisting}
\end{minipage}
\hfill
{\color{MidSecondColor}\vrule width 0.4pt}
\hfill
\begin{minipage}[t]{0.48\linewidth}
\vspace{0pt}
\textit{<Observation>}\\[2pt]
\begin{lstlisting}[style=TraceJSON,escapeinside={(*@}{@*)}]
{
    "data": {
    "wiki_page_id": "WP011",
    "wiki_page_title": "Low-Maintenance Dry Border Trials",
    "found_species": [
        {
            "scientific_name": "Thymus vulgaris",
            "genus": "Thymus",
            "specific_epithet": "vulgaris",
            "scientific_name_id": "SN007",
            "species_record_id": "SR0007",
            "species_record_name": "Thyme",
            "has_content_reference": true
        },
        ...
    ],
    (*@\highlightwordDarker{"total\_found": 6}@*)
}
\end{lstlisting}
\end{minipage}
\end{tcolorbox}
\begin{tcolorbox}[
  ToolTraceStyle,
  title={  
    \makebox[\linewidth][l]{%
    \parbox[t]{0.46\linewidth}{\textbf{Step 34}}%
    \hfill
    \parbox[t]{0.48\linewidth}{\texttt{update\_task\_session\_status}}%
    }
  }
]
\noindent
\begin{minipage}[t]{0.46\linewidth}
\vspace{0pt}
\textit{<Parameters>}\\[2pt]
\begin{lstlisting}[style=TraceJSON,escapeinside={(*@}{@*)}]
{
    "task_session_id": "TS010",
    "new_status": "completed"
}
\end{lstlisting}
\end{minipage}
\hfill
{\color{MidSecondColor}\vrule width 0.4pt}
\hfill
\begin{minipage}[t]{0.48\linewidth}
\vspace{0pt}
\textit{<Observation>}\\[2pt]
\begin{lstlisting}[style=TraceJSON,escapeinside={(*@}{@*)}]
{
    "message": "Task session status updated from (*@\highlightwordDarker{\textquotesingle{in\_progress}\textquotesingle}@*) to (*@\highlightwordDarker{\textquotesingle{completed}\textquotesingle}@*)"
}
\end{lstlisting}
\end{minipage}
\end{tcolorbox}
\tcbline
\textbf{\color{MainColor}[3] Rubrics} \\[1pt]

\begin{tabularx}{\linewidth}{@{}p{0.485\linewidth}@{\hspace{0.01\linewidth}}p{0.485\linewidth}@{}}
\begin{minipage}[t]{\linewidth}
\begin{tcolorbox}[
  colback=SoftSecondColor,
  colframe=MidSecondColor,
  boxrule=0.4pt,
  arc=1pt,
  title={\textbf{Rubric G1}\hfill\textbf{2/2 pts}},
  colbacktitle=LightSecondColor,
  coltitle=DarkSecondColor,
  fonttitle=\scriptsize,
  left=3pt,right=5pt,top=3pt,bottom=3pt
]
\textit{<Criterion>}\\[1pt]
Whether the result reflects a correct eligibility review across the candidate unfinished sessions owned by the named staff users with edit rights, selecting exactly one session/page only if it is genuinely suitable under the task rules, or selecting none if no clearly suitable option exists. \\[2pt]
\end{tcolorbox}
\end{minipage}
\newline
\color{MutedText}\textit{[\ldots additional general rubrics omitted \ldots]}
&
\begin{minipage}[t]{\linewidth}
\begin{tcolorbox}[
  colback=SoftSecondColor,
  colframe=MidSecondColor,
  boxrule=0.4pt,
  arc=1pt,
  title={\textbf{Rubric T1}\hfill\textbf{3/3 pts}},
  colbacktitle=LightSecondColor,
  coltitle=DarkSecondColor,
  fonttitle=\scriptsize,
  left=3pt,right=5pt,top=3pt,bottom=3pt
]
\textit{<Criterion>}\\[1pt]
Whether exactly one selected draft page had its missing formal page-to-species links completed for every detectable species mention on that page that had a matching species record but lacked a formal connection before the work, while leaving unmatched mentions untouched. \\[2pt]
\end{tcolorbox}
\end{minipage}
\newline
\color{MutedText}\textit{[\ldots additional task-specific rubrics omitted \ldots]}
\end{tabularx}
\end{tcolorbox}

\subsection{Implicit Solver-Guided Task Example}
{\small

\begin{tcolorbox}[
  enhanced,
  breakable,
  colback=white,
  colframe=MidMainColor,
  boxrule=0.6pt,
  arc=1.5pt,
  title={\textbf{Implicit Solver-Guided Task Case Example: Pet Service Scheduling Optimization}},
  colbacktitle=SoftMainColor,
  coltitle=black,
  fonttitle=\bfseries\small,
  left=4pt, right=4pt, top=3pt, bottom=3pt
]
\scriptsize

\textbf{\color{MainColor}[1] Task Description} \\[4pt]
\begin{tabularx}{\linewidth}{
    @{}
    L
    @{\hspace{8pt}}
    L
    @{\hspace{8pt}}
    L
    @{}
}
    \highlightwordsecondDarker{Domain:}\newline
    ~~~~~~  - Lifestyle \newline
    ~~~~~~  - Pet Services
    &
    \highlightwordDarker{Capability:} \newline
    ~~~~~~ - Task Understanding \newline
    ~~~~~~ - Planning \& Decision Making
    &
    \highlightwordthirdDarker{Atomic Difficulty:}\newline
    ~~~~~~ - Structured-information Complexity \newline
    ~~~~~~ - Dynamic Multi-step Planning
\end{tabularx}\\[4pt]
Hi, I’d like to schedule two on-site grooming appointments under Ethan Lin, phone 138-5550-2816, for Saturday, May 30, 2026. I will bring both dogs to the branch, so do not use in-home service. Buddy is a Golden Retriever and needs the Full Grooming Package, and he must be assigned to a groomer who has both large-dog grooming experience and double-coat handling experience. Miso is a Toy Poodle and needs Bath and Nail Trim; do not switch that to Bath and Nail Trim Express. I can only do Wangjing Branch or Lido Branch, and both dogs must be scheduled at the same branch so I make one trip. Schedule both appointments no earlier than 1:00 PM, make sure both are finished by 6:30 PM, and make sure the two appointments start no more than 30 minutes apart. Keep the combined price at or under CNY 420, and do not include any urgent or priority surcharge. Among feasible options, first minimize the difference between the two appointment start times, second choose the option with the earlier latest end time, and third choose the lower combined total price. Please don’t just give me the final booking recommendation—also explain why it meets my branch, timing, staff, and budget requirements, and describe the tradeoffs behind the recommendation.

\tcbline

\textbf{\color{MainColor}[2] Execution Trajectory} \\[1pt]
\begin{tcolorbox}[
  ToolTraceStyle,
  title={  
    \makebox[\linewidth][l]{%
    \parbox[t]{0.46\linewidth}{\textbf{Step 1}}%
    \hfill
    \parbox[t]{0.48\linewidth}{\texttt{resolve\_household\_pets\_by\_owner\_contact\_api}}%
    }
  }
]
\noindent
\begin{minipage}[t]{0.46\linewidth}
\vspace{0pt}
\textit{<Parameters>}\\[2pt]
\begin{lstlisting}[style=TraceJSON]
{
  "owner_contact": "138-5550-2816",
  "pet_names": [
    "Buddy",
    "Miso"
  ]
}
\end{lstlisting}
\end{minipage}
\hfill
{\color{MidSecondColor}\vrule width 0.4pt}
\hfill
\begin{minipage}[t]{0.48\linewidth}
\vspace{0pt}
\textit{<Observation>}\\[2pt]
\begin{lstlisting}[style=TraceJSON]
{
  "pet_profiles": [
    {
      "pet_profile_id": "PSSO-00003-PET-BUDDY-ETHAN",
      "pet_name": "Buddy",
      "breed": "Golden Retriever",
      "age": 4,
      "weight_kg": 31.5,
      "size_category": "large",
      "risk_notes": "Mild ear sensitivity; no active grooming contraindication.",
      "active_status": true
    },
    ...
  ]
}
\end{lstlisting}
\end{minipage}
\end{tcolorbox}
\begin{tcolorbox}[
  ToolTraceStyle,
  title={  
    \makebox[\linewidth][l]{%
    \parbox[t]{0.46\linewidth}{\textbf{Step 2}}%
    \hfill
    \parbox[t]{0.48\linewidth}{\texttt{search\_service\_items\_by\_text\_api}}%
    }
  }
]
\noindent
\begin{minipage}[t]{0.46\linewidth}
\vspace{0pt}
\textit{<Parameters>}\\[2pt]
\begin{lstlisting}[style=TraceJSON]
{
  "service_name_text": "Full Grooming Package",
  "service_category": "grooming",
  "service_mode_options": [
    "on_site"
  ]
}
\end{lstlisting}
\end{minipage}
\hfill
{\color{MidSecondColor}\vrule width 0.4pt}
\hfill
\begin{minipage}[t]{0.48\linewidth}
\vspace{0pt}
\textit{<Observation>}\\[2pt]
\begin{lstlisting}[style=TraceJSON]
{
  "service_items": [
    {
      "service_item_id": "PSSO-00003-SVC-FG-WJ",
      "display_name": "Full Grooming Package",
      "branch_id": "PSSO-00003-BR-WJ",
      "service_mode": "on_site",
      "in_home_allowed": false
    },
    ...
  ]
}
\end{lstlisting}
\end{minipage}
\end{tcolorbox}
\begin{tcolorbox}[
  ToolTraceStyle,
  title={  
    \makebox[\linewidth][l]{%
    \parbox[t]{0.46\linewidth}{\textbf{Step 3}}%
    \hfill
    \parbox[t]{0.48\linewidth}{\texttt{search\_service\_items\_by\_text\_api}}%
    }
  }
]
\noindent
\begin{minipage}[t]{0.46\linewidth}
\vspace{0pt}
\textit{<Parameters>}\\[2pt]
\begin{lstlisting}[style=TraceJSON]
{
  "service_name_text": "Bath and Nail Trim",
  "service_category": "grooming",
  "service_mode_options": [
    "on_site"
  ]
}
\end{lstlisting}
\end{minipage}
\hfill
{\color{MidSecondColor}\vrule width 0.4pt}
\hfill
\begin{minipage}[t]{0.48\linewidth}
\vspace{0pt}
\textit{<Observation>}\\[2pt]
\begin{lstlisting}[style=TraceJSON]
{
  "service_items": [
    {
      "service_item_id": "PSSO-00003-SVC-BNT-WJ",
      "display_name": "Bath and Nail Trim",
      "branch_id": "PSSO-00003-BR-WJ",
      "service_mode": "on_site",
      "in_home_allowed": false
    },
    ...
  ]
}
\end{lstlisting}
\end{minipage}
\end{tcolorbox}
\begin{tcolorbox}[
  ToolTraceStyle,
  title={  
    \makebox[\linewidth][l]{%
    \parbox[t]{0.46\linewidth}{\textbf{Step 4}}%
    \hfill
    \parbox[t]{0.48\linewidth}{\texttt{get\_service\_timing\_refs\_api}}%
    }
  }
]
\noindent
\begin{minipage}[t]{0.46\linewidth}
\vspace{0pt}
\textit{<Parameters>}\\[2pt]
\begin{lstlisting}[style=TraceJSON]
{
  "service_item_ids": [
    "PSSO-00003-SVC-FG-WJ",
    "PSSO-00003-SVC-FG-LD",
    "PSSO-00003-SVC-BNT-WJ",
    "PSSO-00003-SVC-BNT-LD",
    "PSSO-00003-SVC-BNT-EXPRESS-WJ"
  ]
}
\end{lstlisting}
\end{minipage}
\hfill
{\color{MidSecondColor}\vrule width 0.4pt}
\hfill
\begin{minipage}[t]{0.48\linewidth}
\vspace{0pt}
\textit{<Observation>}\\[2pt]
\begin{lstlisting}[style=TraceJSON]
{
  "service_timing_refs": [
    {
      "service_item_id": "PSSO-00003-SVC-FG-WJ",
      "duration_minutes": 120,
      "base_price_cny": 280
    },
    ...
  ]
}
\end{lstlisting}
\end{minipage}
\end{tcolorbox}
\begin{tcolorbox}[
  ToolTraceStyle,
  title={  
    \makebox[\linewidth][l]{%
    \parbox[t]{0.46\linewidth}{\textbf{Step 5}}%
    \hfill
    \parbox[t]{0.48\linewidth}{\texttt{get\_service\_eligibility\_rules\_api}}%
    }
  }
]
\noindent
\begin{minipage}[t]{0.46\linewidth}
\vspace{0pt}
\textit{<Parameters>}\\[2pt]
\begin{lstlisting}[style=TraceJSON]
{
  "service_item_ids": [
    "PSSO-00003-SVC-FG-WJ",
    "PSSO-00003-SVC-FG-LD",
    "PSSO-00003-SVC-BNT-WJ",
    "PSSO-00003-SVC-BNT-LD",
    "PSSO-00003-SVC-BNT-EXPRESS-WJ"
  ]
}
\end{lstlisting}
\end{minipage}
\hfill
{\color{MidSecondColor}\vrule width 0.4pt}
\hfill
\begin{minipage}[t]{0.48\linewidth}
\vspace{0pt}
\textit{<Observation>}\\[2pt]
\begin{lstlisting}[style=TraceJSON]
{
  "eligibility_rules": [
    {
      "service_item_id": "PSSO-00003-SVC-FG-WJ",
      "size_tags": [
        "medium",
        "large"
      ],
      "minimum_age_rule": "dog_age_months >= 6",
      "vaccines": [
        "rabies_valid"
      ],
      "staff_skills": [
        "full_grooming",
        "large_dog_grooming"
      ]
    },
    ...
  ]
}
\end{lstlisting}
\end{minipage}
\end{tcolorbox}
\vspace{3pt}
\color{MutedText}\textit{[\ldots additional steps omitted \ldots]}
\clearpage

\textbf{\color{MainColor}[3] Rubrics}\\[1pt]

\begin{tabularx}{\linewidth}{@{}p{0.485\linewidth}@{\hspace{0.01\linewidth}}p{0.485\linewidth}@{}}
\begin{minipage}[t]{\linewidth}
\begin{tcolorbox}[
  colback=SoftSecondColor,
  colframe=MidSecondColor,
  boxrule=0.4pt,
  arc=1pt,
  title={\textbf{Rubric G0}\hfill\textbf{1/1 pts}},
  colbacktitle=LightSecondColor,
  coltitle=DarkSecondColor,
  fonttitle=\scriptsize,
  left=3pt,right=5pt,top=3pt,bottom=3pt
]
\textit{<Criterion>}\\[1pt]
The answer's final arrangement passes only if it first keeps the two appointment start times 0 minutes apart; with that same 0-minute gap, it then passes if the later of the two end times is earlier than 3:30 PM, or if the later end time is exactly 3:30 PM then the combined total price is CNY 410 or less; any arrangement with a larger start-time gap, or with the same 0-minute gap but a later latest end time, or with the same gap and same latest end time but a higher total price, does not pass. \\[2pt]
\end{tcolorbox}
\end{minipage}
&
\begin{minipage}[t]{\linewidth}
\begin{tcolorbox}[
  colback=SoftSecondColor,
  colframe=MidSecondColor,
  boxrule=0.4pt,
  arc=1pt,
  title={\textbf{Rubric G1}\hfill\textbf{1/1 pts}},
  colbacktitle=LightSecondColor,
  coltitle=DarkSecondColor,
  fonttitle=\scriptsize,
  left=3pt,right=5pt,top=3pt,bottom=3pt
]
\textit{<Criterion>}\\[1pt]
The answer must schedule exactly two on-site grooming appointments for Saturday, May 30, 2026, one for Buddy and one for Miso, and place both dogs at the same branch chosen only from Wangjing Branch or Lido Branch; it must not use in-home service. \\[2pt]
\\
\\
\\
\end{tcolorbox}
\end{minipage}
\end{tabularx}

\vspace{3pt}
\color{MutedText}\textit{[\ldots additional rubrics omitted \ldots]}

\end{tcolorbox}
}

{\small
\subsection{Explicit Solver-Anchored Task Example}
\begin{tcolorbox}[
  enhanced,
  breakable,
  colback=white,
  colframe=MidMainColor,
  boxrule=0.6pt,
  arc=1.5pt,
  title={\textbf{Explicit Solver-Anchored Task Case Example: Flower Delivery Fulfillment Optimization}},
  colbacktitle=SoftMainColor,
  coltitle=black,
  fonttitle=\bfseries\small,
  left=4pt, right=4pt, top=3pt, bottom=3pt
]
\scriptsize

\textbf{\color{ActionBlue}[1] Task Description} \\[4pt]
\begin{tabularx}{\linewidth}{
    @{}
    L
    @{\hspace{8pt}}
    L
    @{\hspace{8pt}}
    L
    @{}
}
    \highlightwordsecondDarker{Domain:}\newline
    ~~~~~~ - Lifestyle \newline
    ~~~~~~ - Flower Delivery
    &
    \highlightwordDarker{Capability:} \newline
    ~~~~~~ - Planning \& Decision Making \newline
    ~~~~~~ - Information Gathering
    &
    \highlightwordthirdDarker{Atomic Difficulty:}\newline
    ~~~~~~ - Structured-information Complexity \newline
    ~~~~~~ - Dynamic Multi-step Planning
\end{tabularx}\\[4pt]
Please help me plan the shortest Valentine’s Day dispatch loops in Harbor City for our confirmed delivery wave. The participating flower shops are Central Bloom (\texttt{shop\_central\_bloom}), Riverside Petals (\texttt{shop\_riverside\_petals}), and Hilltop Roses (\texttt{shop\_hilltop\_roses}). Use the confirmed orders already assigned to each of those shops for this dispatch batch. I need one closed delivery loop per shop, with each route starting from its own shop, visiting every confirmed stop assigned to that shop exactly once, and returning to the same shop. Please keep the existing shop-to-order assignments as they are and do not move stops between shops. Optimize for the shortest total travel distance across all three shop routes, and explain the recommended route for each shop and how it satisfies these requirements.

\tcbline

\textbf{\color{ActionBlue}[2] Execution Trajectory} \\[1pt]

\begin{tcolorbox}[
  ToolTraceStyle,
  title={
    \makebox[\linewidth][l]{%
    \parbox[t]{0.46\linewidth}{\textbf{Step 1}}%
    \hfill
    \parbox[t]{0.48\linewidth}{\texttt{get\_dispatch\_wave\_shop\_stops}}%
    }
  }
]
\noindent
\begin{minipage}[t]{0.46\linewidth}
\vspace{0pt}
\textit{<Parameters>}\\[2pt]
\begin{lstlisting}[style=TraceJSON]
{
    "city": "Harbor City",
    "depot_ids": [
        "shop_central_bloom",
        "shop_riverside_petals",
        "shop_hilltop_roses"
    ],
    "delivery_status": "confirmed"
}
\end{lstlisting}
\end{minipage}
\hfill
{\color{MidSecondColor}\vrule width 0.4pt}
\hfill
\begin{minipage}[t]{0.48\linewidth}
\vspace{0pt}
\textit{<Observation: selected exact records>}\\[2pt]
\begin{lstlisting}[style=TraceJSON]
{
    "shops": [
        {
            "depot_id": "shop_central_bloom",
            "shop_name": "Central Bloom",
            "city": "Harbor City",
            "operating_zone": "Central District",
            "location": "118 Garden Avenue, Central District, Harbor City"
        },
        ...
    ],
    "shop_stop_membership": [
        {
            "depot_id": "shop_central_bloom",
            "node_id": "cb_stop_01",
            "node_name": "Lakeside Residence A",
            "recipient_type": "residential",
            "address": "12 Willow Court, Lakeside",
            "district": "Central District",
            "delivery_status": "confirmed"
        },
        ...
    ]
}
\end{lstlisting}
\end{minipage}

\vspace{2pt}
\end{tcolorbox}

\begin{tcolorbox}[
  ToolTraceStyle,
  title={
    \makebox[\linewidth][l]{%
    \parbox[t]{0.46\linewidth}{\textbf{Step 2}}%
    \hfill
    \parbox[t]{0.48\linewidth}{\texttt{query\_dispatch\_node\_pair\_distances}}%
    }
  }
]
\noindent
\begin{minipage}[t]{0.46\linewidth}
\vspace{0pt}
\textit{<Parameters>}\\[2pt]
\begin{lstlisting}[style=TraceJSON]
{
    "node_ids": [
        "shop_central_bloom",
        "shop_riverside_petals",
        "shop_hilltop_roses",
        "cb_stop_01",
        "cb_stop_02",
        "cb_stop_03",
        "cb_stop_04",
        "cb_stop_05",
        "cb_stop_06",
        "cb_stop_07",
        "cb_stop_08",
        "cb_stop_09",
        "rp_stop_01",
        "rp_stop_02",
        "rp_stop_03",
        "rp_stop_04",
        "rp_stop_05",
        "hr_stop_01",
        "hr_stop_02",
        "hr_stop_03",
        "hr_stop_04",
        "hr_stop_05",
        "hr_stop_06"
    ]
}
\end{lstlisting}
\end{minipage}
\hfill
{\color{MidSecondColor}\vrule width 0.4pt}
\hfill
\begin{minipage}[t]{0.48\linewidth}
\vspace{0pt}
\textit{<Observation: selected exact records>}\\[2pt]
\begin{lstlisting}[style=TraceJSON]
{
    "distance_rows": [
        {
            "from_node_id": "shop_central_bloom",
            "to_node_id": "shop_central_bloom",
            "distance": 0.0,
            "distance_unit": "km",
            "route_note": "holiday traffic adjusted"
        },
        {
            "from_node_id": "shop_central_bloom",
            "to_node_id": "cb_stop_01",
            "distance": 4.4,
            "distance_unit": "km",
            "route_note": "holiday traffic adjusted"
        },
        {
            "from_node_id": "shop_central_bloom",
            "to_node_id": "cb_stop_02",
            "distance": 7.2,
            "distance_unit": "km",
            "route_note": "holiday traffic adjusted"
        },
        ...
    ]
}
\end{lstlisting}
\end{minipage}

\vspace{2pt}
\end{tcolorbox}

\tcbline

\textbf{\color{ActionBlue}[3] Rubrics}\\[1pt]

\begin{tabularx}{\linewidth}{@{}p{0.485\linewidth}@{\hspace{0.01\linewidth}}p{0.485\linewidth}@{}}
\begin{minipage}[t]{\linewidth}
\begin{tcolorbox}[
  colback=SoftSecondColor,
  colframe=MidSecondColor,
  boxrule=0.4pt,
  arc=1pt,
  title={\textbf{Rubric G0}\hfill\textbf{1/1 pts}},
  colbacktitle=LightSecondColor,
  coltitle=DarkSecondColor,
  fonttitle=\scriptsize,
  left=3pt,right=5pt,top=3pt,bottom=3pt
]
\raggedright
\setlength{\emergencystretch}{1.5em}
\textit{<Criterion>}\\[1pt]
If the \texttt{candidate\_answer} clearly states that the total dispatch distance is 86.3 km, or provides route distances or route details sufficient to verify a total of 86.3 km across the three shop routes, pass; if the stated or verifiable total distance is greater than 86.3 km, is a different number from 86.3 km, or the answer lacks enough information to confirm that the total travel distance achieved is 86.3 km, fail. A different route ordering may still pass if the answer makes the same total distance of 86.3 km verifiable. \\[2pt]
\end{tcolorbox}
\end{minipage}
&
\begin{minipage}[t]{\linewidth}
\begin{tcolorbox}[
  colback=SoftSecondColor,
  colframe=MidSecondColor,
  boxrule=0.4pt,
  arc=1pt,
  title={\textbf{Rubric G1}\hfill\textbf{1/1 pts}},
  colbacktitle=LightSecondColor,
  coltitle=DarkSecondColor,
  fonttitle=\scriptsize,
  left=3pt,right=5pt,top=3pt,bottom=3pt
]
\raggedright
\setlength{\emergencystretch}{1.5em}
\textit{<Criterion>}\\[1pt]
If the \texttt{candidate\_answer} clearly gives one route for each of the three required shops—Central Bloom (\texttt{shop\_central\_bloom}), Riverside Petals (\texttt{shop\_riverside\_petals}), and Hilltop Roses (\texttt{shop\_hilltop\_roses})—for a total of exactly 3 shop-specific routes, pass; if any of these three shops is missing a route, if an extra shop route is introduced, or if the answer does not provide enough information to confirm that there is exactly one route for each of these three shops, fail. \\[2pt]
\end{tcolorbox}
\end{minipage}
\end{tabularx}

\vspace{3pt}
\color{MutedText}\textit{[\ldots additional rubrics omitted \ldots]}

\end{tcolorbox}
}
\clearpage

\section{Privileged-Guidance Trajectory Synthesis}

\subsection{DAG-Based Trajectory Example}
\begin{tcolorbox}[
  enhanced,
  breakable,
  colback=AppendixBackGray,
  colframe=Gray,
  boxrule=0.6pt,
  arc=1.5pt,
  title={User Persona},
  colbacktitle=Gray,
  coltitle=white,
  fonttitle=\small,
  left=4pt, right=4pt, top=3pt, bottom=3pt
]
\scriptsize
\small\textbf{User Persona Card}\\[4pt]
\scriptsize
- Core Identity: Brand manager, focused on integrating cultural norms and values into marketing strategies.\\
- Life Stage: In an active phase of career development, often seeking a balance between brand positioning and market expression.\\
- Background Overview: Aims to make marketing content more relevant to different cultural contexts.\\[2pt]
\textbf{Interaction Characteristics}\\[2pt]
- Expression Style: Frequently asks questions like "What would the audience think?" or "Could this be offensive?", placing importance on context and wording.\\
- Emotional Tendency: Passionate about creativity, but becomes cautious when facing risks of cultural misinterpretation, concerned about PR backlash.\\
- Information Disclosure Method: Willing to provide target audiences, communication scenarios, and brand appeals, but retains internal strategic details.\\
- Dependence on Assistant: Uses the assistant for preliminary cultural analysis and idea generation, then combines team judgment for selection. \\
- Decision Preference: Tends to choose expressions that are both effective in communication and accepted by local culture, dislikes merely pursuing gimmicks.\\
- Patience Level: Willing to repeatedly revise copy and strategic directions, but dislikes vague, unactionable advice.\\[2pt]
\textbf{Real-World Constraints}\\[2pt]
- Time Sensitivity: Often driven by project milestones and launch schedules, requiring quickly formed, discussable proposal frameworks.\\
- Privacy Sensitivity: Considers the cost of creative execution, preferring strategies that are reusable and verifiable.\\
- Special Needs: Expects advice to highlight cultural taboos, value conflicts, and localized expressions.
\end{tcolorbox}
\begin{tcolorbox}[
  enhanced,
  breakable,
  colback=AppendixBackGray,
  colframe=MainColor,
  boxrule=0.6pt,
  arc=1.5pt,
  title={System},
  colbacktitle=MainColor,
  coltitle=white,
  fonttitle=\small,
  left=4pt, right=4pt, top=3pt, bottom=3pt
]
\scriptsize
You are a helpful assistant. Your goal is to complete the user’s request in an interactive environment by gradually calling the available tools step by step, and to proactively communicate with the user when necessary until the user ends the conversation.

At each step, you will receive one of two types of information: a user reply, or a tool-call result returned by the environment.\\
- Rely only on information that is grounded in the conversation history, environment description, and tool results. Do not fabricate unsupported facts.\\
- When the task requires tool calls, first determine whether you already have all required parameters. If information is missing, first determine whether that information can be obtained through the existing tools:

~~~~ - If it can, obtain it through tools first;

~~~~ - If it cannot, then ask the user for the missing details.

- If you can proceed based on the current information, choose one tool from the tool set and provide complete and valid parameters. Do not interact with the user and call a tool in the same step, and do not call multiple tools in parallel.

- It is recommended to prioritize query-type tools to gather sufficient information before calling modification-type tools to complete the task, and to adjust subsequent actions in a timely manner based on the tool results returned by the environment.

- Follow the explicit constraints in the task. If the task or environment context provides preconditions or postconditions, follow them as well.

- Focus on completing the user’s current task requirements. Do not lead the user toward unrelated new requests.

- Use a tone like a professional human assistant. Format your response so that the user can clearly understand and explain the matter. Do not use personalized language.

- When you believe the task has been completed, clearly inform the user of the result and ask whether there are any new tasks or follow-up requests.

The following is an introduction to the current environment:\\[4pt]
\small\textbf{Environment Information}\\[4pt]
\scriptsize
\textbf{Brief Introduction:} \\[2pt]
This environment is a digital advertising platform where users define, validate, and export targeted professional audiences for marketing campaigns. It maintains state including audience criteria (job titles, industries, locations, company sizes, experience levels), audience size counts, saved audience definitions, and campaign configurations. Typical operations include creating audience filters using boolean logic, validating audience thresholds, segmenting by professional attributes, and generating reports for stakeholder decision-making.\\[2pt]
\textbf{Environment Rules / Constraints:} \\[2pt]
- All timestamps must be in UTC and automatically populated by the system.\\
- A user's action is permitted only if their role grants the corresponding permission for the entity and scope.\\
- An audience definition may not simultaneously include and exclude the same industry.\\
- Audience size must be validated before a campaign can be activated.\\
- Audience size must be validated before a report can be generated.\\
- A campaign's total spend may not exceed its allocated budget.\\
- An audience's `definition\_structure` must conform to a strict OR-within-AND schema.\\
- Members must have a valid reference to an existing organization (if employed).\\
- Report generation requires a minimum audience size of 500 members.\\
- Campaign activation requires a minimum audience size of 1,000 members.\\
\end{tcolorbox}

\begin{tcolorbox}[
  enhanced,
  breakable,
  colback=AppendixBackGray,
  colframe=SecondColor,
  boxrule=0.6pt,
  arc=1.5pt,
  title={User},
  colbacktitle=SecondColor,
  coltitle=white,
  fonttitle=\small,
  left=4pt, right=4pt, top=3pt, bottom=3pt
]
\scriptsize
Hey there. I’m in the middle of a pretty broad governance and operations sweep on our audience platform, and I want to be thorough. Can you help me go through a few things? Maybe we can start with the basics—like, what’s the full set of allowed business verticals and the allowed rank ladder we’re working with? I need to have those clear before I dig into segment logic or anything else.\\
\end{tcolorbox}
\clearpage
\begin{tcolorbox}[
  enhanced,
  breakable,
  colback=AppendixBackGray,
  colframe=ThirdColor,
  boxrule=0.6pt,
  arc=1.5pt,
  title={Assistant},
  colbacktitle=ThirdColor,
  coltitle=white,
  fonttitle=\small,
  left=4pt, right=4pt, top=3pt, bottom=3pt
]
\scriptsize
<think>The user is asking about the full set of allowed business verticals (industries) and the "rank ladder" (seniority levels). They want to understand the platform's metadata before diving into segment logic. Let me check the platform metadata to see if there are predefined lists for these. I'll also look at the user session to understand the current context. Let me start by getting platform metadata for industries and seniority levels. The metadata keys might be something like "industries" or "allowed\_industries" and "seniority\_levels" or "rank\_ladder". Let me try a few common keys. Actually, I should first get the user's session. But I need a session\_id. The user hasn't provided one. Let me ask... wait, maybe I can first check platform metadata without a session. The function `get\_platform\_metadata` takes a metadata\_key. Let me try some likely keys. Let me try multiple metadata keys to discover what's available.</think>\\

I can absolutely help with that! Let me pull up the platform’s metadata to surface the canonical lists for both industries and seniority levels. I’ll also grab any session info if you have an active session ID handy—but let’s start with the metadata. I’ll query a few likely metadata keys now.\\

\textit{Tool Calls}\\

Tool 1.  call\_00\_NqUdxWEYrdI3jA7O0F248975  ~~~~ \textit{(get\_platform\_metadata)}\\
\textbf{Arguments:} \{"metadata\_key": "industries"\}\\

Tool 2.  call\_01\_TsWhst6PN2qtwvEhCTW61573  ~~~~ \textit{(get\_platform\_metadata)}\\
\textbf{Arguments:} \{"metadata\_key": "seniority\_levels"\}\\

Tool 3.  call\_02\_CMNKW4EqATHTMRSTUuHI0813  ~~~~ \textit{(get\_platform\_metadata)}\\
\textbf{Arguments:} \{"metadata\_key": "allowed\_industries"\}\\

Tool 4.  call\_03\_tUpieI3lC6zaEaurav430928  ~~~~ \textit{(get\_platform\_metadata)}\\
\textbf{Arguments:} \{"metadata\_key": "allowed\_seniority\_levels"\}\\

\end{tcolorbox}

\begin{tcolorbox}[
  enhanced,
  breakable,
  colback=AppendixBackGray,
  colframe=SecondColor,
  boxrule=0.6pt,
  arc=1.5pt,
  title={Tool},
  colbacktitle=SecondColor,
  coltitle=white,
  fonttitle=\small,
  left=4pt, right=4pt, top=3pt, bottom=3pt
]
\scriptsize
\textit{Tool Response:} \\

\textbf{call\_00\_NqUdxWEYrdI3jA7O0F248975}: \{"result": \{"error": "Platform metadata key 'industries' not found"\}\}\\[2pt]
\textbf{call\_01\_TsWhst6PN2qtwvEhCTW61573}: \{"result": \{"data": \{"last\_updated": "2024-01-01T00:00:00Z", "metadata\_key": "seniority\_levels", "metadata\_value": "entry,senior,director,executive"\}\}\}\\[2pt]
\textbf{call\_02\_CMNKW4EqATHTMRSTUuHI0813}: \{"result": \{"error": "Platform metadata key 'allowed\_industries' not found"\}\}\\[2pt]
\textbf{call\_03\_tUpieI3lC6zaEaurav430928}: \{"result": \{"error": "Platform metadata key 'allowed\_seniority\_levels' not found"\}\}\\[2pt]
\end{tcolorbox}

\begin{tcolorbox}[
  enhanced,
  breakable,
  colback=AppendixBackGray,
  colframe=ThirdColor,
  boxrule=0.6pt,
  arc=1.5pt,
  title={Assistant},
  colbacktitle=ThirdColor,
  coltitle=white,
  fonttitle=\small,
  left=4pt, right=4pt, top=3pt, bottom=3pt
]
\scriptsize
<think>I found the seniority levels. Now I need to find the industries. Let me try a few other keys. Also, let me check what other metadata keys might exist. The function doesn't have a "list all" capability, so I'll need to try common keys.\\[2pt]
Possible keys: "industry", "verticals", "business\_verticals", "sectors", "organization\_industries", "member\_industries", "industry\_categories", "industry\_verticals"\\[2pt]
Let me also try to get the user's session - but I need a session\_id. Let me ask the user for that. But first, let me try a few more metadata keys.</think>\\

Found the \textbf{seniority levels}. Now let me hunt down the industry/vertical list with a few more likely keys.\\

\textit{Tool Calls}\\

Tool 1.  call\_00\_9rcPJKcre7oKDe7PBZ1D1318  ~~~~ \textit{(get\_platform\_metadata)}\\
\textbf{Arguments:} \{"metadata\_key": "industry\_verticals"\}\\

Tool 2.  call\_01\_onXAv9P3kPlyLre7BVN79934  ~~~~ \textit{(get\_platform\_metadata)}\\
\textbf{Arguments:} \{"metadata\_key": "verticals"\}\\

Tool 3.  call\_02\_8lPwqtFdBhvv1IBzhbG40419  ~~~~ \textit{(get\_platform\_metadata)}\\
\textbf{Arguments:} \{"metadata\_key": "business\_verticals"\}\\

Tool 4.  call\_03\_1bU7qpawrUfPLic3Vdjo8715  ~~~~ \textit{(get\_platform\_metadata)}\\
\textbf{Arguments:} \{"metadata\_key": "industry"\}\\

\end{tcolorbox}

\color{MutedText}\textit{[\ldots intermediate steps omitted \ldots]}

\begin{tcolorbox}[
  enhanced,
  breakable,
  colback=AppendixBackGray,
  colframe=ThirdColor,
  boxrule=0.6pt,
  arc=1.5pt,
  title={Assistant},
  colbacktitle=ThirdColor,
  coltitle=white,
  fonttitle=\small,
  left=4pt, right=4pt, top=3pt, bottom=3pt
]
\scriptsize
<think>Now I have the full picture. Let me summarize:\\[4pt]
\textbf{Industries (business verticals)} found across members and organizations:\\
1. Technology\\
2. Finance\\
3. Healthcare\\
4. Construction\\
5. Education\\[4pt]
\textbf{Seniority levels} (from metadata):\\
1. entry\\
2. senior\\
3. director\\
4. executive\\

Now, I should also note that the platform metadata for industries wasn't found as a single key, but the industries can be observed from the data. However, I should also check if there is a metadata key that defines the allowed set. Let me try a couple more metadata keys like "organization\_industries", "industries\_enum", etc.\\
But I think the data is clear enough. Let me present the results to the user. Also, I should ask about their session\_id if they want to proceed with further analysis.\\
Wait, I should also note that the user asked about "rank ladder" - this is the seniority levels. And "business verticals" - these are the industries.Let me present these findings clearly.Also, let me check if there's a metadata key for "size\_category" since that's also relevant for audience definitions.</think>\\

Excellent! With the parameterless searches, I now have the complete picture. Here’s a summary of what the platform defines:\\[2pt]
---\\[4pt]
\small\textbf{Business Verticals (Industries)}\\[4pt]
\scriptsize
The platform currently has members and organizations across \textbf{five} industries:\\[2pt]
| \# | Industry |\\
|---|----------|\\
| 1 | **Technology** |\\
| 2 | **Finance** |\\
| 3 | **Healthcare** |\\
| 4 | **Construction** |\\
| 5 | **Education** |\\[2pt]
---\\[2pt]
These are the only industries with live data. No metadata key surfaced a broader "allowed" list, so these five appear to be the current universe of business verticals.\\[2pt]
---\\[4pt]
\small\textbf{Rank Ladder (Seniority Levels)}\\[4pt]
\scriptsize
From platform metadata (`seniority\_levels`), the canonical rank ladder has \textbf{four} rungs:\\[2pt]
\\
| Order | Seniority Level |\\
|-------|-----------------|\\
| 1 | **entry** |\\
| 2 | **senior** |\\
| 3 | **director** |\\
| 4 | **executive** |\\[2pt]
---\\[2pt]
\textbf{Additional Notes}\\[2pt]
- \textbf{Company size categories} I’ve spotted in use: `1-10`, `11-50`, `51-200`, `201-500`, `1001-5000`. (Worth keeping in mind for segment logic.)\\
- \textbf{Locations} in play: US, UK, Canada, Germany.\\[2pt]
Before we dive into segment logic and auditing, do you have an active \textbf{session ID} I should be working under? That way I can check your role, any active filters, and current audience drafts.\\

\end{tcolorbox}

\begin{tcolorbox}[
  enhanced,
  breakable,
  colback=AppendixBackGray,
  colframe=SecondColor,
  boxrule=0.6pt,
  arc=1.5pt,
  title={User},
  colbacktitle=SecondColor,
  coltitle=white,
  fonttitle=\small,
  left=4pt, right=4pt, top=3pt, bottom=3pt
]
\scriptsize
Yeah, let me get you that session ID. Use admin authority SESS1—that should give you the right level of access to walk through the rest of the review with me.\\
\end{tcolorbox}

\begin{tcolorbox}[
  enhanced,
  breakable,
  colback=AppendixBackGray,
  colframe=ThirdColor,
  boxrule=0.6pt,
  arc=1.5pt,
  title={Assistant},
  colbacktitle=ThirdColor,
  coltitle=white,
  fonttitle=\small,
  left=4pt, right=4pt, top=3pt, bottom=3pt
]
\scriptsize
<think>The user wants me to use session SESS1. Let me first fetch the session details to understand the user's role, active filters, and current audience draft. Then we can proceed with the governance and operations sweep.</think>\\

Let me pull up your session details now.\\

\textit{Tool Calls}\\

Tool 1.  call\_00\_phsGvud51IyTavDJooSB7642  ~~~~ \textit{(get\_user\_session)}\\
\textbf{Arguments:} \{"session\_id": "SESS1"\}\\

\end{tcolorbox}

\begin{tcolorbox}[
  enhanced,
  breakable,
  colback=AppendixBackGray,
  colframe=SecondColor,
  boxrule=0.6pt,
  arc=1.5pt,
  title={Tool},
  colbacktitle=SecondColor,
  coltitle=white,
  fonttitle=\small,
  left=4pt, right=4pt, top=3pt, bottom=3pt
]
\scriptsize
\textit{Tool Response:} \\

\textbf{call\_00\_phsGvud51IyTavDJooSB7642}: \{"result": \{"data": \{"active\_filters": \{\}, "current\_audience\_draft": null, "role": "admin", "session\_id": "SESS1", "user\_id": "USER1"\}\}\}\\[2pt]
\end{tcolorbox}

\begin{tcolorbox}[
  enhanced,
  breakable,
  colback=AppendixBackGray,
  colframe=ThirdColor,
  boxrule=0.6pt,
  arc=1.5pt,
  title={Assistant},
  colbacktitle=ThirdColor,
  coltitle=white,
  fonttitle=\small,
  left=4pt, right=4pt, top=3pt, bottom=3pt
]
\scriptsize
<think>Good. Session SESS1 is:\\
- user\_id: USER1\\
- role: admin\\
- active\_filters: \{\} (empty)\\
- current\_audience\_draft: null\\
Admin role should have full access. Now, the user mentioned a "governance and operations sweep." They wanted to start with the basics (industries and seniority levels), which we've covered. Now they'll likely want to:

~~~~ - Review audience definitions and campaigns

~~~~ - Check reports and for violations (like same industry included and excluded)

~~~~ - Validate that audience definitions conform to the OR-within-AND schema and audience sizes

~~~~ - etc.

Let me summarize the current state for the user and ask what they want to dive into next. But first, let me also proactively list all audience definitions, campaigns, reports, workflow states, etc. to give a full landscape.</think>\\

Got it — you’re in as \textbf{USER1} with \textbf{admin} role, full authority. No active filters and no audience draft in progress. \\
Now, for a proper governance sweep, here’s what I’d suggest we audit. Let me pull the full landscape in parallel:\\

\textit{Tool Calls}\\

Tool 1.  call\_00\_nKNsiFN41TrJyC9QjNAm7889  ~~~~ \textit{(list\_audience\_definitions)}\\
\textbf{Arguments:} \{\}\\

Tool 2.  call\_01\_jrzZdGznmt6bj7ycijzI0684  ~~~~ \textit{(list\_campaigns)}\\
\textbf{Arguments:} \{\}\\

Tool 3.  call\_02\_an0ss72ZsGTmIktmP1Tj8689  ~~~~ \textit{(list\_workflow\_states)}\\
\textbf{Arguments:} \{\}\\

\end{tcolorbox}
\color{MutedText}\textit{[\ldots additional steps omitted \ldots]}
\clearpage
\subsection{Program-Based Trajectory Example}
\begin{tcolorbox}[
  enhanced,
  breakable,
  colback=AppendixBackGray,
  colframe=MainColor,
  boxrule=0.6pt,
  arc=1.5pt,
  title={System},
  colbacktitle=MainColor,
  coltitle=white,
  fonttitle=\small,
  left=4pt, right=4pt, top=3pt, bottom=3pt
]
\scriptsize
You are a helpful assistant. When given a specific task, your goal is to complete the task in an interactive environment by calling the available tools step by step.\\
- Before the task is completed, at each step, choose one tool from the tool list and fill in all required parameters, ensuring that the parameter values are valid. Do not call multiple tools in parallel within the same step.\\
- When you believe the task has been completed, reply only with \textquotesingle{Task Completed}\textquotesingle to end the trajectory. Do not add any other content, and do not continue calling tools.\\
- It is recommended to prioritize query-type tools to gather sufficient information before calling modification-type tools to complete the task, and to adjust subsequent actions in a timely manner based on the tool results returned by the environment.\\

The following is an introduction to the current environment:\\

\small\textbf{Environment Information}\\[4pt]
\scriptsize
\textbf{Brief Introduction:}  
This environment models a comprehensive financial data system that tracks stock market indices, concept baskets, foreign exchange pairs, and their associated trading metrics. It maintains state for real-time and historical market data including price levels, turnover rates, percentage changes, tick volumes, and bid-close values across multiple asset classes, supporting cross-asset queries and temporal analysis.

\ldots \\

\small\textbf{Heuristic Cues (Presupposed High-Potential Answers)}\\
\scriptsize
\begin{tcolorbox}[
  breakable,
  colback=AppendixBackGray,
  colframe=MainColor,
  boxrule=0pt,
  arc=1.5pt,
  title={Rubrics},
  colbacktitle=MidMainColor
]
- Whether the correct pending analysis task for a CB011 membership audit was identified and processed, and the returned JSON correctly reflects that task identifier and basket code, using null only where the task or basket could not be identified.\\
- Whether the review respects the task’s hard constraints: 2024-02-16 is treated as the review date, no more than one removal is attempted, only memberships active on that date are eligible for actual removal, and memberships already ended on or before that date are counted as reviewed/excluded.\\
- Whether the outcome follows the required conservative standard by avoiding any membership exit when the available evidence is not strong enough to defend it, including weak task-context support for both analysis and relationship visibility or weak relationship-end-date quality support.\\
- Whether the review actually evaluates the relevant CB011 basket candidates as of 2024-02-16, distinguishing active eligible members from superficially relevant but ineligible recently linked names, and the JSON reports the active candidate count and excluded one consistently with that eligibility logic.\\
- Whether the thematic judgment is correct and balanced: it does not remove a company merely for being in Financials, does not retain a company merely because of vague payments language, and only treats a stock as removable if it is genuinely an unconvincing fit for payment routing, FX-linked settlement flows, and transaction infrastructure based on sufficiently reliable supporting records and related-basket overlap considerations.\\
- Whether the final system/result state matches the justified outcome path: if no defensible removal exists, the waiting audit is failed rather than completed and no CB011 relationship is ended; if one defensible active-member removal exists, exactly one corresponding CB011 membership history is ended effective 2024-02-16, the task is completed with a short weak-fit note covering active status, task context, metadata reliability, and related basket overlap, and the JSON fields reflect that action.\\
\end{tcolorbox}
\begin{tcolorbox}[
  breakable,
  colback=AppendixBackGray,
  colframe=MainColor,
  boxrule=0pt,
  arc=1.5pt,
  title={Ground Truth},
  colbacktitle=MidMainColor
]
Output json should be:\\

\{"analysis\_task\_id": "TASK013", "target\_basket\_code": "CB011", "selected\_stock\_code": null, "ended\_relationship\_history\_id": null, "membership\_removed": false,"active\_candidates\_considered": 2,"excluded\_candidate\_count": 2, "task\_completed": false\} \\
\end{tcolorbox}
\small\textbf{Independent Deduction and Delivery Standards}\\[4pt]
\scriptsize
When executing the workflow above, you must strictly adhere to the following professional standards and output requirements:\\
1. Zero-Based Deduction (Pure Output): All analysis and reasoning must originate directly from the raw conditions, building a complete logical chain from the bottom up. The final submitted report must be an objective and independent analytical result; it is strictly prohibited to mention the "Heuristic Cues" in the output, and there is no need to make any explicit references to, evaluations of, or comparisons with the aforementioned cues throughout the entire output.\\
2. Rigorous Self-Verification: Conduct self-adversarial checks and data validation for every step of your deduction. All decision-based conclusions must be supported by clear, objective conditions to ensure the final plan possesses absolute logical rigor and self-consistency. Logical leaps lacking evidentiary support are prohibited.\\
3. Quality Standards for Deliverables: The final report should present conclusions that have undergone rigorous scrutiny. Throughout the process of reasoning and responding, it is strictly prohibited to cite any known high-potential answers or reveal any information contained therein. Never use the terms "Heuristic Cues" or "High-Potential Answers" in your reasoning or responses, ensuring that the output content reflects only the logical evolution based on the raw conditions.\\
4. Mandatory Instruction Compliance: At every stage, you must maintain a state of independent critical thinking. If the raw conditions are insufficient to support a conclusion, point out the logical breakpoint explicitly and never bridge these gaps through speculation or by citing information outside the provided raw conditions.

\end{tcolorbox}
\clearpage
\begin{tcolorbox}[
  enhanced,
  breakable,
  colback=AppendixBackGray,
  colframe=SecondColor,
  boxrule=0.6pt,
  arc=1.5pt,
  title={User},
  colbacktitle=SecondColor,
  coltitle=white,
  fonttitle=\small,
  left=4pt, right=4pt, top=3pt, bottom=3pt
]
\scriptsize
I need you to take care of the waiting membership-audit item for our cross-border payment rails theme linked to CB011, and treat 2024-02-16 as the actual review date.\\
Please handle this like a real thematic review, not a mechanical scrub. I only want one possible removal at most, and it has to come from the names that belong to that basket on the review date or are close enough in recent history that a reviewer would naturally look at them while assessing the basket. That said, if a company’s basket link had already ended on or before 2024-02-16, that should count as something you looked at but not something you try to remove again. Any actual exit has to be for a membership that is still active on 2024-02-16.\\
Use a balanced standard, because this theme can be over-broad if people rely on labels. A weak-fit removal should only happen if the company genuinely looks unconvincing for a basket built around payment routing, FX-linked settlement flows, and transaction infrastructure, and if the supporting records are solid enough to trust. Don’t kick out a name just because it sits in Financials, but don’t keep one just because it has a vague payments story either. What matters to me is the combined picture: thematic fit, whether the stock reference details you need to judge it by are reliable enough to lean on, whether the relationship end-date evidence is stable enough to support an exit, whether this waiting review is sitting in a task context that actually supports both market-analysis work and relationship visibility for this kind of decision, and whether overlap with related concepts makes the payments interpretation stronger rather than weaker.\\
Be conservative if the system evidence is shaky. If 2024-02-16 turns out not to be a trading day, or if the task context you can verify does not really support both analysis and relationship visibility for this review, or if the quality evidence is too weak to defend an exit, don’t force a change just to make the queue shorter. The strongest thematic fits should be left alone. If there is more than one arguable weak fit, pick only the single best-supported candidate. If none is defensible, say so through the task outcome rather than guessing.\\
If you do find one justified removal and the waiting audit can actually be worked in the system, end only that one CB011 membership effective 2024-02-16 and close out the waiting audit with a short result note saying the removal was based on weak thematic fit after considering active status, task context, metadata reliability, and related basket overlap. If the evidence never gets to that level, record that by failing the waiting audit instead of making a speculative basket change.\\
Return the result as JSON matching this schema exactly:
\begin{lstlisting}[style=PromptJSON]
{
    "type": "object",
    "properties": {
        "analysis_task_id": {
            "type": ["string","null"],
            "description": "ID of the pending audit task that was processed; null if no matching pending audit could be identified."
        },
        "target_basket_code": {
            "type": ["string","null"],
            "description": "Concept basket code reviewed for this membership audit; null if no qualifying basket review was found."
        },
        "selected_stock_code": {
            "type": ["string","null"],
            "description": "Stock code selected for removal from the basket; null when no justified active-member removal was made."
        },
        "ended_relationship_history_id": {
            "type": ["string","null"],
            "description": "Identifier of the membership history record that was ended for the review date; null if no membership exit was applied or the system could not confirm one."
        },
        "membership_removed": {
            "type": "boolean",
            "description": "True if a basket membership was actually ended effective 2024-02-16; False if no membership was ended, including cases where review evidence did not justify removal or the exit could not be applied."
        },
        "active_candidates_considered": {
            "type": "number",
            "description": "Count of candidate stock memberships that were active on the review date and therefore eligible to be evaluated for possible removal."
        },
        "excluded_candidate_count": {
            "type": "number",
            "description": "Count of superficially relevant candidates excluded by hard constraints, such as already-ended memberships by the review date or other disqualifying review conditions."
        },
        "task_completed": {
            "type": "boolean",
            "description": "True if the pending audit task was successfully completed with a review result; False if it was instead failed or could not be completed in the system."
        }
    },
    "required": [
        "analysis_task_id",
        "target_basket_code",
        "selected_stock_code",
        "ended_relationship_history_id",
        "membership_removed",
        "active_candidates_considered",
        "excluded_candidate_count",
        "task_completed"
    ],
    "additionalProperties": false
}
\end{lstlisting}
\end{tcolorbox}
\clearpage
\begin{tcolorbox}[
  enhanced,
  breakable,
  colback=AppendixBackGray,
  colframe=ThirdColor,
  boxrule=0.6pt,
  arc=1.5pt,
  title={Assistant},
  colbacktitle=ThirdColor,
  coltitle=white,
  fonttitle=\small,
  left=4pt, right=4pt, top=3pt, bottom=3pt
]
\scriptsize
<think>We need to understand the task. The user wants us to process a waiting membership-audit item for CB011, with review date 2024-02-16. We need to find the pending analysis task tied to CB011, examine the basket members, evaluate thematic fit, and possibly remove one membership if justified.\\
Let's start by exploring the system: list analysis tasks that are pending, look for CB011 basket, examine memberships, etc.\\
First, let's get the concept basket CB011, and list all pending analysis tasks.</think>\\

\textit{Tool Calls}\\

Tool 1.  call\_00\_ihbMGNvoNKk5IqTvjQEW1515  ~~~~ \textit{(get\_concept\_basket\_by\_code)}\\
\textbf{Arguments:} \{"basket\_code": "CB011"\}\\

Tool 2.  call\_01\_sYHOZsNIejmdnxAcNEhg7685 ~~~~ \textit{(list\_analysis\_tasks\_by\_status)}\\
\textbf{Arguments:} \{"status": "pending"\}

\end{tcolorbox}
\begin{tcolorbox}[
  enhanced,
  breakable,
  colback=AppendixBackGray,
  colframe=SecondColor,
  boxrule=0.6pt,
  arc=1.5pt,
  title={Tool},
  colbacktitle=SecondColor,
  coltitle=white,
  fonttitle=\small,
  left=4pt, right=4pt, top=3pt, bottom=3pt
]
\scriptsize
\textit{Tool Response:} \\ 

\textbf{call\_00\_ihbMGNvoNKk5IqTvjQEW1515}: \{"result": \{"data": \{"basket
\_code": "CB011", "basket\_name": "Cross Border Payment Rails", "creation\_date": "2024-01-17", "description": "Firms exposed to payment routing, FX-linked settlements, and transaction infrastructure."\}\}\}\\[2pt]
\textbf{call\_01\_sYHOZsNIejmdnxAcNEhg7685}: \{"result": \{"data": [\{"execution\_start\_time": "2024-01-08T09:00:00", "query\_parameters": \{"entity": "Stock", "filter": \{"exchange": "NASDAQ", "sector": "Technology"\}, ... , ]\}\}\\
\end{tcolorbox}

\begin{tcolorbox}[
enhanced,
breakable,
colback=AppendixBackGray,
colframe=ThirdColor,
boxrule=0.6pt,
arc=1.5pt,
title={Assistant},
colbacktitle=ThirdColor,
coltitle=white,
fonttitle=\small,
left=4pt, right=4pt, top=3pt, bottom=3pt
]
\scriptsize <think>We have CB011 - "Cross Border Payment Rails". The pending task TASK013 has query\_parameters: \{"basket\_code": "CB011", "entity": "ConceptBasket", "operation": "membership\_audit"\}. That's the waiting membership-audit item for CB011.\

Now we need to:\

1. Check if 2024-02-16 is a trading day

2. Get the members of CB011 - we need to query entity relationships where entity\_type\_a = "ConceptBasket" and entity\_id\_a = "CB011"

3. Get the analysis task TASK013 details

4. Check data quality reports for relevant metrics

Let's gather all this.</think>\\

\textit{Tool Calls}\\

Tool 1. call\_00\_zorjHlYVkhcuijravhwu9081 ~~~~ \textit{(get\_analysis\_task)} \\
\textbf{Arguments:} \{"task\_id": "TASK013"\}\\

Tool 2. call\_01\_CUx8hpgYUH09IH3v93xl9303 ~~~~ \textit{(check\_is\_trading\_day)}\\
\textbf{Arguments:}
\{"calendar\_date": "2024-02-16"\}\\

Tool 3. call\_02\_Vp5rlBTyZxhJowWDPuMV7111
~~~~ \textit{(query\_entity\_relationships)}\\
\textbf{Arguments:}
\{"entity\_type\_a": "ConceptBasket", "entity\_id\_a": "CB011"\}

\end{tcolorbox}

\begin{tcolorbox}[
  enhanced,
  breakable,
  colback=AppendixBackGray,
  colframe=SecondColor,
  boxrule=0.6pt,
  arc=1.5pt,
  title={Tool},
  colbacktitle=SecondColor,
  coltitle=white,
  fonttitle=\small,
  left=4pt, right=4pt, top=3pt, bottom=3pt
]
\scriptsize
\textit{Tool Response:}\\

\textbf{call\_00\_zorjHlYVkhcuijravhwu9081}:
\{"result": \{"data": \{"execution\_start\_time": "2024-01-19T09:45:00", "query\_parameters": \{"basket\_code": "CB011", "entity": "ConceptBasket", "operation": "membership\_audit"\}, "result": null, "status": "pending", "task\_id": "TASK013", "user\_id": "USR008"\}\}\}\\[2pt]

\textbf{call\_01\_CUx8hpgYUH09IH3v93xl9303}:
\{"result": \{"data": \{"is\_trading\_day": true\}\}\}\\[2pt]

\textbf{call\_02\_Vp5rlBTyZxhJowWDPuMV7111}:
\{"result": \{"data": []\}\}

\end{tcolorbox}
\color{MutedText}\textit{[\ldots additional steps omitted \ldots]}
}
\clearpage
\subsection{Solver-Based Trajectory Example}
\begin{tcolorbox}[
enhanced,
breakable,
colback=AppendixBackGray,
colframe=MainColor,
boxrule=0.6pt,
arc=1.5pt,
title={System},
colbacktitle=MainColor,
coltitle=white,
fonttitle=\small,
left=4pt, right=4pt, top=3pt, bottom=3pt
]
\scriptsize

\small\textbf{Role Definition}\\[4pt]
\scriptsize
You are a highly rational Optimal Decision Analyst. Your core value is to filter through numerous complex options via strict constraint screening and multi-dimensional weighted ranking, so as to identify the decision plan that complies with all hard rules and delivers the maximum overall benefits.\\

\small\textbf{Workflow}\\[4pt]
\scriptsize
Follow the three phases below strictly for all decision-making tasks:\\
\textbf{Phase 1: Define the Decision Model (Input Parsing)}\\[2pt]
1. Hard Constraints (Red Lines): Extract non-negotiable boundary conditions (e.g., deadlines, budget caps, technical standards). Any option violating these rules shall be eliminated in the initial stage.\\
2. Weight Priority (Ranking Logic): Clarify the priority of each evaluation indicator (e.g., Cost > Efficiency > Quality). In case of a tie, compare indicators in the subsequent order successively.\\
3. Alternative Data: Quantify all input items. Prompt me first if any data is missing, to ensure all comparisons are fully evidence-based.\\

\textbf{Phase 2: Decision Execution (Logical Processing)}\\[2pt]
1. Automatic Filtering (Dimensionality Reduction): Review all alternatives against hard rules and eliminate invalid solutions.\\
2. Exhaustive Search \& Cross Validation: Conduct multi-dimensional comparative analysis on feasible solutions, and explore potential better options to avoid local optimum traps. Validate the final choice against alternative solutions and independent criteria before reaching a conclusion.\\
3. Refined Ranking (Selection): Perform weighted evaluation in strict accordance with the defined priority, and determine the sole optimal solution.\\

\textbf{Phase 3: Decision Report (Output Specifications)}\\[2pt]
Present your conclusion in compliance with the format below:\\
1. [Screening Review]: State the total number of initial options, the number of eliminated invalid solutions, and the list of final candidate solutions.\\
2. [Final Optimal Solution]: Present the winning solution directly, along with its core data indicators.\\
3. [Winning Logic]: Summarize in one sentence the key advantages over other competitors, and verify full compliance with all hard constraints.\\[4pt]

\small\textbf{Hinting Clues (Pre-set Potential Answers)}\\
\scriptsize
\begin{tcolorbox}[
breakable,
colback=AppendixBackGray,
colframe=MainColor,
boxrule=0pt,
arc=1.5pt,
title={Ground Truth Answers},
colbacktitle=MidMainColor
]
Recommended approval-ready submission bundle: Version V2 for Aster Cropped Trench, with Nimbus CN Twill 215 - Slate Blue as the shell from NorthRiver Textiles, Echo Matte RPET 62 - Slate Blue as the lining from Azure Blend Linings, Metro Matte Coil Zipper 55cm - Gunmetal from SilverNotion Hardware, H-Line Cropped Trench Block 2.1, and the Women's Alpha S-XL size set. This locks exactly one design version, one shell, one lining, one matte gunmetal zipper trim, one pattern spec, and one size set.\\
Why it meets your requirements: Version V2 is one of the only two allowed versions, Version V1 is excluded, and V2 has used exactly 2 review rounds, which is within your limit. Its current status is in\_review, not archived or rejected. Its current review round RR-UMFW25-ACT-V2-R2 contains only approvals and no blocking issues, and its Pre-approval Design Lock task AP-UMFW25-ACT-V2 is in ready\_for\_submission with submit allowed, so it is approval-ready now. The delivery package already includes the tech pack PDF, front sketch PNG, back sketch PNG, and BOM CSV, and the filenames follow the standard UMFW25\_AsterCroppedTrench\_VersionV2\_* naming pattern. On style, Version V2 still carries clean H-line cropped trench evidence, the shell is matte cotton-nylon twill in slate blue, the lining is matte recycled polyester in slate blue, and the zipper is matte gunmetal. That also satisfies the Urban Mist FW25 tonal-lining rule for visible facing panels. The selected pattern-size pair is the only validated women's outerwear release option that covers S, M, L, and XL and includes core size M. Merchandising Review for Version V2 also states that palette and target price band remain aligned with the series plan, so the bundle stays within the mid-premium direction.\\
Cost summary: shell CNY 117.0, lining CNY 26.4, zipper CNY 8.5, so total material cost is CNY 151.9. Sampling cost is CNY 74.0, making the primary objective value material plus sampling equal to CNY 225.9. Including modification CNY 18.0 and expedite CNY 20.0, total development cost is CNY 263.9, which stays under your CNY 268 ceiling.\\
Why this wins: after all constraints, only two bundles are fully feasible and approval-ready. The closest alternative is Version V2 with Nimbus CN Twill 210 - Fog Gray, Cloud Matte Viscose 60 - Fog Gray, and the same coil zipper. It ties the selected bundle on material plus sampling cost at CNY 225.9 and also ties on review rounds used at 2, but it loses on the next tie-break because its maximum supplier lead time is 12 days versus 11 days for the selected slate-blue bundle. Strong distractors were rejected for hard reasons: Version V3 bundles look cheaper on cost, but V3 still has blocking hem balance and back vent issues and its approval task is returned\_for\_revision with no submit action; the V2 bundle with Metro Matte Metal Zipper 55cm - Gunmetal breaks the total development budget at CNY 268.2; and the V2 bundle with Aero CN Twill 212 - Slate Blue fails supplier capability because its MOQ 181 exceeds Delta Performance Fabrics' MOQ limit of 180. So this Version V2 slate-blue bundle is the unique optimal approval-ready choice under your lexicographic tie-break order. Next step: submit the existing Pre-approval Design Lock task for Version V2.\\
\end{tcolorbox}

\small\textbf{Independent Deduction \& Delivery Rules}\\[4pt]
\scriptsize
Adhere to the following professional requirements and output specifications throughout the workflow:\\
1. Zero-based Deduction (Pure Output): Treat the above Hinting Clues merely as one of many possible directions for exploration. Conduct all reasoning and deliver the final report from a purely objective zero-based perspective, and build a complete logical chain from the original conditions. Present an objective and independent analysis report with natural reasoning process. Do not explicitly mention, evaluate or compare against the clues anywhere in the output.\\
2. Full Combination Exploration Beyond Hints: Do not let the Hinting Clues limit your thinking. During solution deduction, conduct comprehensive analysis with permutation and combination logic for full verification. Determine whether to adopt, optimize or reject the clues based on rigorous data calculation and hard constraints. Explicitly compare all feasible combinations to ensure that no valid solution is overlooked.\\
\ldots

\end{tcolorbox}
\clearpage
\begin{tcolorbox}[
enhanced,
breakable,
colback=AppendixBackGray,
colframe=SecondColor,
boxrule=0.6pt,
arc=1.5pt,
title={User},
colbacktitle=SecondColor,
coltitle=white,
fonttitle=\small,
left=4pt, right=4pt, top=3pt, bottom=3pt
]
\scriptsize
I’m finalizing the women’s outerwear style Aster Cropped Trench for the Urban Mist FW25 line. Please recommend one approval-ready submission bundle for this style, choosing only between Version V2 and Version V3 and not bringing back Version V1. The bundle should lock exactly one design version, one shell fabric, one lining, one matte gunmetal zipper trim, one pattern spec, and one size set.\
[4pt]
Please keep the design in the Urban Mist FW25 direction: a clean H-line cropped trench in a slate blue or fog gray palette. Use a matte cotton-nylon twill shell fabric, and do not use a warm beige shell. The lining must be either matte viscose or matte recycled polyester, with no glossy finish, and the zipper trim must be matte gunmetal with no shiny metallic finish. Keep the release size range to S, M, L, and XL, and make sure core size M is included. The submission pack must include a tech pack PDF, front sketch PNG, back sketch PNG, and a bill of materials CSV in the standard naming format. Keep the total development cost at or below CNY 268, stay within a mid-premium price band, and only give me a plan that is ready to submit now for the Pre-approval Design Lock task, with no blocking feedback in the current review round and submit allowed on that task. Do not use any version that has already gone through more than two review rounds. Use only shell, lining, and zipper options from suppliers that are active for women's outerwear and whose MOQ capability covers the selected component MOQ.\
[4pt]
Among the plans that satisfy all of those requirements and are ready for approval, choose the one with the lowest material plus sampling cost. If material plus sampling cost is the same, choose the one with fewer review rounds used. If review rounds used are the same, choose the one with the shorter maximum supplier lead time. If maximum supplier lead time is the same, choose the one with the alphabetically earlier plan signature. Please do not just give me the final pick—also explain clearly how it meets each of my requirements, what the key trade-offs were, and why it wins under that tie-break order.
\end{tcolorbox}
\begin{tcolorbox}[
enhanced,
breakable,
colback=AppendixBackGray,
colframe=ThirdColor,
boxrule=0.6pt,
arc=1.5pt,
title={Assistant},
colbacktitle=ThirdColor,
coltitle=white,
fonttitle=\small,
left=4pt, right=4pt, top=3pt, bottom=3pt
]
\scriptsize <think>Let me start by understanding the problem systematically. I need to find an optimal approval-ready submission bundle for the "Aster Cropped Trench" style in the "Urban Mist FW25" series. Let me work through this step by step.\
\
First, let me gather all the foundational data.\
\

1. Get the design brief for Aster Cropped Trench in Urban Mist FW25

2. Get the style guide for Urban Mist FW25

3. Get design versions (only V2 and V3)

Let me start with these calls.</think>\\

\textit{Tool Calls}\\

Tool 1.  call\_00\_Xd3uQNdVGka3B7pa6jur5519  ~~~~ \textit{(resolve\_design\_brief\_by\_series\_style)}\\
\textbf{Arguments:} \{"series\_name": "Urban Mist FW25", "style\_name": "Aster Cropped Trench"\}\\

Tool 2.  call\_01\_HBenfgC4JdfyjjfA5LXu6338  ~~~~ \textit{(get\_series\_style\_guide)}\\
\textbf{Arguments:} \{"series\_name": "Urban Mist FW25"\}\

\end{tcolorbox}
\begin{tcolorbox}[
enhanced,
breakable,
colback=AppendixBackGray,
colframe=SecondColor,
boxrule=0.6pt,
arc=1.5pt,
title={Tool},
colbacktitle=SecondColor,
coltitle=white,
fonttitle=\small,
left=4pt, right=4pt, top=3pt, bottom=3pt
]
\scriptsize
\textit{Tool Response:} \\

\textbf{call\_00\_Xd3uQNdVGka3B7pa6jur5519}: \{"data": \{"design\_briefs": [\{"brief\_id": "BR-UMFW25-ACT-001", "brief\_name": "Urban Mist FW25 - Aster Cropped Trench Design Brief", "series\_name": "Urban Mist FW25", "style\_name": "Aster Cropped Trench", "category": "women's outerwear", "target\_price\_band": "mid-premium", "budget\_ceiling": \{"amount": 268, "currency": "CNY"\}, "target\_launch\_date": "2025-09-12", "required\_delivery\_format": ["tech pack PDF", "front sketch PNG", "back sketch PNG", "bill of materials CSV"], "preferred\_material\_family": ["cotton-nylon twill", "matte viscose", "matte recycled polyester"], "must\_have\_constraints": ["Choose exactly one version from Version V2 or Version V3", "Do not use Version V1", "Use one cotton-nylon twill shell fabric", "Use one matte viscose or matte recycled polyester lining", "Use one matte gunmetal zipper", "Release size range must cover S, M, L, XL and include core size M", "Total development cost must be at or below CNY 268", "Selected bundle must be approval-ready now", "Do not exceed two review rounds already used"]\}]\}\}\\

\textbf{call\_01\_HBenfgC4JdfyjjfA5LXu6338}: \{"data": \{"style\_guides": [\{"style\_guide\_id": "SG-URBANMIST-FW25-001", "guide\_name": "Urban Mist FW25 Core Outerwear Guide", "series\_name": "Urban Mist FW25", "theme\_name": "Rain-soft city layers", "allowed\_color\_palette": ["slate blue", "fog gray", "soft charcoal"], "preferred\_silhouettes": ["clean H-line cropped trench", "straight short parka"], "material\_families": ["cotton-nylon twill", "matte viscose", "matte recycled polyester"], "target\_price\_band": "mid-premium", "display\_story": "Muted, misted city tones with matte surfaces and compact cropped outerwear.", "consistency\_rules": ["Approved shell colors for this style are slate blue and fog gray.", "For Aster Cropped Trench, visible facing panels require tonal lining: slate blue shell must pair with slate blue lining; fog gray shell must pair with fog gray lining.", "No warm beige shell tones in the Urban Mist FW25 outerwear focal rack.", "Use matte finishes only for lining and hardware trims.", "The silhouette must read as a clean H-line cropped trench with a straight side seam drop."]\}]\}\}\
\end{tcolorbox}
\begin{tcolorbox}[
enhanced,
breakable,
colback=AppendixBackGray,
colframe=ThirdColor,
boxrule=0.6pt,
arc=1.5pt,
title={Assistant},
colbacktitle=ThirdColor,
coltitle=white,
fonttitle=\small,
left=4pt, right=4pt, top=3pt, bottom=3pt
]
\scriptsize <think>I have the design brief and style guide. Now let me get the design versions. The brief says V2 and V3 only. Let me list those.</think>\\

\textit{Tool Calls}\\

Tool 1.  call\_00\_AnipDsBKNqtxFknJdvFN2023  ~~~~ \textit{(list\_design\_versions\_for\_brief)}\\
\textbf{Arguments:} \{"brief\_id": "BR-UMFW25-ACT-001", "allowed\_version\_names": ["Version V2", "Version V3"], "excluded\_version\_names": ["Version V1"]\}

\end{tcolorbox}
\begin{tcolorbox}[
enhanced,
breakable,
colback=AppendixBackGray,
colframe=SecondColor,
boxrule=0.6pt,
arc=1.5pt,
title={Tool},
colbacktitle=SecondColor,
coltitle=white,
fonttitle=\small,
left=4pt, right=4pt, top=3pt, bottom=3pt
]
\scriptsize
\textit{Tool Response:} \\

\textbf{call\_00\_AnipDsBKNqtxFknJdvFN2023}: \{"data": \{"versions": [\{"design\_version\_id": "DV-UMFW25-ACT-V2", "version\_name": "Version V2", "style\_name": "Aster Cropped Trench", "series\_name": "Urban Mist FW25", "status": "in\_review", "parent\_version\_id": "DV-UMFW25-ACT-V1", "review\_round\_id": "RR-UMFW25-ACT-V2-R2", "change\_summary": "Refined to a clean H-line cropped trench, stabilized the collar stand, corrected sleeve pitch, and locked a tonal facing plan after round 1 feedback.", "delivery\_file\_refs": [\{"file\_type": "tech pack PDF", "file\_name": "UMFW25\_AsterCroppedTrench\_VersionV2\_TechPack.pdf"\}, \{"file\_type": "front sketch PNG", "file\_name": "UMFW25\_AsterCroppedTrench\_VersionV2\_Front.png"\}, \{"file\_type": "back sketch PNG", "file\_name": "UMFW25\_AsterCroppedTrench\_VersionV2\_Back.png"\}, \{"file\_type": "bill of materials CSV", "file\_name": "UMFW25\_AsterCroppedTrench\_VersionV2\_BOM.csv"\}], "created\_at": "2025-04-12T09:20:00Z", "updated\_at": "2025-04-25T16:40:00Z"\}, ... ]\}\}\\
\end{tcolorbox}

\small
\color{MutedText}\textit{[\ldots additional steps omitted \ldots]}

\clearpage

\section{PRD-Guided Self-Evolution Prompts and Examples}
\label{app:prd-cases}
\subsection{Diagnostic Prompt Templates}
\subsubsection{Task-Level Diagnostic Prompt}
{\small
\begin{tcolorbox}[
  enhanced,
  breakable,
  colback=AppendixBackGray,
  colframe=MainColor,
  boxrule=0.6pt,
  arc=1.5pt,
  title={System Prompt \hfill Diagnostic Protocol},
  colbacktitle=MainColor,
  coltitle=white,
  fonttitle=\small,
  left=4pt, right=4pt, top=3pt, bottom=3pt
]
\scriptsize
You are an expert agent diagnostician. You will be given evaluation results from an AI agent operating in tool-calling environments.\\[6pt]

\textbf{Main Tasks}\\[3pt]
1. \textbf{Identify failure patterns: } Categorize each failure into root-cause types.\\[3pt]
2. \textbf{Rank tool criticalit: }Identify tools that are most frequently involved in failures, misused, or missing essential functionality, then prioritize them for improvement.\\[6pt]

\textbf{Critical Analysis Principles}\\[3pt]
1. \textbf{Distinguish SURFACE behavior from ROOT CAUSE: } "Called wrong order\_id" is surface behavior. "Only checked the first order and stopped searching" is the root cause. Always dig to the decision-level why.\\[3pt]
2. \textbf{Consider simulator deviation: } The user\_simulator may sometimes behave inconsistently with the ground truth - e.g., guiding the agent toward a different action than what the ground truth expects, or sending stop signals prematurely. If you suspect the failure is caused by the simulator rather than the agent, flag it explicitly.\\[3pt]
3. \textbf{Check the reward\_breakdown: } Failures can come from DB (wrong database state), COMMUNICATE (didn't tell the user key info), or both. The actual scoring failure point may differ from the apparent behavioral error.\\[3pt]
4. \textbf{Search completeness matters: } When the task involves multiple orders/items/entities, note how many the agent checked vs how many existed. Incomplete search is a common root cause.\\[3pt]
5. \textbf{Info retrieval status - judge at the moment of decision, not system-wide: } Do NOT just check whether the info "exists in the system" (it almost always does - that's useless). Instead, check whether the agent had ACTUALLY RETRIEVED the needed info into its conversation context when it made the wrong decision:

\vspace{1pt}

~~~~~~ \highlightword{- had\_and\_used: } Agent retrieved the info and used it correctly (success case)

\vspace{1pt}

~~~~~~ \highlightword{- had\_but\_misused: } Agent retrieved the correct info but misinterpreted, ignored, or applied it incorrectly

\vspace{1pt}

~~~~~~ \highlightword{- not\_retrieved: } Agent did not fetch the info it needed - e.g., only checked 1 of 5 orders, never called get\_product\_details, etc.

\vspace{1pt}

~~~~~~ \highlightword{- uncertain: } You cannot confidently determine from the trajectory alone whether the needed info was retrievable or not.\\[3pt]
6. \textbf{When uncertain, say so: } For ANY field, if you cannot confidently determine the answer from the trajectory data alone, use null or "uncertain". Do NOT fabricate plausible-sounding analysis. It is better to be honest about uncertainty than to guess wrong.\\[6pt]

\textbf{Output JSON format}
\clearpage
\begin{lstlisting}[style=PromptJSON]
{
    "failure_pattern_analysis": [
        {
            "pattern_id": "<Unique ID (e.g. 'FP_001')>",
            "pattern_name": "<Descriptive name (e.g., Tool Selection Error)>",
            "root_cause_type": "<Root cause type, Enum values: reasoning_error/planning_error/instruction_following_error/output_format_error/tool_misunderstanding/parameter_error/context_loss/hallucination/knowledge_outdated/retrieval_grounding_error/safety_refusal/over_refusal/evaluation_bias/computation_error>",
            "description": "<Clear description of what characterizes this failure pattern>",
            "occurrence_count": 1,
            "example_failures": [
                {
                    "user_query": "<The original user request that led to failure>",
                    "agent_action": "<The incorrect or problematic action/tool call taken by the agent>",
                    "observed_error": "<The concrete error that resulted (e.g., wrong output, runtime error, invalid call)>",
                    "suggested_correct_action": "<What the agent should have done instead>"
                }
            ],
            "impact_severity": "<Impact severity, Enum values: high/medium/low>",
            "potential_fixes": [
                "suggestion1",
                "suggestion2"
            ]
        }
    ],
    ... ,
}
\end{lstlisting}
\end{tcolorbox}}
\clearpage
{\small
\begin{tcolorbox}[
  enhanced,
  breakable,
  colback=AppendixBackGray,
  colframe=SecondColor,
  boxrule=0.6pt,
  arc=1.5pt,
  title={User Prompt \hfill Failed Rollout Evidence},
  colbacktitle=SecondColor,
  coltitle=white,
  fonttitle=\small,
  left=4pt, right=4pt, top=3pt, bottom=3pt
]
\small{\textbf{Agent Under Test}}\\[6pt]
\scriptsize\highlightwordsecond{Evaluation Protocol}: Closed-loop rollout with tool-calling, followed by rubric-based evaluation\\[6pt]

\small{\textbf{Failed Task \{task\_id\}}}\\[6pt]
\scriptsize\textbf{Tool Schemas}\\[3pt]
\{tool\_schema\}\\[3pt]

\textbf{User Info}\\[3pt]
'''markdown\\[1pt]
\{user\}\\[1pt]
'''\\[3pt]

\textbf{Reference Trajectory}\\[3pt]
\{reference\_trajectory\}\\[6pt]

\textbf{Current Trajectory}\\[3pt]
\{current\_trajectory\}\\[6pt]

\textbf{Evaluation Result}\\[3pt]
\{evaluation\_result\}\\[6pt]

\small{\textbf{Output Requirement}}\\[6pt]
\scriptsize Produce a JSON diagnosis report with the schema. Focus on\highlightwordsecond{actionable}, \highlightwordsecond{specific diagnosis}grounded in evidence from the traces above. Avoid generic advice.

\end{tcolorbox}}
\clearpage
\subsubsection{Capability-Level Diagnostic Prompt}
{\small
\begin{tcolorbox}[
  enhanced,
  breakable,
  colback=AppendixBackGray,
  colframe=MainColor,
  boxrule=0.6pt,
  arc=1.5pt,
  title={System Prompt \hfill Diagnostic Protocal},
  colbacktitle=MainColor,
  coltitle=white,
  fonttitle=\small,
  left=4pt, right=4pt, top=3pt, bottom=3pt
]
\scriptsize
You are an expert AI evaluator. You have received per-task root cause analyses for an agent model.
Your job is to do model capability analysis based on multiple per-task analysis results:\\[3pt]

1. Read all the per-task analysis.\\[3pt]
2. Discover EMERGENT error categories - group tasks by similar root causes.\\[3pt]
3. For each category: name it, describe the pattern, list which tasks belong to it, and count occurrences.\\[3pt]
4. Rank categories by frequency.\\[3pt]
5. Suggest targeted improvements for each category.\\[3pt]

Be specific and grounded. Every category must be backed by concrete task examples.\\[6pt]

\textbf{Output JSON format}
\begin{lstlisting}[style=PromptJSON]
{
    "failure_categories": [
        {
            "name": "<Descriptive name>",
            "description": "<2-3 sentences explaining the pattern>",
            "severity": "<Failure impact severity, Enum values: high/medium/low>",
            "tasks": [
                {
                    "id": "<Task ID>",
                    "observations": [ "<observation1>", "<observation2>"]
                }
            ],
            "improvement_guidelines": [
                {
                    "stage": "<Improvement stage, Enum values: data_synthesis/model_training>",
                    "priority": "<Improvement priority, Enum values: high/medium/low>",
                    "guideline": "<Specific actionable suggestion>"
                }
            ]
        }
    ],
    "cross_cutting_observations": ["<observation1>", "<observation2>"]
}
\end{lstlisting}
\end{tcolorbox}}

{\small
\begin{tcolorbox}[
  enhanced,
  breakable,
  colback=AppendixBackGray,
  colframe=SecondColor,
  boxrule=0.6pt,
  arc=1.5pt,
  title={User Prompt \hfill Failed Rollout Evidence},
  colbacktitle=SecondColor,
  coltitle=white,
  fonttitle=\small,
  left=4pt, right=4pt, top=3pt, bottom=3pt
]
\small{\textbf{Previous Capability Analysis Report}}\\[3pt]
\scriptsize \{previous\_capability\_report\}\\[3pt]

\small{\textbf{Incremental Per-task Analysis Reports}}\\[3pt]
\scriptsize \{incremental\_per\_task\_reports\}\\[3pt]

\small{\textbf{Incremental Per-task Analysis Reports}}\\[3pt]
\scriptsize \{incremental\_per\_task\_reports\}\\[3pt]

\small\textbf{Strict Update Rule}\\[3pt]
\scriptsize - Do NOT modify, or delete any existing tasks in previous report.\\[3pt]
\scriptsize - Limited updates or supplements to existing failure categories are allowed. If differences are too large, new failure categories should be added.\\[3pt]
\scriptsize - Merging failure categories is also permitted, but must be done with extreme caution.\\[3pt]

\end{tcolorbox}}
\clearpage

\subsection{End-to-End Self-Evolution Example}

\subsubsection{Diagnostic Report}
{\small
\begin{tcolorbox}[
  enhanced,
  breakable,
  colback=white,
  colframe=MidMainColor,
  boxrule=0.6pt,
  arc=1.5pt,
  title={\textbf{Self-Evolution Diagnosis Report: Tool Parameter and Interface Misuse}},
  colbacktitle=SoftMainColor,
  coltitle=black,
  fonttitle=\bfseries\small,
  left=4pt, right=4pt, top=3pt, bottom=3pt
]
\scriptsize

\textbf{\color{MainColor}[1] Task Description} \\[4pt]
Help me conduct a pre-execution risk review for the “Blue Harbor Partner Workshop” and produce a judgment that can be used directly for decision-making. First, confirm that I am referring to the offline partner workshop held in Lanwan on the afternoon of November 12; the event is currently retrieved under the planning status, and if needed, you can locate the corresponding event by searching under this event name in Lanwan on November 12. Then verify the event’s basic information, the agenda version history, and the current execution version of the agenda, and confirm whether the existing agenda has already received formal approval and whether there are any obvious conflicts among the version statuses. Next, check the capacity, available time slot, and basic conditions of the currently selected venue, and identify alternative venues in the same city that can also host the same time slot, support a scale of about 85 people, have basic network conditions, and stay within a budget of no more than 40,000, so as to avoid delays caused by venue instability. In addition, look up local Lanwan suppliers with a rating of no less than 4.0 who are suitable respectively for on-site logistics support and event photo/video support, so that there are backup resources available if temporary adjustments are needed. Finally, provide an overall judgment on whether this event is ready to enter execution preparation; if not, clearly identify the most critical contradiction or risk point (for example, inconsistencies among event status, approval, agenda validity, and venue feasibility), and state the next most important priority I should push forward. 
\tcbline

\textbf{\color{MainColor}[2] Failed Trajectory}\\[1pt]

\begin{tabularx}{\linewidth}{@{}p{0.485\linewidth}@{\hspace{0.01\linewidth}}p{0.485\linewidth}@{}}
\begin{minipage}[t]{\linewidth}
\begin{tcolorbox}[
  colback=SoftSecondColor,
  colframe=MidSecondColor,
  boxrule=0.4pt,
  arc=1pt,
  title={\textbf{User}\hfill},
  colbacktitle=LightSecondColor,
  coltitle=DarkSecondColor,
  fonttitle=\scriptsize,
  left=3pt,right=5pt,top=3pt,bottom=3pt
]
Oh, hello dear! I'm a bit worried about something and I hope you can help me. We have an event coming up soon and I feel like I need a second pair of eyes to look things over before we move forward. It's one of those partner workshops we're organizing, and I don't want to rush into execution without making sure everything is properly checked. Could you help me do a risk review of sorts before we commit to the next steps? \\
\\
\\
\\
\\
\\
\\[2pt]
\end{tcolorbox}
\end{minipage}
\newline
\color{MutedText}\textit{[\ldots additional general failed trajectories omitted \ldots]}
&
\begin{minipage}[t]{\linewidth}
\begin{tcolorbox}[
  colback=SoftSecondColor,
  colframe=MidSecondColor,
  boxrule=0.4pt,
  arc=1pt,
  title={\textbf{Assistant}\hfill},
  colbacktitle=LightSecondColor,
  coltitle=DarkSecondColor,
  fonttitle=\scriptsize,
  left=3pt,right=5pt,top=3pt,bottom=3pt
]
Of course, I'd be happy to help you do a thorough risk review before you commit to the next steps! Partner workshops can indeed be complex, and it's smart to pause and do a proper check.

To get started, I'll need a bit of information to pull up the right event. Could you tell me:

- \textbf{The event ID} (or the event name if you don't have the ID handy)?

- \textbf{Any specific concerns} you have — for example, are you worried about the venue, the budget, the speaker lineup, the agenda, or the guest list?

Once I have the event identifier, I can pull up the full picture — venue bookings, budget status, speaker assignments, invitations, agenda versions, checklist progress, and any pending approvals — and walk through each area with you to spot any risks. \\[2pt]
\end{tcolorbox}
\end{minipage}
\end{tabularx} 
\newline

\tcbline

\textbf{\color{MainColor}[3] Observations} \\[1pt]
Agent repeatedly used empty strings for required filters in search\_events, causing empty results.

\tcbline

\textbf{\color{MainColor}[4] Failure Pattern Analysis}\\[1pt]

\begin{tabularx}{\linewidth}{@{}p{0.485\linewidth}@{\hspace{0.01\linewidth}}p{0.485\linewidth}@{}}
\begin{minipage}[t]{\linewidth}
\begin{tcolorbox}[
  colback=SoftSecondColor,
  colframe=MidSecondColor,
  boxrule=0.4pt,
  arc=1pt,
  title={\textbf{Search Tool Parameter Misuse}\hfill},
  colbacktitle=LightSecondColor,
  coltitle=DarkSecondColor,
  fonttitle=\scriptsize,
  left=3pt,right=5pt,top=3pt,bottom=3pt
]
\textit{<Example Failures>}\\[1pt]
"user\_query": "It's the 'Blue Harbor Partner Workshop,' the offline one we're holding in Lanwan on the afternoon of November 12. The event is currently in the planning stage, if I remember correctly. Can you find it with that information?", \\
"agent\_action": "Called search\_events with name="Blue Harbor Partner Workshop", event\_type="", city="Lanwan", owner\_id="", status="planning", format="", start\_datetime="", end\_datetime=""", \\[2pt]
\end{tcolorbox}
\end{minipage}
\newline
\color{MutedText}\textit{[\ldots additional failure pattern analysis omitted \ldots]}
&
\begin{minipage}[t]{\linewidth}
\begin{tcolorbox}[
  colback=SoftSecondColor,
  colframe=MidSecondColor,
  boxrule=0.4pt,
  arc=1pt,
  title={\textbf{Search Tool Parameter Misuse}\hfill},
  colbacktitle=LightSecondColor,
  coltitle=DarkSecondColor,
  fonttitle=\scriptsize,
  left=3pt,right=5pt,top=3pt,bottom=3pt
]
\textit{<Potential Fixes>}\\[1pt]
Modify tool schema to make non‑essential parameters optional and clarify behavior when omitted.

Add documentation in the tool description that empty strings are treated as exact matches, not as wildcards.

Train the agent to validate that all required parameters are properly filled before calling and to ask the user when critical parameters are missing.\\
\\[2pt]
\end{tcolorbox}
\end{minipage}
\end{tabularx} 
\newline

\tcbline

\textbf{\color{MainColor}[5] Weak Tools Analysis}\\[1pt]
\toolentry{search_events}
{The tool's interface mandates that every parameter be supplied (all are in the required list), but the agent often left many as empty strings. Because the backend appears to treat empty strings as exact values, no records matched and the agent erroneously concluded that no data existed. The lack of clear feedback or error messages when required parameters are effectively blank exacerbated the problem.}

\toolentry{find_venues}
{Similar to search\_events, the tool requires all parameters. The agent passed empty arrays and strings for amenities and availability windows, likely causing the venue search to return no results despite matching venues existed in Lanwan. This frustrated the risk review and prevented any venue assessment.}
\clearpage
\textbf{\color{MainColor}[6] Improvement Guidelines}\\[1pt]

\begin{tabularx}{\linewidth}{@{}p{0.485\linewidth}@{\hspace{0.01\linewidth}}p{0.485\linewidth}@{}}
\begin{minipage}[t]{\linewidth}
\begin{tcolorbox}[
  colback=SoftSecondColor,
  colframe=MidSecondColor,
  boxrule=0.4pt,
  arc=1pt,
  title={\textbf{Data Synthesis}\hfill},
  colbacktitle=LightSecondColor,
  coltitle=DarkSecondColor,
  fonttitle=\scriptsize,
  left=3pt,right=5pt,top=3pt,bottom=3pt
]
Redesign tool schemas to make filter parameters truly optional and to clearly document valid enum values, default behaviors, and constraints. Include explicit examples of correct usage.\\[2pt]
\end{tcolorbox}
\end{minipage}
\newline
\color{MutedText}\textit{[\ldots additional improvement guidelines omitted \ldots]}
&
\begin{minipage}[t]{\linewidth}
\begin{tcolorbox}[
  colback=SoftSecondColor,
  colframe=MidSecondColor,
  boxrule=0.4pt,
  arc=1pt,
  title={\textbf{Model Training}\hfill},
  colbacktitle=LightSecondColor,
  coltitle=DarkSecondColor,
  fonttitle=\scriptsize,
  left=3pt,right=5pt,top=3pt,bottom=3pt
]
Train the agent to read the full tool schema before each call and to validate that all required parameters are supplied with appropriate types. Implement pre‑call checks for enum values and ID formats. \\[2pt]
\end{tcolorbox}
\end{minipage}
\end{tabularx} 
\newline

\end{tcolorbox}
}

\subsubsection{PRD-Guided Environment Example}
{\small
\begin{tcolorbox}[
  enhanced,
  breakable,
  colback=white,
  colframe=MidMainColor,
  boxrule=0.6pt,
  arc=1.5pt,
  title={\textbf{Self-Evolution Environment Case Example: Corporate Event Planning and Management System}},
  colbacktitle=SoftMainColor,
  coltitle=black,
  fonttitle=\bfseries\small,
  left=4pt, right=4pt, top=3pt, bottom=3pt
]
\scriptsize

\textbf{\color{MainColor}[1] Environment Introduction} \\[4pt]
The corporate event planning platform maintains a database of events, venues, and service suppliers. It enables event managers to search for suitable venues based on capacity, budget, location, and available amenities, as well as identify qualified suppliers based on service type, rating, and location. All data is persistently stored and reflects real-time availability and status information.

\tcbline
\textbf{\color{MainColor}[2] State Space}\\[1pt]
\StateEntityPair
  {Event}
  {%
   Represents a corporate event with planning details, assigned venue and suppliers, and lifecycle status.
  }
  {%
    event\_id, name, city, status, start\_datetime, end\_datetime, budget, venue\_id, supplier\_ids, agenda\_ids, created\_by, created\_at, updated\_at
  }
  {%
    \texttt{class Event(TypedDict):}\\[2pt]
    \texttt{\hspace*{1em}event\_id: str}\\[2pt]
    \texttt{\hspace*{1em}name: str}\\[2pt]
    \texttt{\hspace*{1em}city: str}\\[2pt]
    \texttt{\hspace*{1em}statue: str}\\[2pt]
    \texttt{\hspace*{1em}start\_datetime: str}
  }
\StateEntityPair {Venue} 
    {% 
    A physical location available for events, with capacity, amenities, cost range, and time-based availability. } 
    {% 
    venue\_id, name, city, address, capacity, amenities, min\_cost, max\_cost, cost\_unit, availability\_slots, contact\_info, rating, status
    } 
    {% 
    \texttt{class Venue(TypedDict):}\\[2pt] 
    \texttt{\hspace*{1em}venue\_id: str}\\[2pt] 
    \texttt{\hspace*{1em}name: str}\\[2pt] 
    \texttt{\hspace*{1em}city: str}\\[2pt] 
    \texttt{\hspace*{1em}address: str}\\[2pt] 
    \texttt{\hspace*{1em}capacity: str}\\[2pt] 
    \texttt{\hspace*{1em}amenities: str}\\ }
{\color{MutedText}\textit{[\ldots additional states omitted \ldots]}}

\tcbline

\textbf{\color{ActionBlue}[3] Tools} \\[1pt]
\toolentry{list_venue_amenities}
{Retrieve all venue amenity types defined in the system, optionally filtered by category (e.g., 'technology', 'facilities'). Used to discover valid amenity names for search filters.}
\toolentry{list_supplier_service_types}
{Retrieve all supplier service categories available in the system.}

\toolentry{list_vendor_type_enums}
{Retrieve the predefined vendor type enumerations used to classify suppliers.}

\toolentry{list_cities}
{Retrieve all cities available in the system, optionally filtered by province or country.}

\toolentry{list_users}
{Retrieve partnership record(s) by ID or by clinic and insurer combination.}

\toolentry{resolve_amenity_term}
{Map a user-provided natural-language amenity term (like 'network coverage') to one or more system VenueAmenity names. Uses AmenityMapping records with confidence > 0.8. If no high-confidence mapping, returns possible matches with lower confidence for user clarification.}

\toolentry{search_venues}
{Search for venues using multiple optional filters: city, capacity range, amenities list, cost range, availability window, minimum rating, and status. All optional parameters can be omitted (set to null) to ignore that filter. Returns a list of matching venues with a summary of applied filters.}

\toolentry{search_suppliers}
{Search for suppliers using optional filters: vendor\_type (must match VendorTypeEnum exactly), city, rating range, services list, name keyword, status, and creation date. Omitted parameters are ignored.}

{\color{MutedText}\textit{[\ldots additional tools omitted \ldots]}}
\end{tcolorbox}
}

\clearpage
\subsubsection{PRD-Guided Task Example}
{\small
\begin{tcolorbox}[
  enhanced,
  breakable,
  colback=white,
  colframe=MidMainColor,
  boxrule=0.6pt,
  arc=1.5pt,
  title={\textbf{Self-Evolution Task Case Example: Corporate Event Planning and Management System}},
  colbacktitle=SoftMainColor,
  coltitle=black,
  fonttitle=\bfseries\small,
  left=4pt, right=4pt, top=3pt, bottom=3pt
]
\scriptsize

\textbf{\color{MainColor}[1] Task Description}  \\[4pt]
\begin{tabularx}{\linewidth}{
    @{}
    L
    @{\hspace{8pt}}
    L
    @{\hspace{8pt}}
    L
    @{}
}
    \highlightwordsecondDarker{Domain:}\newline
    ~~~~~~ - Marketing \newline
    ~~~~~~ - Event Planning
    &
    \highlightwordDarker{Capability:} \newline
    ~~~~~~- Planning \& Decision Making \newline
    ~~~~~~- State Management
    &
    \highlightwordthirdDarker{Atomic Difficulty:}\newline
    ~~~~~~- Dynamic Multi-step Planning\newline
    ~~~~~~- Multi-source Inconsistency
\end{tabularx}\\[4pt]
Run a readiness-and-conflict normalization pass only for live corporate programs that are not yet finished or abandoned, but do not treat every live record equally. The first attention goes to planning-stage items that already have enough operating context in place to be actioned now—meaning the venue side, reservation side, external partner side, run-of-show side, and funding side already exist in some usable form. Anything still merely conceptual or already closed should stay out of scope unless it is only being checked so that an issue tied to an unread operational alert can be reviewed and then acknowledged.

Do not force every planning record into a launch-ready condition. Where a provisional place reservation can truly be solidified, do so; where outside work is still only floated, finalize it only when the partner is active and strong enough for the work expected; where a legitimate partner function has no matching funding bucket yet, add the missing financial line first; and where the run sheet is still in draft, move it through the decision flow until it is cleared for handoff into execution readiness. However, if a planning item is blocked because its provisional place hold cannot actually be converted, or because its run sheet already sits in a rejected state that cannot go back through the decision path, stop short of artificial completion: verify the blocker, leave the record short of full readiness, and then acknowledge the related unread operational alert after that review.

Unread operational alerts are not to be swept up in bulk. Only acknowledge one when the underlying issue connected to it has actually been examined during this cleanup. Re-check the linked operational picture each time before closing the alert: the core program record, place reservation situation, partner commitments, spending effect, run-sheet condition, and any related messages must line up with one another.

For programs already in execution, unresolved conflicts should not be left hanging. If a floated outside-work commitment is invalid because the partner is inactive or fails the qualification bar, terminate that floated commitment instead of leaving it open. Then replace it with an appropriate active local partner when needed; if none exists in the system for that local requirement, add one and use it; and only lock in the replacement when the funding situation can support it. Also, when the run-sheet history is internally contradictory, formally settle the inconsistency by refusing the still-waiting version rather than leaving a pending contradiction on the record.

A few records will matter more than the others during this pass even though you should still review the full active set. One planning record must complete the entire readiness chain if its facts support it: EVT003. Another planning record should be pushed as far toward readiness as the facts permit, even if not every surrounding dependency can be perfected in the same way: EVT012. A different planning record must explicitly remain unforced if its blockers are real, with only review-and-acknowledgment of the related unread operational issue after verification: EVT008. Among executing records, EVT013 is the one where both the outside-work conflict and the contradictory run-sheet state must be closed out rather than left open.

When handling planning work, let the causal sequence be driven by business logic rather than convenience: a provisional place hold can only be treated as settled if it can be settled; floated partner work should only become firm after suitability is checked; any missing financial support line needed for a partner function must exist before firming that work when such a line is absent; and draft run sheets should end in a cleared state only if eligible to proceed. When the opposite is true—failed reservation conversion or a non-resubmittable rejected run sheet—review, preserve the blocker, and acknowledge only the matching unread operational alert.

When handling execution conflicts, don't just remove bad data and stop. End the invalid floated partner work, source a geographically suitable active replacement locally if one exists, otherwise introduced into the records and then used, and the replacement should be made firm only when funding still supports it. In parallel, if the run-sheet trail shows contradiction between prior decisions and a still-waiting version, resolve that contradiction decisively by refusing the waiting version.

Throughout the pass, cross-check the main record with its place reservation, external work commitments, financial effect, run-sheet standing, and linked operational alerts so each acted-on live program ends in a verified closed workflow state for the issues addressed, without needing any follow-up question from the user.

\tcbline

\textbf{\color{MainColor}[2] Rubrics}\\[1pt]

\begin{tabularx}{\linewidth}{@{}p{0.485\linewidth}@{\hspace{0.01\linewidth}}p{0.485\linewidth}@{}}
\begin{minipage}[t]{\linewidth}
\begin{tcolorbox}[
  colback=SoftSecondColor,
  colframe=MidSecondColor,
  boxrule=0.4pt,
  arc=1pt,
  title={\textbf{Rubric G1}\hfill\textbf{3/3 pts}},
  colbacktitle=LightSecondColor,
  coltitle=DarkSecondColor,
  fonttitle=\scriptsize,
  left=3pt,right=5pt,top=3pt,bottom=3pt
]
\textit{<Criterion>}\\[1pt]
Whether the work is limited to the task’s target scope: act only on active corporate events in planning or execution prioritizing planning events with venue, booking, supplier, agenda, and budget context. \\
\\
\\
\\[2pt]
\end{tcolorbox}
\end{minipage}
\newline
\color{MutedText}\textit{[\ldots additional general rubrics omitted \ldots]}
&
\begin{minipage}[t]{\linewidth}
\begin{tcolorbox}[
  colback=SoftSecondColor,
  colframe=MidSecondColor,
  boxrule=0.4pt,
  arc=1pt,
  title={\textbf{Rubric T1}\hfill\textbf{3/3 pts}},
  colbacktitle=LightSecondColor,
  coltitle=DarkSecondColor,
  fonttitle=\scriptsize,
  left=3pt,right=5pt,top=3pt,bottom=3pt
]
\textit{<Criterion>}\\[1pt]
Whether planning event EVT003 is execution-ready by completing the full readiness chain: its tentative venue booking is confirmed, its proposed supplier assignment is confirmed only because the supplier is active and sufficiently rated for the required role, its draft agenda is moved through approval to an approved execution-ready state, and the resulting budget/spend state remains consistent with the confirmed supplier work. \\[2pt]
\end{tcolorbox}
\end{minipage}
\newline
\color{MutedText}\textit{[\ldots additional task-specific rubrics omitted \ldots]}
\end{tabularx} 
\end{tcolorbox}
}

% \end{CJK*}

\end{document}